\documentclass[11pt,reqno]{amsproc}
\allowdisplaybreaks
\numberwithin{equation}{section}
\usepackage{cite}
\usepackage{color}
\usepackage{multirow}
\usepackage{fullpage}
\usepackage{graphicx}
\usepackage{textcomp}
\usepackage{subfigure}
\usepackage{enumerate}
\usepackage{lineno}
\usepackage{algorithmic}
\usepackage{algorithm}
\usepackage[debug=false, colorlinks=true, pdfstartview=FitV,
linkcolor=blue, citecolor=blue, urlcolor=blue]{hyperref}
\usepackage[semicolon,square,authoryear]{natbib}

\usepackage[doipre={DOI:~}]{uri}
\usepackage[scriptsize]{caption}

\usepackage{morefloats}

\newlength{\drop}
\definecolor{amethyst}{rgb}{0.6, 0.4, 0.8}
\definecolor{burgundy}{rgb}{0.5, 0.0, 0.13}

\title{\textbf{A deep learning modeling framework to capture 
mixing patterns in reactive-transport systems}}

\author{\textbf{N.~V.~Jagtap}, \textbf{M.~K.~Mudunuru}, and \textbf{K.~B.~Nakshatrala}}

\keywords{Deep learning; 
reactive-transport; 
non-negative solutions; 
spatial-temporal forecasting; 
pattern recognition;
convolutional neural networks (CNN); 
long short-term memory (LSTM) networks}

\begin{document}
\begin{titlepage}
  \drop=0.1\textheight
  \centering
  \vspace*{\baselineskip}
  \rule{\textwidth}{1.6pt}\vspace*{-\baselineskip}\vspace*{2pt}
  \rule{\textwidth}{0.4pt}\\[\baselineskip]
       {\Large \textbf{\color{burgundy}
       A deep learning modeling framework to capture 
mixing patterns in reactive-transport systems}}\\[0.3\baselineskip]
       \rule{\textwidth}{0.4pt}\vspace*{-\baselineskip}\vspace{3.2pt}
       \rule{\textwidth}{1.6pt}\\[\baselineskip]
       \scshape
       \vspace{-0.1in}
       An e-print of the paper is available on arXiv. \par
       \vspace*{0.3\baselineskip}
       Authored by \\[0.5\baselineskip]
  {\Large N.~V.~Jagtap\par}
  {\itshape Graduate student, Dept. of Mechanical 
  Engineering, University of Houston, Texas 77204.}\\[0.5\baselineskip]

  {\Large M.~K.~Mudunuru\par}
  {\itshape Earth Scientist, 
  Atmospheric Sciences \& Global Change Division, \\
  Pacific Northwest National Laboratory, 
  Richland, Washington, 99352. \\
  \textbf{phone:} +1-509-375-6645, 
  \textbf{e-mail:} maruti@pnnl.gov}\\[0.5\baselineskip]

  {\Large K.~B.~Nakshatrala\par}
  {\itshape Associate Professor, 
  Department of Civil \& Environmental Engineering, \\
  University of Houston, Houston, Texas 77204. \\
  \textbf{phone:} +1-713-743-4418, \textbf{e-mail:} knakshatrala@uh.edu \\
  \textbf{website:} http://www.cive.uh.edu/faculty/nakshatrala}\\[0.5\baselineskip]
  \vspace{-0.25in}
\begin{figure*}[h]
  \centering
  \subfigure[Ground truth]
    {\includegraphics[width = 0.31\textwidth]
    {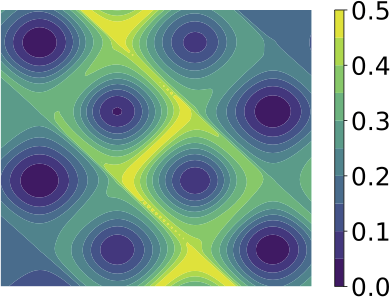}}
  %
  \hspace{0.2in}
  \subfigure[Prediction:~32\% data]
    {\includegraphics[width = 0.26\textwidth]
    {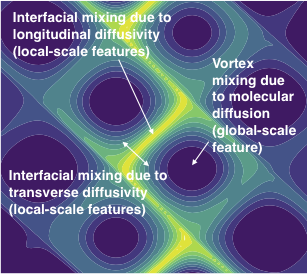}}
  %
  \hspace{0.2in}
  \subfigure[Prediction:~64\% data]
    {\includegraphics[width = 0.26\textwidth]
    {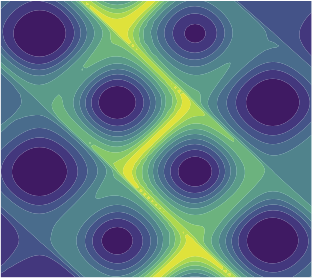}}

  \emph{The images in this figure 
  compare the ground truth (left) and 
  the predictions (middle and right) 
  from the proposed framework at the 
  end of the simulation (i.e., time 
  = 1.0). The middle image is a forecast 
  based on training using the first 32\% 
  of input data; training data is provided 
  until t = 0.32 and the model predicts 
  for the rest of the simulation time. 
  Likewise, the right image provides the 
  forecast based on the 64\% of the input 
  data. For capturing local-scale mixing 
  features (e.g., interfacial mixing due 
  to anisotropy), 32\% input data will 
  suffice to make predictions within 10\% error. 
  In addition, to capture global-scale mixing 
  features (e.g., mixing near vortices caused 
  by molecular diffusion) with similar accuracy, 
  one will need approximately twice 
  the amount of data.}
\end{figure*}

  \vfill
  {\scshape 2021} \\
  {\small Computational \& Applied Mechanics Laboratory} \par
\end{titlepage}

\begin{abstract}
Prediction and control of chemical mixing are vital for many scientific areas such as subsurface reactive transport, climate modeling, combustion, epidemiology, and pharmacology. Due to the complex nature of mixing in heterogeneous and anisotropic media, the mathematical models related to this phenomenon are not analytically tractable. Numerical simulations often provide a viable route to predict chemical mixing accurately. However, contemporary modeling approaches for mixing cannot utilize available spatial-temporal data to improve the accuracy of the future prediction and can be compute-intensive, especially when the spatial domain is large and for long-term temporal predictions. To address this knowledge gap, we will present in this paper a deep learning (DL) modeling framework applied to predict the progress of chemical mixing under fast bimolecular reactions. This framework uses convolutional neural networks (CNN) for capturing spatial patterns and long short-term memory (LSTM) networks for forecasting temporal variations in mixing. By careful design of the framework---placement of non-negative constraint on the weights of the CNN and the selection of activation function, the framework ensures non-negativity of the chemical species at all spatial points and for all times. Our DL-based framework is fast, accurate, and requires minimal data for training. The time needed to obtain a forecast using the model is a fraction ($\approx \mathcal{O}(-6)$) of the time needed to obtain the result using a high-fidelity simulation. To achieve an error of 10\% (measured using the infinity norm) for capturing local-scale mixing features such as interfacial mixing, only 24\% to 32\% of the sequence data for model training is required. To achieve the same level of accuracy for capturing global-scale mixing features, the sequence data required for model training is 64\% to 70\% of the total spatial-temporal data. Hence, the proposed approach---a fast and accurate way to forecast long-time spatial-temporal mixing patterns in heterogeneous and anisotropic media---will be a valuable tool for modeling  reactive-transport in a wide range of applications.
\end{abstract}

\maketitle

\vspace{-0.15in}


\section*{ABREVIATIONS}
\begin{quote}
CNN: Convolutional Neural Networks, 
DL: Deep Learning, 
FEM: Finite Element Method, 
LSTM: Long Short-Term Memory, 
RNN: Recurrent Neural Network 
\end{quote}

\section{INTRODUCTION AND MOTIVATION}
Reactive-transport equations arise in a wide range of scientific domains like epidemiology \citep{viguerie2020diffusion}, combustion \citep{naoumov2019chemical}, hypersonic flows \citep{yang2020reacting, ZHANG2020118812, 2018_Ren_PoF, Sabelnikov2018}, subsurface energy \citep{xiao2018reactive}, contamination remediation \citep{ARBOGAST199619}, air and water quality research \citep{zhu2002environmental}, and  population dynamics \citep{petrovskii2003exactly}. Although the underlying mathematical models in these applications are similar, the associated spatial and temporal scales are much different. For example, the spatial domains in combustion and hypersonic flows are small: an internal-combustion engine for the former case while the exterior surface of an atmospheric reentry capsule for the latter. In contrast, the spatial domains involved in subsurface remediation and epidemics could range from a few tens to several hundreds of miles. The reaction time scales in combustion are much less than a second while those associated with an epidemic could be very long (months to years). For complex geometries, presence of nonlinearities, and different spatial/temporal scales, analytical treatment of mathematical models arising in the said applications is not viable. Thus, it is customary to resort to numerical simulations.

Finite element \citep{codina2000stabilized,2016_Mudunuru_Nakshatrala_JCP_v305_p448_p493}, finite volume \citep{TAMBUE20103957}, and lattice Boltzmann methods \citep{GAO20179} are the contemporary numerical methods to solve reactive-transport equations. But these methods suffer from two major drawbacks. \emph{First}, they do not have an ability to exploit the available data in certain cases to improve prediction accuracy. To elaborate, consider the modeling of an epidemic. During an epidemic, spatial-temporal evolution data gathered by health agencies for the disease is often available \citep{viguerie2020diffusion, jha2020bayesian}. However, the traditional numerical methods cannot utilize such data to improve the accuracy of future predictions; the input to these methods will be limited to data such as the boundary and initial conditions of the solution fields.

The \emph{second} drawback is that reactive-transport simulations using traditional numerical methods are in general compute-intensive for practical problems \citep{AMIKIYA2018155}. For example, numerical solutions of a turbulent combustion problem will require small time-steps for accurate predictions to capture the high-speed traveling reaction front and involved multiple chemical species \citep{hundsdorfer2013numerical}. On the other hand, for subsurface remediation-related applications or for time-critical scenarios like chemical spills, the evolution of the chemical species for short and long times needs to be understood. Also, the spatial degrees of freedom associated with the subsurface remediation problems could be $\mathcal{O}(6)$ to $\mathcal{O}(9)$\footnote{$\mathcal{O}(n)$ means order of $10^{n}$.} \citep{chang2017large}. It takes hours of simulation time to resolve such spatial and temporal scales adequately even on state-of-the-art supercomputers---despite using thousands of processors, as the calculations in the time domain cannot be parallelized.

The \emph{central aim} of this paper is to present a deep learning (DL) framework that overcomes the said deficiencies of the conventional numerical methods. The efficacy of the framework is demonstrated by applying it to a reactive-transport problem. DL is a subset of machine learning that employs multiple neural layers to extract features successively from the input. An attractive feature of DL-based methods is that they can take advantage of the spatial-temporal data available for a limited portion of the total time of interest. This limited-duration data, obtained from physical experiments or high-fidelity numerical simulations can train a DL model. The resulting trained model can then predict accurate results over the remaining duration of interest much faster than the traditional numerical schemes. Such a model will be extremely useful for real-life situations like a pandemic or simulations requiring multiple runs by varying the simulation parameters of interest to assess their impact on the end results. Once the model is developed, the computational cost for training the model and getting results for the remaining duration of interest will be a fraction of the time required to get the entire solution using the traditional numerical methods. In short, the developed model enables seamless integration of modeling and data at scale, and significantly reduces the time required to get the solution.

The two main DL architectures used in the current framework are \emph{convolutional neural networks} (CNNs) and \emph{long short-term memory} (LSTM) network. CNNs have existed since 1980s \citep{schmidhuber2015deep} and are particularly suitable for processing data that has a grid-like topology; e.g., images that are in-fact a 2-dimensional grid of pixels. Their popularity and application has increased tremendously in the last decade due to exponential growth in computing speed as well as volume of data. CNNs offer distinct advantages over the dense/fully connected neural layers that CNNs learn the local patterns in their input features spaces whereas the dense neural layers can only learn the global features. 
The multiple layers of a CNN can learn the hierarchical patterns.
For instance, the first convolution neural layer may learn small local patterns.
A second convolution layer may learn broader set of abstract patterns comprised of the features from the first layer, and so on \citep{Goodfellow-et-al-2016, chollet2017deep}. 
In the current model, CNNs are used for spatial feature extraction and prediction of reactive-mixing patterns.

A LSTM network is a type of recurrent neural network (RNN), and was developed in the late 1990s \citep{schmidhuber2015deep} while researching the solutions for the well-known vanishing gradient problem in deep neural networks. CNNs are well-suited for processing a grid-like topology while RNNs are ideal for processing a sequence of values. LSTMs preserve the information across many time-steps by preventing the older signals from gradually vanishing during passing of information across the neural network layers \citep{Goodfellow-et-al-2016, chollet2017deep}. In the current model, the LSTM networks are used for temporal feature extraction and prediction. Thus, CNN and LSTM together are used for predicting spatial and temporal evolution of reactive-mixing. 
 
The proposed DL-based framework is validated by applying it to a fast bimolecular reaction problem. High-fidelity simulation results were obtained using finite element method for the entire time domain of interest. The DL-based model is trained using only portion of the total simulation time. The trained model is then used to predict subsequent spatial-temporal evolution of the chemical species. The evolution of the chemical species predicted by this model is compared with the simulation results to assess the accuracy of the model. Sensitivity analysis is performed with respect to amount of training data and various DL algorithm parameters to finalize the hyperparameter values for generating results.

The novel features of the proposed framework are: 
\begin{enumerate}[(a)]
\item \emph{data incorporation}: utilizes the available
spatio-temporal species data for accurate prediction of evolution of the species;   
\item \emph{fast}: computationally attractive because it requires 
  smaller time-to-solution compared to high-fidelity
  simulations; 
\item \emph{physically realistic}: ensures non-negative concentration
  fields at all times and at all spatial points unlike many traditional numerical schemes; 
\item \emph{accurate}: provides good accuracy, comparable to high-fidelity
numerical solutions;
\item \emph{predictive}: captures a wide range of reactive mixing; preferential mixing that
  manifests under small-scale velocity features, as
  well as incomplete mixing that manifests under large-scale
  velocity features;
\item \emph{requires minimal training data}: requires a small
  amount of training data compared to contemporary
  data-driven methods; and
\item \emph{extendable}: can be incorporated into
  existing simulators with minimal changes to the
  computer code---an attractive feature for application
  scientists. 
\end{enumerate}

Bereft of such a computational framework, forecasting for reactive-transport applications involving chaotic mixing will be prohibitively time-consuming, and addressing subsurface remediation problems under uncertainties would remain elusive. This paper serves as a proof-of-concept by showing the efficacy of the approach using a two component system; however, our approach is applicable even to those problems involving many chemical species.

An outline of the rest of this article is as follows. We will consider bimolecular reactions as the prototypical reactive-transport model and present the associated governing equations (\S\ref{Sec:S2_DLRT_GE}). The details about the proposed deep learning based modeling framework for transient reactive-transport will follow this presentation (\S\ref{Sec:S3_DLRT_Framework}). The predictive capabilities and computational performance of the proposed framework will be illustrated using representative numerical examples (\S\ref{Sec:S4_DLRT_NR}). The paper will end with concluding remarks along with a discussion on potential extensions of this work (\S\ref{Sec:S5_DLRT_CR}).

\section{GOVERNING EQUATIONS:~REACTIVE-TRANSPORT}
\label{Sec:S2_DLRT_GE}
Let $\Omega$ be a bounded domain with piecewise smooth boundary 
$\partial \Omega = \overline{\Omega} - \Omega$, where a superposed 
bar denotes the set closure. A spatial point is denoted by $\mathbf{x} \in 
\overline{\Omega}$, and the spatial gradient and divergence 
operators are denoted by $\mathrm{grad}[\cdot]$ and 
$\mathrm{div}[\cdot]$, respectively. The unit outward normal 
vector to the boundary is denoted by $\widehat{\mathbf{n}}(\mathbf{x})$. 
The time is denoted by $t \in [0,\mathcal{I}]$, where $\mathcal{I}$ is the 
length of the time interval of interest. 

Consider a bimolecular reaction of the following form:
\begin{align}
  \label{Eqn:Bimolecular_fast_reaction}
    n_A A + n_B B \longrightarrow n_C C 
\end{align} 
where $n_A$, $n_B$ and $n_C$ are stoichiometric
coefficients; $A$ and $B$ are the reactants; and 
$C$ is the product of the reaction. We denote the 
molar concentrations of the chemical species by 
$c_A(\mathbf{x},t)$, $c_B(\mathbf{x},t)$ and 
$c_C(\mathbf{x},t)$, respectively.

The boundary is divided into two complementary parts.
$\Gamma^{\mathrm{D}}$ is that part of the boundary on
which concentrations of the chemical species (i.e., Dirichlet 
boundary conditions) are prescribed. $\Gamma^{\mathrm{N}}$
is the part of the boundary on which diffusive fluxes
of the chemical species (i.e., Neumann boundary conditions) 
are prescribed.
Note that
\begin{align}
  \Gamma^{\mathrm{D}} \cup \Gamma^{\mathrm{N}}
  = \partial \Omega
  \quad \mathrm{and} \quad
  \Gamma^{\mathrm{D}} \cap
  \Gamma^{\mathrm{N}} = \emptyset
\end{align}

In the absence of non-reactive volumetric sources 
and advection, the equations that govern the fate 
of the reactants and the product in a bimolecular 
reaction can be written as follows:
\begin{subequations}
  \label{Eqn:DRs_for_A_B_C}
  \begin{alignat}{2}
    \label{Eqn:DRs_for_A}
    &\frac{\partial c_A}{\partial t} - \mathrm{div}[\mathbf{D}
      (\mathbf{x},t) \, \mathrm{grad}[c_A]] =
    -n_{\small{A}} \,r(c_A,c_B)
    &&\quad \mathrm{in} \; \Omega \times ]0,
    \mathcal{I}[ \\
    \label{Eqn:DRs_for_B}
    &\frac{\partial c_B}{\partial t} - \mathrm{div}[\mathbf{D}
    (\mathbf{x},t) \, \mathrm{grad}[c_B]] =  -
    n_{\small{B}} \, r(c_A, c_B)
    &&\quad \mathrm{in} \; \Omega \times ]0,
    \mathcal{I}[ \\
    \label{Eqn:DRs_for_C}
    &\frac{\partial c_C}{\partial t} - \mathrm{div}[\mathbf{D}
    (\mathbf{x},t) \, \mathrm{grad}[c_C]] = +
    n_{\small{C}} \, r(c_A, c_B)
    &&\quad \mathrm{in} \; \Omega \times ]0,
    \mathcal{I}[ \\
    \label{Eqn:DRs_for_Dirchlet}
    &c_i(\mathbf{x},t) = c^{\mathrm{p}}_i(\mathbf{x},t)
    &&\quad \mathrm{on} \; \Gamma^{\mathrm{D}} \times ]0,
    \mathcal{I}[ \quad (i = A, \, B, \, C) \\
    \label{Eqn:DRs_for_Neumann}
    -&\widehat{\mathbf{n}}(\mathbf{x}) \bullet
    \mathbf{D} (\mathbf{x},t) \, \mathrm{grad}[c_i] 
    = h^{\mathrm{p}}_i(\mathbf{x},t)
    &&\quad \mathrm{on} \;
    \Gamma^{\mathrm{N}} \times ]0, \mathcal{I}[
      \quad (i = A, \, B, \, C) \\
      \label{Eqn:DRs_for_IC}
      &c_i(\mathbf{x},t=0) = c^{0}_i(\mathbf{x})
      &&\quad \mathrm{in} \; \overline{\Omega}
      \quad (i = A, \, B, \, C)
  \end{alignat}
\end{subequations}
where $c^{0}_i(\mathbf{x})$ is the initial
concentration of the $i$-th chemical species,
$\mathbf{D}(\mathbf{x},t)$ is the anisotropic
dispersion tensor, and $r(c_A,c_B)$ is the
reactive-part of the volumetric source. $c^{\mathrm{p}}_i(\mathbf{x},t)$
is the prescribed (molar) concentration of the
$i$-th chemical species on $\Gamma^{\mathrm{D}}$.
$h^{\mathrm{p}}_i(\mathbf{x},t)$ is the prescribed
diffusive flux of the $i$-th chemical species on
$\Gamma^{\mathrm{N}}$.

\subsection{Invariants and fast reactions}
\label{SubSec:S2_FastRxns_Invs}
One can simplify the solution procedure by introducing invariants, 
as done in several papers in the literature; for example, see 
\citep{nakshatrala2013numerical}. 
The name ``invariants" is apparent from the fact that pure transport equations govern the fate of these reaction 
invariants (i.e., without any reaction term) rather than dispersion-reaction equations. For a bimolecular equation, 
one can find twelve reaction invariants. However, only two of them are (linearly) independent. Herein we will take 
the following two independent invariants: 
\begin{align}
  &c_F(\mathbf{x},t) = c_A(\mathbf{x},t) + \frac{n_A}{n_C} c_C(\mathbf{x},t) \\
  &c_G(\mathbf{x},t) = c_B(\mathbf{x},t) + \frac{n_B}{n_C} c_C(\mathbf{x},t) 
\end{align}

The fate of these two invariants are governed 
by the following equations: 
\begin{subequations}
  \begin{alignat}{2}
    \label{Eqn:Diffusion_for_F}
    &\frac{\partial c_F}{\partial t} - \mathrm{div}[\mathbf{D}
    (\mathbf{x},t) \, \mathrm{grad}[c_F]] = 0
    &&\quad \mathrm{in} \; \Omega \times ]0, \mathcal{I}[ \\
    \label{Eqn:Diffusion_for_Dirchlet_F}
    &c_F(\mathbf{x},t) = c_F^{\mathrm{p}}(\mathbf{x},t) :=
    c^{\mathrm{p}}_A(\mathbf{x},t) + \left( \frac{n_A}{n_C}
    \right) c^{\mathrm{p}}_C(\mathbf{x},t) 
    &&\quad \mathrm{on}
    \; \Gamma^{\mathrm{D}} \times ]0, \mathcal{I}[ \\
    \label{Eqn:Diffusion_for_Neumann_F}
    & \left(-\mathbf{D} (\mathbf{x},t) \, \mathrm{grad}[c_F]
    \right) \bullet \widehat{\mathbf{n}}(\mathbf{x}) =  h^{\mathrm{p}}
    _F(\mathbf{x},t) := h^{\mathrm{p}}_A(\mathbf{x},t) +
    \left( \frac{n_A}{n_C} \right) h^{\mathrm{p}}_C(\mathbf{x},t) 
    &&\quad \mathrm{on} \; \Gamma^{\mathrm{N}} \times ]0,
    \mathcal{I}[ \\
    \label{Eqn:Diffusion_for_IC_F}
    &c_F(\mathbf{x},t=0) = c^{0}_F(\mathbf{x}) :=
    c^{0}_A(\mathbf{x}) + \left( \frac{n_A}{n_C} \right)
    c^{0}_C(\mathbf{x}) 
    &&\quad \mathrm{in} \; \overline{\Omega}
  \end{alignat}
\end{subequations}
and
\begin{subequations}
  \begin{alignat}{2}
    \label{Eqn:Diffusion_for_G}
    &\frac{\partial c_G}{\partial t} - \mathrm{div}[\mathbf{D}
    (\mathbf{x},t) \, \mathrm{grad}[c_G]] = 0
    &&\quad \mathrm{in} \; \Omega \times ]0, \mathcal{I}[ \\
    \label{Eqn:Diffusion_for_Dirchlet_G}
    &c_G(\mathbf{x},t) = c^{\mathrm{p}}_G(\mathbf{x},t) :=
    c^{\mathrm{p}}_B (\mathbf{x},t) + \left( \frac{n_B}{n_C}
    \right) c^{\mathrm{p}}_C (\mathbf{x},t) 
    &&\quad \mathrm{on}
    \; \Gamma^{\mathrm{D}} \times ]0, \mathcal{I}[ \\
    \label{Eqn:Diffusion_for_Neumann_G}
    & \left(-\mathbf{D} (\mathbf{x},t) \, \mathrm{grad}[c_G]
    \right) \bullet \widehat{\mathbf{n}}(\mathbf{x}) = h^{\mathrm{p}}_
    G(\mathbf{x},t) := h^{\mathrm{p}}_B(\mathbf{x},t) +
    \left( \frac{n_B}{n_C} \right) h^{\mathrm{p}}_C(\mathbf{x},t) 
    &&\quad \mathrm{on} \; \Gamma^{\mathrm{N}} \times ]0,
    \mathcal{I}[ \\
    \label{Eqn:Diffusion_for_IC_G}
    &c_G(\mathbf{x},t=0) = c^{0}_G(\mathbf{x}) := c^{0}_B
    (\mathbf{x}) + \left( \frac{n_B}{n_C} \right) c^{0}_
    C(\mathbf{x}) 
    &&\quad \mathrm{in} \; \overline{\Omega}
  \end{alignat}
\end{subequations}

In this paper, we assume that the bimolecular reaction is 
fast---the time-scale of the reaction is faster than that 
of the dispersion process. This assumption implies that at 
any spatial point and at instance of time, concentration 
of only one of the reactants $A$ or $B$ could be non-zero. 
Said differently, if both $c_A$ and $c_B$ are non-zero, the 
reaction will proceed further instantaneously until one of 
the reactants is depleted completely.

For \emph{fast} bimolecular reactions, the solution procedure 
can be further simplified by noting that reactants $A$ and $B$ cannot 
coexist. Once $c_F(\mathbf{x},t)$ and $c_G(\mathbf{x},t)$ are 
known, the concentrations of the reactants and the product are 
calculated as follows: 
\begin{subequations}
  \label{Eqn:Fast_A_B_C}
  \begin{align}
    \label{Eqn:Fast_A}
    &c_A(\mathbf{x},t) = \mathrm{max} \left[c_F(\mathbf{x},t)
      - \left(\frac{n_A}{n_B}\right) c_G(\mathbf{x},t), \, 0
      \right] \\
    \label{Eqn:Fast_B}
    &c_B(\mathbf{x},t) = \left( \frac{n_B}{n_A} \right) \;
    \mathrm{max}\left[- c_F(\mathbf{x},t) +
      \left(\frac{n_A}{n_B} \right) c_G(\mathbf{x},t), \, 0 \right] \\
    \label{Eqn:Fast_C}
    &c_C(\mathbf{x},t) = \left( \frac{n_C}{n_A} \right) \;
    \left(c_F(\mathbf{x},t) - c_A(\mathbf{x},t) \right)
  \end{align}
\end{subequations}

Appendix provides details of the  non-negative finite element formulation used to generate the high-fidelity simulation data employed herein.
To summarize, the mathematical model considered 
in this paper makes the following assumptions: 
\begin{enumerate}[(i)]
\item Advection is neglected.
\item Non-reactive volumetric source
  is neglected for all chemical species.
\item The dispersion tensor is assumed 
to be the same for all the chemical 
species involved in the reaction.
\item Only the diffusive part of the
  flux is prescribed on the boundary.
\item The time-scale of the reaction
  is much faster than the time-scale
  associated with the dispersion process. 
\end{enumerate}

Some prior works \citep{vesselinov2018unsupervised,mudunuru2019mixing,ahmmed2020comparative} 
have used a similar bimolecular reaction model, and utilized unsupervised and supervised machine learning methods to perform data mining in the simulation data. Specifically, these works have discovered hidden features, estimated relative importance of reaction-transport model input parameters, and emulated key quantities of interest (e.g., degree of mixing, product yield, species decay). However, these works did not capture the spatial-temporal mixing patterns, which is the main uniqueness of this paper
\section{PROPOSED DEEP LEARNING MODELING FRAMEWORK}
\label{Sec:S3_DLRT_Framework}
In this section, we will present our proposed deep learning framework. 
The details on the neural architecture of our non-negative CNN-LSTM 
model will be outlined first. We will then describe the training 
process for ingesting reactive-transport simulation data to make 
predictions at future time-steps.

\subsection{Deep learning model architecture}
\label{SubSec:S3_DL_Model_Architecture}
It is well known that many of the traditional finite element formulations 
for the reactive-transport equations suffer from a deficiency: they could 
produce negative concentrations \citep{2017_Mudunuru_Nakshatrala_MAMS}.
For example, the standard single-field Galerkin formulation produces negative values and
spurious node-to-node oscillations for the primary variables in advection- and 
reaction-dominated diffusion equations. 
The negative concentrations are unphysical and can erode the solution accuracy in the entire domain. 
Additionally, if these erroneous numerical solutions are used to train deep learning models, then they could produce inaccurate forecasts. 
Therefore, ensuring non-negative concentrations in the ground truth and during DL-model training/predictions is essential.

The framework developed herein ensures non-negativity of the predicted concentrations through
careful selection of the trainable parameters.
Figure~\ref{Fig:DL_RT_Workflow} shows a pictorial description of the proposed deep learning model.
Figure~\ref{Fig:DL_RT_Workflow}(a) shows a schematic of the non-negative CNN-LSTM architecture, which is developed to forecast the evolution of spatial patterns in the concentration of product $C$.
The inputs to the convolutional layers are an image stack as shown in Figure~\ref{Fig:STSF_ConcC}.
The convolutional layers learn the underlying representations (e.g., feature maps) of the mixing process based on the observed patterns in the concentration, velocity field, and anisotropic dispersion at the previous time-steps.
The last dense layers along with the LSTM units predict the evolution of the reactive-mixing based on the learned patterns (e.g., features maps or representations) from convolutional layers.
Figure~\ref{Fig:DL_RT_Workflow}(b) shows a schematic of prediction at future-times based on the trained non-negative CNN-LSTM model.
Concentrations (i.e., ground truth data obtained from the non-negative FEM) at initial time-steps along with velocity field and dispersion tensor are used to train the deep learning model. 
As described in Figure~\ref{Fig:STSF_ConcC}, to forecast the product $C$ concentration at subsequent time-steps, we input these image stacks to the trained non-negative CNN-LSTM model.  

\textsf{Keras} and \textsf{Tensorflow} packages \citep{chollet2017deep} are used to build our CNN-LSTM model (see Figure~\ref{Fig:DL_RT_Workflow}(a)).
The convolutional neural layers in our model are trained to find a set of transformations that turn the input image stack into simpler and more general feature maps that represent the reactive-mixing process.
The convolution operator learned by a filter that is applied at every spatial point on the input image stack is a local linear combination of neighboring points of that input. 
Pooling and activation operations are convex operations.
These are applied to the convolution operations, which make our CNN a form of high-dimensional composition of spline operators \citep{balestriero2018spline}.

Following \citet{Goodfellow-et-al-2016}, the CNN layers in the proposed model can be mathematically described as follows: Let us assume $L^{r-1}$ is a $(r-1)$-th neuron layer of size $\mathcal{R} \times \mathcal{R}$ followed by a convolutional layer. 
A filter $F^{r-1}$ of size $\mathcal{M} \times \mathcal{M}$ is applied to $L^{r-1}$.
Then, our convolutional layer output will be of size $(\mathcal{N} - \mathcal{M} + 1) \times (\mathcal{N} - \mathcal{M} + 1)$.
This convolution operation is given as follows:
\begin{equation}
  \label{Eqn:Conv_2D}
  s_{ij} = \displaystyle \sum_{p=0}^{M-1} 
  \displaystyle \sum_{q=0}^{M-1} f_{pq} \, 
  l^{r-1}_{(i+p)(j+q)}  \quad \mathrm{where} \; \;  
  f_{pq} \in F^{r-1}, \; l^{r-1}_{ij} \in L^{r-1}
\end{equation}

Nonlinearity is then applied to $s_{ij}$ to get the next neuron layer $L^{r}$. 
This operation is given by $l^{r}_{ij} = \sigma(s_{ij})$, where $l^{r}_{ij} \in L^{r}$ and $\sigma$ is the sigmoid function.
There is no learning in the max-pooling layers.
The pooling operation simply take certain $k \times k$ region within the $(\mathcal{N} - \mathcal{M} + 1) \times (\mathcal{N} - \mathcal{M} + 1)$ and outputs a single value.
This value is the maximum in that $k \times k$ region. 
In our case, for a given neuron layer $L^{r}$, max-pooling operation will output a $\frac{(\mathcal{N} - \mathcal{M} + 1)}{k} \times \frac{(\mathcal{N} - \mathcal{M} + 1)}{k}$ 
layer. This is because each $k \times k$ matrix is reduced to just a single value through the max function.

The LSTM layers are primarily composed of three gates -- input gate, forget gate, and output gate.
Sigmoid activation functions are used in these gates to give non-negative values at different states.
Input gate tells us the new information to be stored in the LSTM cell state.
Forget gate tells us the features to be throw away or irrelevant.
The output gate provides the activation required to estimate the 
final output of the LSTM block at time-step $j$. Mathematically, 
the flow of information from these gates can be written as follows: 
\begin{subequations}
  \begin{align}
    \label{Eqn:LSTM_Cell_1}
    \mathbf{f}_j &= \sigma(\mathbf{W}_f \mathbf{z}_j + 
    \mathbf{U}_f \mathbf{h}_{j-1} + \mathbf{b}_f) \\
    \label{Eqn:LSTM_Cell_2}
    \mathbf{i}_j &= \sigma(\mathbf{W}_i \mathbf{z}_j + 
    \mathbf{U}_i \mathbf{h}_{j-1} + \mathbf{b}_i) \\
    \label{Eqn:LSTM_Cell_3}
    \mathbf{o}_j &= \sigma(\mathbf{W}_o \mathbf{z}_j + 
    \mathbf{U}_o \mathbf{h}_{j-1} + \mathbf{b}_o) \\
    \label{Eqn:LSTM_Cell_4}
    \tilde{\mathbf{a}}_j &= \tanh(\mathbf{W}_a \mathbf{z}_j + 
    \mathbf{U}_a \mathbf{h}_{j-1} + \mathbf{b}_a) \\
    \label{Eqn:LSTM_Cell_5}
    \mathbf{a}_j &= \mathbf{f}_j \odot \tilde{\mathbf{a}}_{j-1} + 
    \mathbf{i}_j \odot \tilde{\mathbf{a}}_j \\
    \label{Eqn:LSTM_Cell_6}
    \mathbf{h}_j &= \mathbf{o}_j \odot \tanh(\mathbf{a}_j)
  \end{align}
\end{subequations}
where $\odot$ denotes the Hadamard product \citep{horn2012matrix}.
At time-step $j = 0$, $\mathbf{a}_j = 0$ and $\mathbf{h}_j = 0$.
$\mathbf{z}_j$ is the input vector to the LSTM cell.
$\mathbf{h}_j$ is the hidden state vector, which is also known as output vector of the LSTM cell.
$\mathbf{f}_j$, $\mathbf{i}_j$, and $\mathbf{o}_j$ are the activation vectors for forget, input, and output gates, respectively.
$\tilde{\mathbf{a}}_j$ is the LSTM cell input activation vector.
$\mathbf{a}_j$ is the LSTM cell state vector.
$\mathbf{W}_{\alpha}$ and $\mathbf{U}_{\alpha}$, $\forall \alpha = f, i, o, \, \mathrm{and}, \, a$, are the weight matrices of the input and recurrent connections.
$\mathbf{b}_{\alpha}$ are the corresponding bias vectors.
$\mathbf{W}_{\alpha}$, $\mathbf{U}_{\alpha}$, and $\mathbf{b}_{\alpha}$ are learned during the training process.

Figure~\ref{Fig:STSF_ConcC} shows a pictorial description of the spatial-temporal sequence forecasting (STSF) of the reactive-transport data. 
STSF is performed using the non-negative CNN-LSTM model that is built on the proposed DL-based model architecture as shown in Figure~\ref{Fig:DL_RT_Workflow}. 

To summarize, our framework ingests data from zero to $N$-th time-step for training the CNN-LSTM model.
The ingested data is an image stack consisting of $c_C(x,y)$, $\mathrm{v}_{x}$, $\mathrm{v}_{y}$, $D_{xx}$, $D_{xy}$, and $D_{yy}$ images at $N$-th time-step. 
$c_C(x,y)$ is the concentration of product $C$ from high-resolution numerical simulation using the non-negative FEM (ground truth).
$\mathrm{v}_{x}$ and $\mathrm{v}_{y}$ are the $x$- and $y$-components of the vortex-based velocity fields given by Eqs.~\eqref{Eqn:Vel_x}--\eqref{Eqn:Vel_y}.
$D_{xx}$, $D_{xy}$, and $D_{yy}$ are the $xx$-, $xy$-, and $yy$-components of the anisotropic dispersion tensor given by Eq.~\ref{Eqn:Aniso_Diff_Tensor_Lit}.
The trained CNN-LSTM model uses these images to produce a sequence of predictions for $c_C(x,y)$ for time-steps greater than $N$.
We also note that the trained CNN-LSTM model constantly ingests data that it produces after $N$-th time-step to make forecasts until the end of simulation time.

\subsection{Non-negative CNN-LSTM model training}
\label{SubSec:S3_DL_Model_Training}
The architecture of our CNN consists of three convolutional neural layers with 16 filters.
We initialize filters with the Glorot uniform initializer, which is also called Xavier uniform initializer \citep{glorot2010understanding}.
The CNN layer weights are constrained to be non-negative and non-linear mapping is based on ReLU activation function \citep{nair2010rectified}.
This allows us to learn non-negative feature maps.
A max-pooling operation is applied after each convolution.
The final CNN layer is flattened and followed by a LSTM layer with 400 units.
This LSTM layer output is then fed to a fully connected layer of size 6561 that has a sigmoid activation function.
Note that the dense layer outputs the concentration at the next time-step and its weights are also constrained to be non-negative.
The output of the dense layer is then reshaped to a 2D concentration image of product $C$.
The entire model is compiled with an Adam optimizer \citep{kingma2014adam} with loss being the mean squared error.
The resulting model has a total of 3,298,785 trainable parameters.
Training our model is accomplished by backpropagation. 
The weights in the CNN-LSTM model are updated iteratively on batches of data from the training set.
The batch size is 6 and the training is terminated after 100 epochs. 
The proposed model is updated after seeing each batch by using gradient descent and backpropagation.
The weights are adjusted to minimize the mean squared error of the proposed deep learning model (i.e., through the gradient of the error with respect to the weights). Note that we performed sensitivity analysis with respect to model parameters like LSTM units, batch size and epochs to finalize these hyperparameters used to generate results presented in this paper.

\section{REPRESENTATIVE NUMERICAL RESULTS}
\label{Sec:S4_DLRT_NR}
In this section, we present numerical results to evaluate accuracy 
and efficiency of the proposed framework. We will first describe an 
initial boundary value problem that is used to generate data; this 
problem provides insights into reactive-mixing in subsurface under 
weakly chaotic flow fields (e.g., vortex-based structures). We then 
summarize the input parameters used to generate data and train the 
CNN-LSTM models. Next, we will describe in-detail on the accuracy 
of the proposed DL-based framework. Ground truth and CNN-LSTM model 
predictions are compared to show the predictive capability of these 
models. Finally, we will discuss the computational cost associated 
with training and testing the deep learning models.

\subsection{Reaction tank problem}
\label{SubSec:Reaction_Tank_Problem}
Figure~\ref{Fig:Problem_Description} provides a pictorial description of the initial boundary value problem. 
The computational domain is a square with side length $L = 1$.
Zero flux boundary conditions are enforced on the sides of the domain.
The non-reactive volumetric source $f_i(\mathbf{x}, t)$ is equal to zero for all the chemical species.
Initially species $A$ and species $B$ are segregated such that species $A$ is placed in the left half of the domain while species $B$ is placed in the right half.
The stoichiometric coefficients are selected as $n_A = n_B = n_C = 1$.
The total time of interest is selected as $\mathcal{I} = 1$.
The dispersion tensor is chosen from the subsurface literature \citep{Pinder_Celia} and is given as follows:
\begin{align}
  \label{Eqn:Aniso_Diff_Tensor_Lit}
  \mathbf{D}_{\mathrm{subsurface}}
  (\mathbf{x}) = D_{m} \mathbf{I} +
  \alpha_{T} \|\mathbf{v}\| \mathbf{I} +
  \frac{\alpha_L - \alpha_T}{\|\mathbf{v}\|}
  \mathbf{v} \otimes \mathbf{v}
\end{align}
where $D_m$ is the molecular diffusivity, $\alpha_L$ is the longitudinal diffusivity, $\alpha_T$ is the transverse diffusivity, $\mathbf{I}$ is the identity tensor, $\otimes$ is the tensor product, $\mathbf{v}$ is the velocity vector field, and $\| \bullet \|$ is the Frobenius norm.
As discussed in Sec.~\ref{SubSec:S2_FastRxns_Invs}, we have neglected advection.
A model velocity field is used to define the anisotropic dispersion tensor through the following stream function \citep{2002_Adrover_etal_CCE_v26_p125_p139,2009_Tsang_PRE_v80_p026305,2016_Mudunuru_Nakshatrala_JCP_v305_p448_p493}:
\begin{align}
  \label{Eqn:Div_Free_Stream_Function}
  \psi(\mathbf{x},t) =
  \begin{cases}
    \frac{1}{2 \pi \kappa_f} \left( \sin(2 \pi \kappa_f x)
    - \sin(2 \pi \kappa_f y) + v_0 \cos(2 \pi \kappa_f y)
    \right) &\quad \mathrm{if} \; \nu T \leq t < \left( \nu
    + \frac{1}{2} \right) T  \\
    \frac{1}{2 \pi \kappa_f} \left( \sin(2 \pi \kappa_f x)
    - \sin(2 \pi \kappa_f y) - v_0 \cos(2 \pi \kappa_f x)
    \right) &\quad \mathrm{if} \; \left( \nu + \frac{1}{2}
    \right) T \leq t < \left( \nu + 1 \right) T
  \end{cases}
\end{align}
where $\nu = 0, 1, 2, \cdots$ is an integer. $\kappa_fL$ 
and $T$ are characteristic scales of the flow field. 
Using Eq.~\eqref{Eqn:Div_Free_Stream_Function}, the 
divergence-free velocity field components are given 
as follows:

\begin{align}
  \label{Eqn:Vel_x}
  \mathrm{v}_{x}(\mathbf{x},t) = -\frac{\partial
  \psi}{\partial \mathrm{y}} =
  \begin{cases}
    \cos(2 \pi \kappa_f y) + v_0 \sin(2 \pi \kappa_f y)
    &\quad \mathrm{if} \; \nu T \leq t < \left( \nu +
    \frac{1}{2} \right) T  \\
    \cos(2 \pi \kappa_f y) &\quad \mathrm{if} \;
    \left( \nu + \frac{1}{2} \right) T \leq t <
    \left( \nu + 1 \right) T
  \end{cases}
\end{align}
\begin{align}
  \label{Eqn:Vel_y}
  \mathrm{v}_{y}(\mathbf{x},t) = +\frac{\partial
  \psi}{\partial \mathrm{x}} =
  \begin{cases}
    \cos(2 \pi \kappa_f x) &\quad \mathrm{if} \;
    \nu T \leq t < \left( \nu + \frac{1}{2} \right) T \\
    \cos(2 \pi \kappa_f x) + v_0 \sin(2 \pi \kappa_f x)
    &\quad \mathrm{if} \; \left( \nu + \frac{1}{2} \right)
    T \leq t < \left( \nu + 1 \right) T
  \end{cases}
\end{align}

\subsection{Data generation}
\label{SubSec:S4_Parameters}
A non-negative finite element method described in Sec.~\ref{Sec:Appendix} is used to generate high-resolution data for training and testing the DL-based framework.
Low-order triangular finite element mesh of size $81 \times 81$ with 6,561 degrees-of-freedom is employed for discretizing the domain.
Backward Euler time-stepping scheme with $\Delta t = 0.001$ is used to advance the simulation time from $t = 0.0$ to $t = 1.0$.
Note that in our previous works \citep{nakshatrala2013numerical,2016_Mudunuru_Nakshatrala_JCP_v305_p448_p493,2017_Mudunuru_Nakshatrala_MAMS}, we have performed $h$-convergence and time-step-refinement studies to solve a similar system of equations described in Sec.~\ref{Sec:S2_DLRT_GE}.
The outcome of these studies showed that $81 \times 81$ mesh with $\Delta t = 0.001$ was sufficient to achieve accurate numerical solutions to capture fine-scale spatial-temporal variations caused by anisotropic dispersion.

To demonstrate the applicability of the proposed DL-based framework, we have used a total of four high-resolution numerical simulations to train and test the non-negative CNN-LSTM models.
Each simulation has 1000 images of $c_C$, whose size is $81 \times 81$.
A detailed description of the simulation data (a total of 2315 realizations) generated for machine learning analysis is described in References \citep{vesselinov2018unsupervised,mudunuru2019mixing,ahmmed2020comparative}.
Herein, we use a subset of this massive dataset for testing our proposed framework.
The reaction-diffusion model input parameters associated with this subset are as follows:~$\kappa_fL$ = $\left[2, 3, 4, 5 \right]$, $v_0 = 10^{-1}$, $\frac{\alpha_L}{\alpha_T} = 10^{4}$, $D_m = 10^{-3}$, and $T = 10^{-4}$.
That is, $\kappa_fL$ is varied and other parameters (e.g., $v_0$, $\frac{\alpha_L}{\alpha_T}$, $D_m$, $T$) are kept constant. The resulting data is shown in Figure~\ref{Fig:DL_RT_Pred_kfL2}(a), Figure~\ref{Fig:DL_RT_Pred_kfL3}(a), Figure~\ref{Fig:DL_RT_Pred_kfL4}(a), Figure~\ref{Fig:DL_RT_Pred_kfL5}(a).
The above selected parameters represent reactive-mixing under high-anisotropy.
$\frac{\alpha_L}{\alpha_T} = 10^{4}$ corresponds to a scenario where longitudinal dispersion ($\alpha_\mathrm{L}$ along the streamlines) is $10^{4}$ times higher than transverse dispersion ($\alpha_\mathrm{T}$ across streamlines).
Higher values of $\kappa_fL$ enhance mixing and lower values results in regions/islands where species rarely mix. 
$v_0 = 10^{-1}$ results in very small perturbations in the flow field, thereby, preserving the underlying symmetry in the mixing patterns.
$D_m = 10^{-3}$ corresponds to reasonably high molecular diffusion.
$T = 10^{-4}$ corresponds to highly-oscillating flow field sampled at a frequency of 10 kHz.

The data generated based on the above parameters represents preferential and incomplete reactive-mixing.
The system always starts in the unmixed state.
But depending on the choice of the model parameters, it can result in either preferential mixing or incomplete mixing states.
Incomplete mixing corresponds to a parameter scenario where $\frac{\alpha_L}{\alpha_T} = 10^{4}$ with large-scale vortex structures (e.g., $k_fL \leq 3$; see Figure~\ref{Fig:DL_RT_Pred_kfL2}(a), Figure~\ref{Fig:DL_RT_Pred_kfL3}(a)).
This results in regions/islands (e.g., specifically near the center of the vortex) where product $C$ concentration is zero.
Preferential mixing represents a situation when $\frac{\alpha_L}{\alpha_T} = 10^{4}$ with small-scale vortex structures (e.g., $k_fL \geq 4$, Figure~\ref{Fig:DL_RT_Pred_kfL4}(a), Figure~\ref{Fig:DL_RT_Pred_kfL5}(a)).
In this case, the system is not uniformly mixed due to anisotropy and small-scale features in flow field facilitate mixing along a certain direction.
As a result, training and testing our CNN-LSTM models on this data allows us to see how well they predict different spatial-temporal patterns in mixing under high-anisotropy at late times. 
The reason to choose large anisotropic contrast data is because of its rich information content on the spatial-temporal evolution of reactive-mixing patterns (e.g., incomplete mixing, preferential mixing, interfacial mixing).
Such patterns are not formed under lower anisotropic contrast.
This is because the system tends to move towards uniform or well-mixed state (e.g., when $\frac{\alpha_L}{\alpha_T} \leq 100$) even in the early times \citep{ahmmed2020comparative,mudunuru2019mixing}.

\subsection{Accuracy of mixing patterns under the proposed DL-based framework}
\label{SubSec:S4_Accuracy}
In this subsection, we present detailed analysis and associated accuracy of the trained models.
Prediction accuracy and its metrics are focused on how well the proposed CNN-LSTM models forecast mixing patterns under the combined effects of high anisotropy and different characteristic scales of vortex structures.

Figure~\ref{Fig:DL_RT_Pred_kfL2} compares the ground truth and deep learning model predictions at $t = 1.0$.
The ground truth is based on reactive-transport simulation data generated for $\kappa_fL = 2$.
This value of $\kappa_fL$ correspond to reactive-mixing under large-scale vortex structures.
The proposed scenario also corresponds to incomplete mixing that occurs due to the combined effects of high anisotropy contrast along with large-scale features in velocity field.
Specifically, we can find a total of eight regions/islands in the entire domain (e.g., vortex centers), where progress of mixing and formation of product $C$ is very slow.
At the center of these islands, the value of $c_C(x,y)$ is equal to zero.
We analyze whether our trained CNN-LSTM models can capture such incomplete mixing patterns that occur due to large-scale features in the velocity field.

As mentioned in the prior section, FEM simulation data was obtained from from $t = 0.0$ to $t = 1.0$ for a time increment of 0.001. Figure~\ref{Fig:DL_RT_Pred_kfL2}(a) shows the ground truth at the end of simulation time (at $t = 1.0$).
Figures~\ref{Fig:DL_RT_Pred_kfL2}(b)--(m) show the predictions at $t = 1.0$ from the CNN-LSTM models trained using incrementally larger amount of training data as described in the individual captions. 
To elaborate, the forecast at $t = 1.0$ in Figure~\ref{Fig:DL_RT_Pred_kfL2}(b) is obtained using the CNN-LSTM model trained using the first 8\% (from $t = 0.001$ to $t = 0.08$) of the ground truth data.
The trained model then predicts $c_C(x,y)$ for the rest of the simulation time (i.e., $t = 0.09$, $0.10$, $\cdots$, $1.00$).
The CNN-LSTM model makes this forecast based on the procedure summarized in Figure~\ref{Fig:STSF_ConcC}.
It can be observed in Figure~\ref{Fig:DL_RT_Pred_kfL2} that the forecasts are generally better for higher amounts of training data.
A very accurate prediction is obtained when we use 88\% of the ground truth data.
For this scenario, based on Figure~\ref{Fig:DL_RT_Pred_kfL2_Errors}(k), the maximum possible prediction error is less than 4\% in the entire domain.
Moreover, the maximum possible prediction error to accurately capture interfacial mixing is less than 2\%.
However, qualitatively, we can see that non-negative CNN-LSTM model trained using ground truth data as low as 32\% is able to effectively capture the underlying mixing pattern (e.g., interfacial mixing). 

Figure~\ref{Fig:DL_RT_Pred_kfL2_Errors} compares the prediction errors in the entire domain at $t = 1.0$.
The prediction error is in percentage, which is calculated by taking the difference between the ground truth and the non-negative CNN-LSTM model prediction at every point in the domain.
Positive value of the error shows that the CNN-LSTM model under-predicts the true concentration while negative error shows over prediction.
From Figure~\ref{Fig:DL_RT_Pred_kfL2_Errors}, it is clear that there is a coherent spatial structure in the error.
It also shows where CNN-LSTM model fails to make an accurate prediction.
That is, qualitatively and quantitatively, the error sheds light on the difficulties faced by the CNN-LSTM model on predicting mixing patterns with different amounts of training data.
For instance, with 8\% training data, we can see that the maximum possible value for under predictions and over predictions is approximately 40\% and 16\%, respectively.
In this scenario, high-values of over prediction ($>30\%$) by the CNN-LSTM model are seen in the regions closer to the boundary of the domain (e.g., due to boundary effects).
Similarly, high-values of under prediction ($>10\%$) are seen near the mixing interface (e.g., due to local-scale features).
This is where the barrier is removed and species are allowed to interact and facilitate the progress of chemical reaction.
From Figures~\ref{Fig:DL_RT_Pred_kfL2_Errors}(a)--(l), it is evident that to capture the interfacial mixing (e.g., due to high anisotropy) within an error of 10\% (e.g., measured using the infinity norm), 32\% of ground truth data is sufficient.
To capture global-scale mixing features (e.g., resulting from molecular diffusion) within an error of 10\%, 64\% of ground truth data is needed.
With this amount of training data, the prediction error associated with capturing local-scale features near the interface is reduced by more than 50\%.
Specifically, with 64\% ground truth data, the prediction error to capture the interfacial mixing is less than 4\%.
This level of enhanced accuracy (e.g., using 32\%, 64\%, 88\% data) instills confidence in the proposed CNN-LSTM model architecture to make reasonably good forecasts in both early and late times of reactive-mixing.
With minimal data (between $32\%$ to $64\%$), we can capture various patterns and associated multi-scale features in incomplete mixing arising due to large-scale vortex structures.

Figure~\ref{Fig:DL_RT_Pred_kfL3} shows the ground truth and the trained CNN-LSTM model predictions for $\kappa_fL = 3$.
Similar to $\kappa_fL = 2$ scenario, we have large-scale vortex structures ($\approx 18$ vortices in the domain).
However, the spatial-scale of these vortices for $\kappa_fL = 3$ are smaller compared to $\kappa_fL = 2$.
As a result, the extent of incomplete mixing is reduced when compared to Figure~\ref{Fig:DL_RT_Pred_kfL2}(a).
This is because of the reduced size in the vortices, which improves the interaction between reactants.
As a result, we have a better mixing scenario compared to $\kappa_fL = 2$ even under high anisotropy.
Figure~\ref{Fig:DL_RT_Pred_kfL3}(a) has six regions/islands where progress of mixing and formation of product  $C$ is very slow.
These regions are located primarily near the boundary and at the corners of the domain (e.g., $(x,y) \approx (0.1, 0.25), (0.1, 0.6), (0.25, 0.1), (0.75, 0.9), (0.9, 0.75), (0.9, 0.4)$).
Compared to Figure~\ref{Fig:DL_RT_Pred_kfL2}(a), we have a lesser number of regions/islands where $c_C(x,y) = 0.0$.
Additionally, the interfacial mixing in Figure~\ref{Fig:DL_RT_Pred_kfL3}(a) is stronger than Figure~\ref{Fig:DL_RT_Pred_kfL2}(a).
This is because of larger number of vortices in the domain, which produce enhanced mixing.

Figure~\ref{Fig:DL_RT_Pred_kfL3_Errors} compares the prediction errors at the end of simulation time.
Similar to $\kappa_fL = 2$, it is evident that the model forecasts are better for higher amounts of training data.
The best and very accurate forecast is obtained when we use 80\% of the ground truth data (see Figure~\ref{Fig:DL_RT_Pred_kfL3}(k)).
Unfortunately, 88\% training data model perform poorly due to model over-fitting.
For 80\% training data, the maximum possible prediction error is less than 8\% in the entire domain.
This accuracy in prediction is greater than that of $\kappa_fL = 2$ scenario (e.g., see Figure~\ref{Fig:DL_RT_Pred_kfL2_Errors}(j)).
The maximum possible prediction error to accurately capture interfacial mixing is less than 3\% (double that of $\kappa_fL = 2$).
In a nutshell, qualitatively and quantitatively, we can see that non-negative CNN-LSTM model trained using ground truth data, which is greater than or equal to 72\%  is able to effectively capture various patterns for enhanced mixing.
This includes both interfacial mixing and mixing near the vortices.

Figure~\ref{Fig:DL_RT_Pred_kfL4} shows the mixing patterns under small-scale vortex structures ($\approx 32$ vortices in the domain).
In this scenario, we observe preferential mixing aligned approximately at an angle of $15^{\mathrm{o}}$ anti-clockwise to vertical barrier.
This preferential direction consists of four small-scale vortices, which allow the reactant to mix faster across the interface when compared to $\kappa_fL = 2$ and $3$.
As a result, concentration of product $C$ is non-zero in the entire domain.
However, we can see the formation of the product is not homogeneous 
in the entire domain despite the small-scale features in the velocity field. 
This spatial inhomogeneity of the product formation is because of high 
contrast in anisotropy, which results in preferential mixing patterns.
Qualitatively and quantitatively, the predictions shown in Figures~\ref{Fig:DL_RT_Pred_kfL4}--\ref{Fig:DL_RT_Pred_kfL4_Errors} shed light on how the CNN-LSTM model is able to capture different mixing patterns.
For example, 72--80\% data is reasonably sufficient to capture product formation with 10\% error in the entire domain.
For this level of input data for training and prediction, we can accurately capture the interfacial mixing within 3--4\%.

Figure~\ref{Fig:DL_RT_Pred_kfL5} compares the ground truth and predictions for $\kappa_fL = 5$.
This value of $\kappa_fL$ corresponds to preferential mixing under fine-scale features in the velocity field ($\approx 50$ vortices in the domain).
The angle of preferential mixing is approximately 20--$23^{\mathrm{o}}$ anti-clockwise to vertical barrier.
In this direction, the maximum possible plume width (e.g., the region in the domain that has $c_C(x,y) \geq 0.45$) is approximately 0.2--0.25 times the length of the domain.
Based on Figures~\ref{Fig:DL_RT_Pred_kfL5}(i)--(k), the CNN-LSTM model prediction based on 64--80\% ground truth data is able to capture most of the characteristics of this plume (e.g., plume width, angle of preferential mixing).
Figures~\ref{Fig:DL_RT_Pred_kfL5_Errors}(h)--(j) show that the associated prediction errors (e.g., measured in infinity norm) are less than 5\%.
This accuracy instill confidence in the predictive capability of our models to forecast mixing patterns at late times under the combined effect of high anisotropy and fine-scale features in the velocity field.

\subsection{Computational cost}
\label{SubSec:S4_Comp_Cost}
The computational cost to run a single high-fidelity numerical 
simulation is approximately 26 mins (1560 seconds) on a single 
core and 18 mins (1080 seconds) on a eight core processor 
(Intel(R)Xeon(R) CPU E5-2695 v4 2.10GHz) \citep{mudunuru2019mixing}. 
The simulation dataset was developed using 
high-performance computing (HPC) resources 
\citep{vesselinov2018unsupervised,ahmmed2020comparative,mudunuru2019mixing}.
However, access to leadership class supercomputing resources 
(e.g., NERSC \citep{NERSC2021}, ALCF \citep{ALCF2021}, OLCF \citep{OLCF2021}) 
to researchers is generally limited. Additionally, even after having access 
to such HPC systems, there are various challenges and bottlenecks to generate 
data for longer-periods of simulation times. This includes availability of 
compute time to users (e.g., nodes, machine hours), waiting time in job submission queue, and cap on the maximum amount of wall clock time to run on a supercomputer. 
For example, at NERSC user facility, one has to restart simulations every two days as the cap to run on these machines is 48 hours in a regular queue.
Due to this bottleneck, simulations need to be restarted a lot of times to generate data at future time-steps.
As a result, the overall wall clock time to generate simulation data for multiple realizations can be in the order of weeks to months. 
Our DL-based framework provides an attractive way to overcome these challenges.
Specifically, this computational cost can be avoided by training the proposed DL models on partially generated data to forecast spatial-temporal variables at future times.

We have trained the proposed CNN-LSTM models on a MacBook Pro Laptop (2GHz 2-Core Intel i7 CPU, 8GB DDR4 RAM).
The associated computational cost to train the CNN-LSTM models and make a prediction is shown in Table~\ref{Tab:WCT_Time}.
From this table, it is evident that model training is computationally expensive as the wall clock times are in the $\mathcal{O}(10^{3})$ to $\mathcal{O}(10^{4})$ seconds.
This is expected as CPUs alone (with low-end memory) are really slow for training deep learning models. 
However, once the CNN-LSTM model is developed, the wall clock time to make an inference (i.e., predicting the spatial concentration of product $C$ at the next time-step) is very low.
That is, the inference time on this laptop is in the order of $\mathcal{O}(10^{-1})$ seconds.
This shows that our proposed CNN-LSTM model is fast in addition to its predictive capability.
Our future work involves accelerating the training process using Google Cloud Platform's GPUs and TPUs \citep{bisong2019google}.

{\small
\begin{table}[htbp]
  \centering
  \caption{Training wall clock time (TWCT) and prediction wall clock time (PWCT) for the proposed non-negative CNN-LSTM models.
  The reported TWCT and PWCT are in seconds.
  TWCT is the amount of time taken to ingest the simulation data and perform the training process.
  PWCT is the average amount of time taken to make a spatial prediction of the concentration field per time-step for all the future times.
  As mentioned in Sec.~\ref{SubSec:S3_DL_Model_Training}, CNN-LSTM model are trained using 100 epochs with a batch size of 6.}
  \begin{tabular}{|c|c|c|c|c|c|}\hline
    \textbf{Training} & \multicolumn{4}{|c|}{\textbf{TWCT}} & \textbf{Avg. PWCT} \\ 
    \cline{2-5} 
    \textbf{data}&\textbf{$\kappa_fL=2$}&\textbf{$\kappa_fL=3$}&\textbf{$\kappa_fL=4$}&\textbf{$\kappa_fL=5$}&\textbf{(all $\kappa_fL$)}\\
    \hline\hline
    8\% & 3239 & 3239 & 3138 & 3037 & 0.1325 \\\hline
    16\% & 6339 & 6037 & 6037 & 6037 & 0.1326 \\\hline  
    24\% & 8937 & 9037 &  9238 & 8937 & 0.1327 \\\hline  
    32\% & 11937 & 11837 & 11836 & 11937 & 0.1330 \\\hline
    40\% & 14837 & 14837 & 14837 & 15137 & 0.1333 \\\hline
    48\% & 17837 & 17736 & 17736 & 17837 & 0.1334 \\\hline
    56\% & 20736 & 20636 & 21037 & 20636 & 0.1335 \\\hline
    64\% & 23626 & 23536 & 23636 & 23737 & 0.1343 \\\hline
    72\% & 26737 & 26436 & 26636 & 25835 & 0.1346 \\\hline
    80\% & 29737 & 29737 & 29837 & 29436 & 0.1365 \\\hline 
    88\% & 32536 & 32436 & 32436 & 32436 & 0.1395 \\\hline
    96\% & 35436 & 35336 & 35236 & 35536 & 0.1400 \\\hline
  \end{tabular}
  \label{Tab:WCT_Time}
\end{table}}

Finally, we note that it is straightforward to perform similar analysis for other realizations (e.g., for different parameters in reaction-diffusion model). 
This can be achieved by re-training the proposed non-negative CNN-LSTM models with minimal data on other input parameters. 
The re-training process can be minimal as the weights of the convolutional layers that create the feature maps are not updated.
This is because these CNN layers learned the underlying representation of the reactive-transport process through the proposed training approach.
Specifically, minimal re-training can be performed through transfer learning techniques, where weights of the dense layer connected to the output are only updated.
One should expect similar outcomes as described in this study. 

Below we summarize our results
\begin{itemize}
  \item Our model's prediction accuracy generally increases with the amount of training data.
  \item However, in certain training scenarios, the prediction accuracy is reduced with larger training data.
    This poor performance can be attributed to model over-fitting.
  \item The concentration profile in the interfacial mixing zones is captured with reasonably good accuracy by our non-negative CNN-LSTM model trained using 32\% of the simulation data.   
  \item The high concentration zones are accurately captured within 10\% error (measured in infinity norm) by training our model using 64\% of the simulation data.
  \item The wall clock time to make prediction/inference is $\approx \mathcal{O}(-5)$ to that of running a high-fidelity simulation.
\end{itemize}

\subsection{A potential application}
\label{SubSec:Sec4_Discussion}
Based on the accuracy of the results presented in Sec.~\ref{SubSec:S4_Accuracy}--\ref{SubSec:S4_Comp_Cost}, the proposed DL framework will be an ideal candidate 
for modeling virus transport. Several prior studies 
have used reactive-transport 
for pandemic modeling. Some concrete example include: predict the 
evolution of COVID-19 \citep{viguerie2020diffusion,jha2020bayesian}, and 
capture the transport of microbial pathogens in the terrestrial subsurface (primarily at 
aquatic interfaces) 
\citep{harvey2007transport,hansenterrestrial,bradford2013transport}. 
In the said applications, accurate forecasts of reactive-transport are essential because fluctuating flow regimes (e.g., water table fluctuations, flow fields given by Eqns.~\eqref{Eqn:Vel_x}--\eqref{Eqn:Vel_y}) can enhance virus detachment from solid-water interfaces and air-water interfaces \citep{zhang2012modeling}. 
These flow fields can result in regions (e.g., interfacial mixing areas as shown in Figures~\ref{Fig:DL_RT_Pred_kfL2}, \ref{Fig:DL_RT_Pred_kfL3}, \ref{Fig:DL_RT_Pred_kfL4}, \ref{Fig:DL_RT_Pred_kfL5}) that have high concentrations in drinking water resources, which the 
cited references did not consider. Our DL framework can be applied to solve such problems. 

The non-negative 
CNN-LSTM models would be able to accurately model the spatial-temporal 
evolution of an epidemic such as COVID-19 
and virus transport in porous media. The DL-based model can be trained 
using the initial few days of 
epidemic evolution over a region of interest. The trained model 
can then be used to predict the future evolution of the epidemic. 
This would provide very significant information to the policymakers 
in terms of when and where the epidemic infections 
would be reaching the critical level. This information would also 
help in planning the allocation of resources and in taking early 
actions to suppress the spread of the disease (e.g., lockdown, travel ban).
\section{CONCLUDING REMARKS}
\label{Sec:S5_DLRT_CR}
There is a need for fast, reliable, and physically 
meaningful forecasts of the concentrations of 
chemical species and the evolution of reactive-mixing for diverse engineering
applications such as combustion and pandemic modeling. 
However, high-resolution numerical simulations mostly use the initial and boundary values and cannot utilize any intermediate temporal data to improve accuracy of the predictions. Secondly, high-resolution numerical simulations are often 
expensive, as each simulation may take several hours or 
days to provide spatial-temporal patterns of mixing, 
especially for problems with large spatial and temporal domains.
To address the said need, we have presented a fast and predictive 
framework that can forecast the spatial-temporal evolution of 
reactive-transport using minimal data. Our approach leverages 
recent advances in deep learning techniques---non-negative 
constrained CNNs for capturing spatial spreading and LSTMs 
for temporal evolution. The framework was used to develop 
a new non-negative model, which elegantly overcomes the issue of spurious
negative concentration predictions by traditional numerical methods like FEM.

The DL-based model developed herein needs to be trained using 
only a subset of the data for the total temporal domain of interest. 
The data can be obtained using high-fidelity simulations or experiments.
Our results demonstrate that less than 40\% of the simulation 
data is needed to capture the interfacial mixing (e.g., 
local-scale mixing features due to combined effects of anisotropy 
and vortex structures) within an error of 10\% (as measured using 
the infinity norm). To capture global features at the same level of 
accuracy, less than 70--80\% data is needed. The time needed for a 
forecast is a fraction ($\approx \mathcal{O}(-6)$) of the time needed 
for training. Also, the ratio of the solution time for the proposed deep 
learning framework to that for the high-fidelity simulation 
used to generate data for training the model is $\approx \mathcal{O}(-5)$. 
Thus, the accuracy of the predictions combined with the fast inference 
makes our non-negative CNN-LSTM models attractive for real-time forecasting 
of species concentration in reactive transport problems for a wide range
of engineering applications.   

A plausible future work can be towards accelerating the 
training process of the proposed deep learning models using 
graphical processing units (GPUs) and tensor processing units 
(TPUs) \citep{bisong2019google}. Distributed training (e.g., 
using \textsf{Horovod} \citep{sergeev2018horovod}) and scalable 
machine learning (e.g., using \textsf{DeepHyper} 
\citep{balaprakash2018deephyper}) provide viable routes to 
achieve speedup by splitting the workload to train the 
non-negative CNN-LSTM model among multiple processors (i.e., 
worker nodes). 


\appendix

\section{GENERATION OF NON-NEGATIVE SOLUTIONS}
\label{Sec:Appendix}
For completeness, this section provides details on the 
non-negative single-field formulation used to generate 
data for training and testing the CNN-LSTM models. 

Let the total time interval $[0, \mathcal{I}]$ be discretized into $N$ non-overlapping sub-intervals:
\begin{align}
  [0, \mathcal{I}] = \bigcup_{n = 1}^{N} [t_{n-1}, t_{n}]
\end{align}
where $t_0 = 0$ and $t_N = \mathcal{I}$.
Assuming a uniform time step $\Delta t$, we define:
\begin{align}
  c^{(n)}_F(\mathbf{x}) := c_F(\mathbf{x},t_n)
  \quad \mathrm{and} \quad 
  c^{(n)}_G(\mathbf{x}) := c_G(\mathbf{x},t_n)
\end{align}

Following \citep{nakshatrala2016numerical}, we will first discretize 
with respect to time the governing equations for the invariants; we 
will use the backward Euler time-stepping scheme. The resulting 
time-discrete equations corresponding to 
Eqs.~\eqref{Eqn:Diffusion_for_F}--\eqref{Eqn:Diffusion_for_IC_F} are:
\begin{subequations}
  \begin{alignat}{2}
    \label{Eqn:Decay_Diffusion_for_F}
    &\left(\frac{1}{\Delta t} \right) c^{(n + 1)}_F(\mathbf{x})
    - \mathrm{div} \left[\mathbf{D}(\mathbf{x}) \, \mathrm{grad}
    \left[c^{(n + 1)}_F(\mathbf{x}) \right] \right] =
    f_F(\mathbf{x},t_{n + 1}) + \left( \frac{1}{\Delta t} \right)
    c^{(n)}_F(\mathbf{x}) 
    &&\quad \mathrm{in} \; \Omega \\
    \label{Eqn:Decay_Diffusion_for_Dirchlet_F}
    &c^{(n + 1)}_F(\mathbf{x}) = c_F^{\mathrm{p}}(\mathbf{x},
    t_{n + 1}) := c^{\mathrm{p}}_A(\mathbf{x},t_{n + 1}) +
    \left( \frac{n_A}{n_C} \right) c^{\mathrm{p}}_C(\mathbf{x},t_{n + 1})
    &&\quad \mathrm{on} \; \Gamma^{\mathrm{D}} \\
    \label{Eqn:Decay_Diffusion_for_Neumann_F}
    &\mathbf{n}(\mathbf{x}) \bullet \mathbf{D} (\mathbf{x}) \,
    \mathrm{grad} \left[c^{(n + 1)}_F(\mathbf{x}) \right] =
    h^{\mathrm{p}}_F(\mathbf{x},t_{n + 1}) :=
    h^{\mathrm{p}}_A(\mathbf{x},t_{n + 1}) + \left( \frac{n_A}{n_C}
    \right) h^{\mathrm{p}}_C(\mathbf{x},t_{n + 1}) 
    &&\quad \mathrm{on}\; \Gamma^{\mathrm{N}} \\
    \label{Eqn:Decay_Diffusion_for_IC_F}
    &c_F(\mathbf{x},t_0) = c^{0}_F(\mathbf{x}) :=
    c^{0}_A(\mathbf{x}) + \left( \frac{n_A}{n_C} \right)
    c^{0}_C(\mathbf{x}) 
    &&\quad \mathrm{in} \; \overline{\Omega}
  \end{alignat}
\end{subequations}

Note that one can extend the proposed framework to other time-stepping schemes by following the non-negative procedures given in \citep{nakshatrala2016numerical}.
The standard finite element Galerkin formulation for Eqs.~\eqref{Eqn:Decay_Diffusion_for_F}--\eqref{Eqn:Decay_Diffusion_for_IC_F} is given as follows:~Find $c^{(n + 1)}_F(\mathbf{x}) \in \mathcal{C}^t_F$ with
\begin{align}
  \mathcal{C}^t_F &:= \left\{c_F(\bullet ,t) \in H^{1}(\Omega) \; \big| \;
  c_F(\mathbf{x},t) = c^{\mathrm{p}}_F(\mathbf{x},t) \; \mathrm{on} \;
  \Gamma^{\mathrm{D}}\right\}
\end{align}
such that we have
\begin{align}
  \label{Eqn:Transient_single_field_formulation_diffusion}
  \mathcal{B}^{t} \left(w;c^{(n + 1)}_F \right) = L^{t}_{F}(w)
  \quad \forall w(\mathbf{x}) \in \mathcal{W}
\end{align}
where the bilinear form and linear functional are, respectively, defined as
\begin{subequations}
  \begin{align}
    \mathcal{B}^{t} \left(w;c^{(n + 1)}_F \right) &:=
    \frac{1}{\Delta t} \int \limits_{\Omega} w(\mathbf{x}) \; c^{(n + 1)}_F(\mathbf{x}) \;
    \mathrm{d} \Omega + \int \limits_{\Omega} \mathrm{grad}[w(\mathbf{x})]
    \bullet \mathbf{D}(\mathbf{x}) \; \mathrm{grad}[c^{(n + 1)}_F(\mathbf{x})]
    \; \mathrm{d} \Omega \\
    L^{t}_{F}(w) &:= \frac{1}{\Delta t}\int \limits_{\Omega}  w(\mathbf{x}) \;
    c^{(n)}_F(\mathbf{x}) \; \mathrm{d} \Omega +
    \int \limits_{\Gamma^{\mathrm{N}}} w(\mathbf{x}) \;
    h^{\mathrm{p}}_F(\mathbf{x},t_{n + 1}) \; \mathrm{d} \Gamma
  \end{align}
\end{subequations}
and
\begin{align}
  \mathcal{W} &:= \left\{w(\mathbf{x}) \in H^{1}(\Omega) \; \big| \;
  w(\mathbf{x}) = 0 \; \mathrm{on} \;
  \Gamma^{\mathrm{D}}\right\}
\end{align}
A similar formulation can be written for the invariant $c_G$.

The Galerkin formulation given by Eq.~\eqref{Eqn:Transient_single_field_formulation_diffusion} can be re-written as a minimization problem as follows:
\begin{align}
  \label{Eqn:Minimzation_Problem_Statement_DiffusionDecay}
  \mathop{\mathrm{minimize}}_{c^{(n + 1)}_F(\mathbf{x}) \in
  \mathcal{C}^t_F} & \quad \frac{1}{2} \mathcal{B}^{t}
  \left(c^{(n + 1)}_F; c^{(n + 1)}_F \right) - L^t_{F}
  \left(c^{(n + 1)}_F \right)
\end{align}
However, it has been shown that the Galerkin formulation leads to negative concentrations for 
product $C$ \citep{nakshatrala2013numerical}, which is unphysical. Instead, we will solve the 
minimization problem Eq.~\eqref{Eqn:Minimzation_Problem_Statement_DiffusionDecay} with 
the constraint that $c_F$ is non-negative; that is,
\begin{align}
    c_F \geq 0
\end{align}
A similar minimization problem with non-negative constraint for $c_G$ can be written as follows:
\begin{subequations}
  \begin{align}
    \label{Eqn:Minimzation_Problem_Statement_DiffusionDecay_G}
    \mathop{\mathrm{minimize}}_{c^{(n + 1)}_G(\mathbf{x}) \in
    \mathcal{C}^t_G} & \quad \frac{1}{2} \mathcal{B}^{t}
    \left(c^{(n + 1)}_G; c^{(n + 1)}_G \right) - L^t_{G}
    \left(c^{(n + 1)}_G \right) \\
    \mbox{subject to} &\quad c_G \geq 0
  \end{align}
\end{subequations}

Upon discretizing Eqs.~\eqref{Eqn:Minimzation_Problem_Statement_DiffusionDecay}--\eqref{Eqn:Minimzation_Problem_Statement_DiffusionDecay_G} using low-order finite elements, one arrives at:
\begin{subequations}
  \label{Eqn:NonNegative_Solver_Invariant_F_DiffusionDecay}
    \begin{align}
      \mathop{\mbox{minimize}}_{\mathbf{c}^{(n + 1)}_F \in
      \mathbb{R}^{ndofs}} & \quad \frac{1}{2}  \left \langle
      \mathbf{c}^{(n + 1)}_F; {\mathbf{K}}
      \mathbf{c}^{(n + 1)}_F \right \rangle -
      \frac{1}{\Delta t}  \left \langle
      \mathbf{c}^{(n + 1)}_F; \mathbf{c}^{(n)}_F
      \right \rangle \\
      \mbox{subject to} & \quad \mathbf{0}
      \preceq \mathbf{c}^{(n + 1)}_F  \\
      \mathop{\mbox{minimize}}_{\mathbf{c}^{(n + 1)}_G \in
      \mathbb{R}^{ndofs}} & \quad \frac{1}{2}  \left \langle
      \mathbf{c}^{(n + 1)}_G; {\mathbf{K}}
      \mathbf{c}^{(n + 1)}_G \right \rangle -
      \frac{1}{\Delta t}  \left \langle
      \mathbf{c}^{(n + 1)}_G; \mathbf{c}^{(n)}_G
      \right \rangle \\
      \mbox{subject to} & \quad \mathbf{0}
      \preceq \mathbf{c}^{(n + 1)}_G
    \end{align}
\end{subequations}
where $\langle \bullet; \bullet \rangle$ represents the standard 
inner-product on Euclidean spaces, and ``$ndofs$" denotes the 
(nodal) degrees-of-freedom. $\mathbf{c}^{(n + 1)}_F$ and 
$\mathbf{c}^{(n + 1)}_G$ are the nodal concentration vectors 
for the invariant $c_F$ and $c_G$ at time level $t_{n + 1}$, 
respectively. The coefficient matrix ${\mathbf{K}}$ is positive 
definite, and hence a unique global minimizer exists 
\citep{nagarajan2011enforcing}.

%


\section*{DATASETS}
The datasets used in this study were developed as a part 
previously published study \citep{vesselinov2018unsupervised,
ahmmed2020comparative,mudunuru2019mixing}; these works are 
cited and acknowledged in this paper. 

\section*{ACKNOWLEDGMENTS}
MKM is supported by ExaSheds project during the paper writing 
process, which was supported by the U.S. Department of Energy, 
Office of Science, under Award Number DE-AC02-05CH11231. 
KBN acknowledges the support from the University of Houston 
through High Priority Area Research Seed Grant. 

\bibliographystyle{plainnat}
\bibliography{Master_References/DL_RT_references}


\begin{figure}
  \centering
  \subfigure[Schematic of non-negative CNN-LSTM architecture for spatio-temporal sequence forecasting]
    {\includegraphics[width = 0.95\textwidth]
    {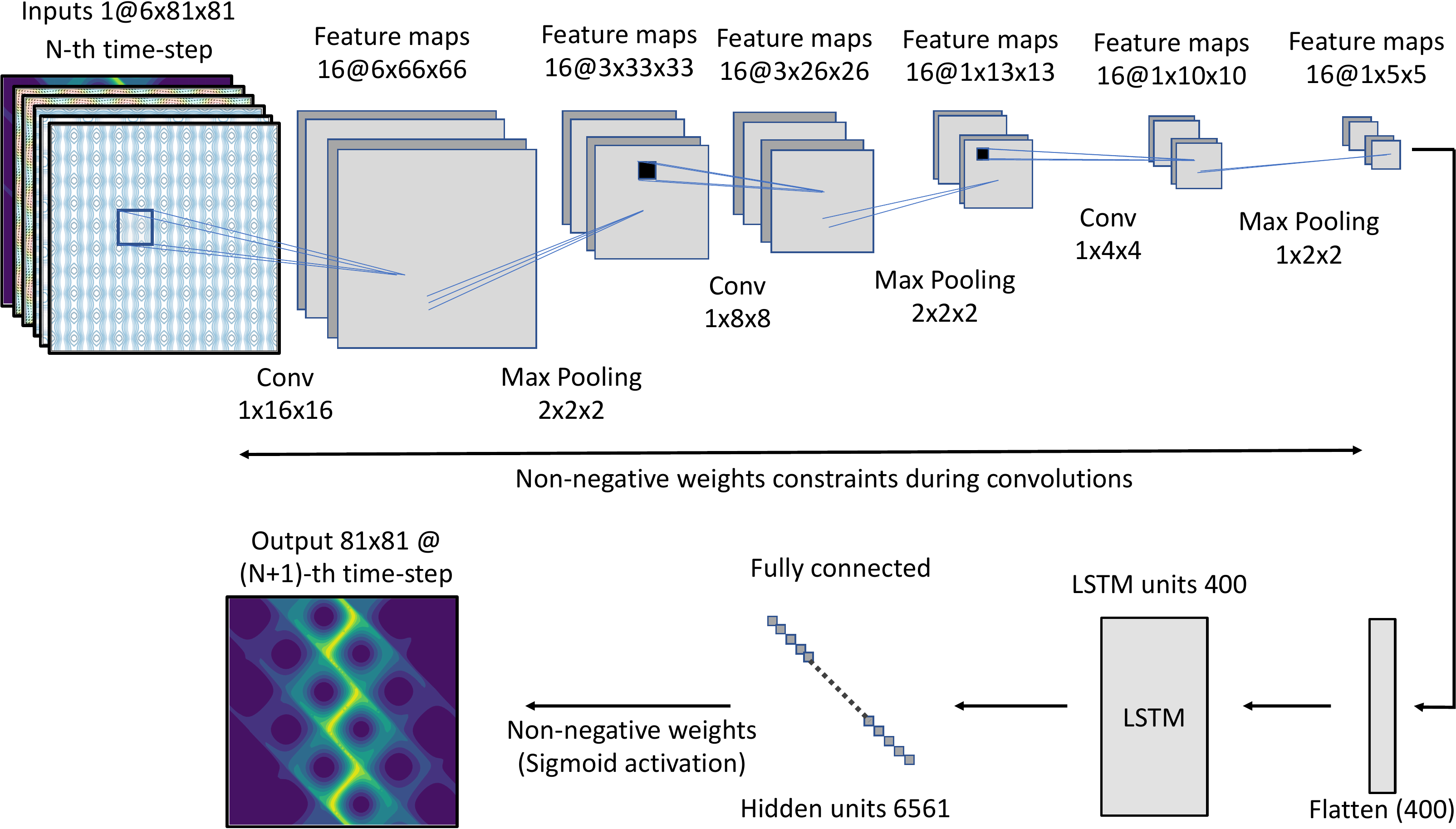}}
  \subfigure[RT predictions at future time-steps based on a trained non-negative CNN-LSTM model]
    {\includegraphics[width = 0.95\textwidth]
    {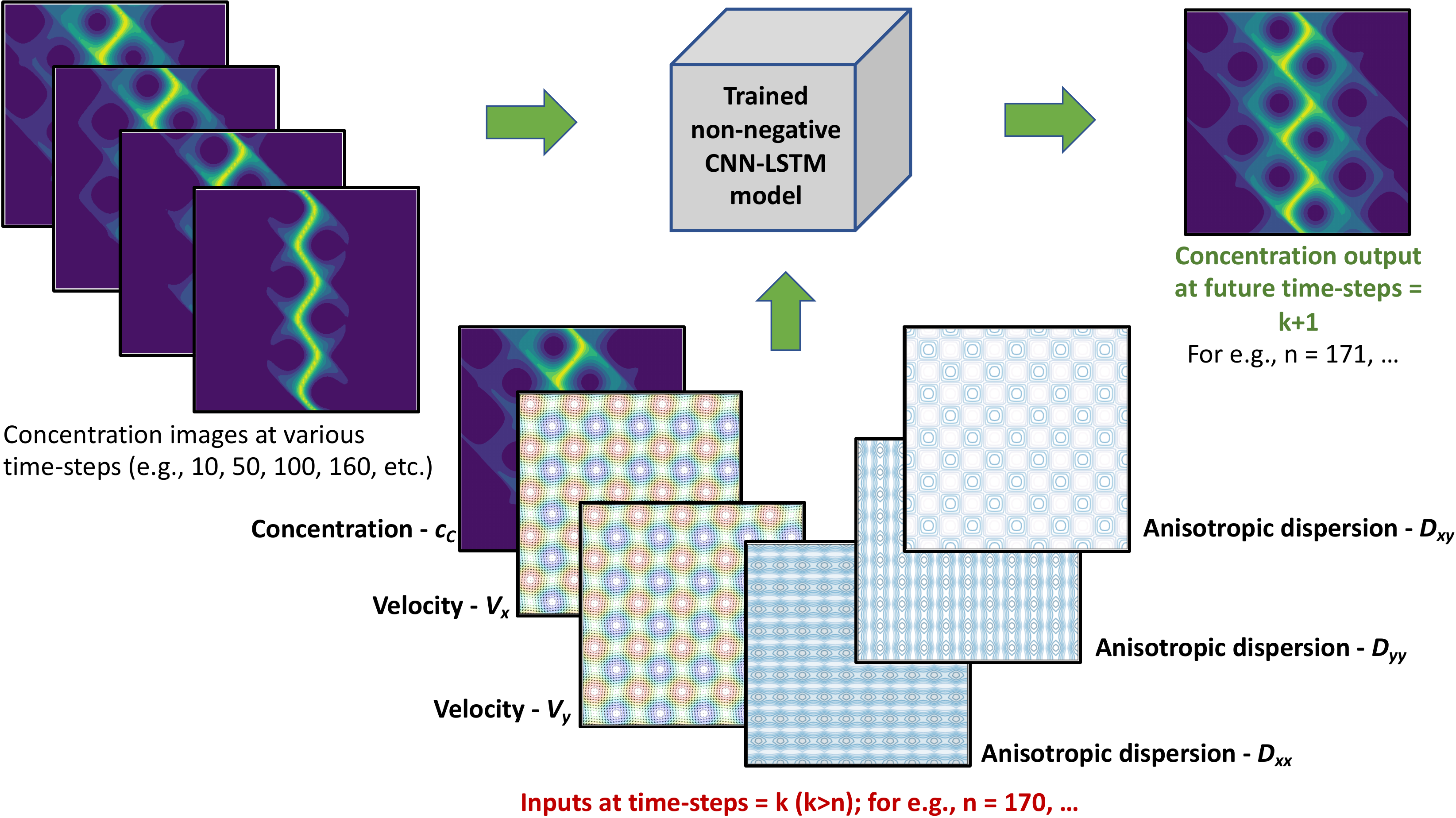}}
  \vspace{-0.15in}
  \caption{\textbf{Proposed deep learning modeling framework:}~A pictorial description of the proposed DL-based framework.
  The top figure shows the non-negative CNN-LSTM architecture and the bottom figure shows the predictions at the future time-steps based on the trained model.   
  \label{Fig:DL_RT_Workflow}}
\end{figure}

\begin{figure}
  \centering
  \includegraphics[width = \textwidth]{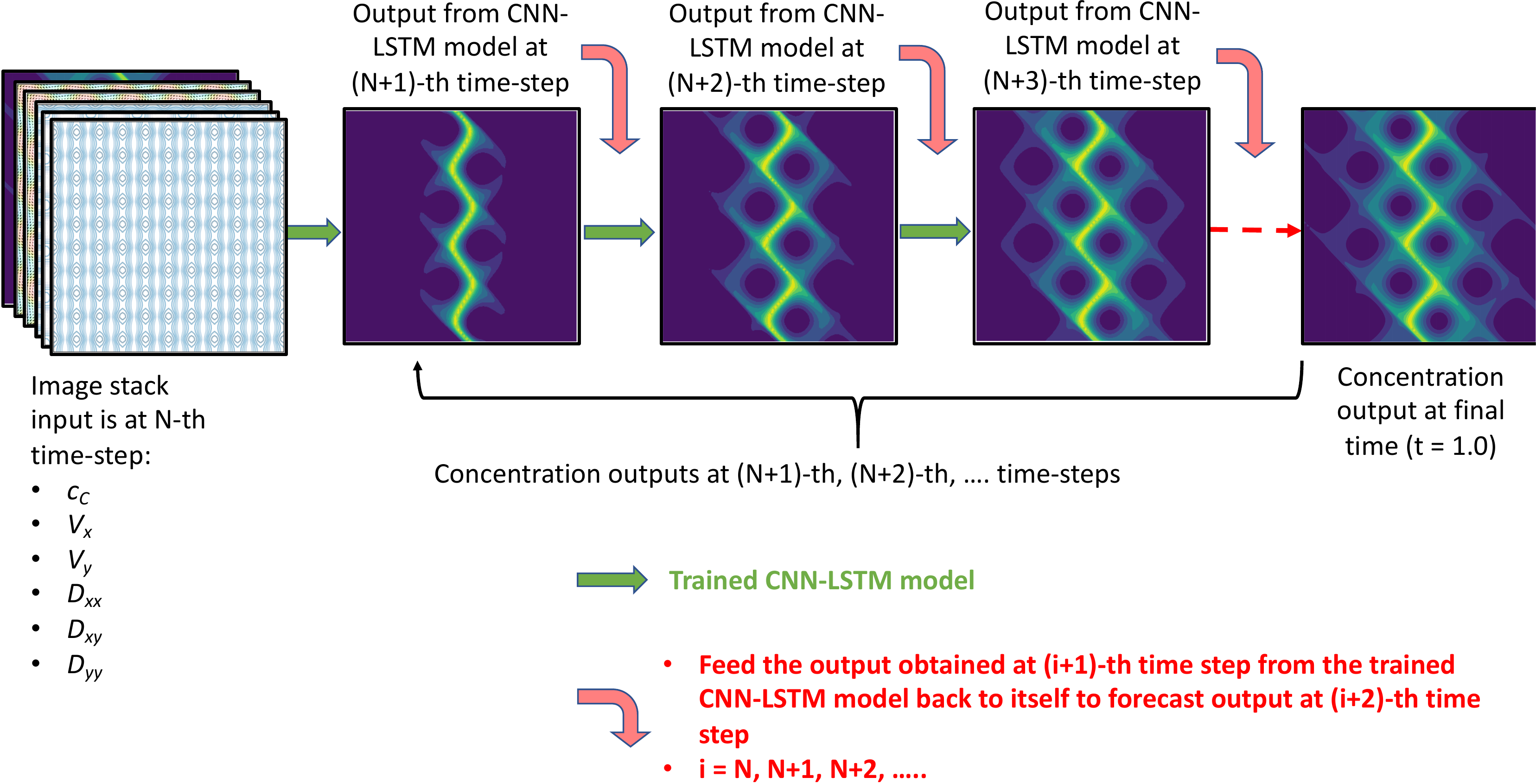}
  \caption{\textbf{Summary of forecasting product concentrations:}~A pictorial description of spatio-temporal sequence forecasting of the reactive-transport data using the trained non-negative CNN-LSTM model.
  The model is trained for the first $N$ time-steps and predictions are made from the $N+1$-th time-step till the end of simulation time. 
  \label{Fig:STSF_ConcC}}
\end{figure}

\begin{figure}
  \centering
  \includegraphics[width=0.6\textwidth]{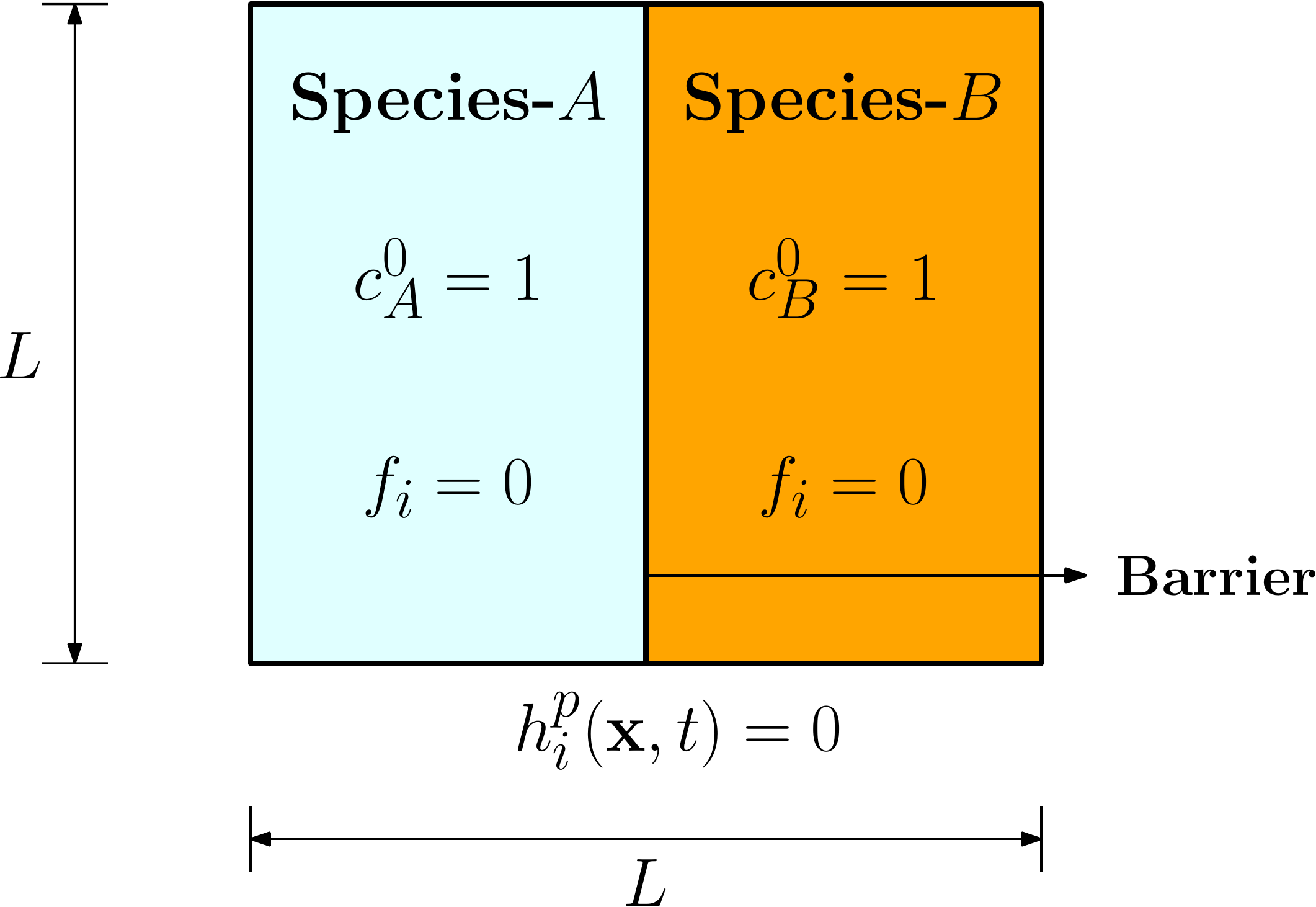}
  \caption{\textbf{Problem description:}~A pictorial
    description of the initial boundary value problem.
    The figure shows the model domain for generating
    reactive-transport simulation data. The length of
    the domain `$L$' is assumed to be equal to 1. 
    Zero flux boundary conditions are enforced on all sides of the domain 
    ($h^{\mathrm{p}}_i(\boldsymbol{x},t) = 0.0$).
    Focus is on fast irreversible bimolecular reactions, where species $A$ and $B$ are initially separated by a barrier.
    They are on the left and right sides of the domain, respectively.
    Initial concentration of $A$ and $B$ are equal to 1.0.
    The species are allowed to mix and react after $t > 0$ according to the velocity field given by Eqs.~\eqref{Eqn:Vel_x}--\eqref{Eqn:Vel_y}. 
    The non-reactive volumetric sources for all the chemical species 
    are assumed to be equal to zero. 
  \label{Fig:Problem_Description}}
\end{figure}

\begin{figure}
  \centering
  \subfigure[Ground truth at $t = 1.0$]
    {\includegraphics[width = 0.285\textwidth]
    {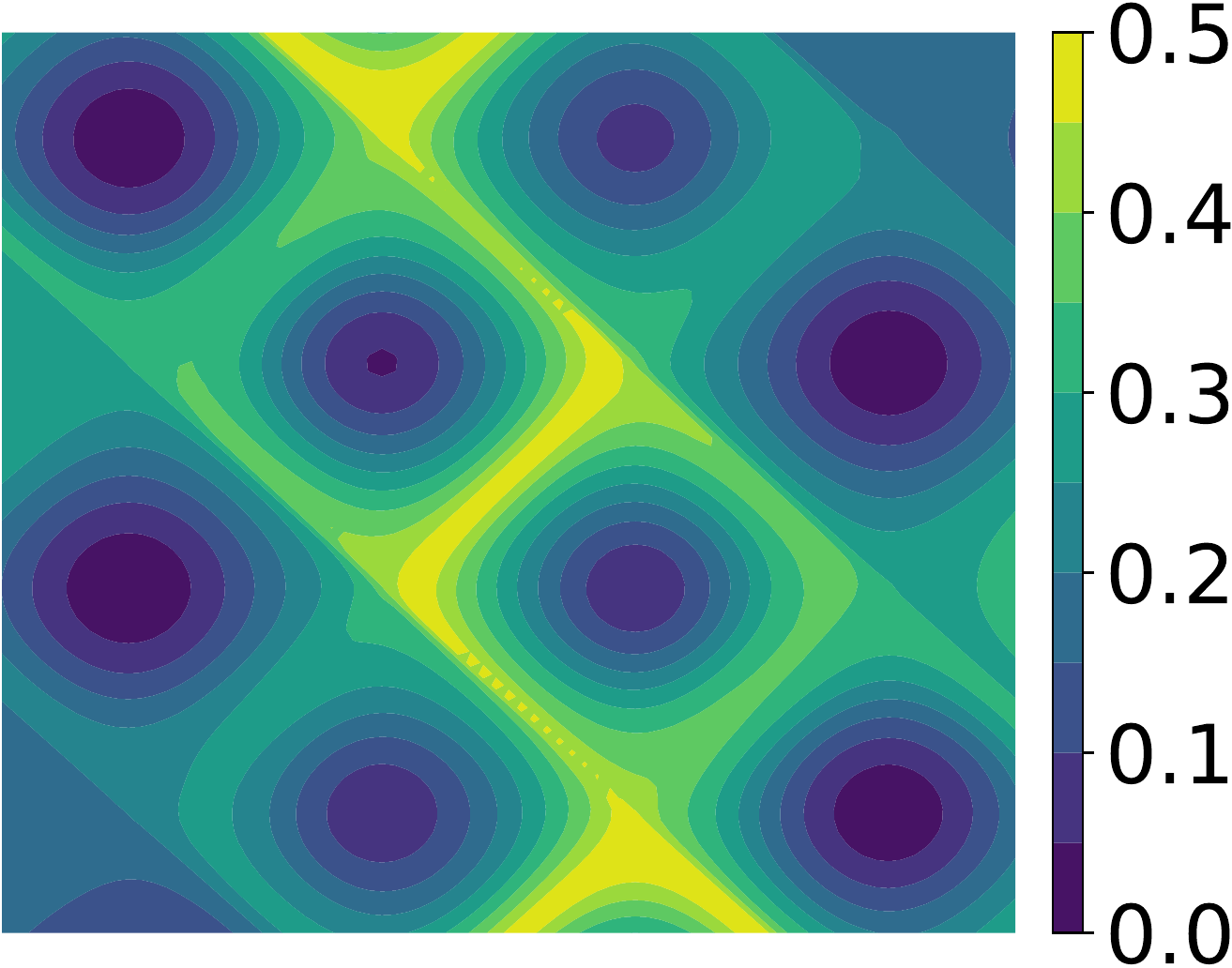}}
  \hspace{3.5in}
  \subfigure[{\tiny Prediction:~8\% data}]
    {\includegraphics[width = 0.2\textwidth]
    {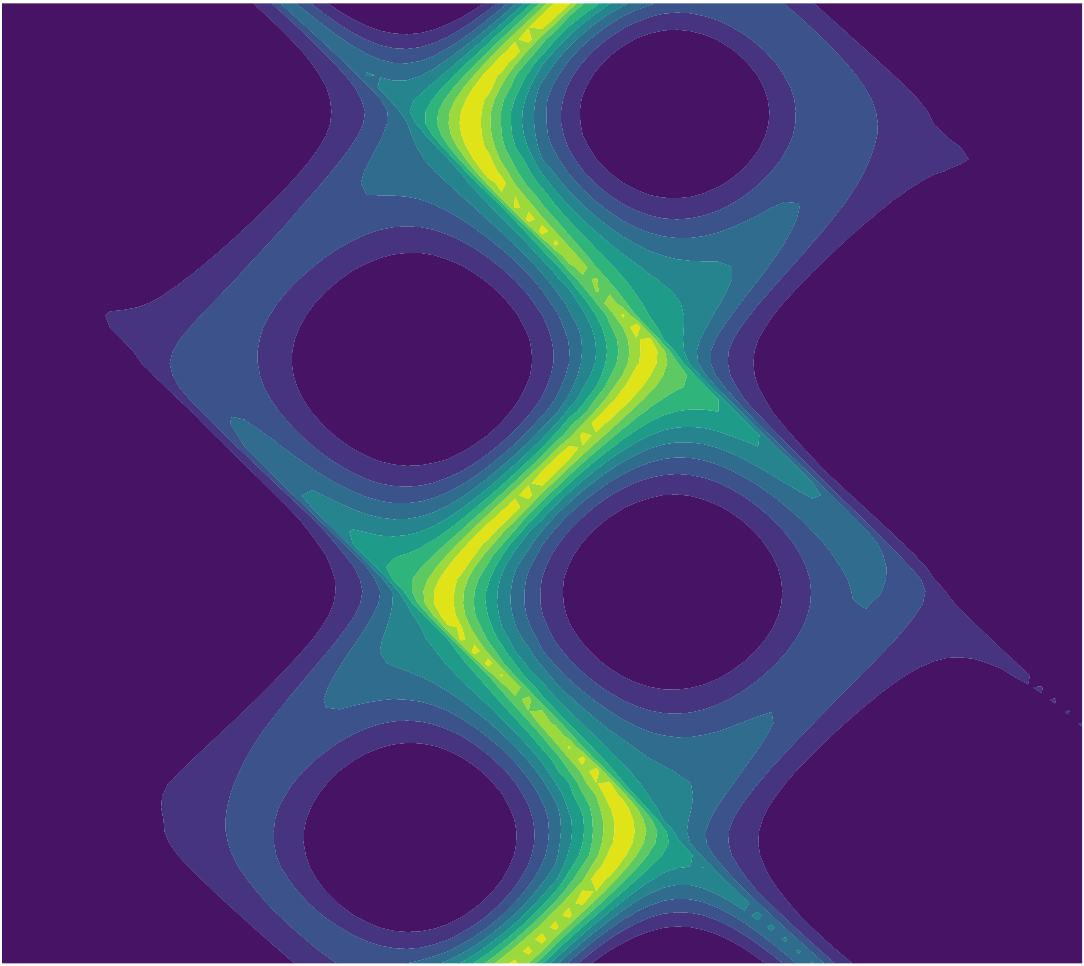}}
  \hspace{0.1in}
  \subfigure[{\tiny Prediction:~16\% data}]
    {\includegraphics[width = 0.2\textwidth]
    {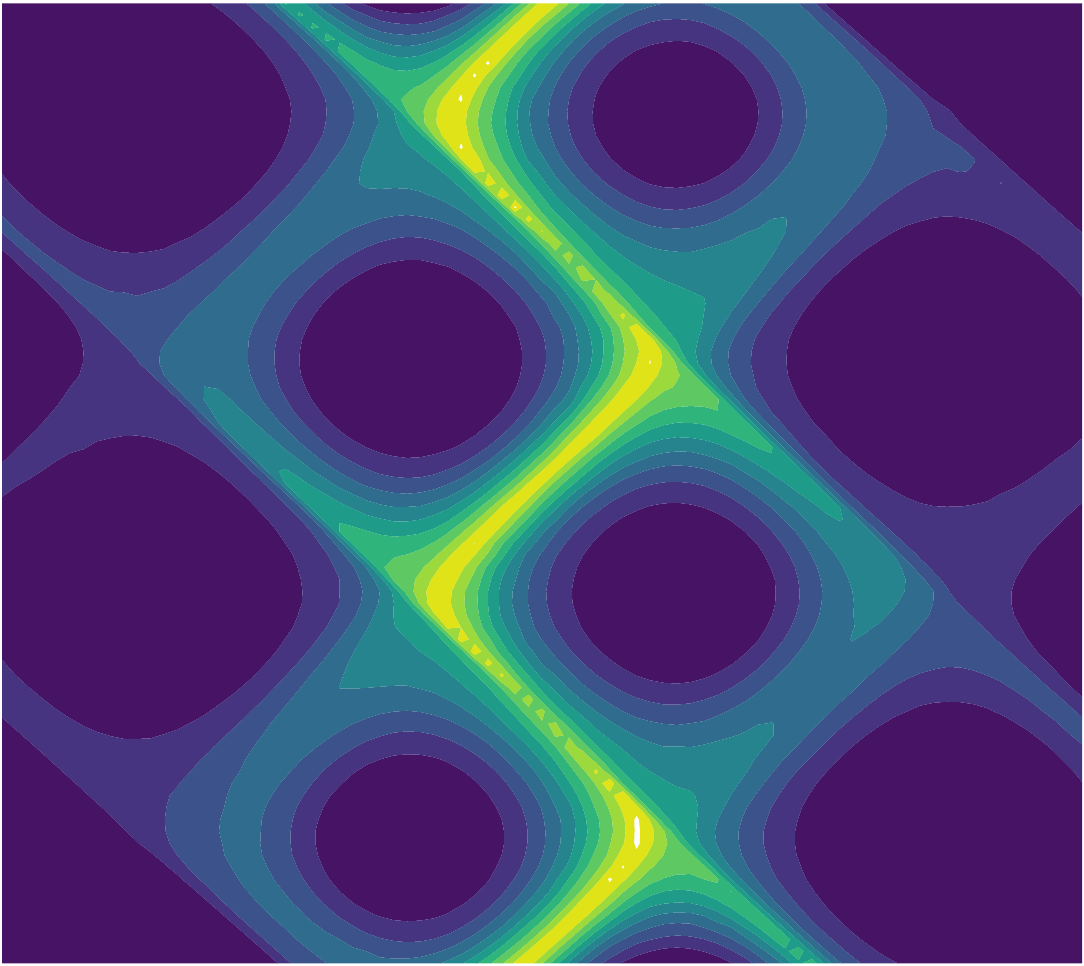}}
  \hspace{0.1in}
  \subfigure[{\tiny Prediction:~24\% data}]
    {\includegraphics[width = 0.2\textwidth]
    {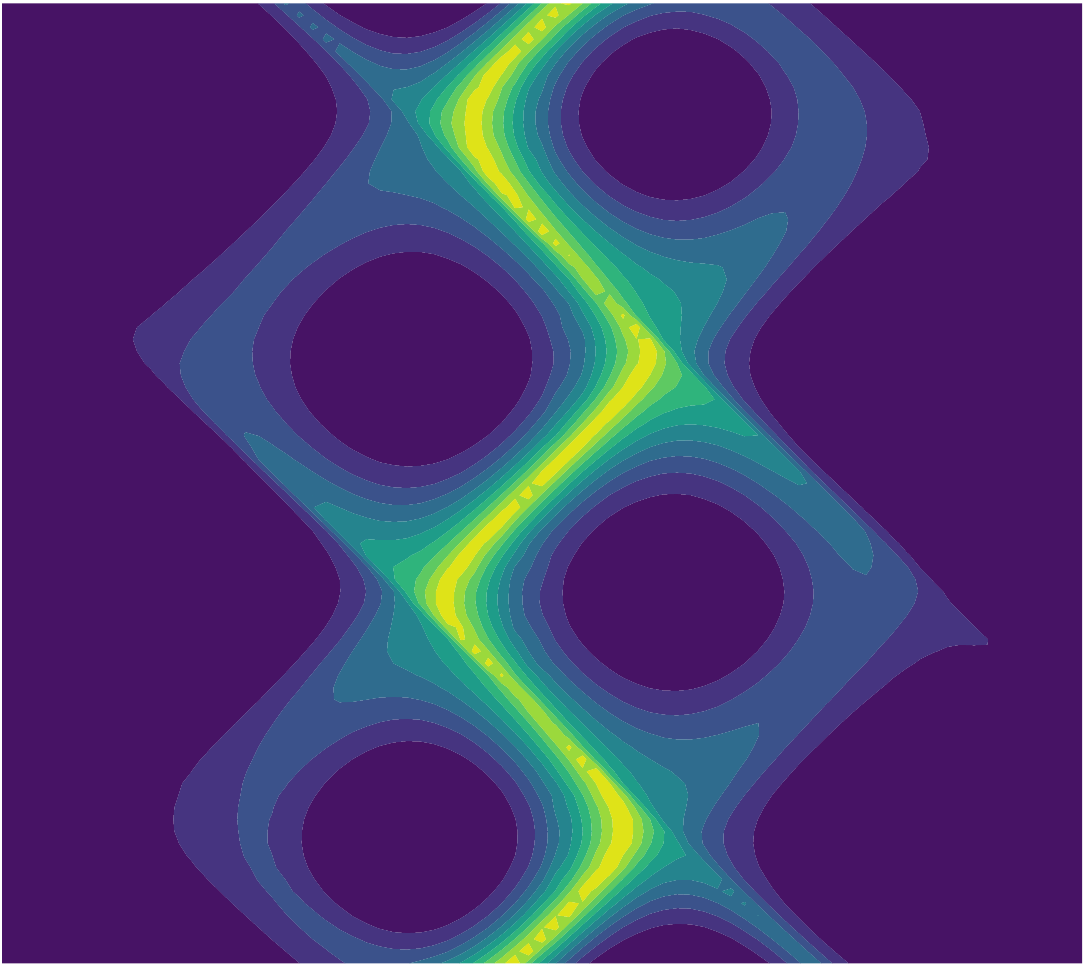}}
  \hspace{0.1in}
  \subfigure[{\tiny Prediction:~32\% data}]
    {\includegraphics[width = 0.2\textwidth]
    {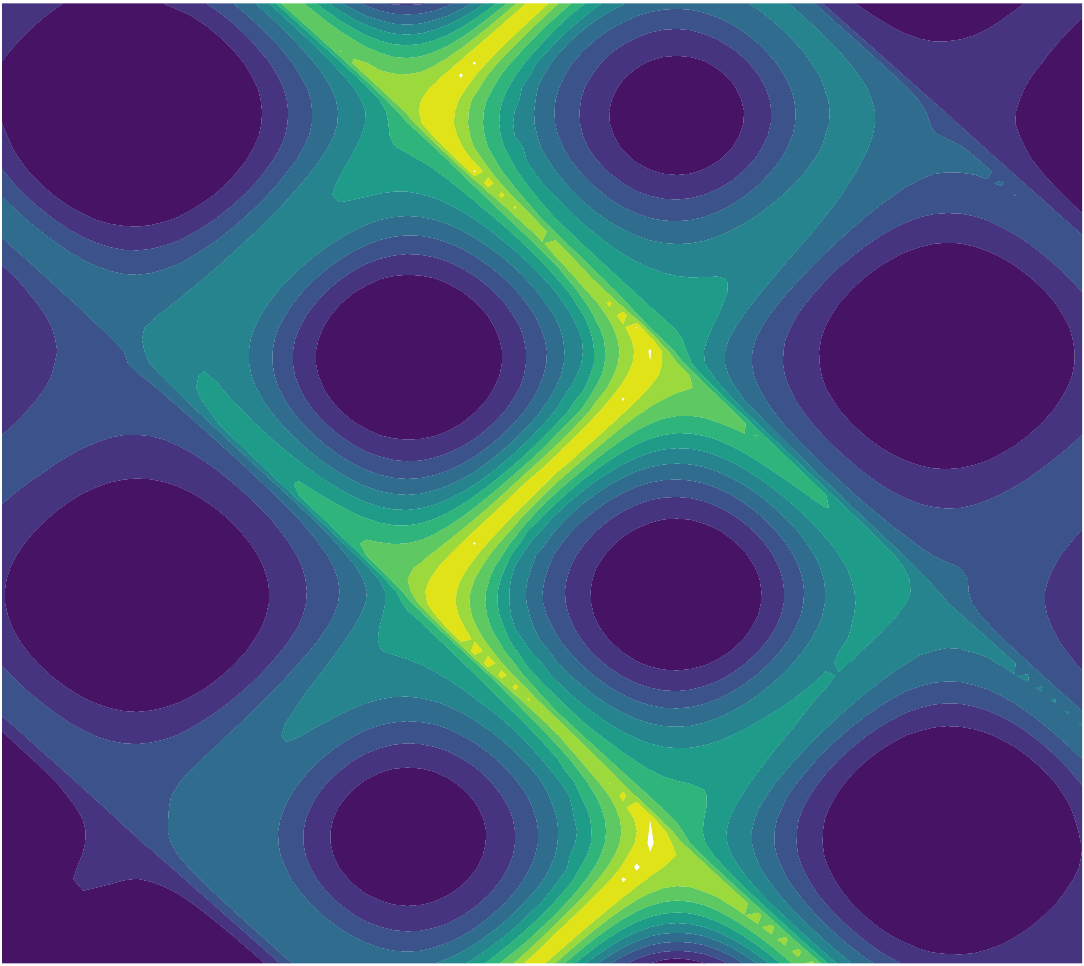}}
  \subfigure[{\tiny Prediction:~40\% data}]
    {\includegraphics[width = 0.2\textwidth]
    {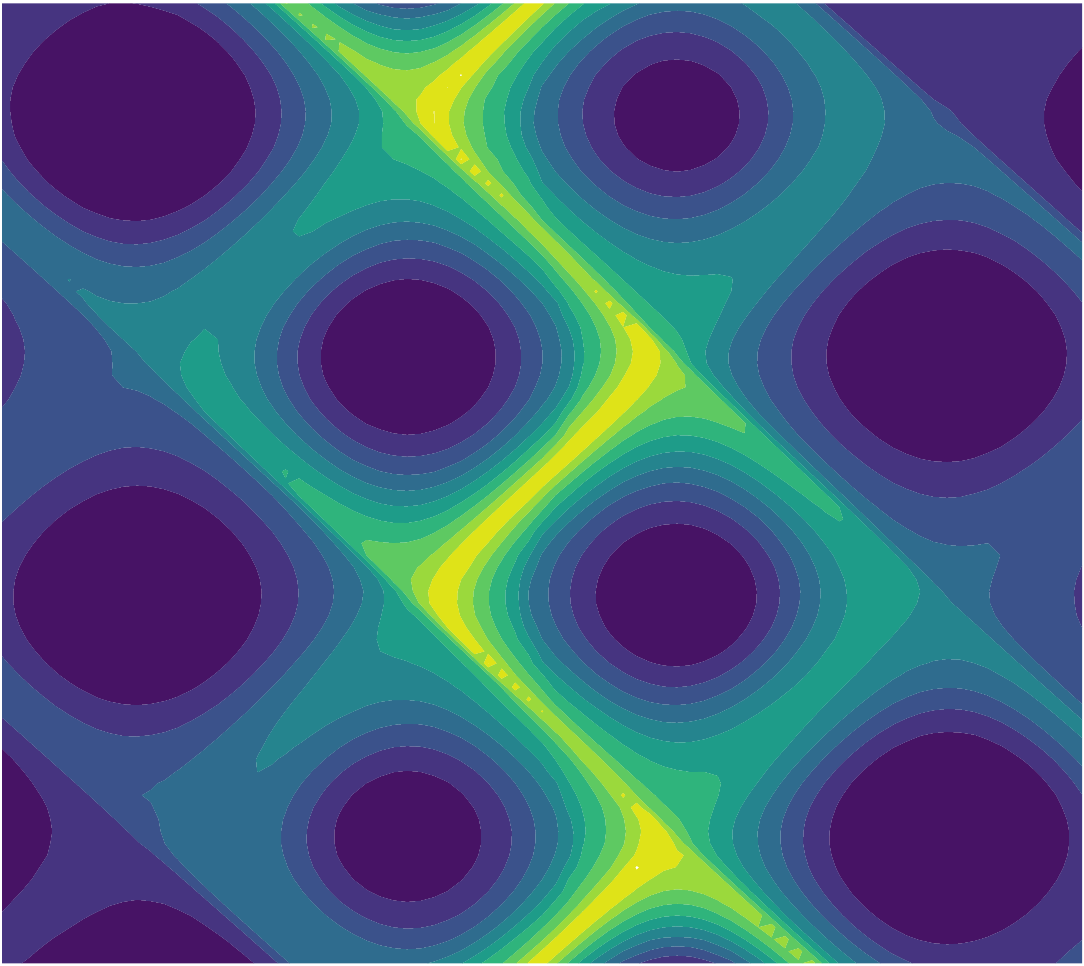}}
  \hspace{0.1in}
  \subfigure[{\tiny Prediction:~48\% data}]
    {\includegraphics[width = 0.2\textwidth]
    {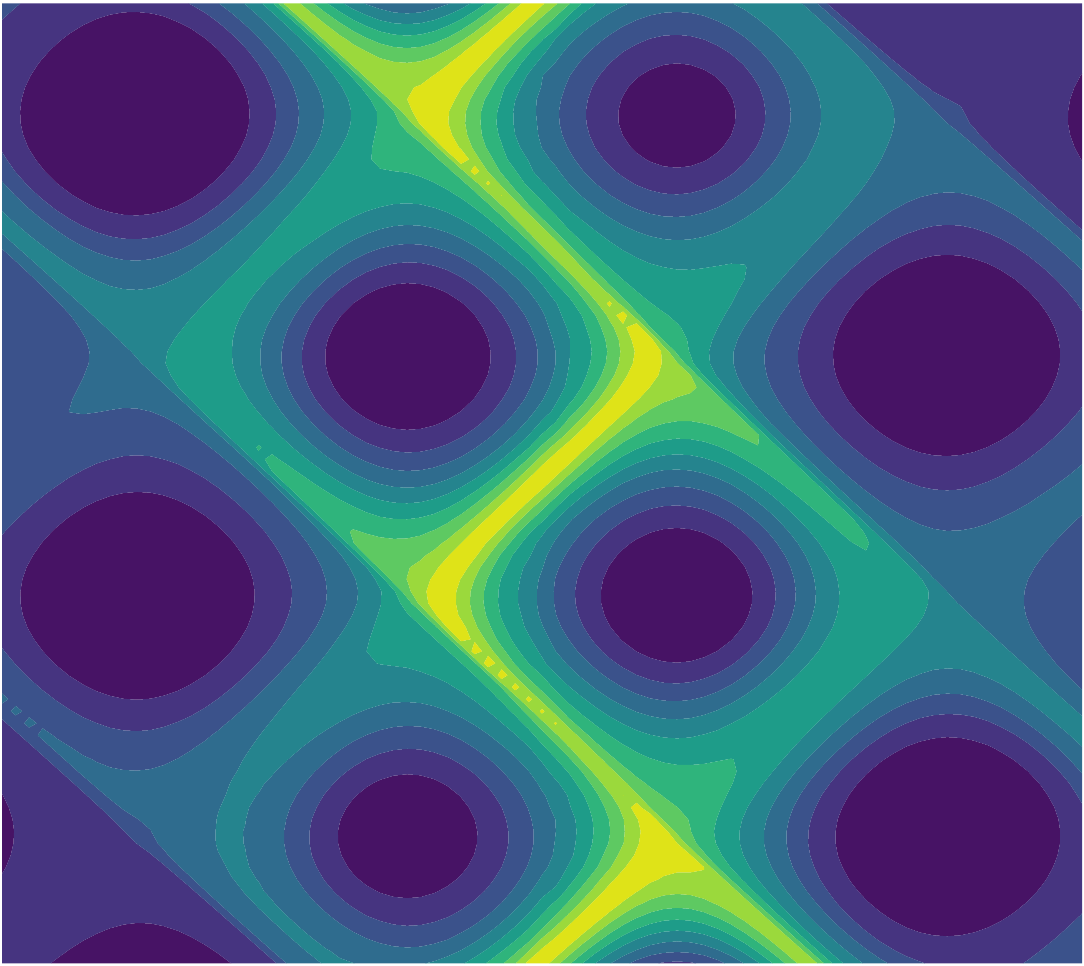}}
  \hspace{0.1in}
  \subfigure[{\tiny Prediction:~56\% data}]
    {\includegraphics[width = 0.2\textwidth]
    {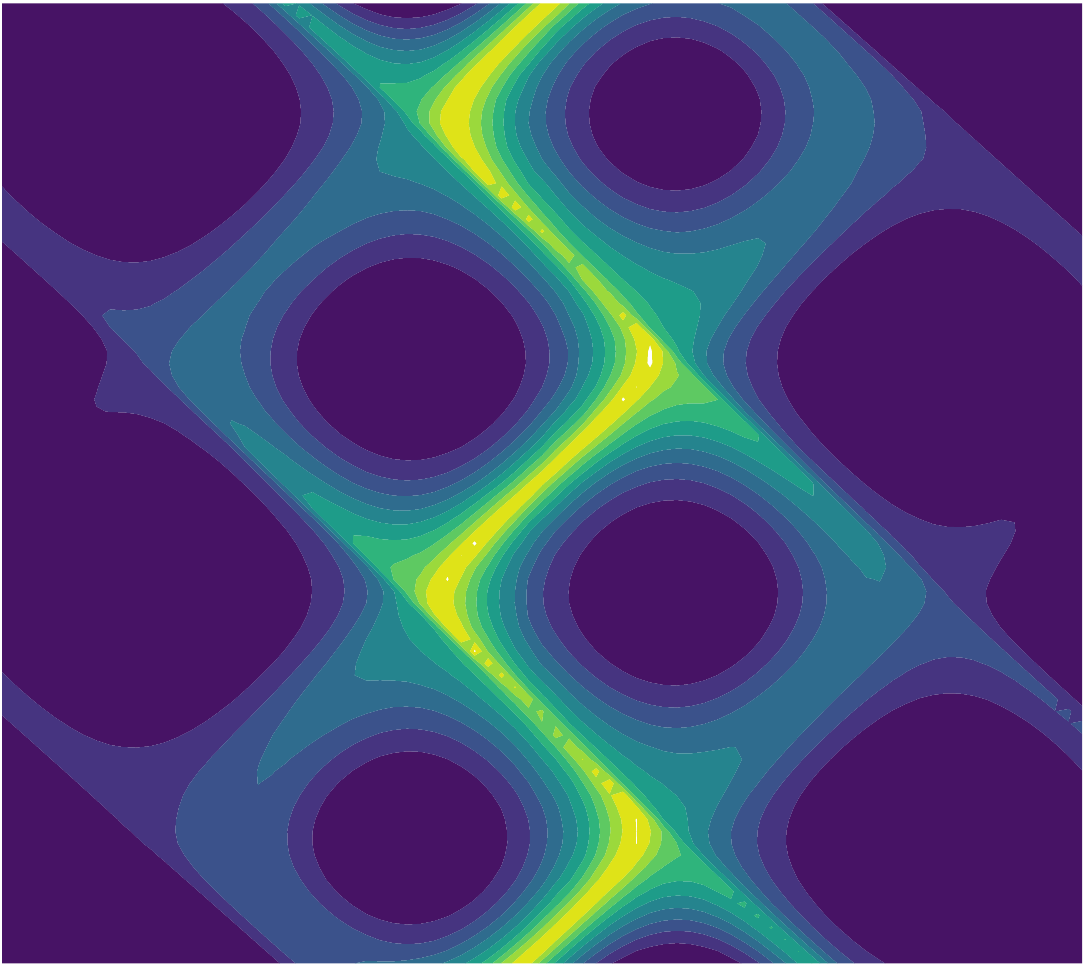}}
  \hspace{0.1in}
  \subfigure[{\tiny Prediction:~64\% data}]
    {\includegraphics[width = 0.2\textwidth]
    {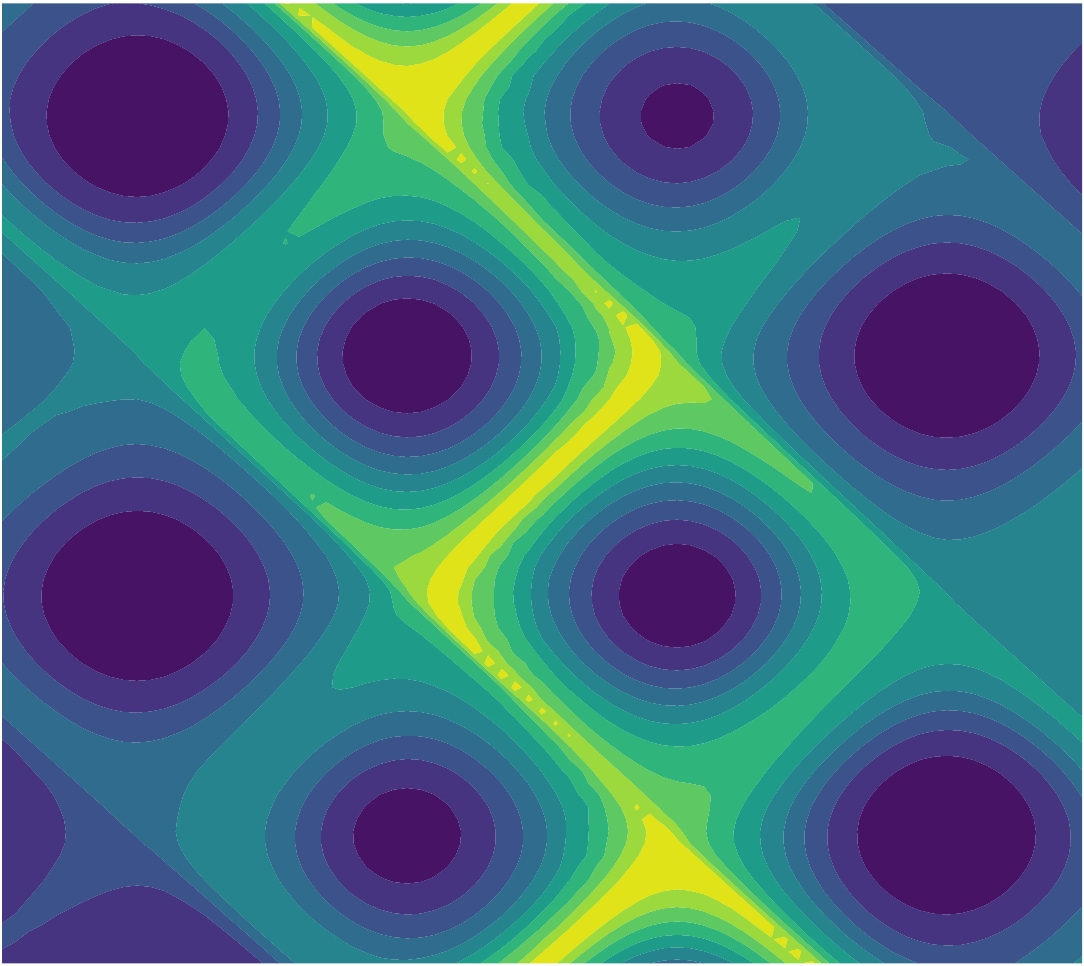}}
  \subfigure[{\tiny Prediction:~72\% data}]
    {\includegraphics[width = 0.2\textwidth]
    {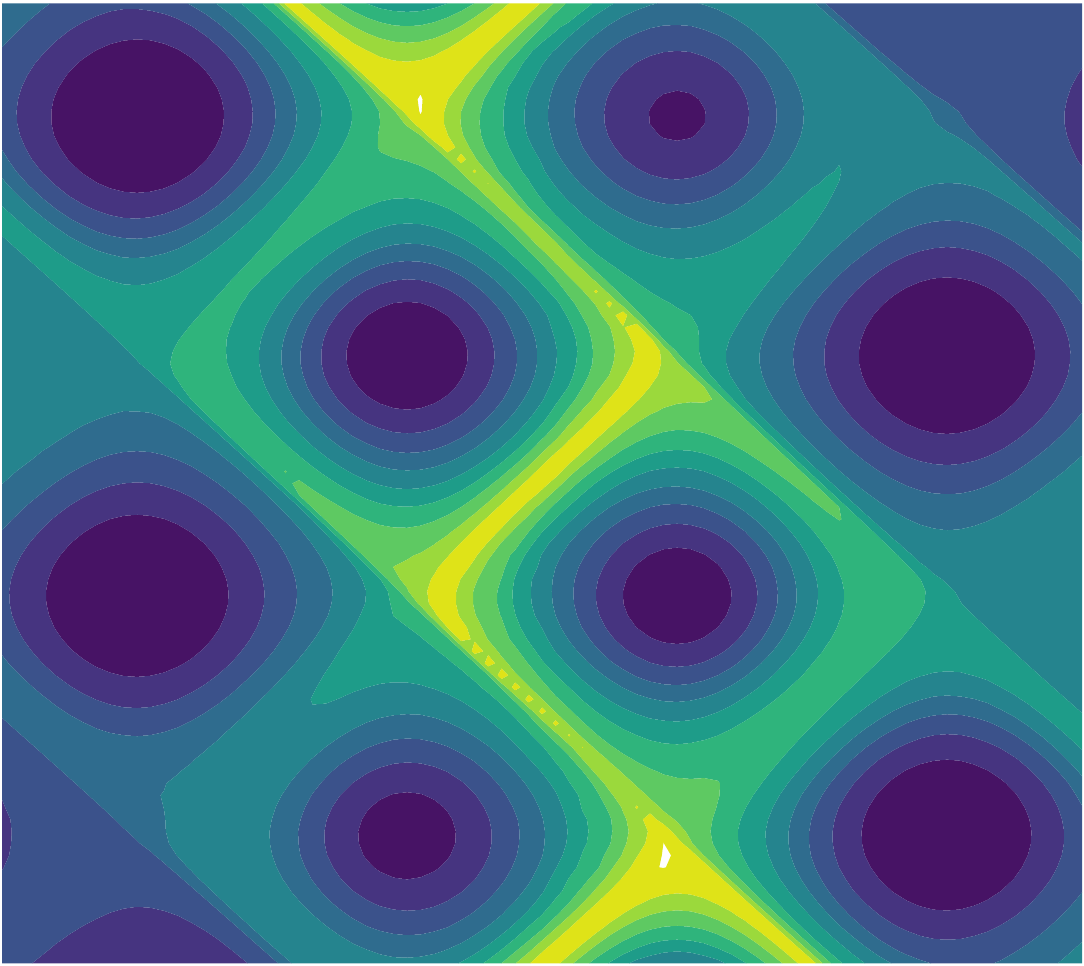}}
  \hspace{0.1in}
  \subfigure[{\tiny Prediction:~80\% data}]
    {\includegraphics[width = 0.2\textwidth]
    {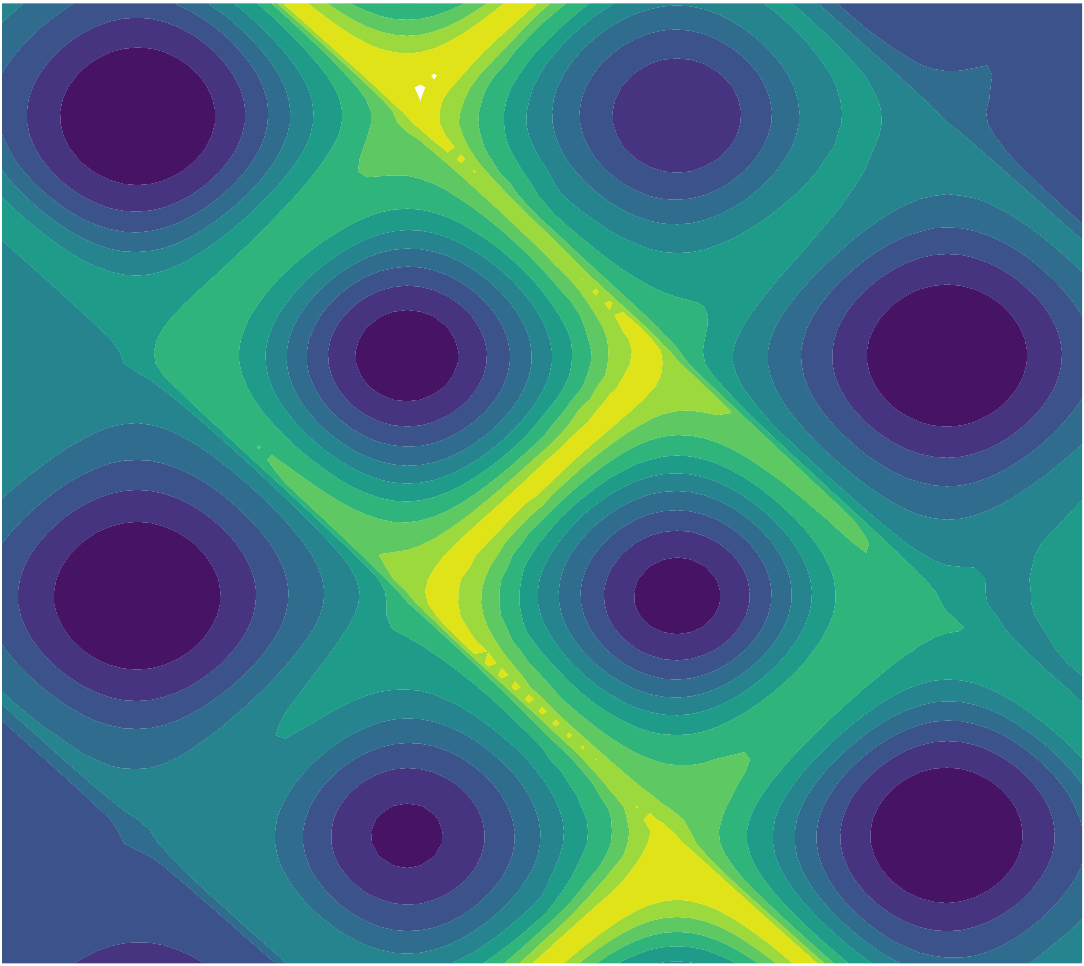}}
  \hspace{0.1in}
  \subfigure[{\tiny Prediction:~88\% data}]
    {\includegraphics[width = 0.2\textwidth]
    {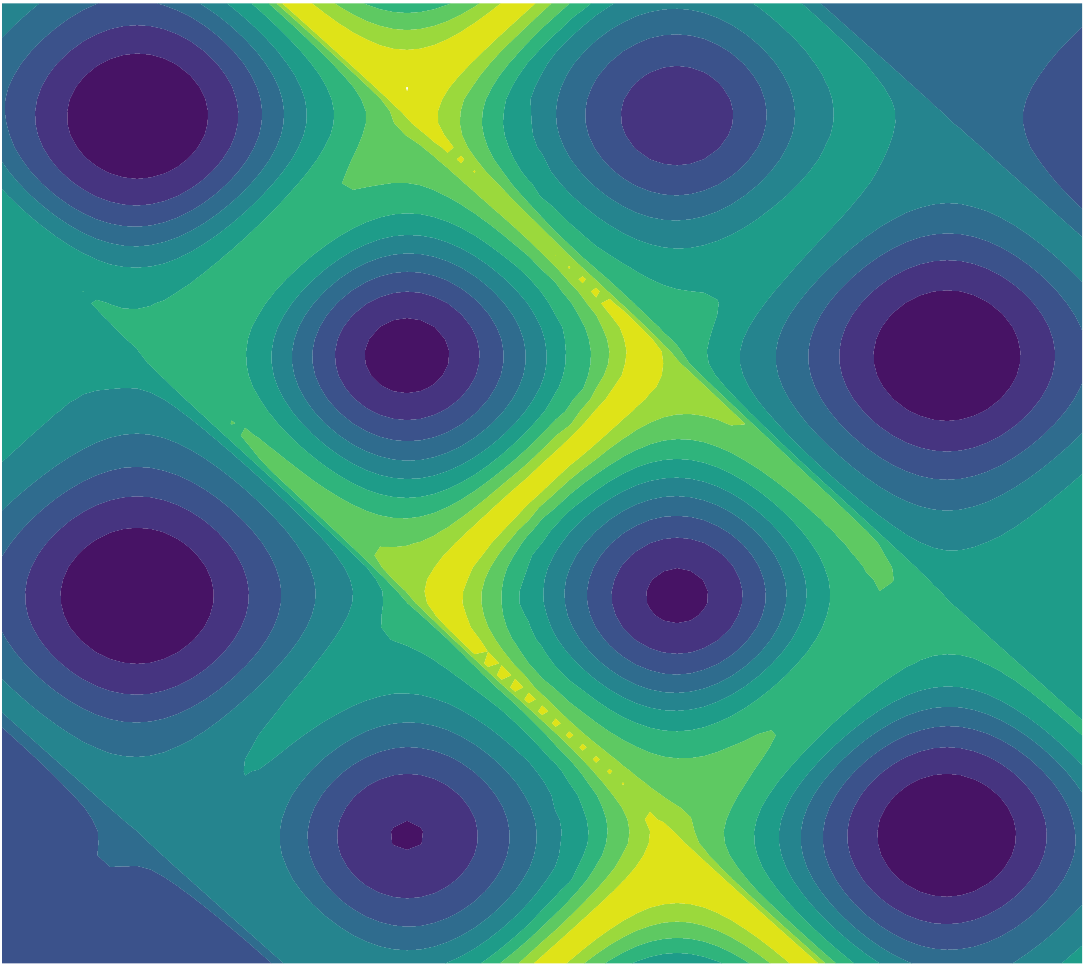}}
  \hspace{0.1in}
  \subfigure[{\tiny Prediction:~96\% data}]
    {\includegraphics[width = 0.2\textwidth]
    {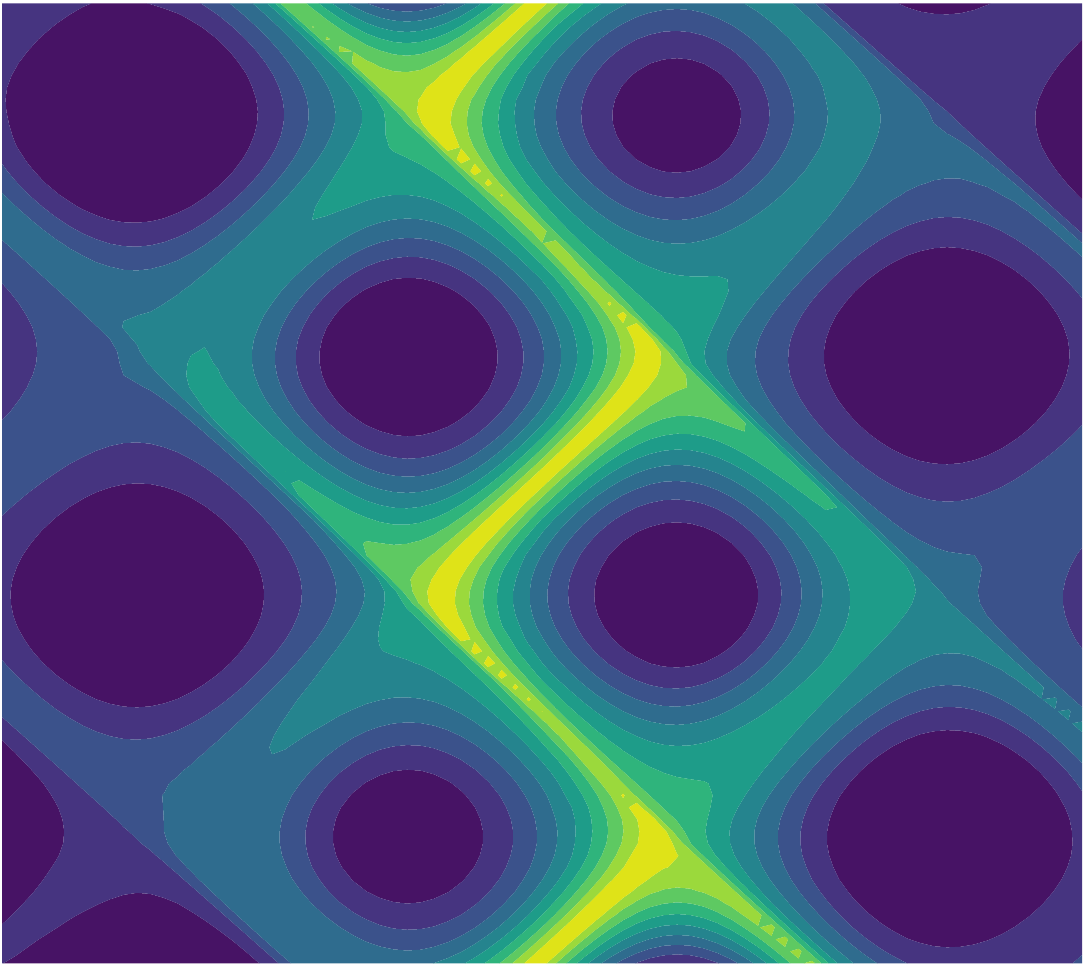}}
  \caption{\textbf{Predictions for $\kappa_fL = 2$:}~This figure shows the ground truth from the non-negative FEM and predictions from the trained non-negative CNN-LSTM models at $t = 1.0$.
  From this figure, it is evident that our deep learning model is able to effectively capture product $C$ evolution with minimal training data (e.g., greater than or equal to 32\%). 
  \label{Fig:DL_RT_Pred_kfL2}}
\end{figure}

\begin{figure}
  \centering
  \subfigure[{\tiny Prediction error:~8\% data}]
    {\includegraphics[width = 0.285\textwidth]
    {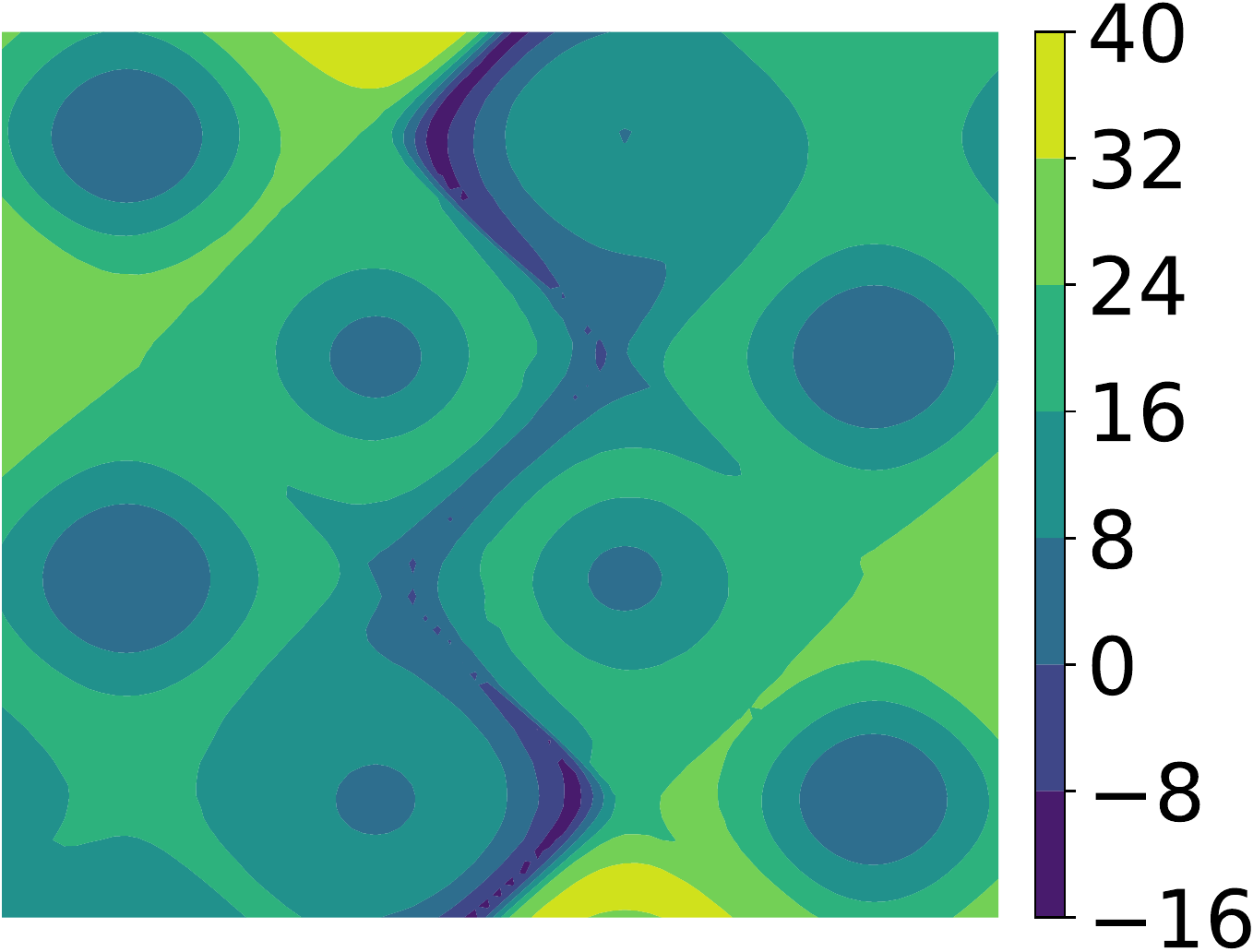}}
  \hspace{0.25in}
  \subfigure[{\tiny Prediction error:~16\% data}]
    {\includegraphics[width = 0.285\textwidth]
    {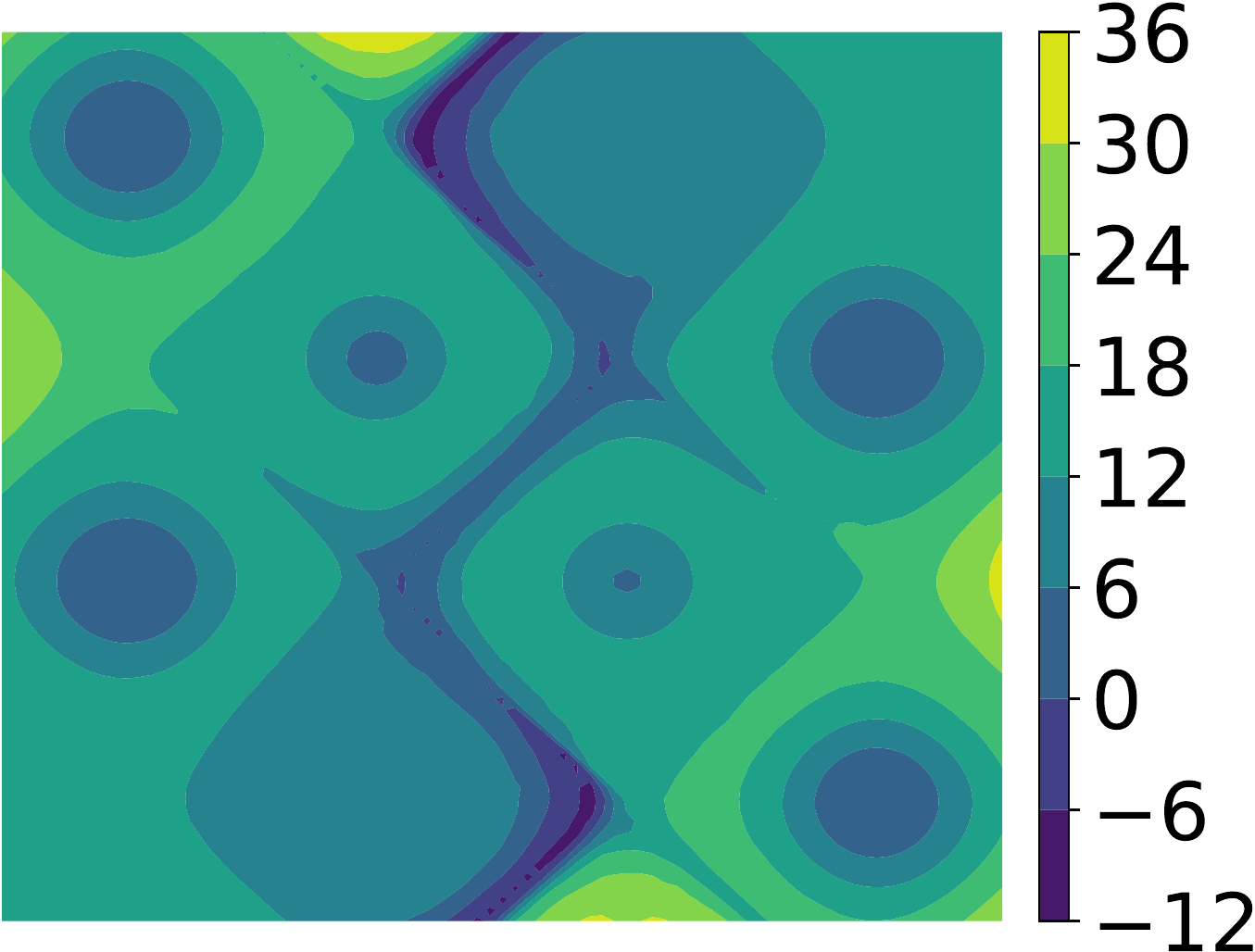}}
  \hspace{0.25in}
  \subfigure[{\tiny Prediction error:~24\% data}]
    {\includegraphics[width = 0.285\textwidth]
    {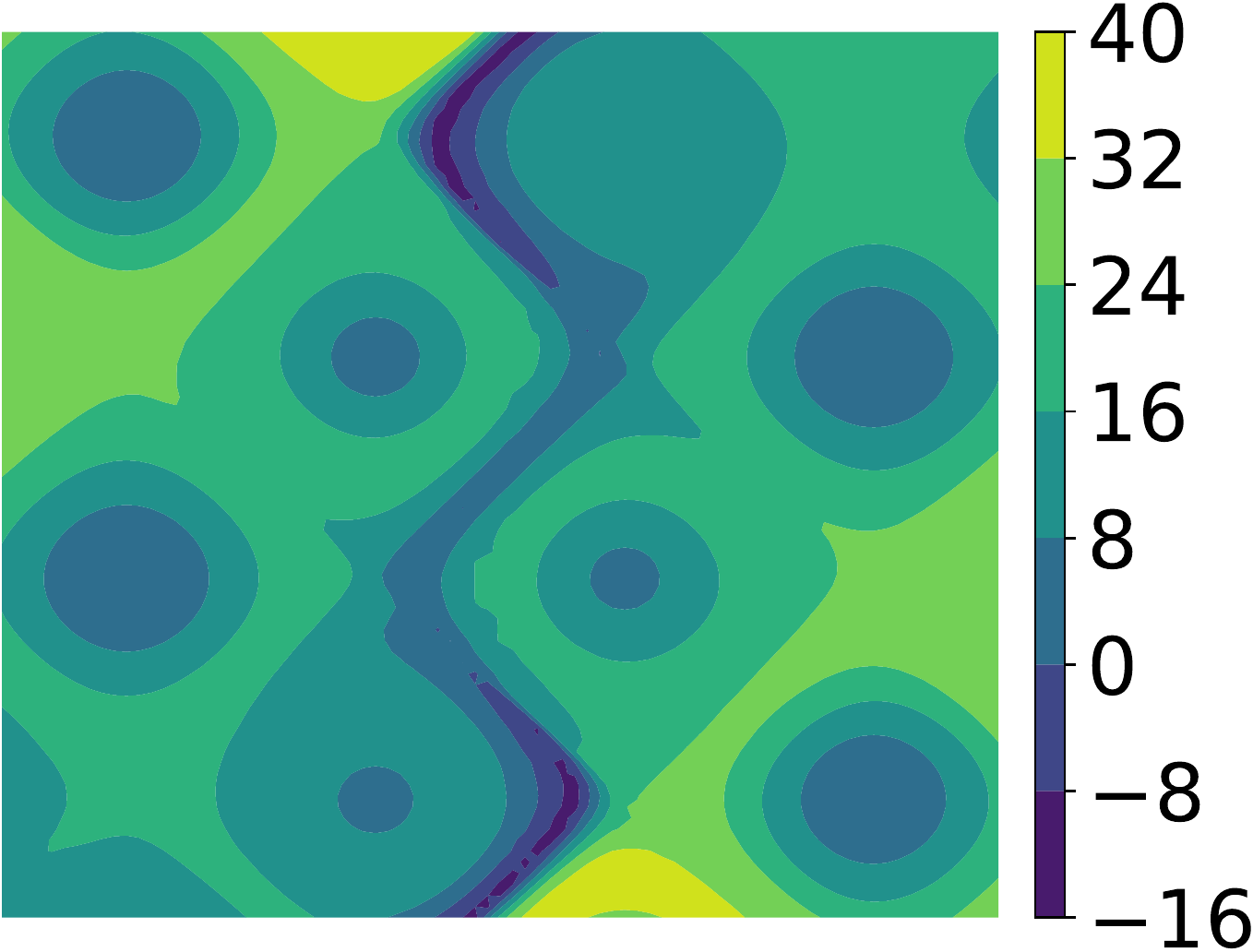}}
  \subfigure[{\tiny Prediction error:~32\% data}]
    {\includegraphics[width = 0.285\textwidth]
    {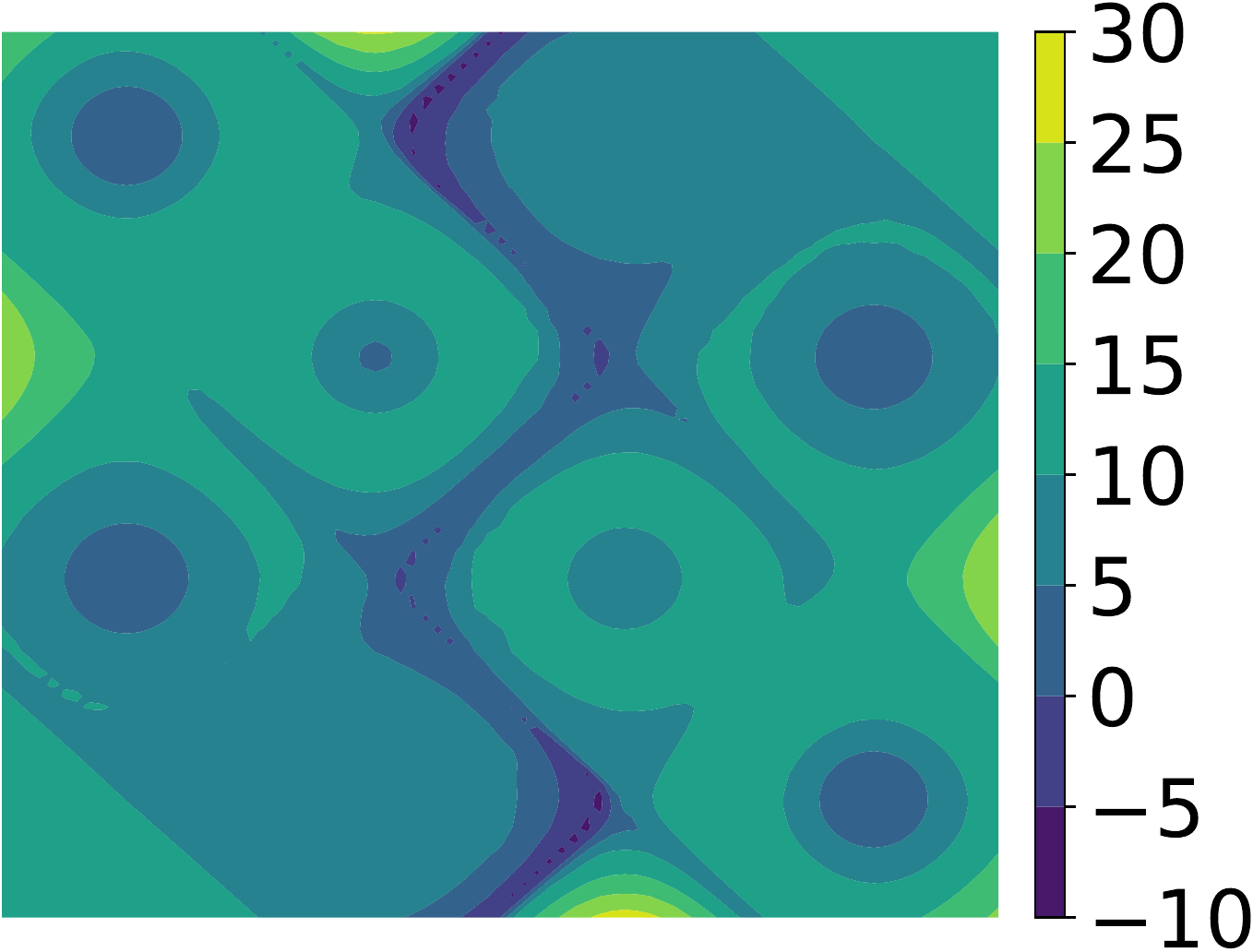}}
  \hspace{0.25in}
  \subfigure[{\tiny Prediction error:~40\% data}]
    {\includegraphics[width = 0.275\textwidth]
    {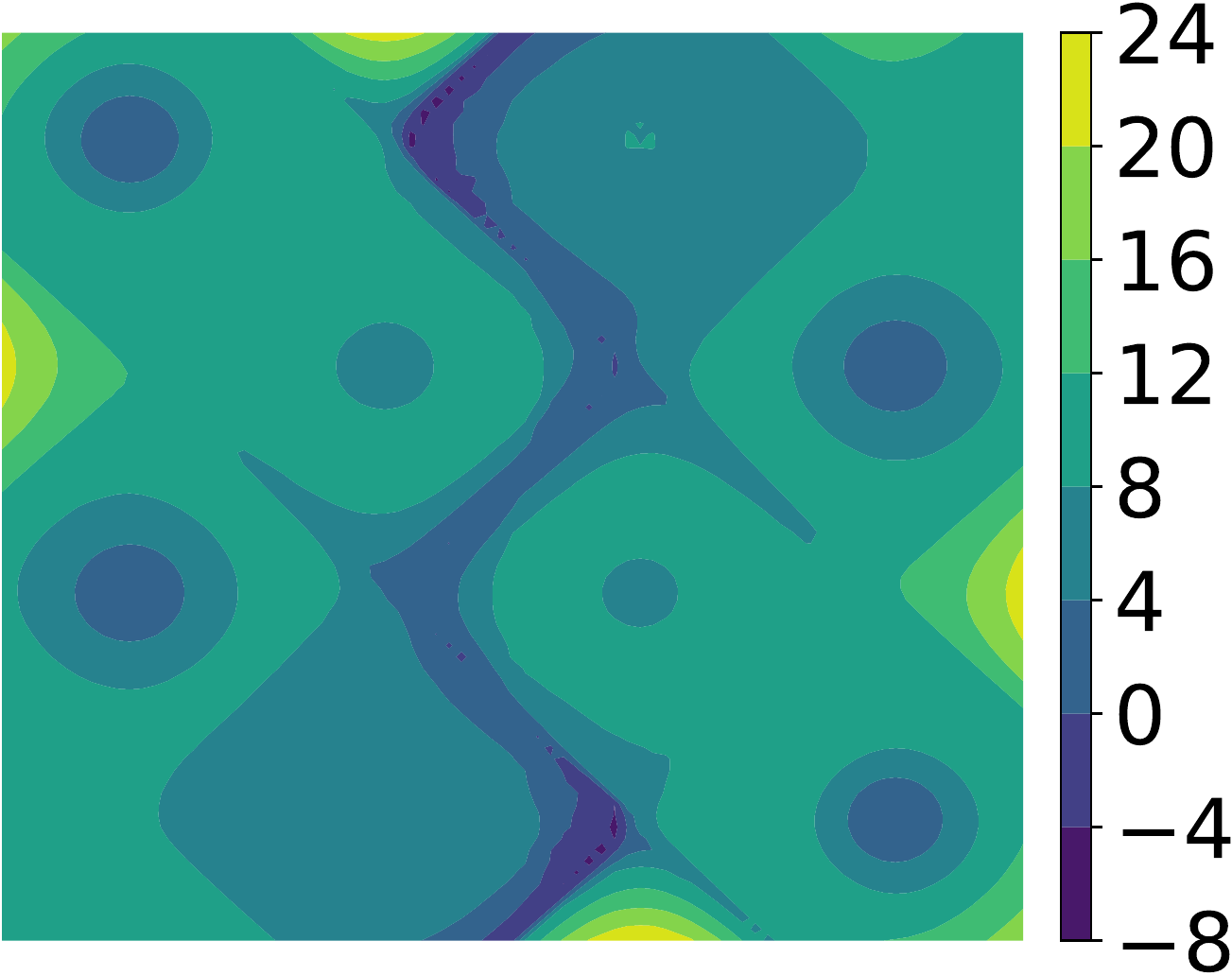}}
  \hspace{0.25in}
  \subfigure[{\tiny Prediction error:~48\% data}]
    {\includegraphics[width = 0.285\textwidth]
    {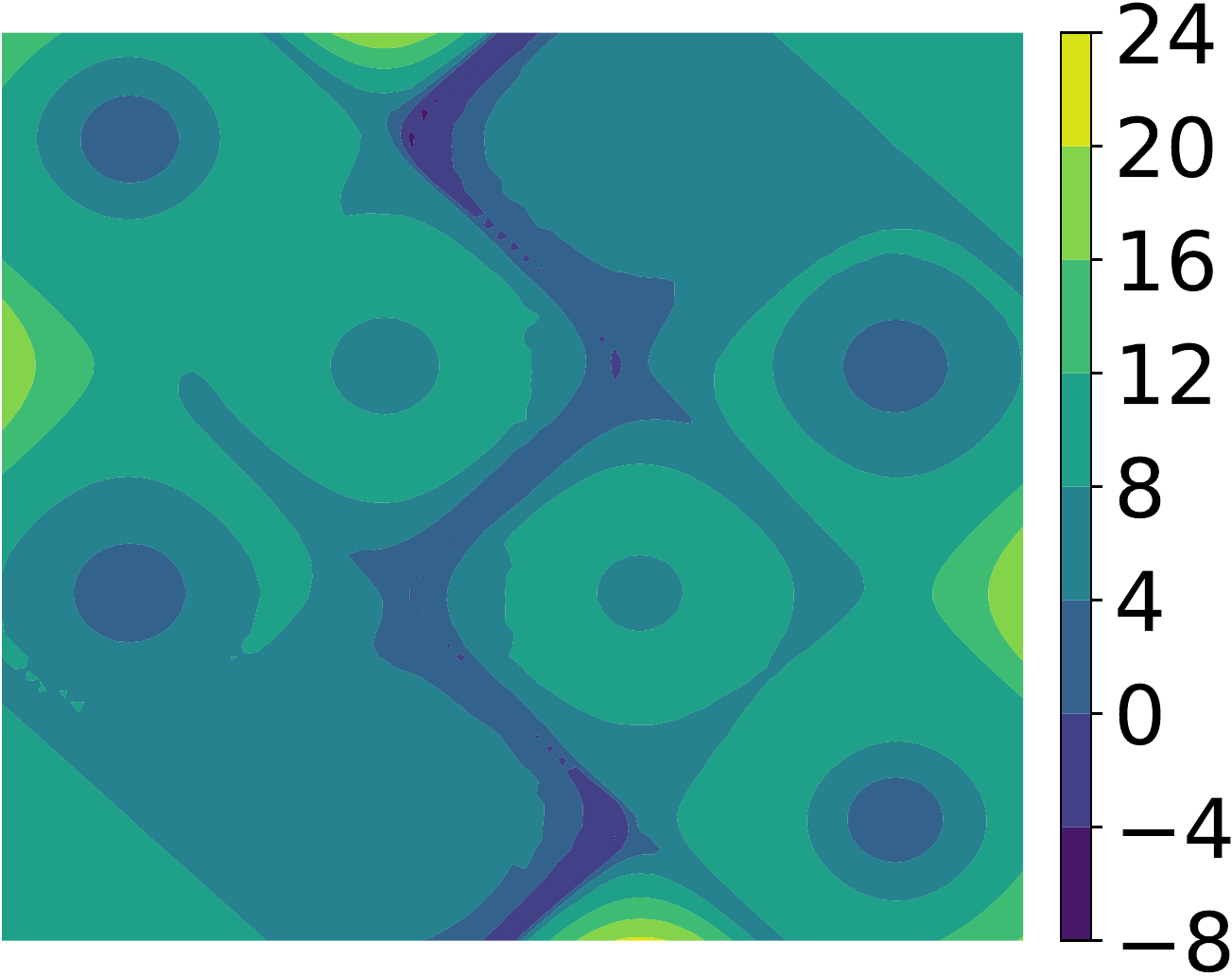}}
  \subfigure[{\tiny Prediction error:~56\% data}]
    {\includegraphics[width = 0.285\textwidth]
    {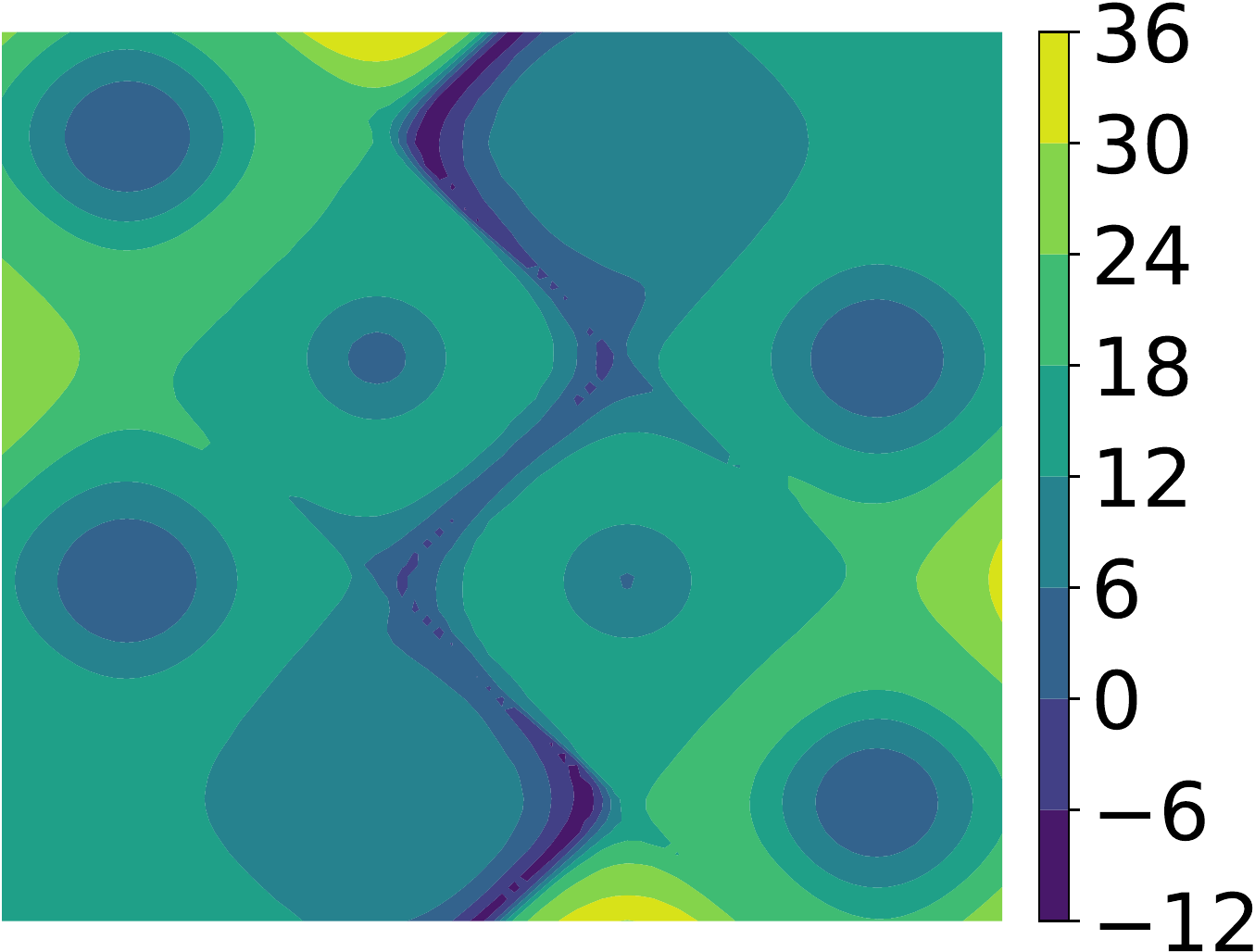}}
  \hspace{0.25in}
  \subfigure[{\tiny Prediction error:~64\% data}]
    {\includegraphics[width = 0.285\textwidth]
    {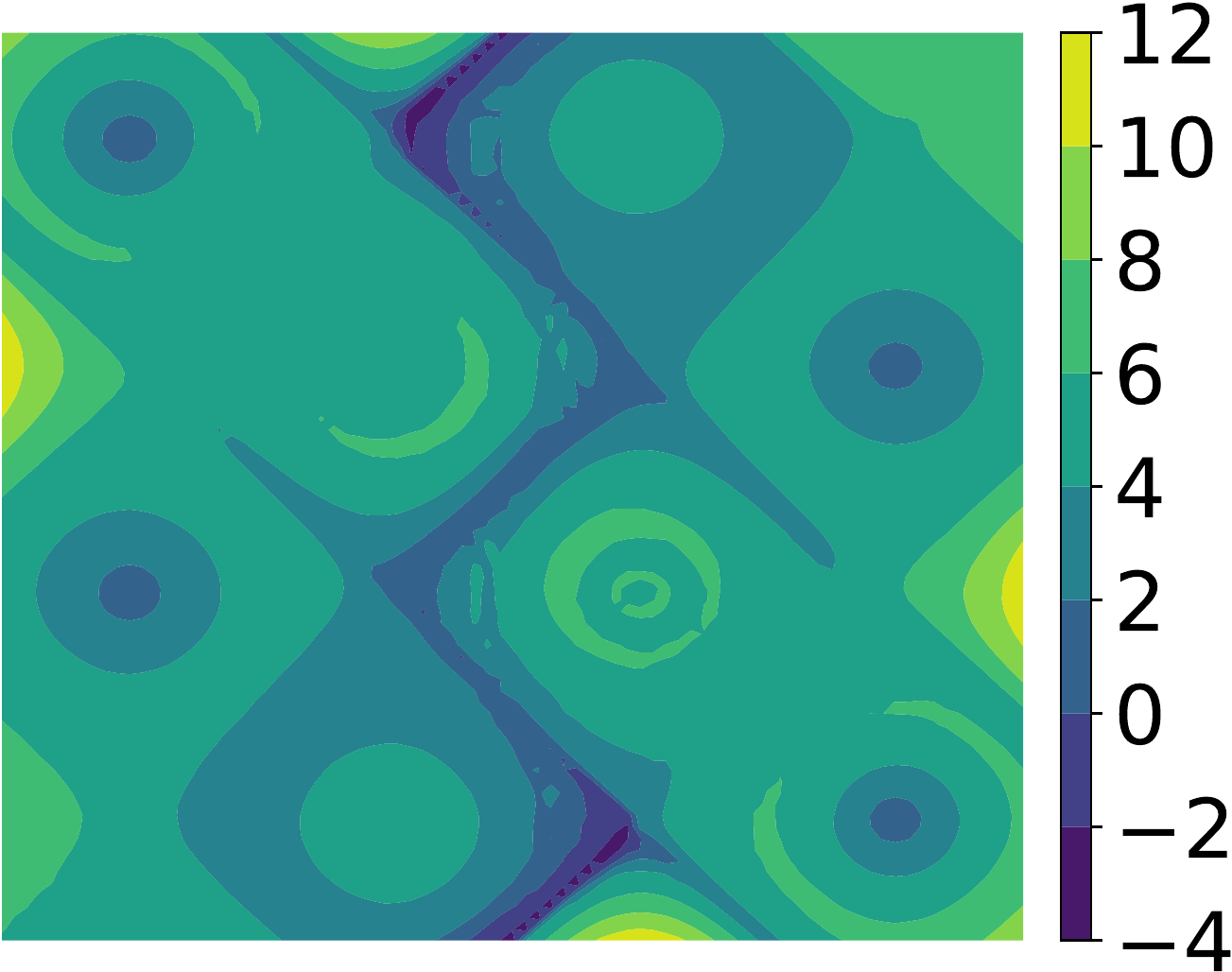}}
  \hspace{0.25in}
  \subfigure[{\tiny Prediction error:~72\% data}]
    {\includegraphics[width = 0.285\textwidth]
    {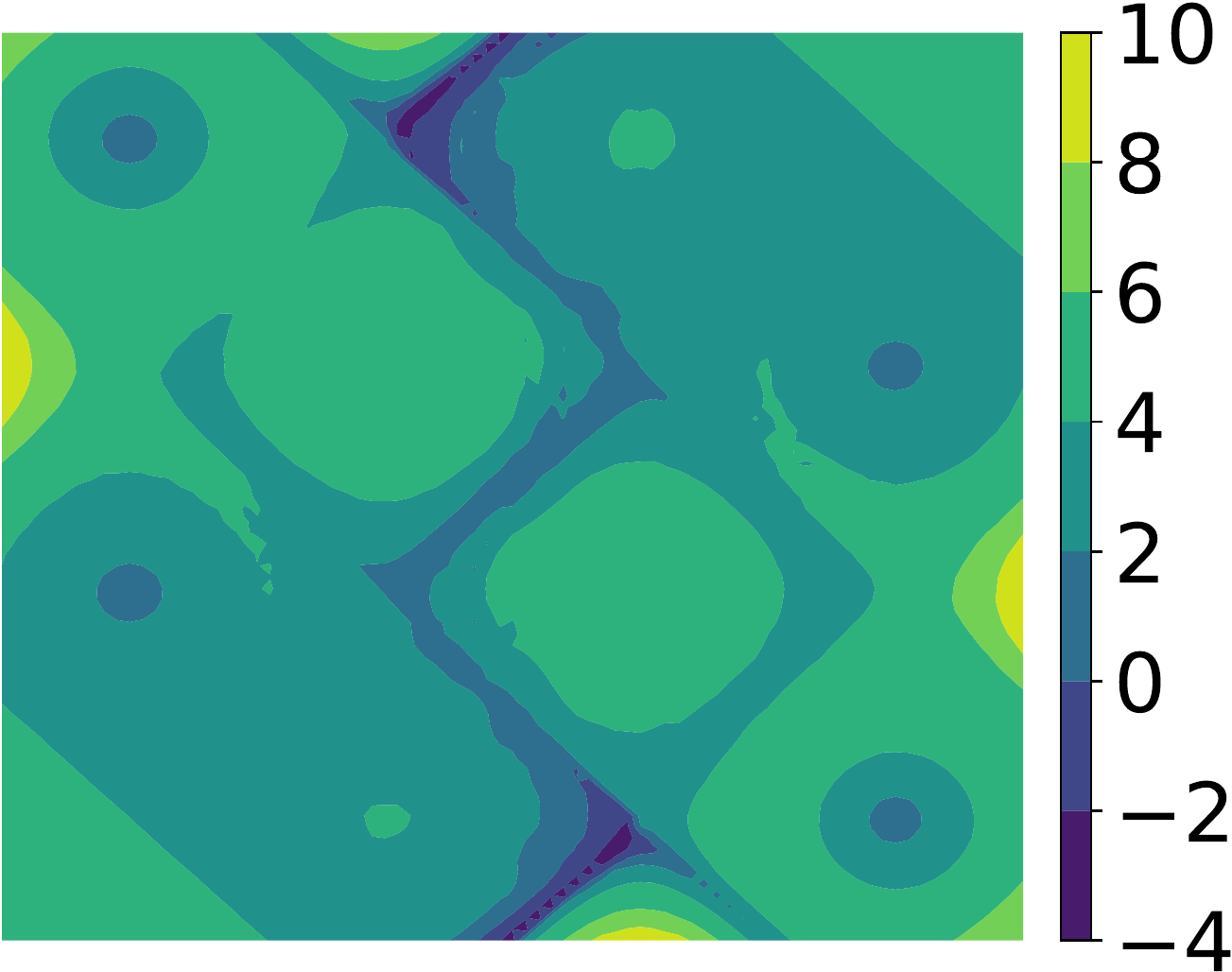}}
  \subfigure[{\tiny Prediction error:~80\% data}]
    {\includegraphics[width = 0.285\textwidth]
    {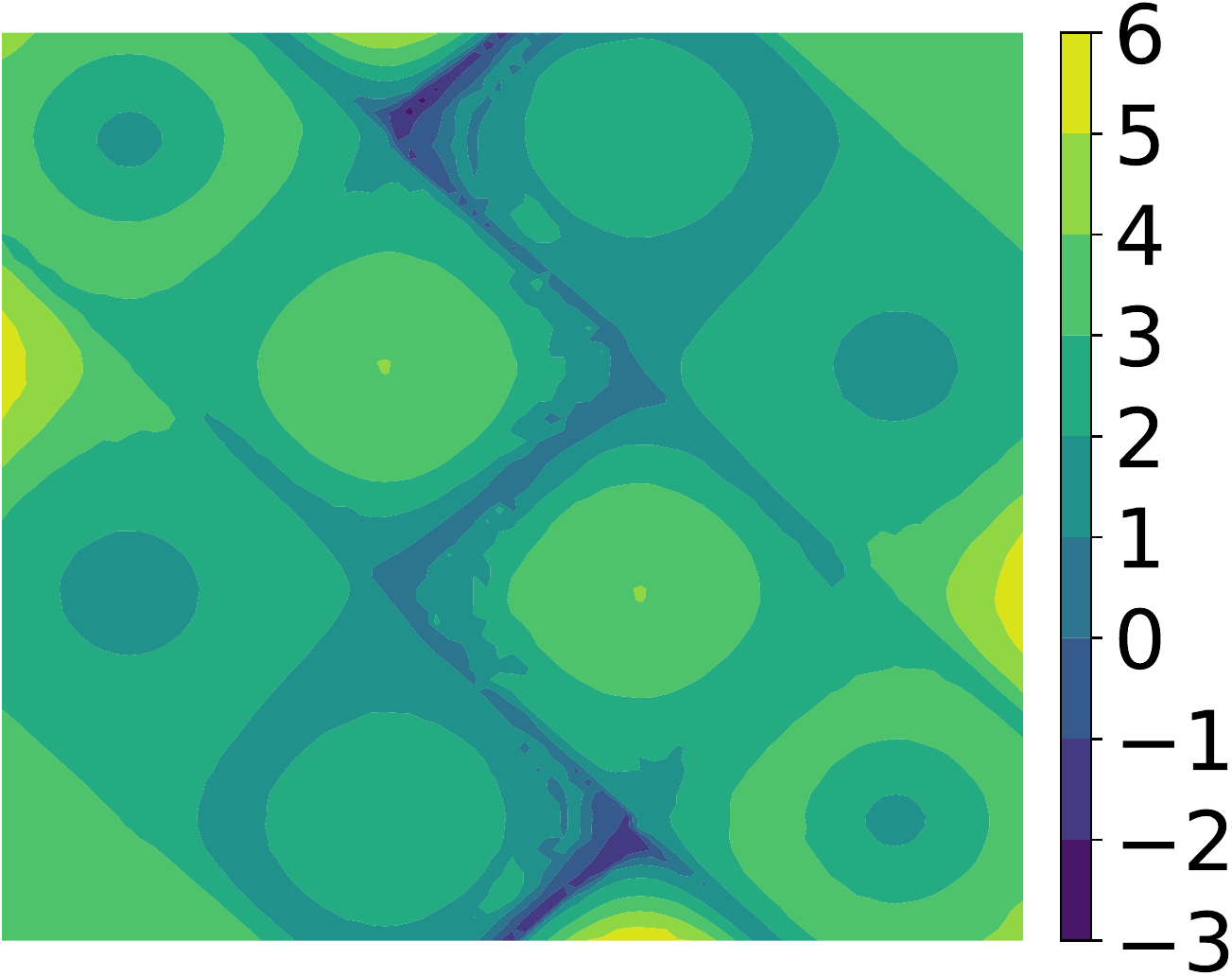}}
  \hspace{0.25in}
  \subfigure[{\tiny Prediction error:~88\% data}]
    {\includegraphics[width = 0.285\textwidth]
    {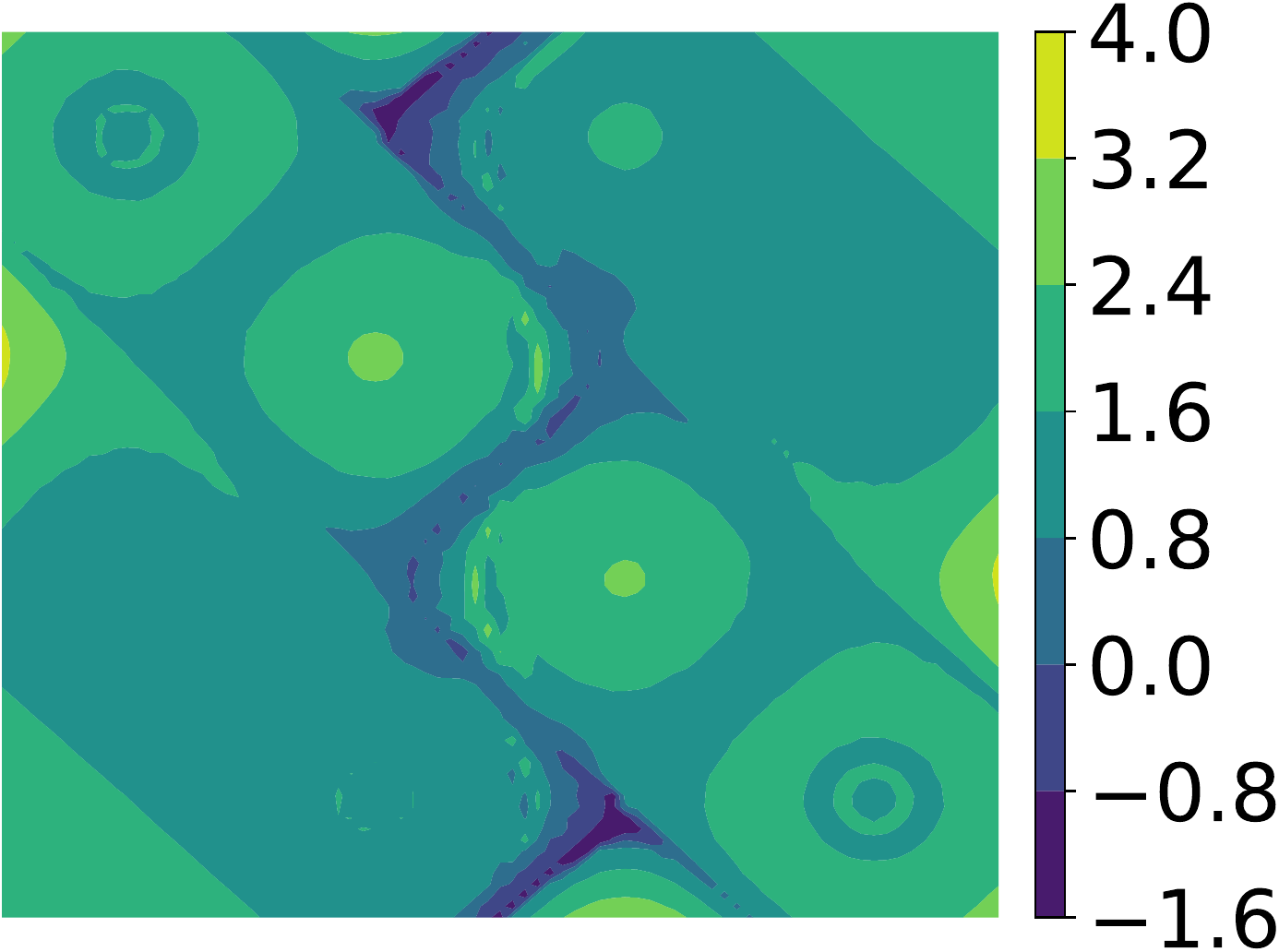}}
  \hspace{0.25in}
  \subfigure[{\tiny Prediction error:~96\% data}]
    {\includegraphics[width = 0.285\textwidth]
    {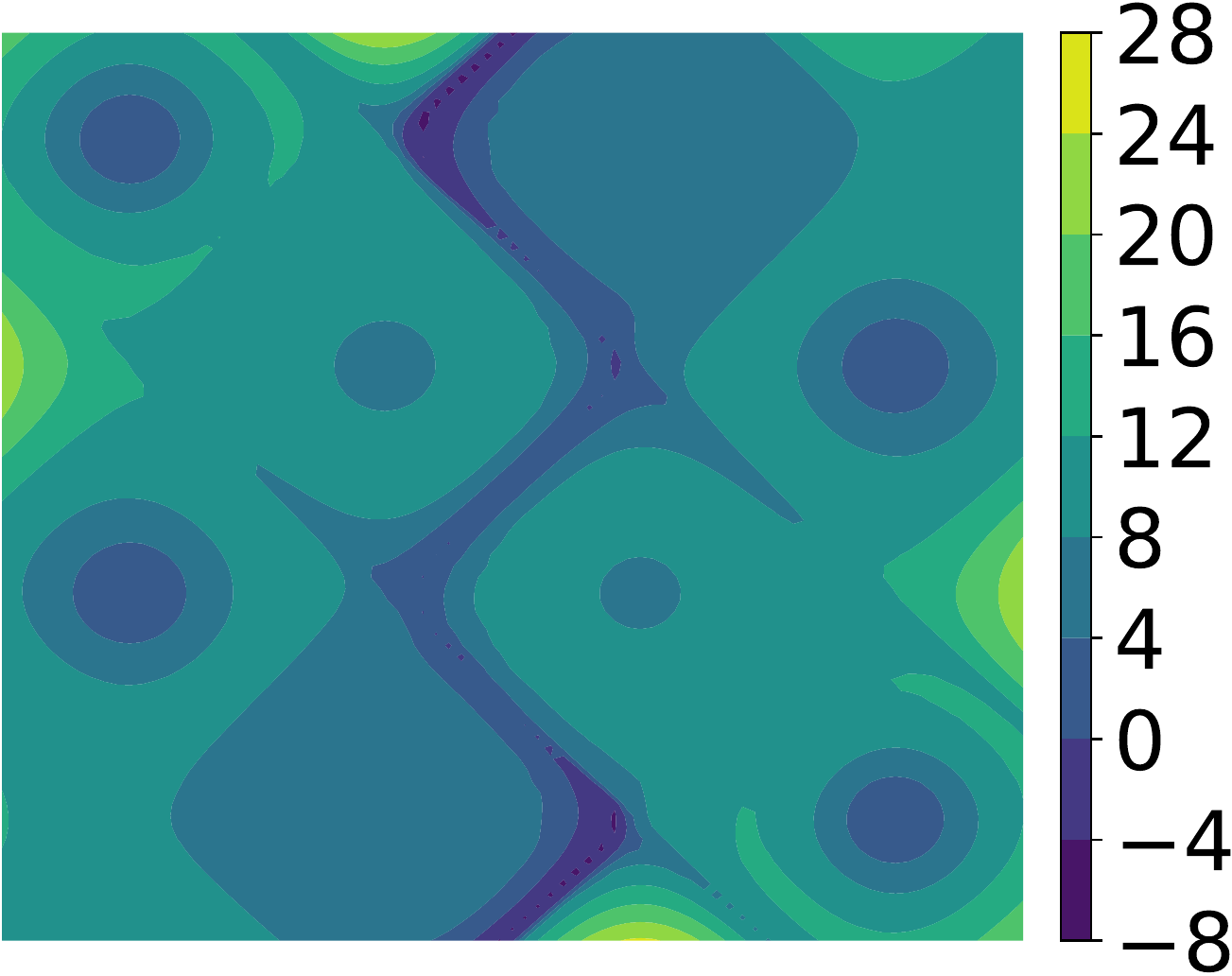}}
  \caption{\textbf{Prediction error in percentage for $\kappa_fL = 2$:}~This figure compares the prediction errors in the entire domain at $t = 1.0$ for different amount of training data.
  Based on the error values (e.g., $\leq 10\%$), it is evident that with less than 64\% of training data, we can accurately capture different mixing patterns (e.g., interfacial mixing, mixing near vortices due to molecular diffusion).
  \label{Fig:DL_RT_Pred_kfL2_Errors}}
\end{figure}

\begin{figure}
  \centering
  \subfigure[Ground truth at $t = 1.0$]
    {\includegraphics[width = 0.285\textwidth]
    {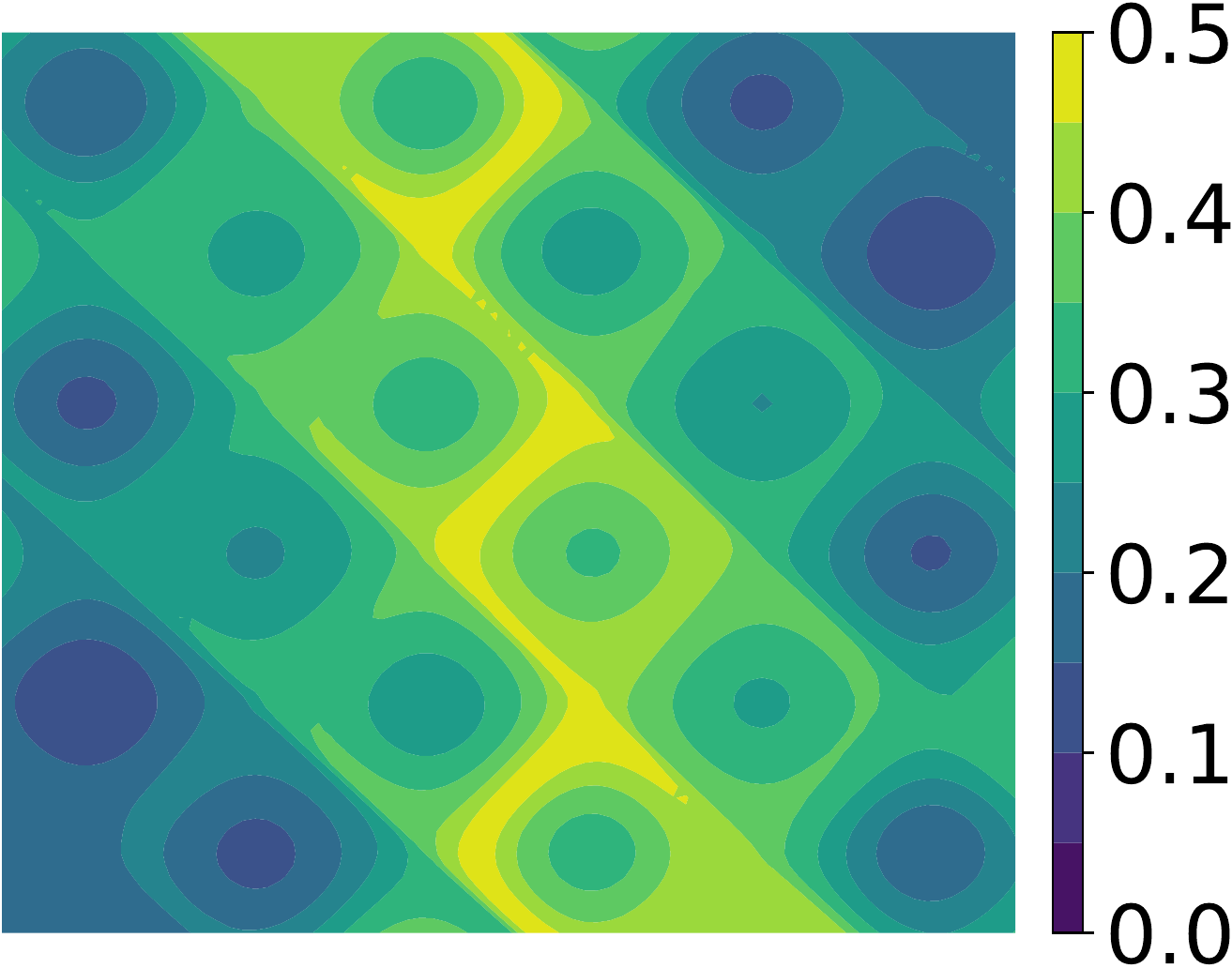}}
  \hspace{3.5in}
  \subfigure[{\tiny Prediction:~8\% data}]
    {\includegraphics[width = 0.2\textwidth]
    {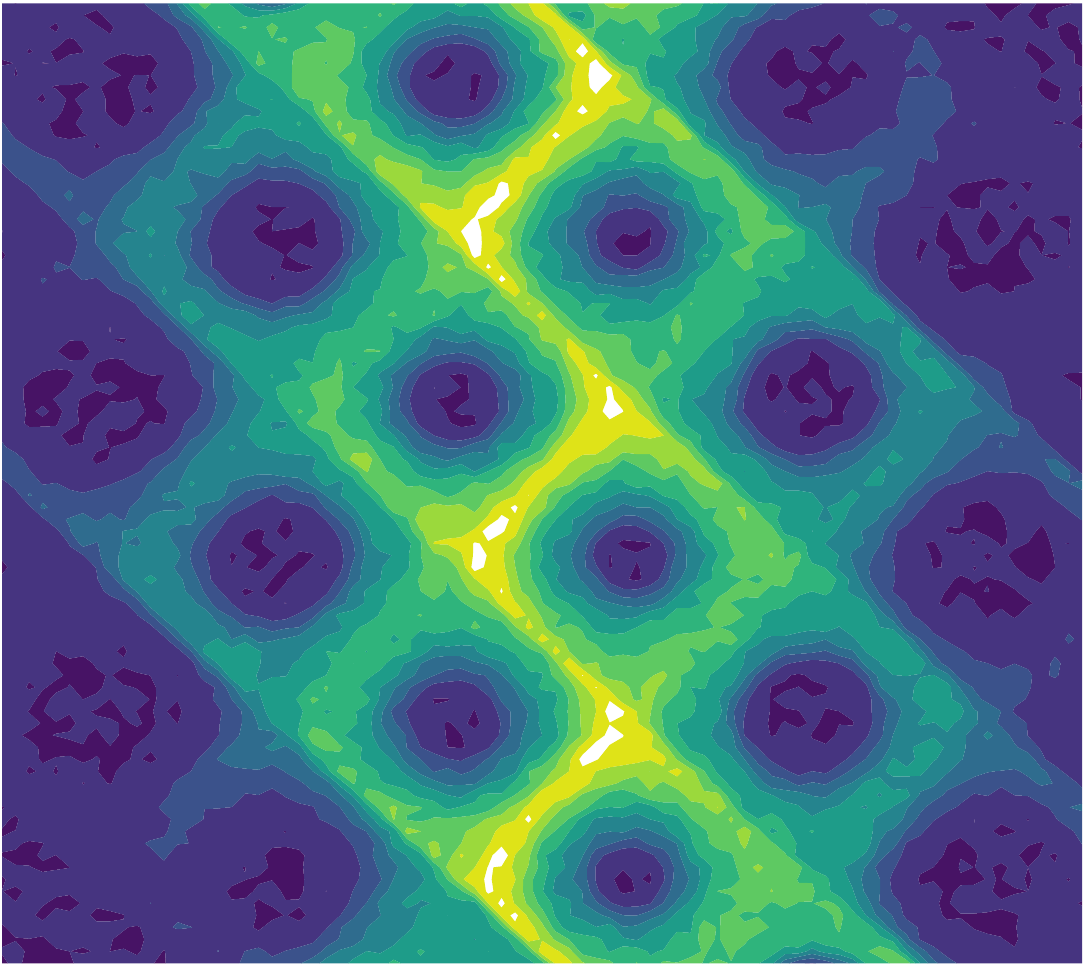}}
  \hspace{0.1in}
  \subfigure[{\tiny Prediction:~16\% data}]
    {\includegraphics[width = 0.2\textwidth]
    {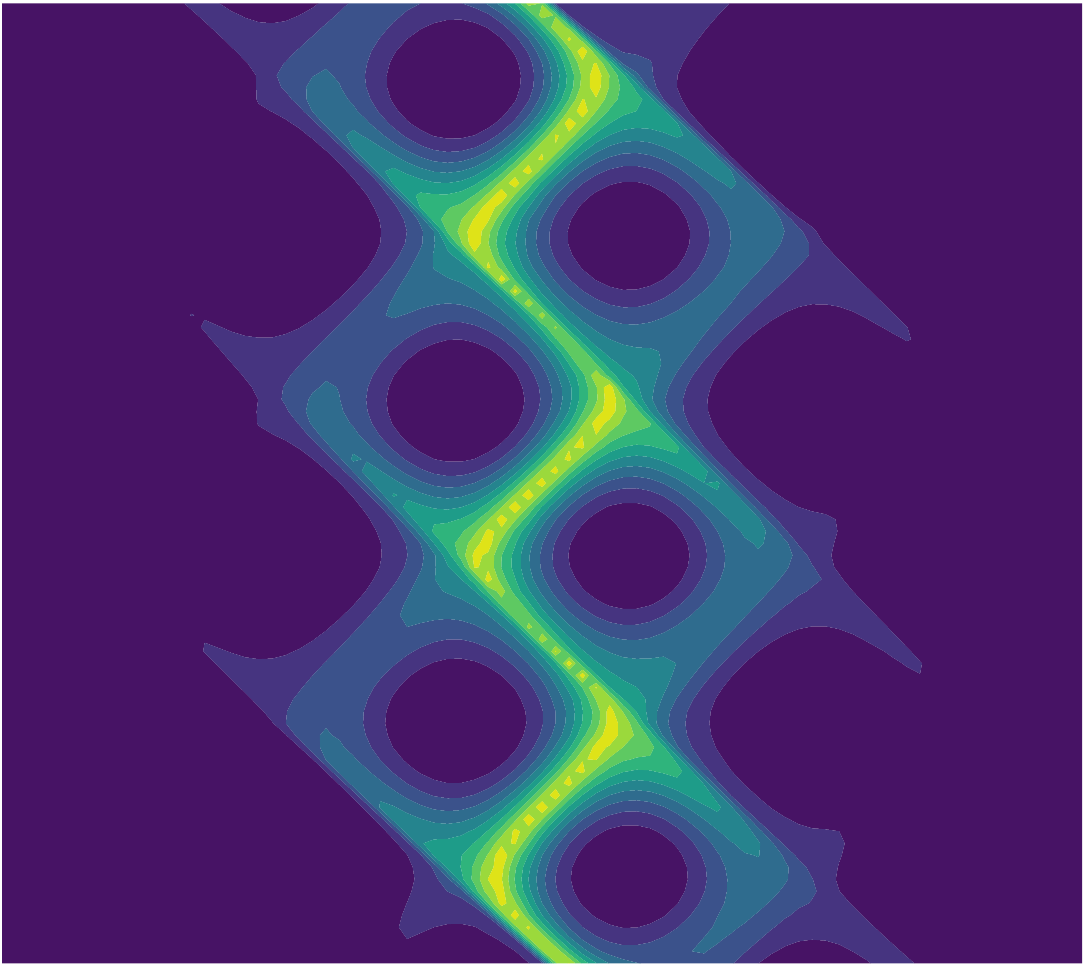}}
  \hspace{0.1in}
  \subfigure[{\tiny Prediction:~24\% data}]
    {\includegraphics[width = 0.2\textwidth]
    {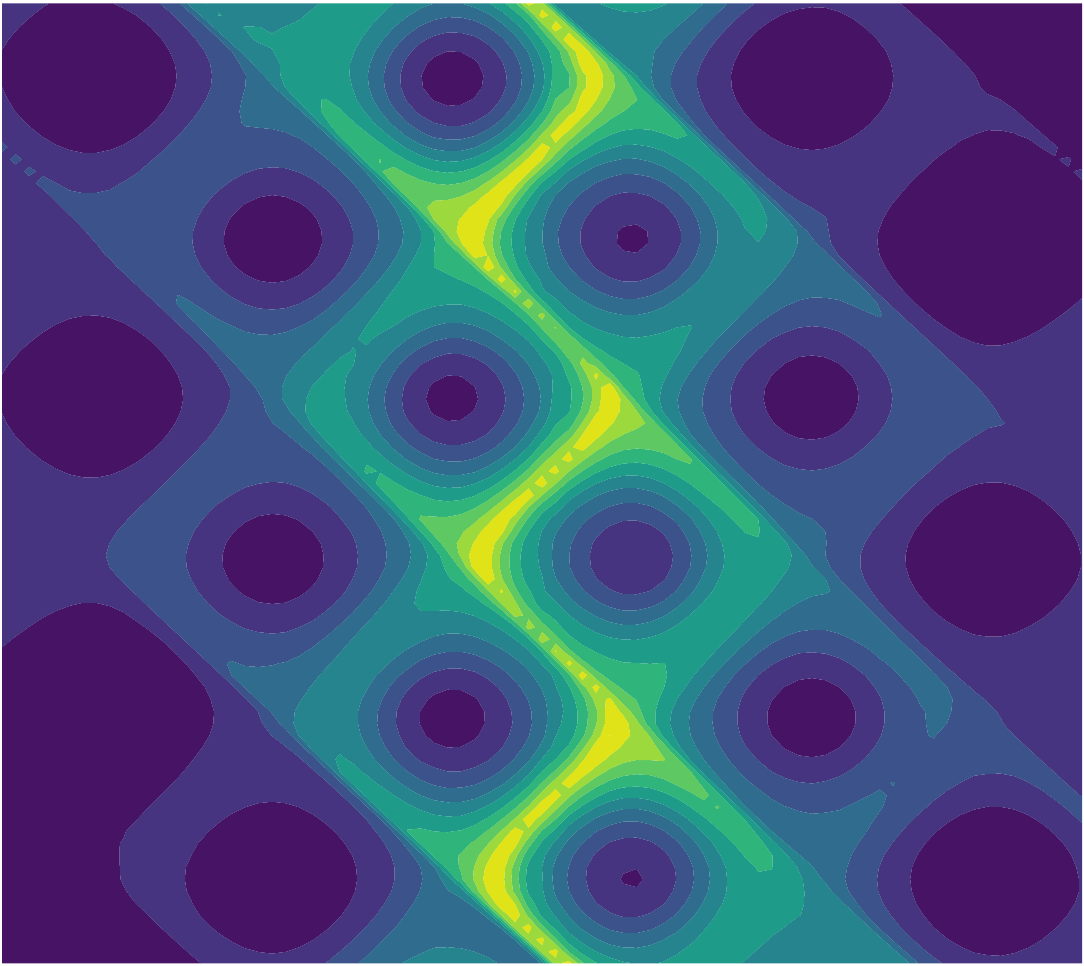}}
  \hspace{0.1in}
  \subfigure[{\tiny Prediction:~32\% data}]
    {\includegraphics[width = 0.2\textwidth]
    {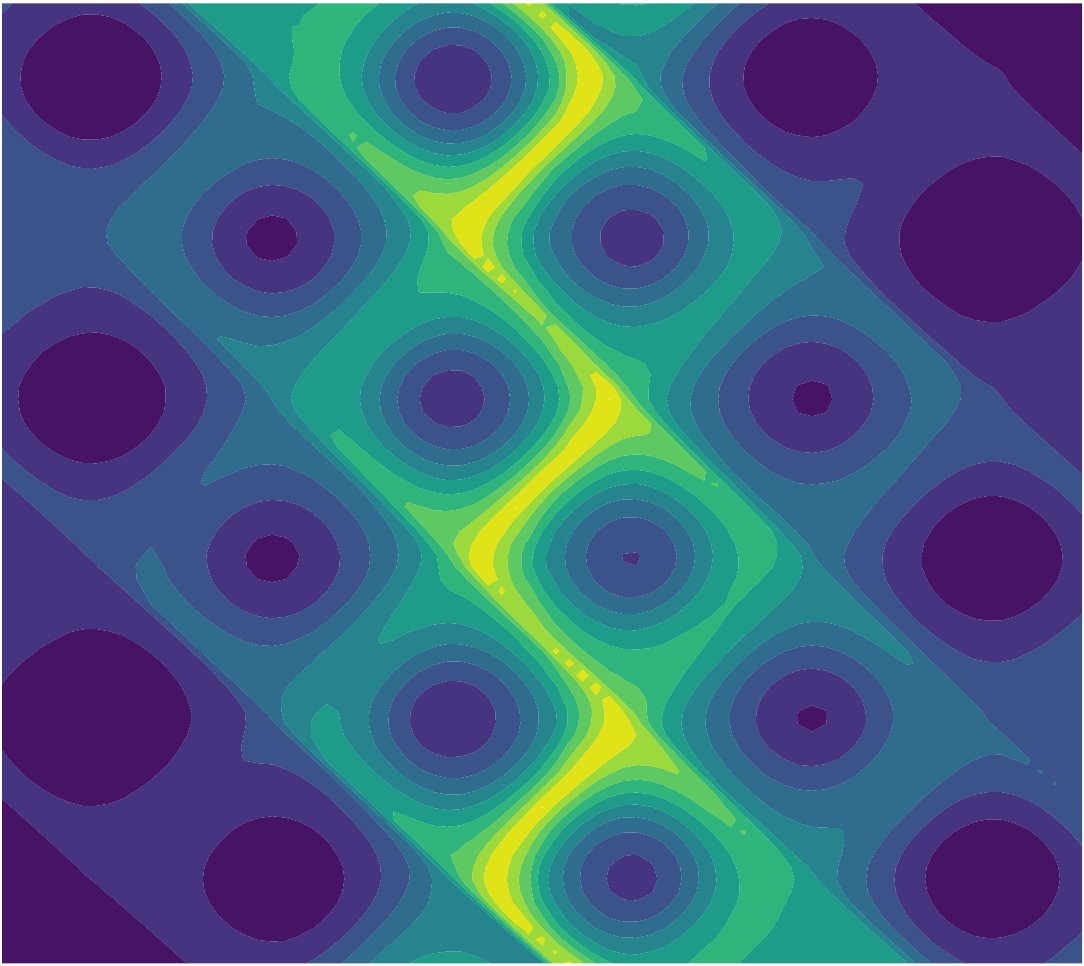}}
  \subfigure[{\tiny Prediction:~40\% data}]
    {\includegraphics[width = 0.2\textwidth]
    {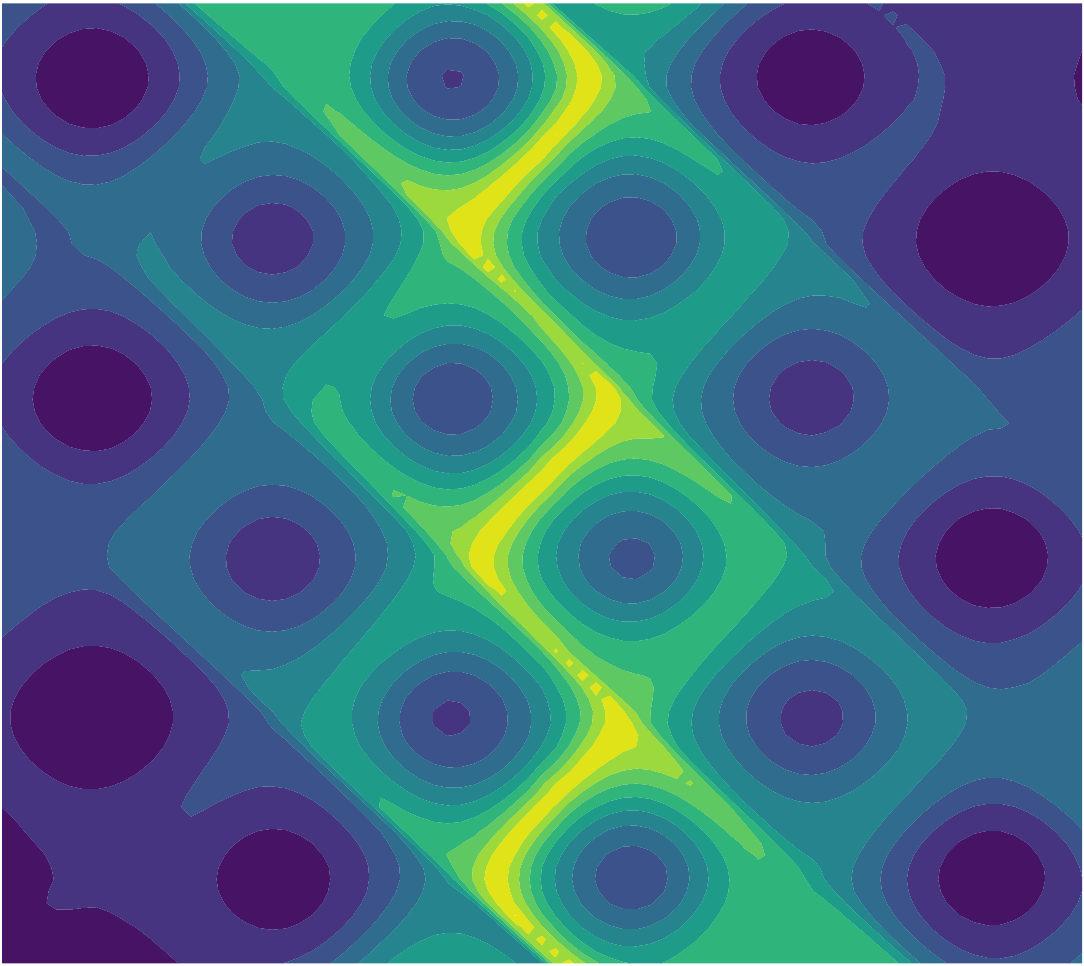}}
  \hspace{0.1in}
  \subfigure[{\tiny Prediction:~48\% data}]
    {\includegraphics[width = 0.2\textwidth]
    {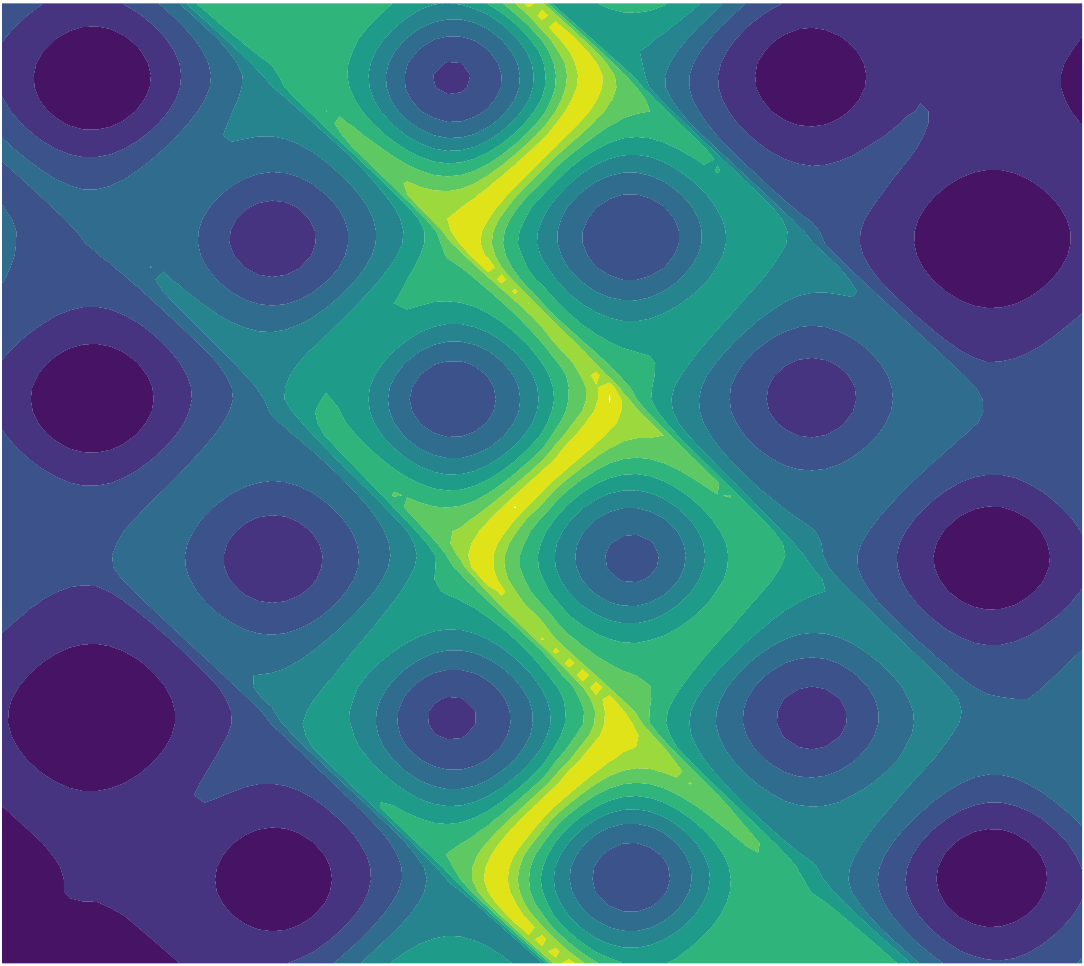}}
  \hspace{0.1in}
  \subfigure[{\tiny Prediction:~56\% data}]
    {\includegraphics[width = 0.2\textwidth]
    {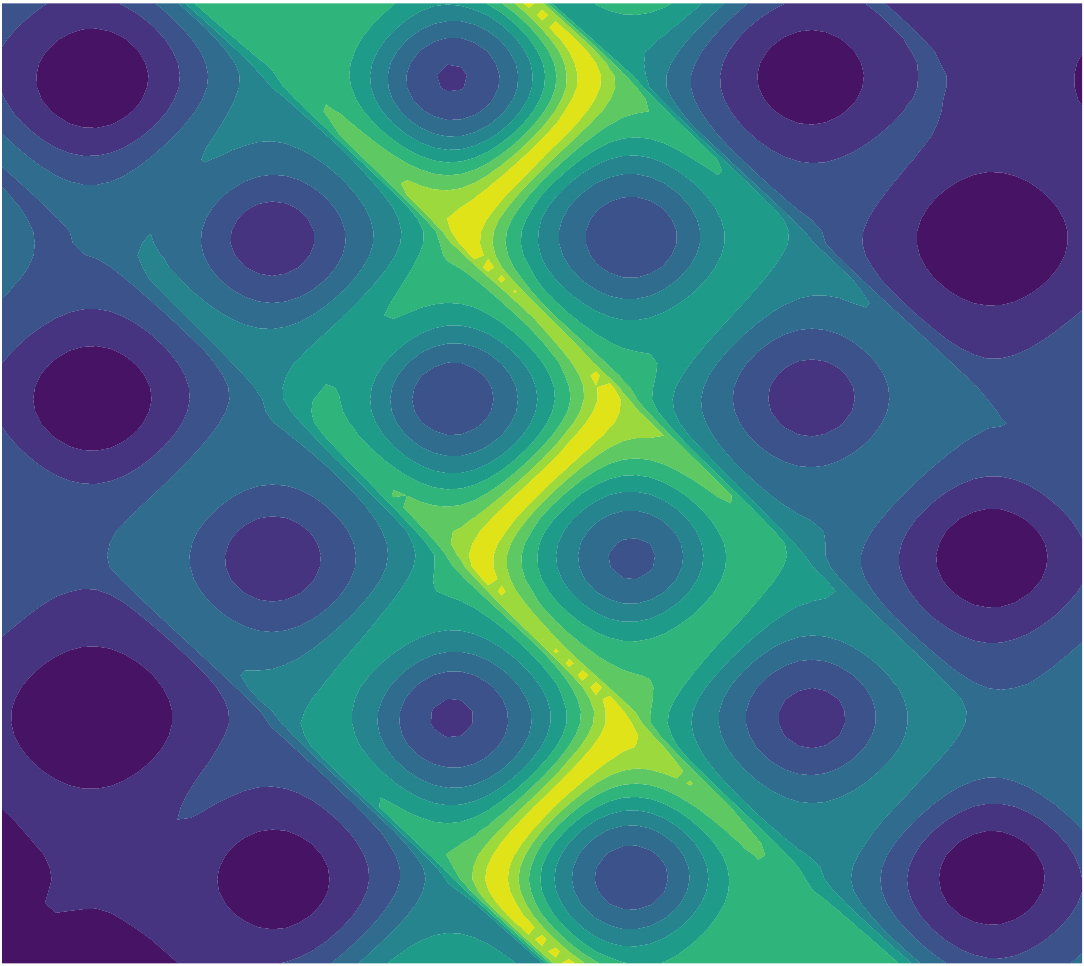}}
  \hspace{0.1in}
  \subfigure[{\tiny Prediction:~64\% data}]
    {\includegraphics[width = 0.2\textwidth]
    {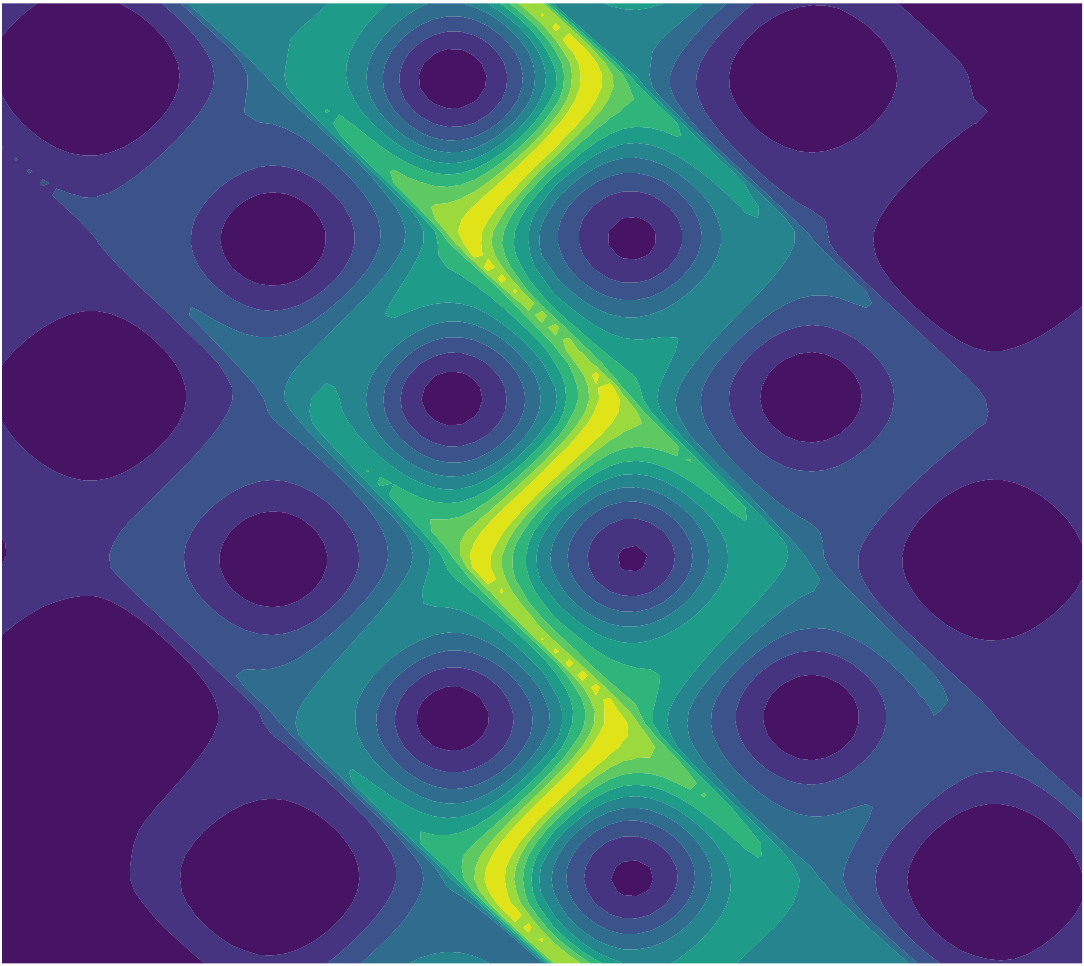}}
  \subfigure[{\tiny Prediction:~72\% data}]
    {\includegraphics[width = 0.2\textwidth]
    {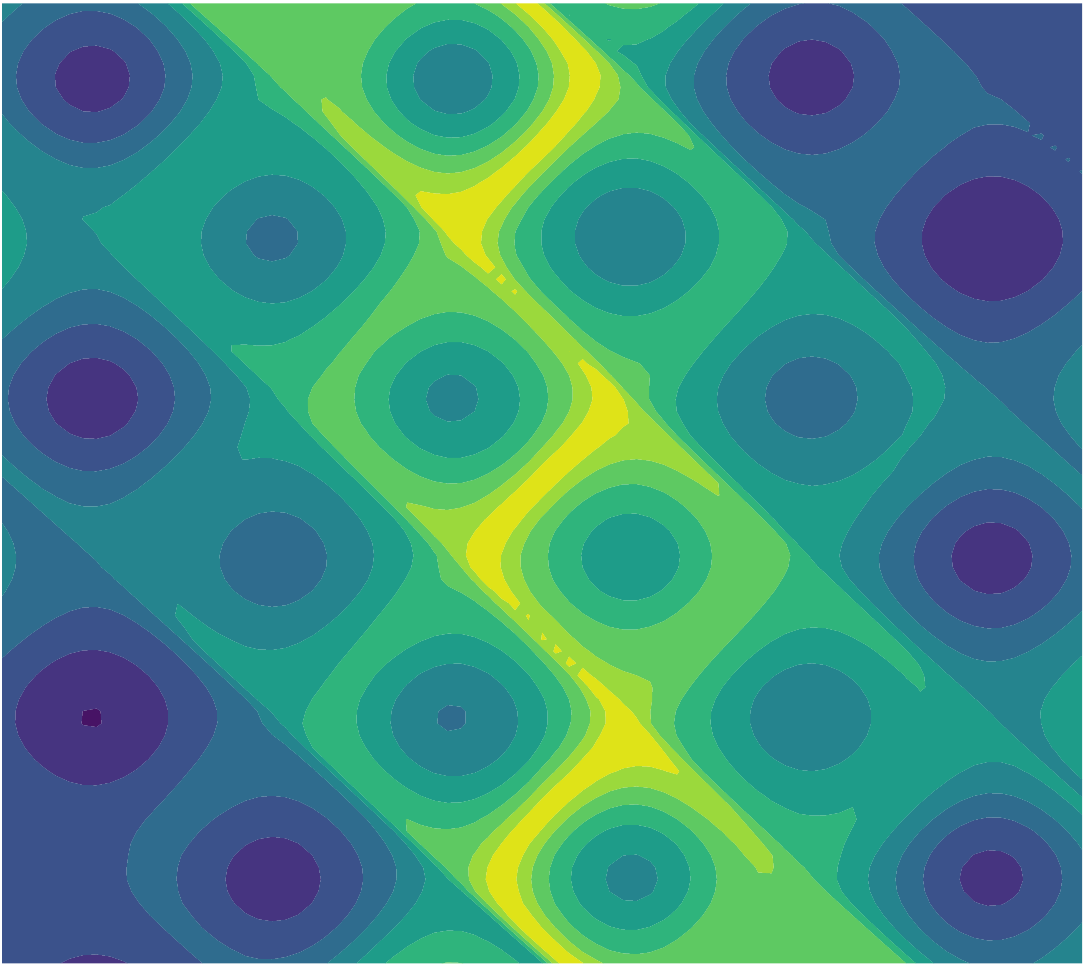}}
  \hspace{0.1in}
  \subfigure[{\tiny Prediction:~80\% data}]
    {\includegraphics[width = 0.2\textwidth]
    {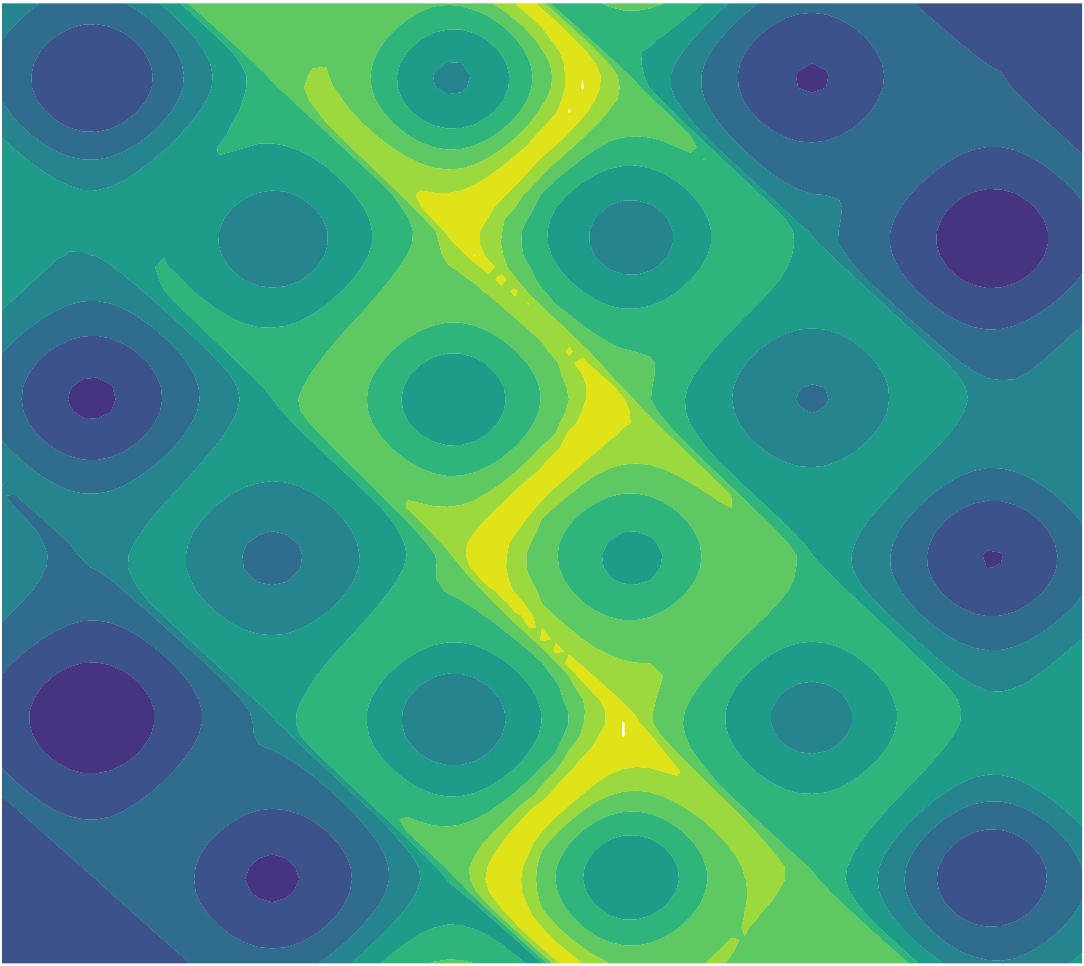}}
  \hspace{0.1in}
  \subfigure[{\tiny Prediction:~88\% data}]
    {\includegraphics[width = 0.2\textwidth]
    {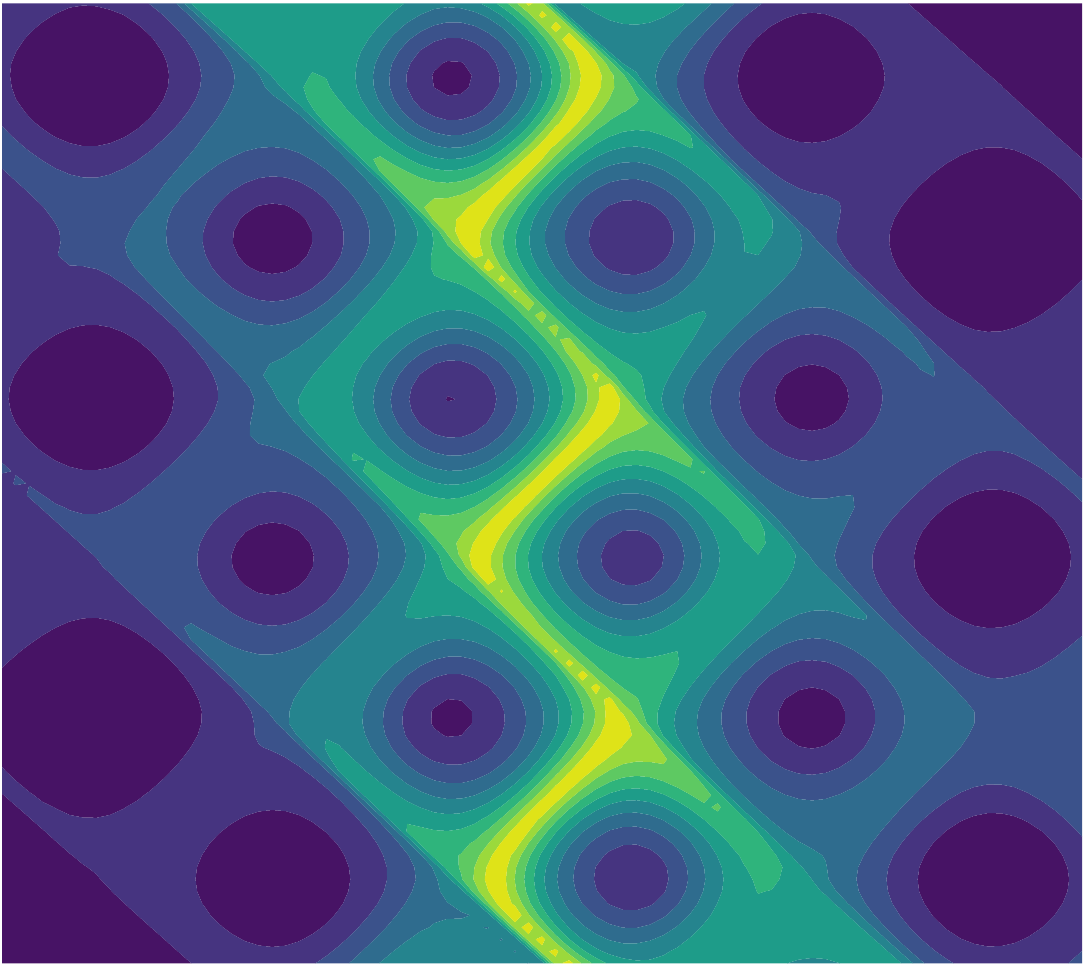}}
  \hspace{0.1in}
  \subfigure[{\tiny Prediction:~96\% data}]
    {\includegraphics[width = 0.2\textwidth]
    {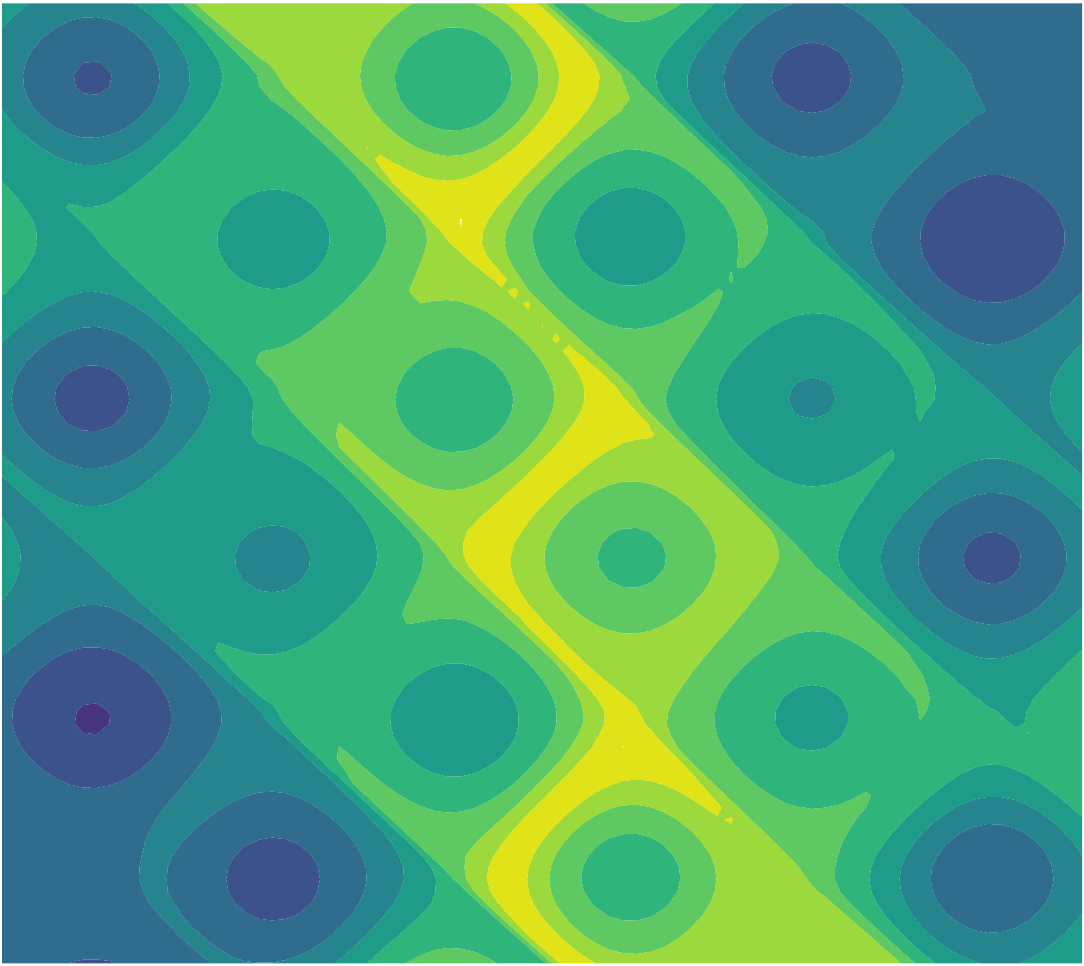}}
  \caption{\textsf{Predictions for $\kappa_fL = 3$:}~This figure compares the ground truth and predictions from the trained non-negative CNN-LSTM models at $t = 1.0$.
  From this figure, it is evident that 80\% data is necessary to accurately capture multi-scale mixing patterns.
  This includes both interfacial mixing and mixing near the vortices.
  \label{Fig:DL_RT_Pred_kfL3}}
\end{figure}

\begin{figure}
  \centering
  \subfigure[{\tiny Prediction error:~8\% data}]
    {\includegraphics[width = 0.295\textwidth]
    {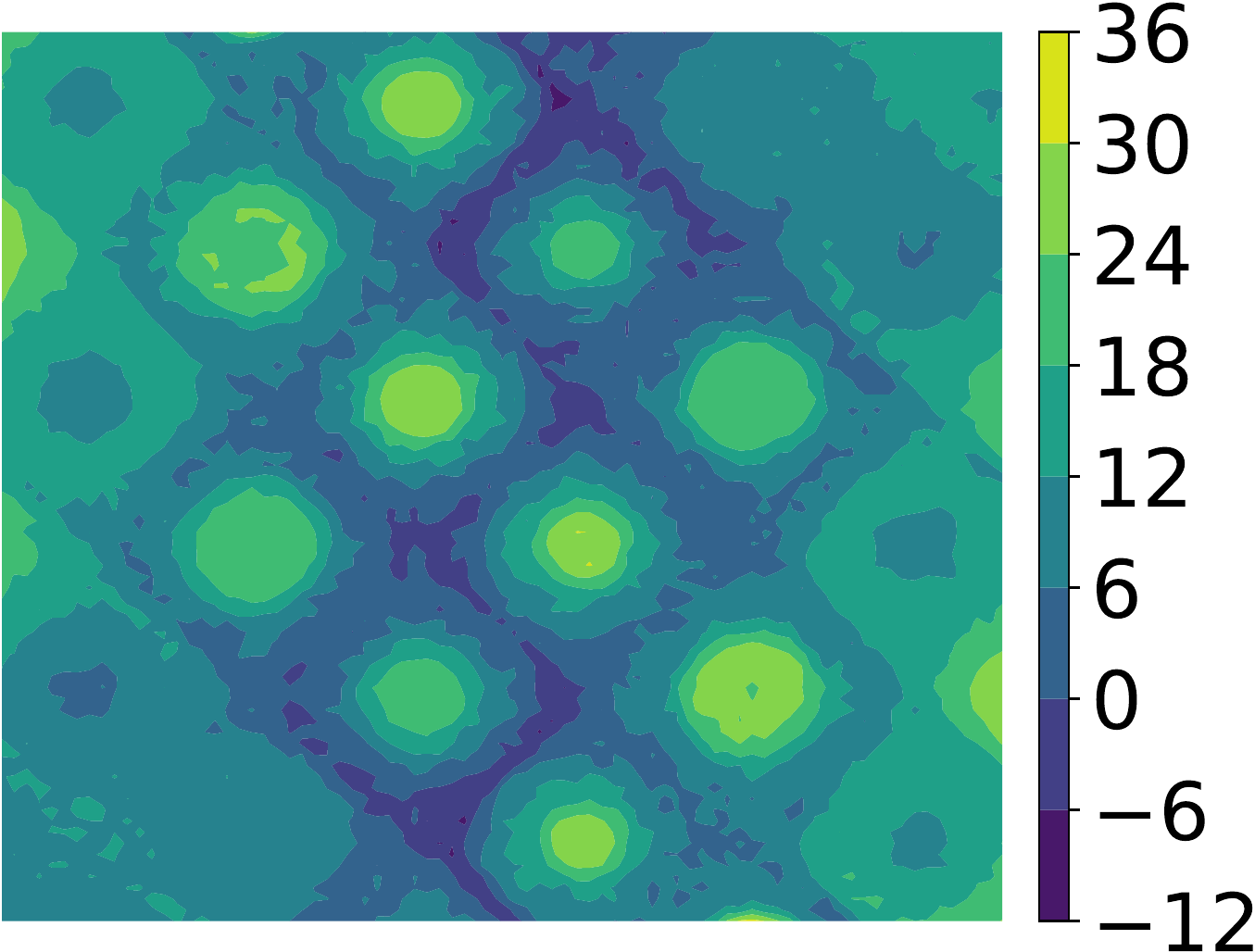}}
  \hspace{0.25in}
  \subfigure[{\tiny Prediction error:~16\% data}]
    {\includegraphics[width = 0.285\textwidth]
    {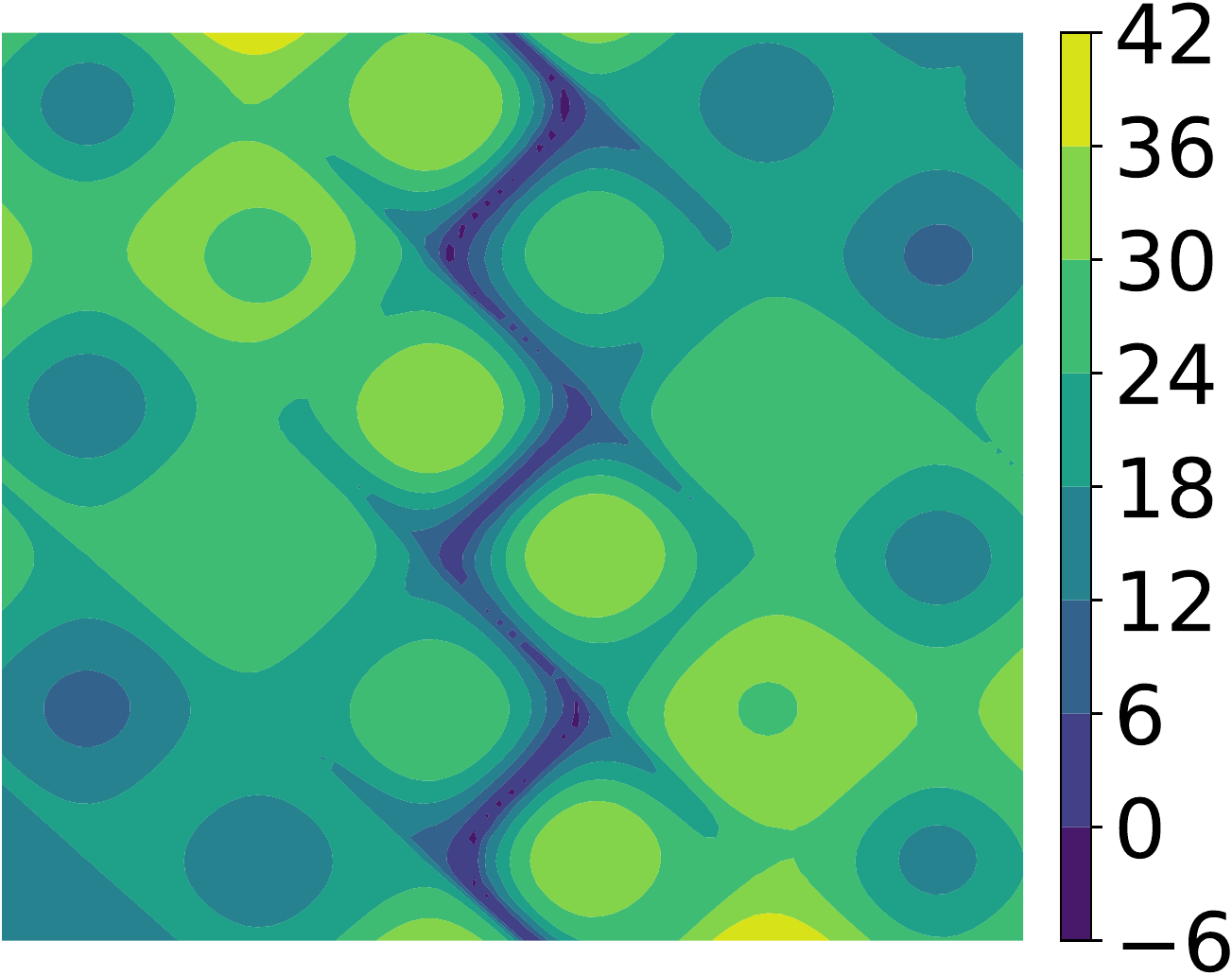}}
  \hspace{0.25in}
  \subfigure[{\tiny Prediction error:~24\% data}]
    {\includegraphics[width = 0.285\textwidth]
    {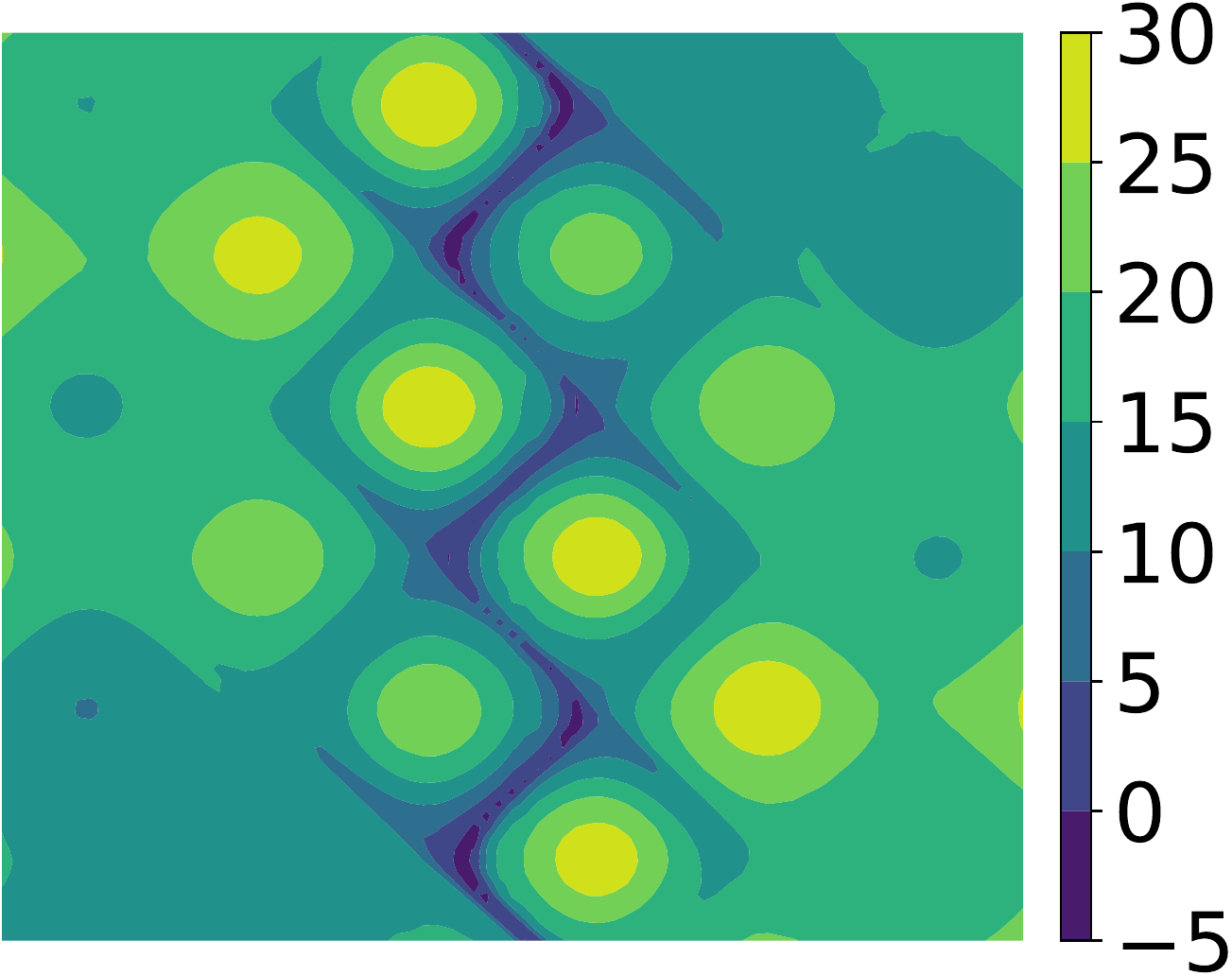}}
  \subfigure[{\tiny Prediction error:~32\% data}]
    {\includegraphics[width = 0.285\textwidth]
    {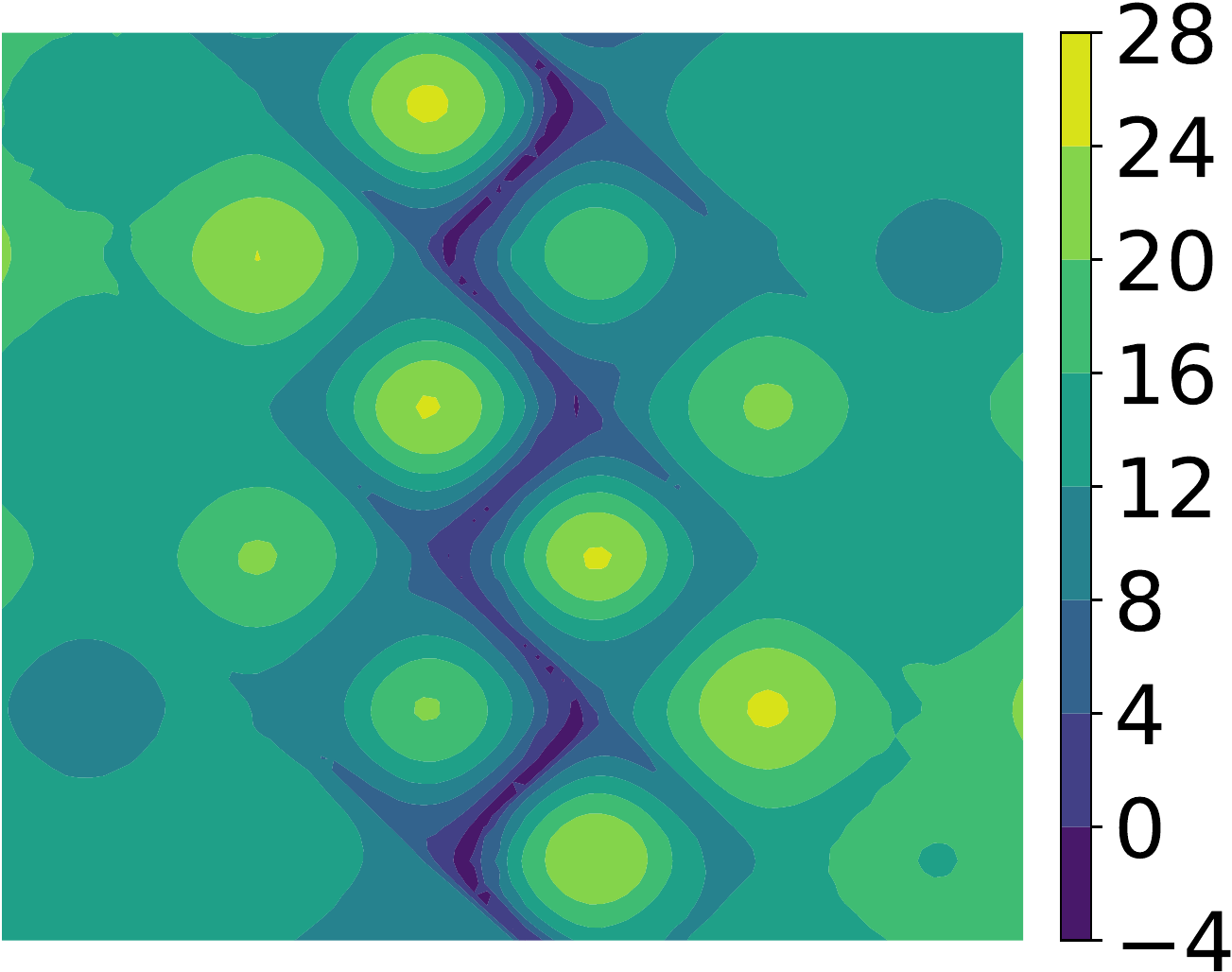}}
  \hspace{0.25in}
  \subfigure[{\tiny Prediction error:~40\% data}]
    {\includegraphics[width = 0.285\textwidth]
    {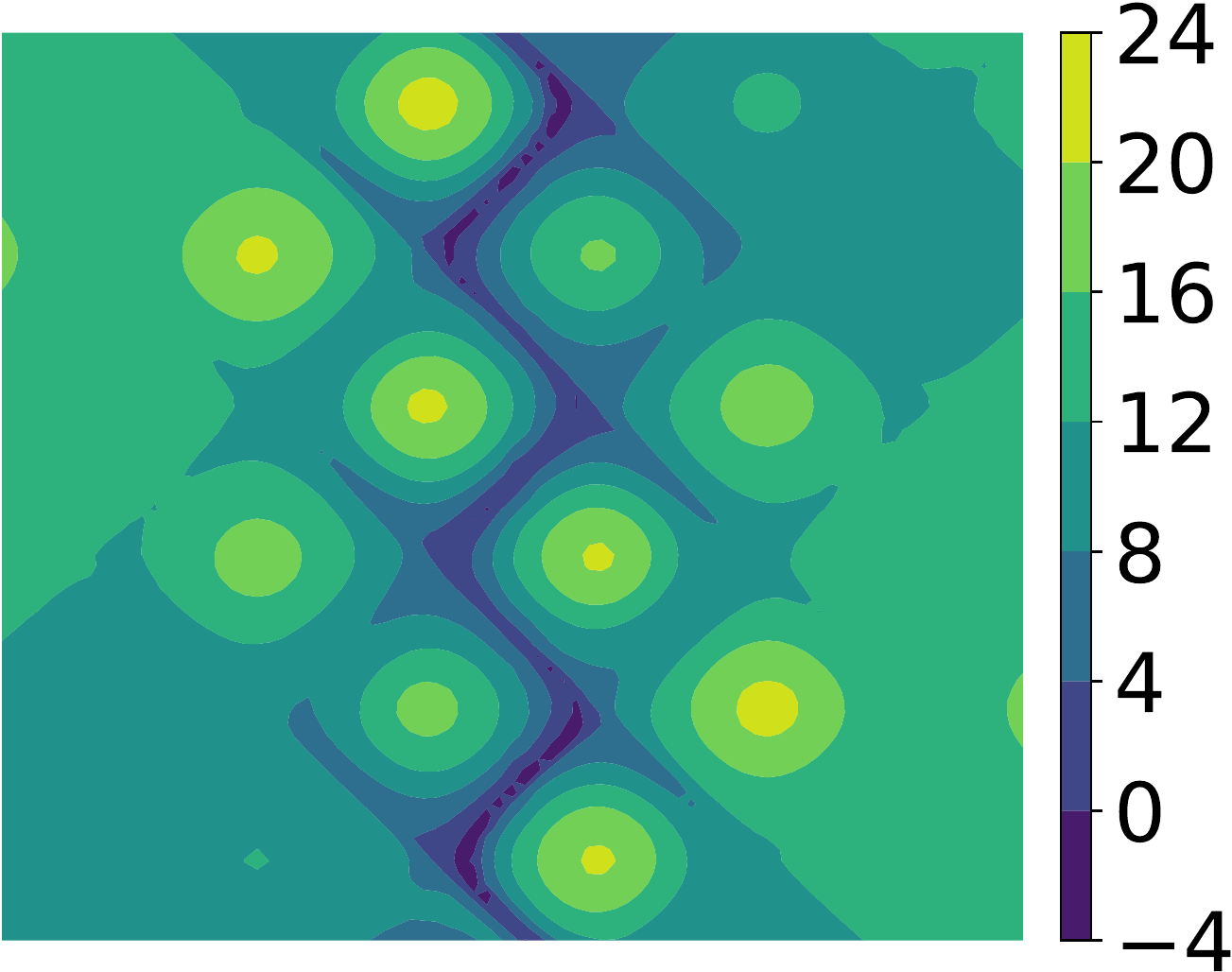}}
  \hspace{0.25in}
  \subfigure[{\tiny Prediction error:~48\% data}]
    {\includegraphics[width = 0.285\textwidth]
    {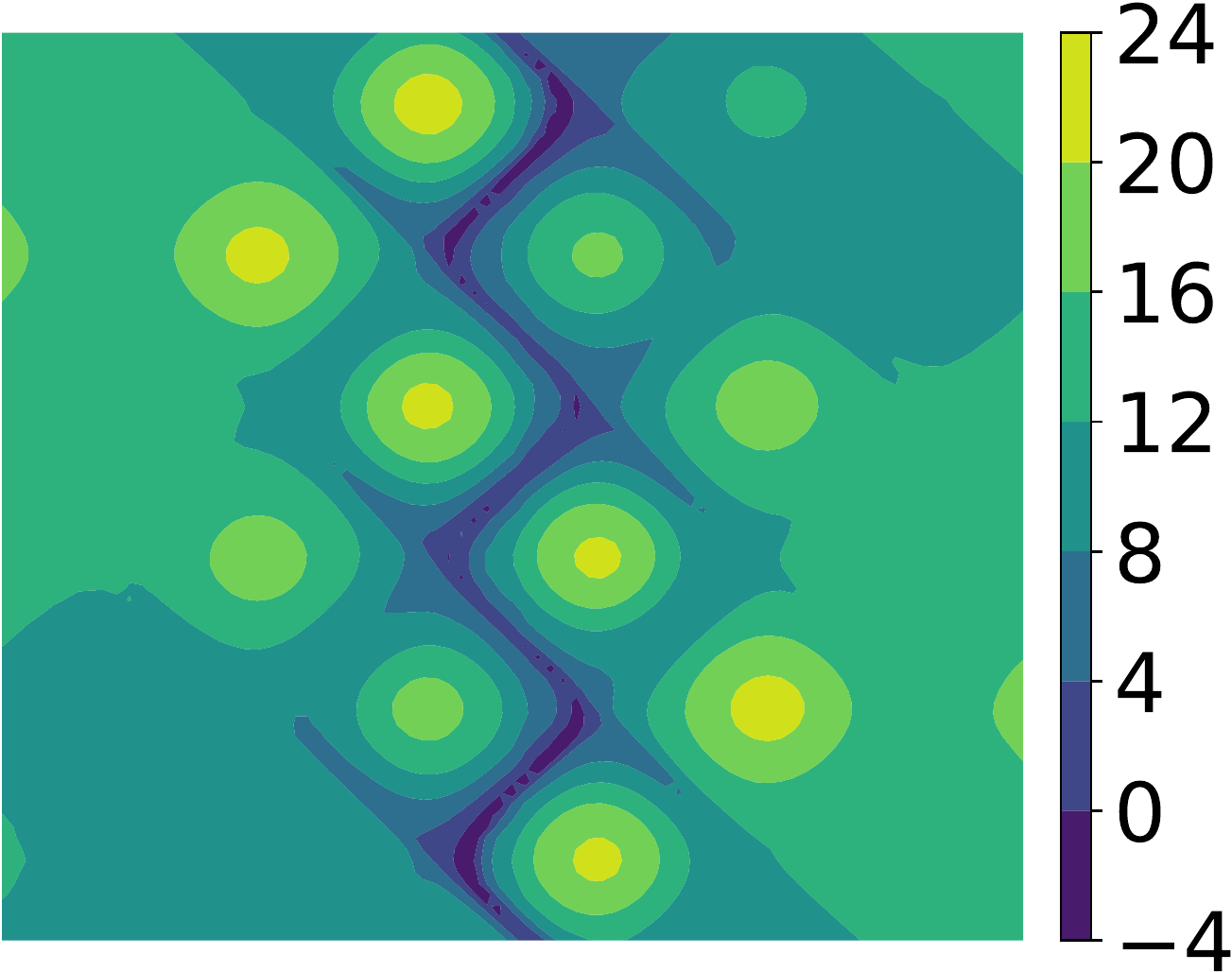}}
  \subfigure[{\tiny Prediction error:~56\% data}]
    {\includegraphics[width = 0.285\textwidth]
    {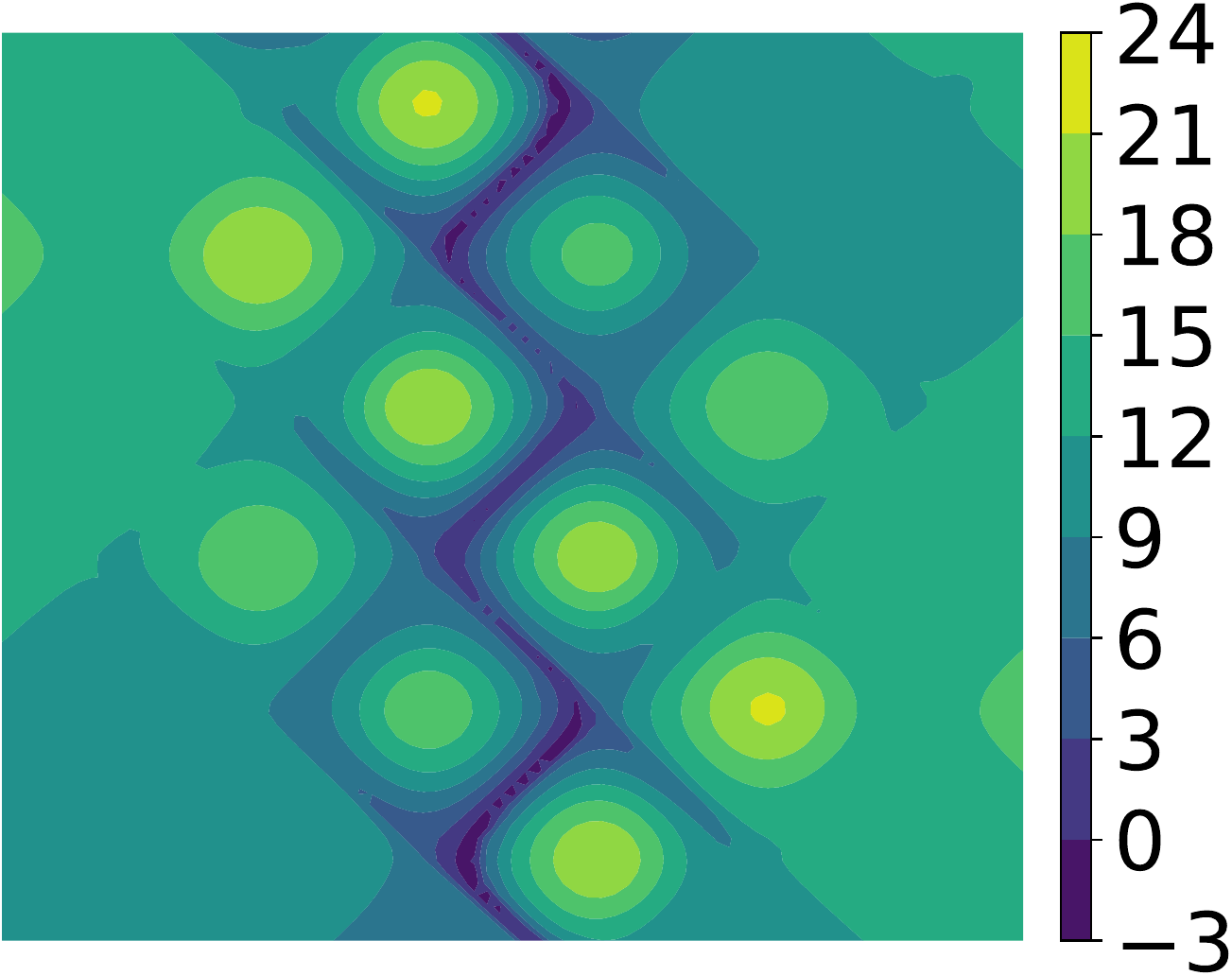}}
  \hspace{0.25in}
  \subfigure[{\tiny Prediction error:~64\% data}]
    {\includegraphics[width = 0.285\textwidth]
    {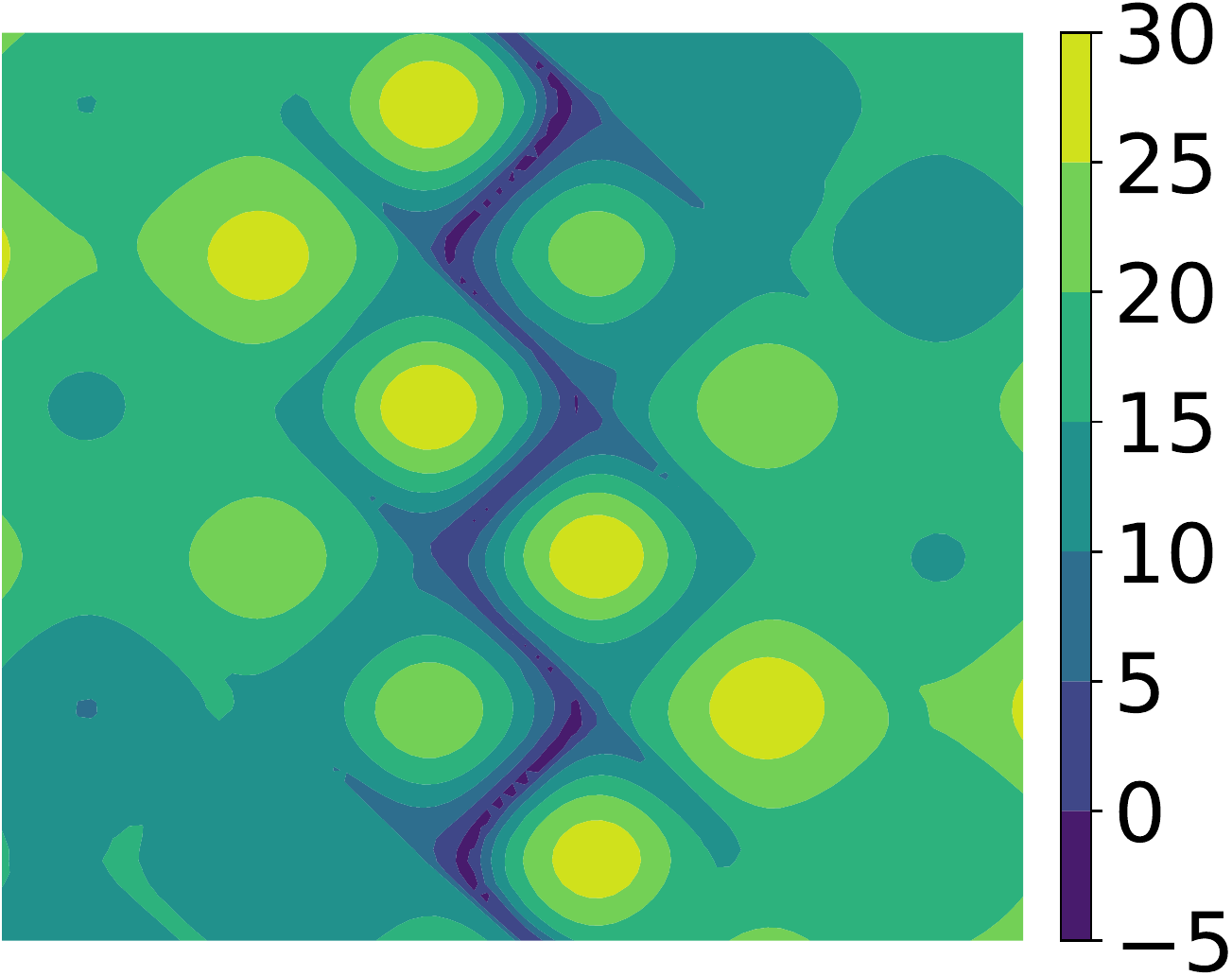}}
  \hspace{0.25in}
  \subfigure[{\tiny Prediction error:~72\% data}]
    {\includegraphics[width = 0.30\textwidth]
    {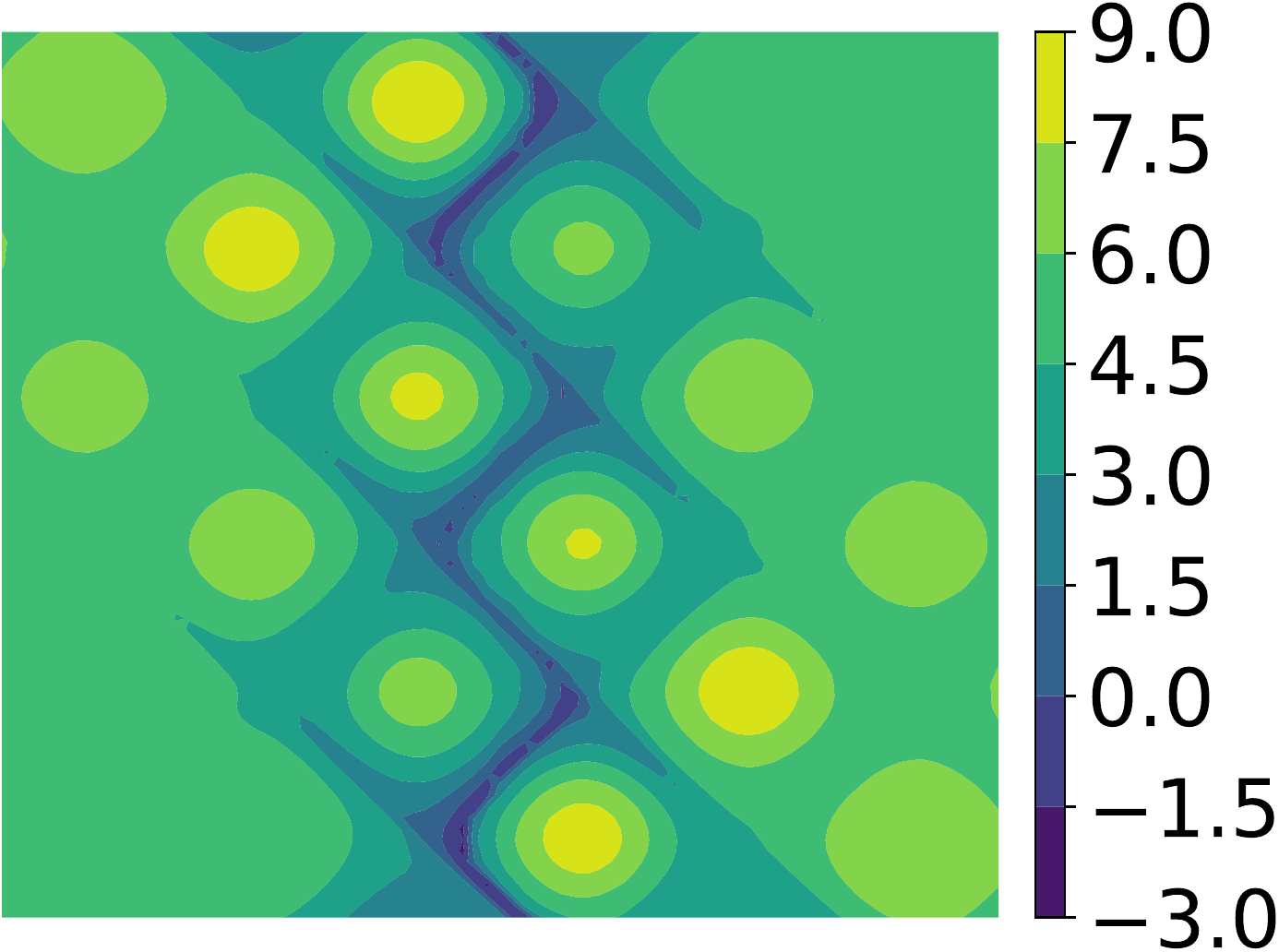}}
  \subfigure[{\tiny Prediction error:~80\% data}]
    {\includegraphics[width = 0.295\textwidth]
    {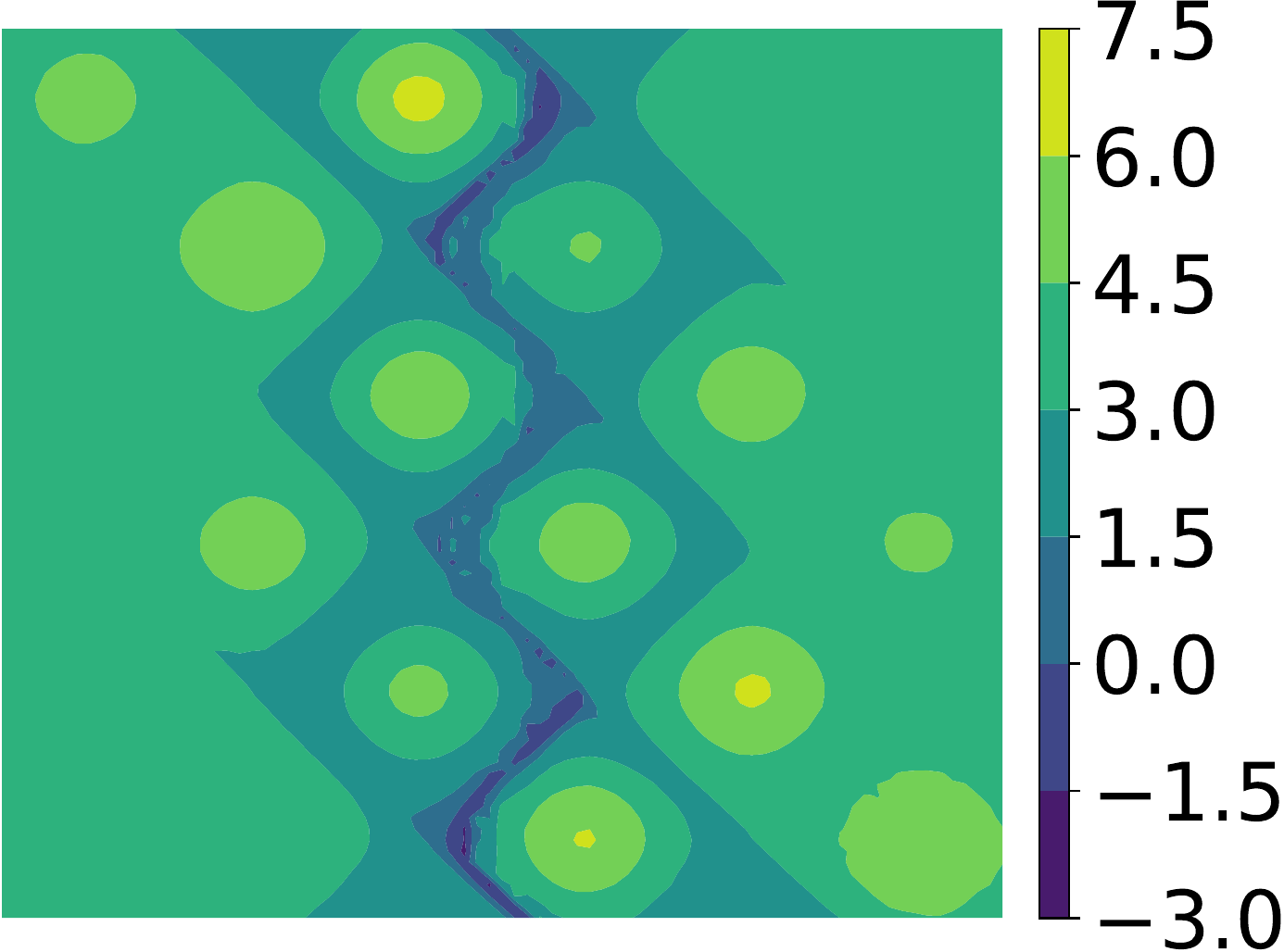}}
  \hspace{0.25in}
  \subfigure[{\tiny Prediction error:~88\% data}]
    {\includegraphics[width = 0.285\textwidth]
    {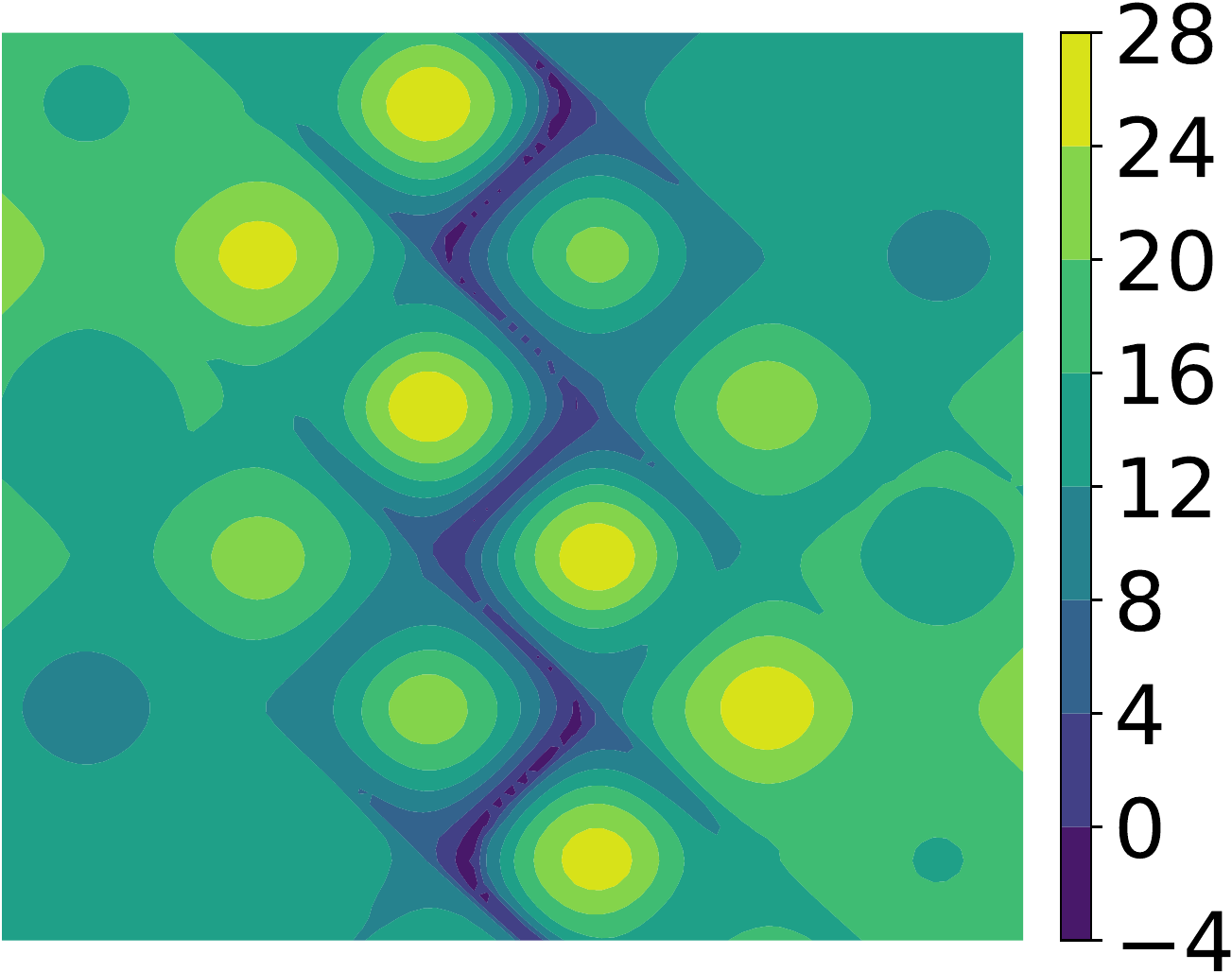}}
  \hspace{0.25in}
  \subfigure[{\tiny Prediction error:~96\% data}]
    {\includegraphics[width = 0.30\textwidth]
    {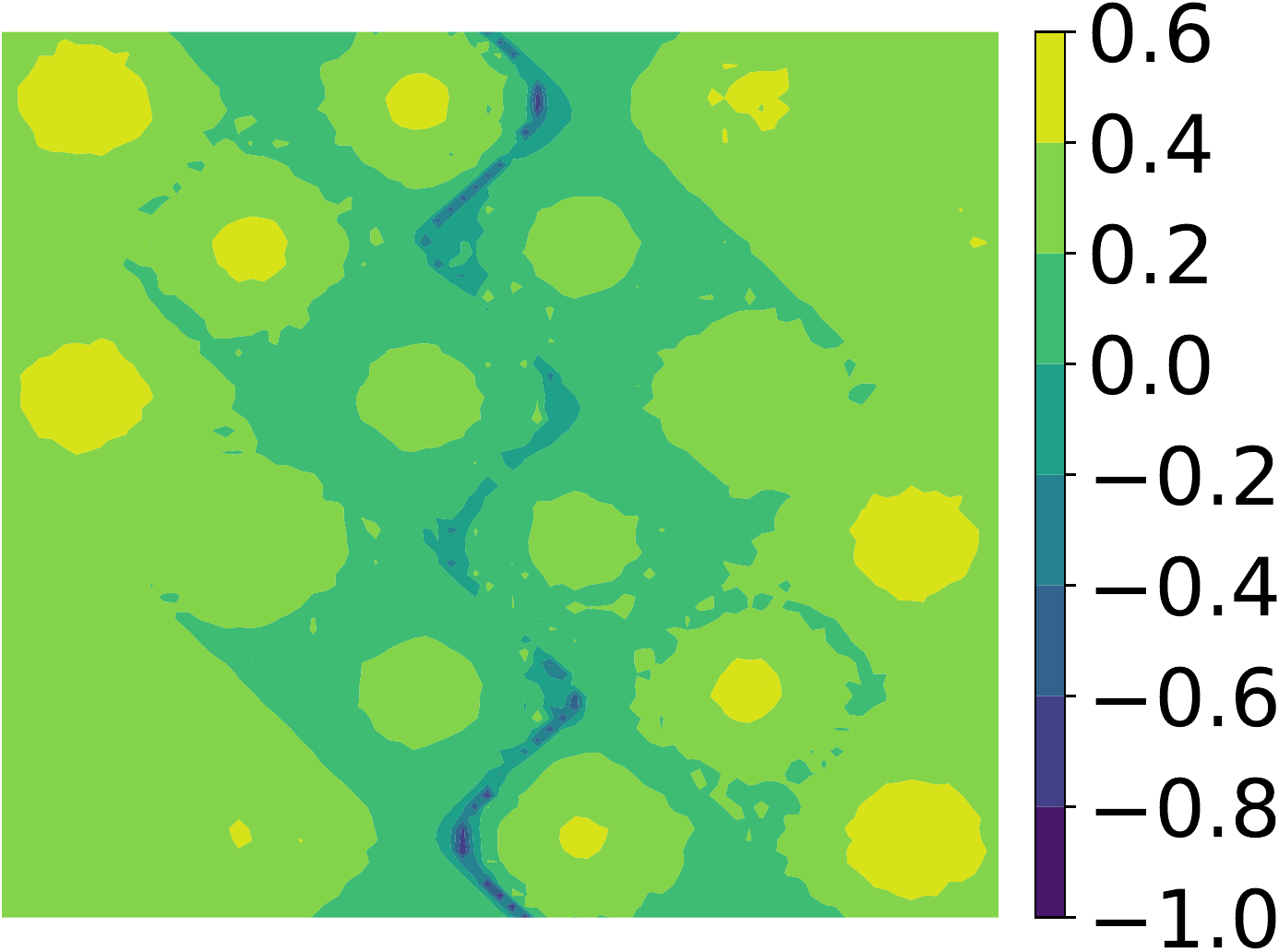}}
  \caption{\textsf{Prediction error in percentage for $\kappa_fL = 3$:}~This figure compares the prediction errors in the entire domain at $t = 1.0$ for different amount of training data.
  Based on the error values (e.g., $\leq 10\%$), it is evident that with less than 80\% of training data is needed to accurately forecast product $C$ formation.
  \label{Fig:DL_RT_Pred_kfL3_Errors}}
\end{figure}

\begin{figure}
  \centering
  \subfigure[Ground truth at $t = 1.0$]
    {\includegraphics[width = 0.285\textwidth]
    {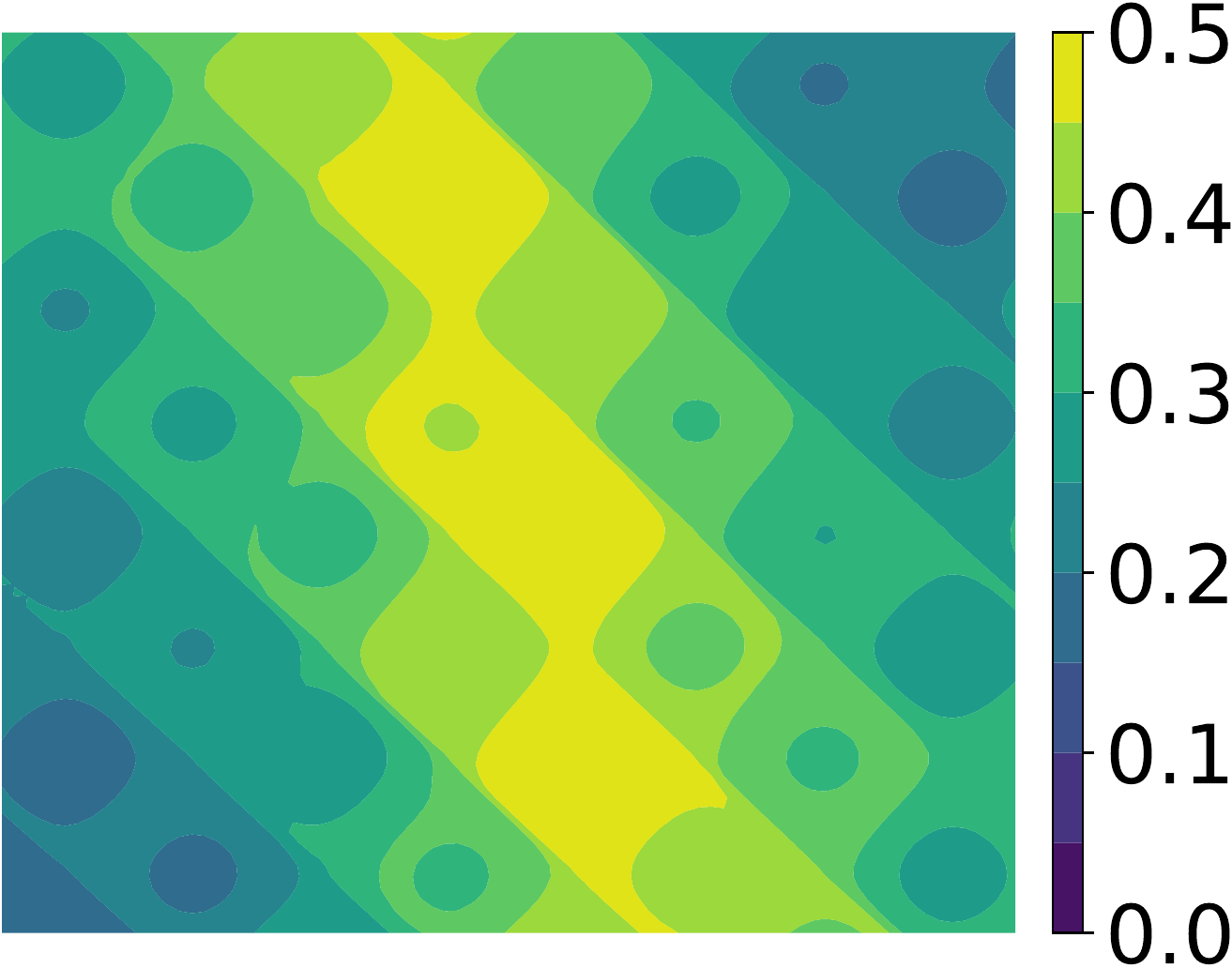}}
  \hspace{3.5in}
  \subfigure[{\tiny Prediction:~8\% data}]
    {\includegraphics[width = 0.2\textwidth]
    {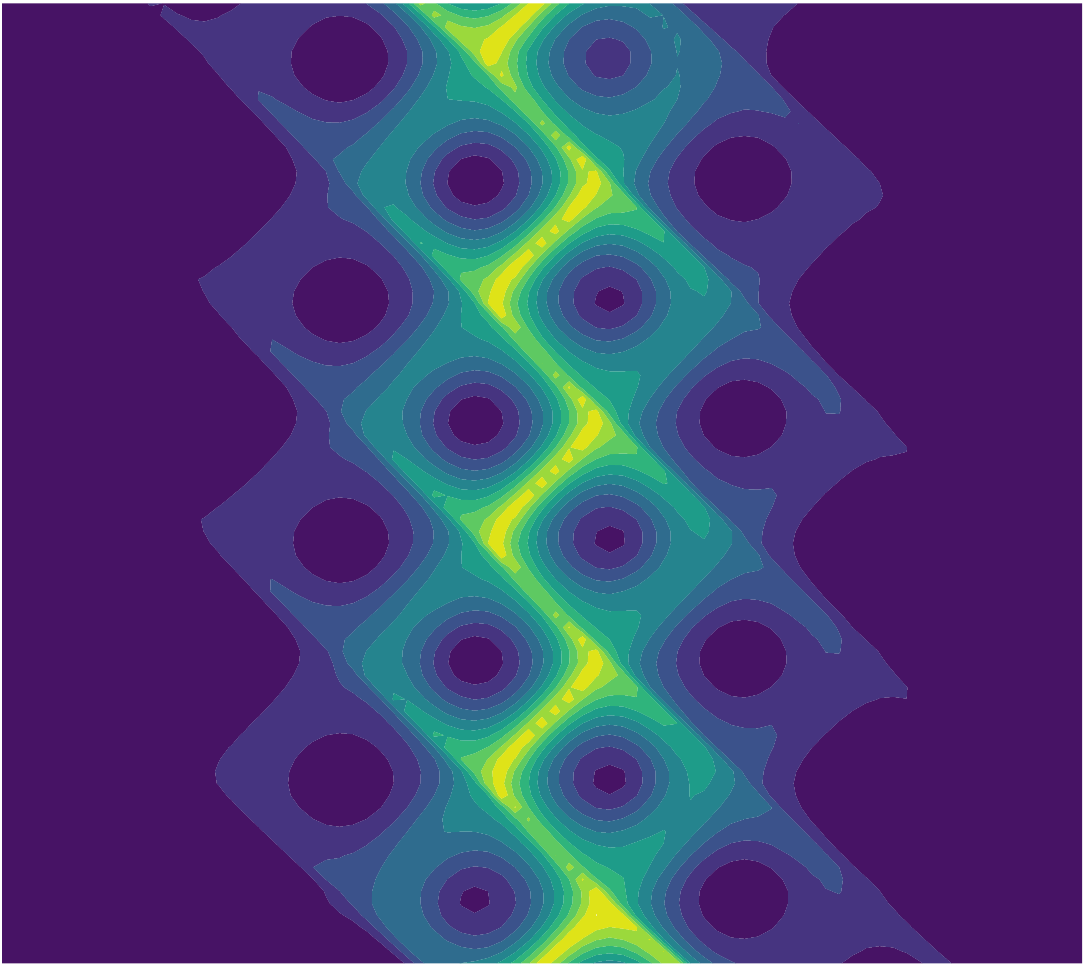}}
  \hspace{0.1in}
  \subfigure[{\tiny Prediction:~16\% data}]
    {\includegraphics[width = 0.2\textwidth]
    {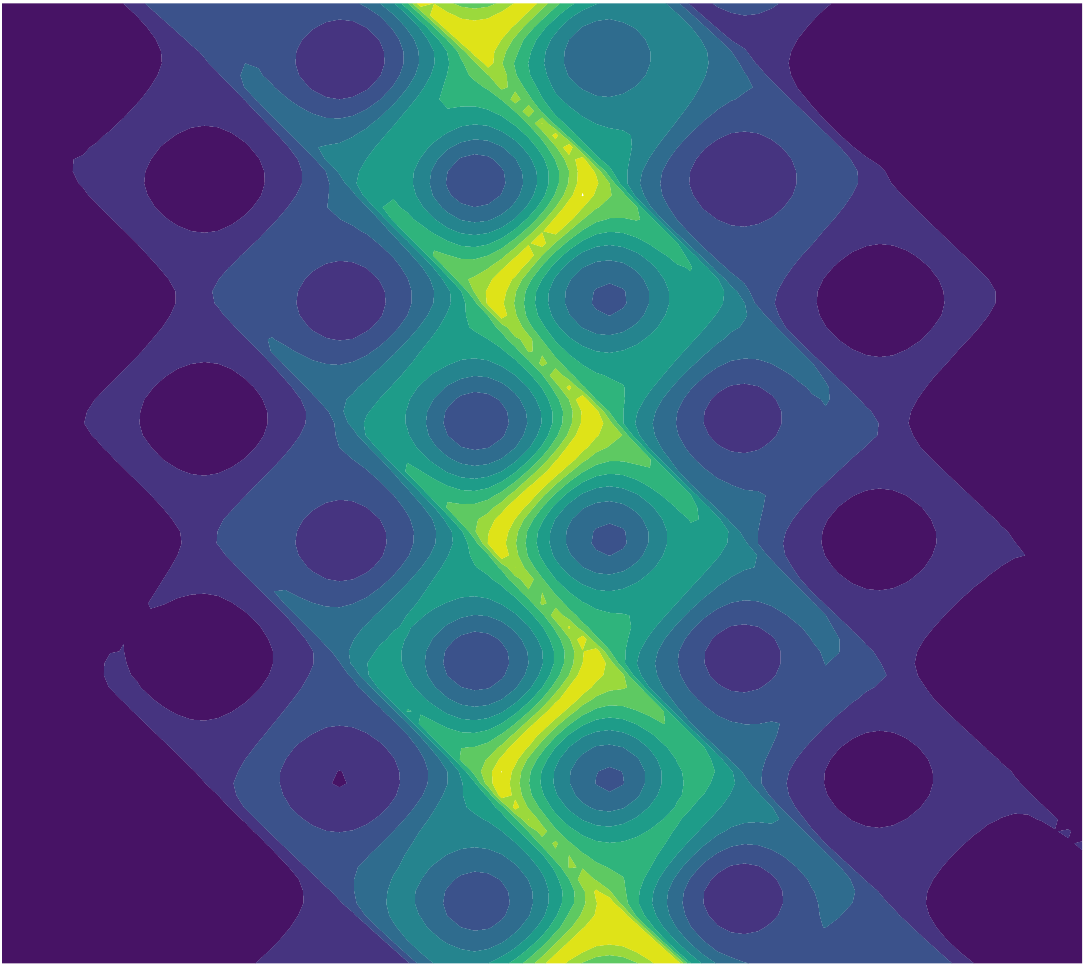}}
  \hspace{0.1in}
  \subfigure[{\tiny Prediction:~24\% data}]
    {\includegraphics[width = 0.2\textwidth]
    {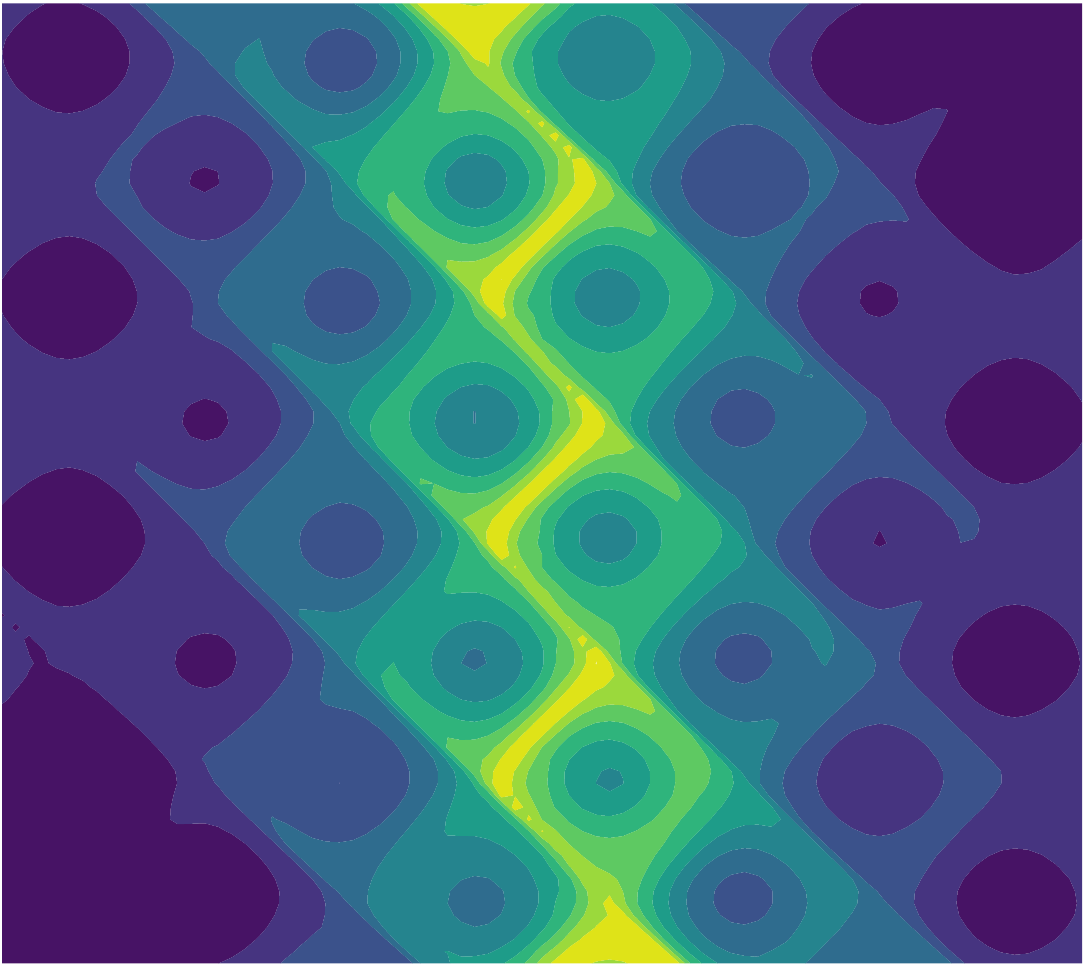}}
  \hspace{0.1in}
  \subfigure[{\tiny Prediction:~32\% data}]
    {\includegraphics[width = 0.2\textwidth]
    {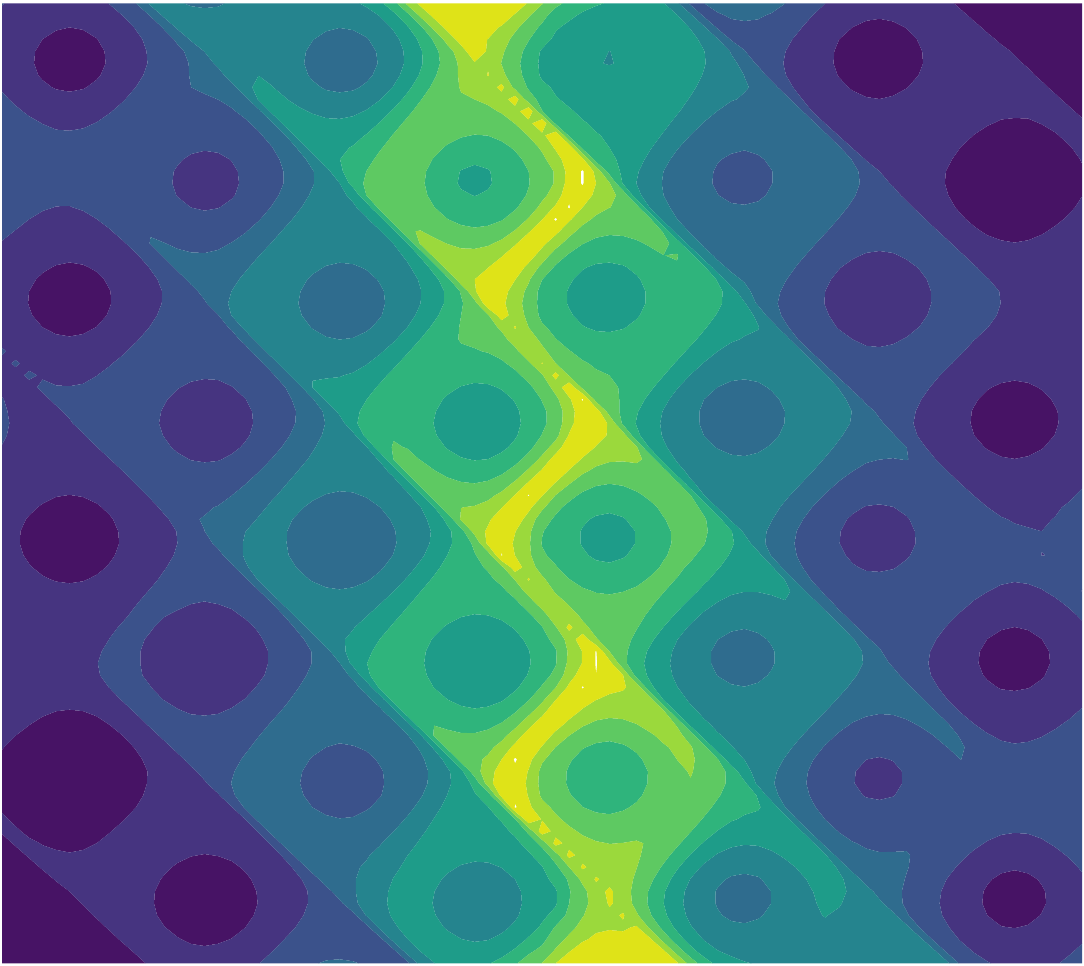}}
  \subfigure[{\tiny Prediction:~40\% data}]
    {\includegraphics[width = 0.2\textwidth]
    {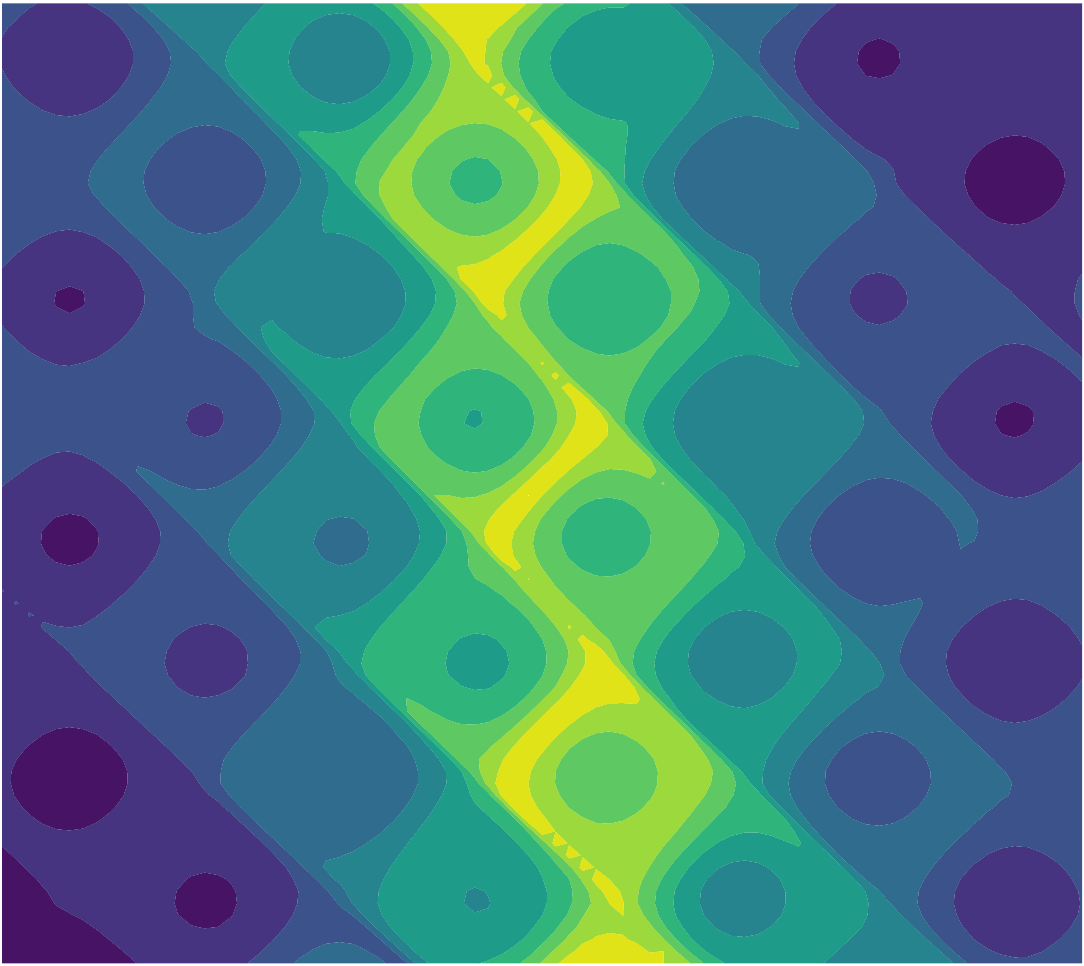}}
  \hspace{0.1in}
  \subfigure[{\tiny Prediction:~48\% data}]
    {\includegraphics[width = 0.2\textwidth]
    {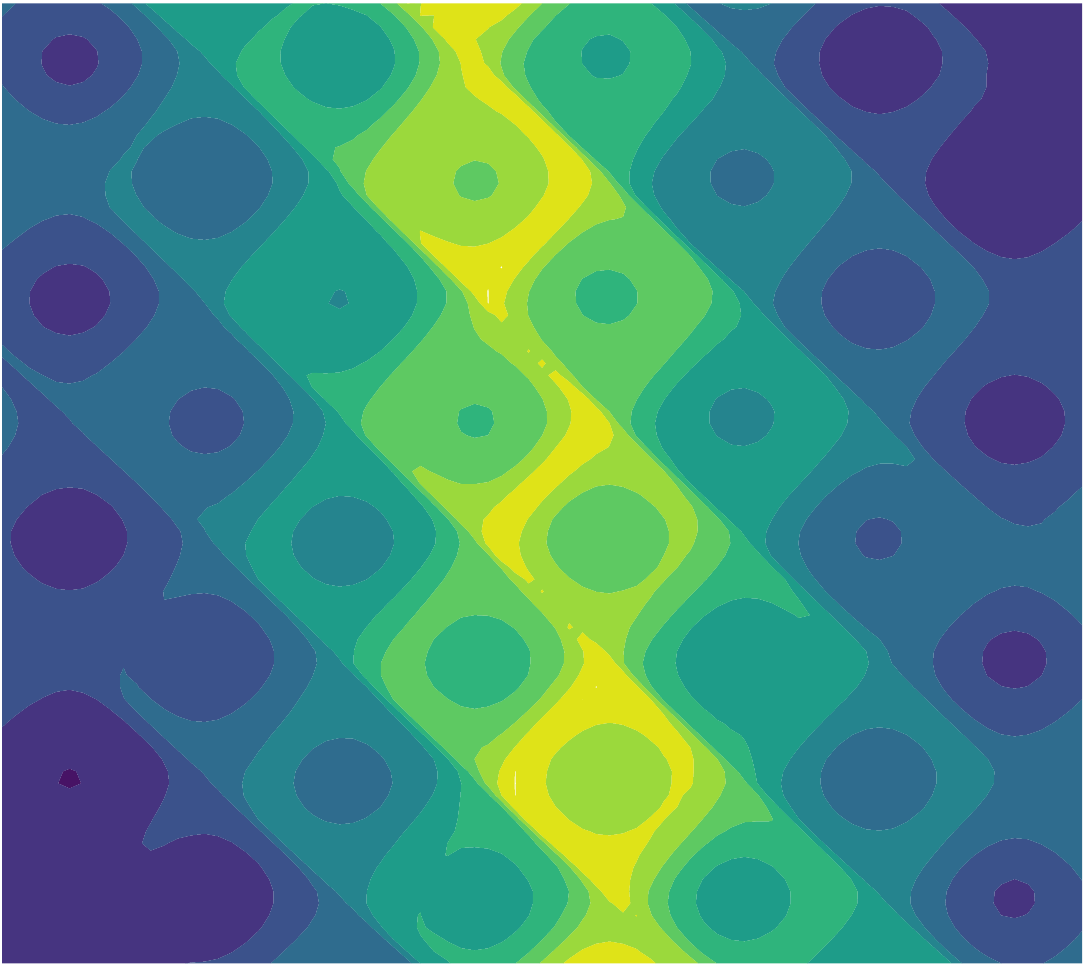}}
  \hspace{0.1in}
  \subfigure[{\tiny Prediction:~56\% data}]
    {\includegraphics[width = 0.2\textwidth]
    {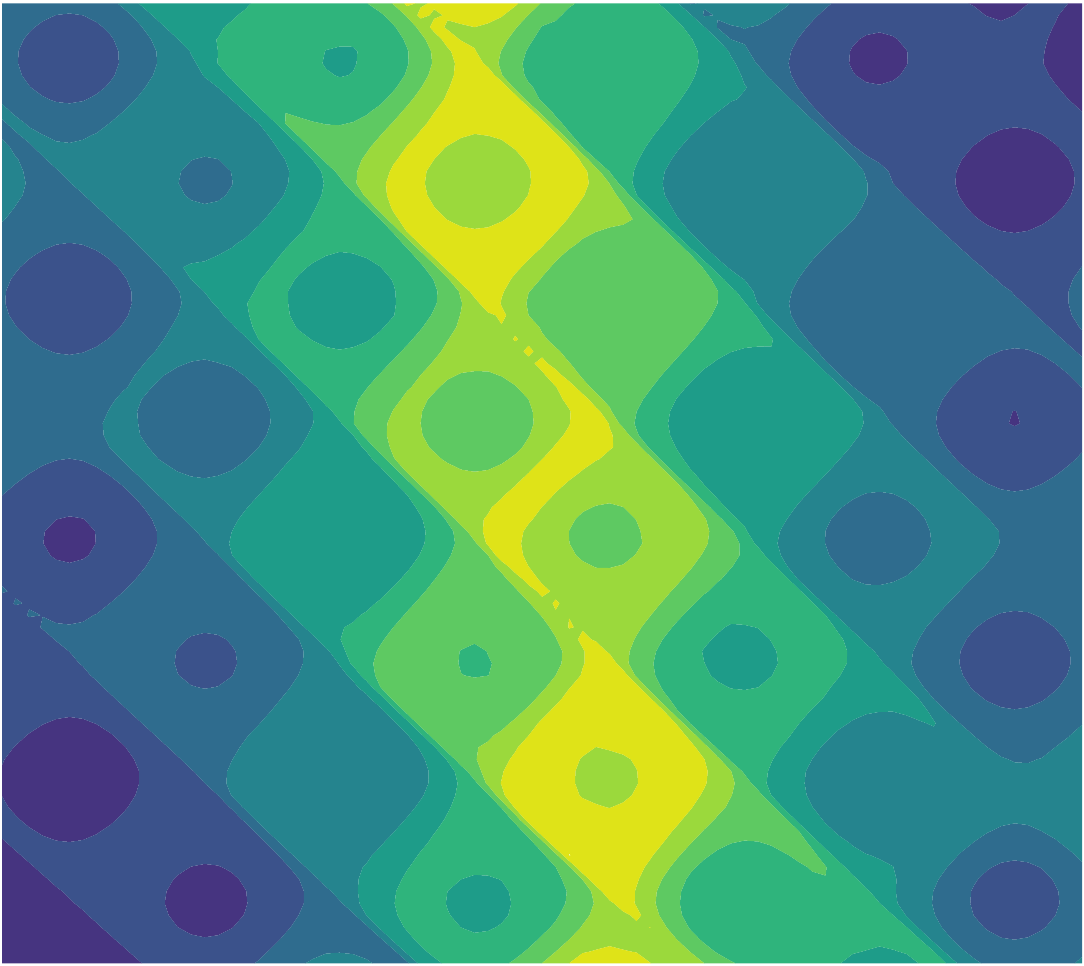}}
  \hspace{0.1in}
  \subfigure[{\tiny Prediction:~64\% data}]
    {\includegraphics[width = 0.2\textwidth]
    {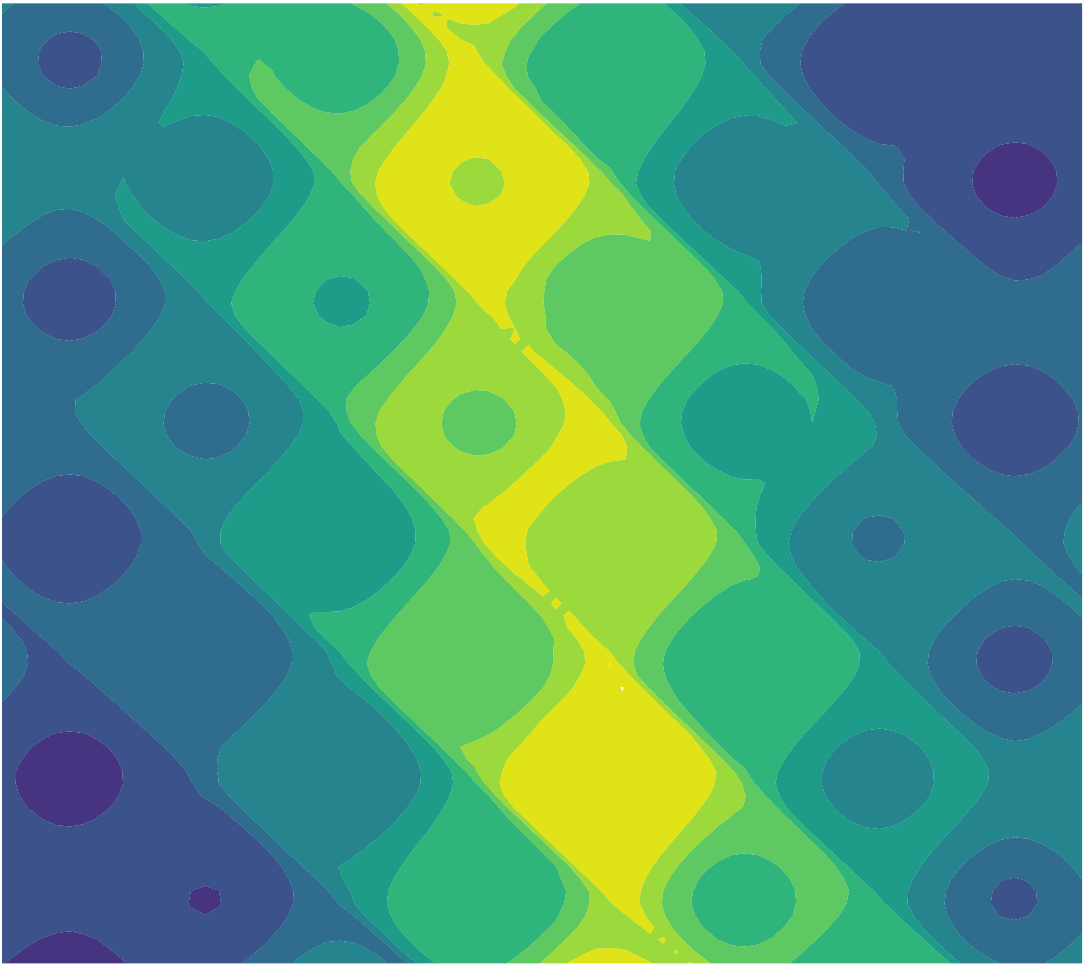}}
  \subfigure[{\tiny Prediction:~72\% data}]
    {\includegraphics[width = 0.2\textwidth]
    {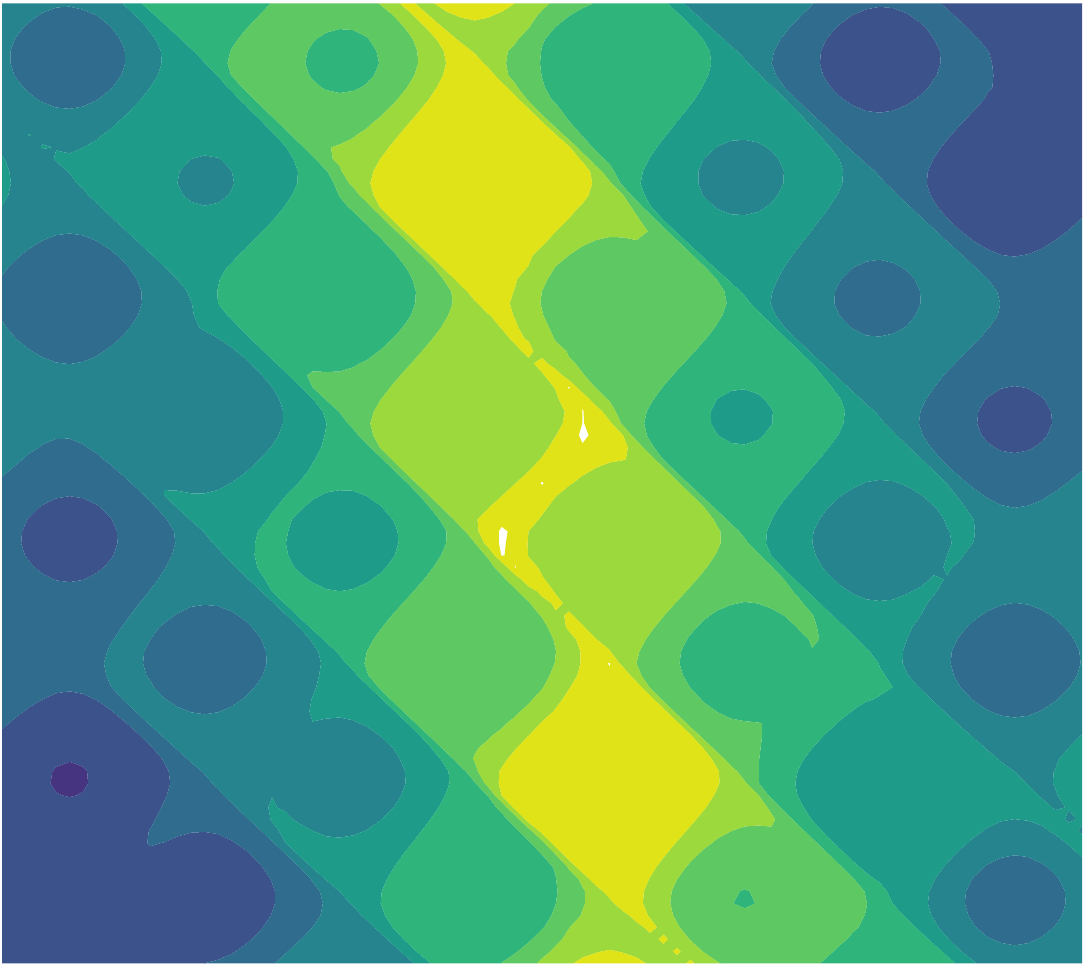}}
  \hspace{0.1in}
  \subfigure[{\tiny Prediction:~80\% data}]
    {\includegraphics[width = 0.2\textwidth]
    {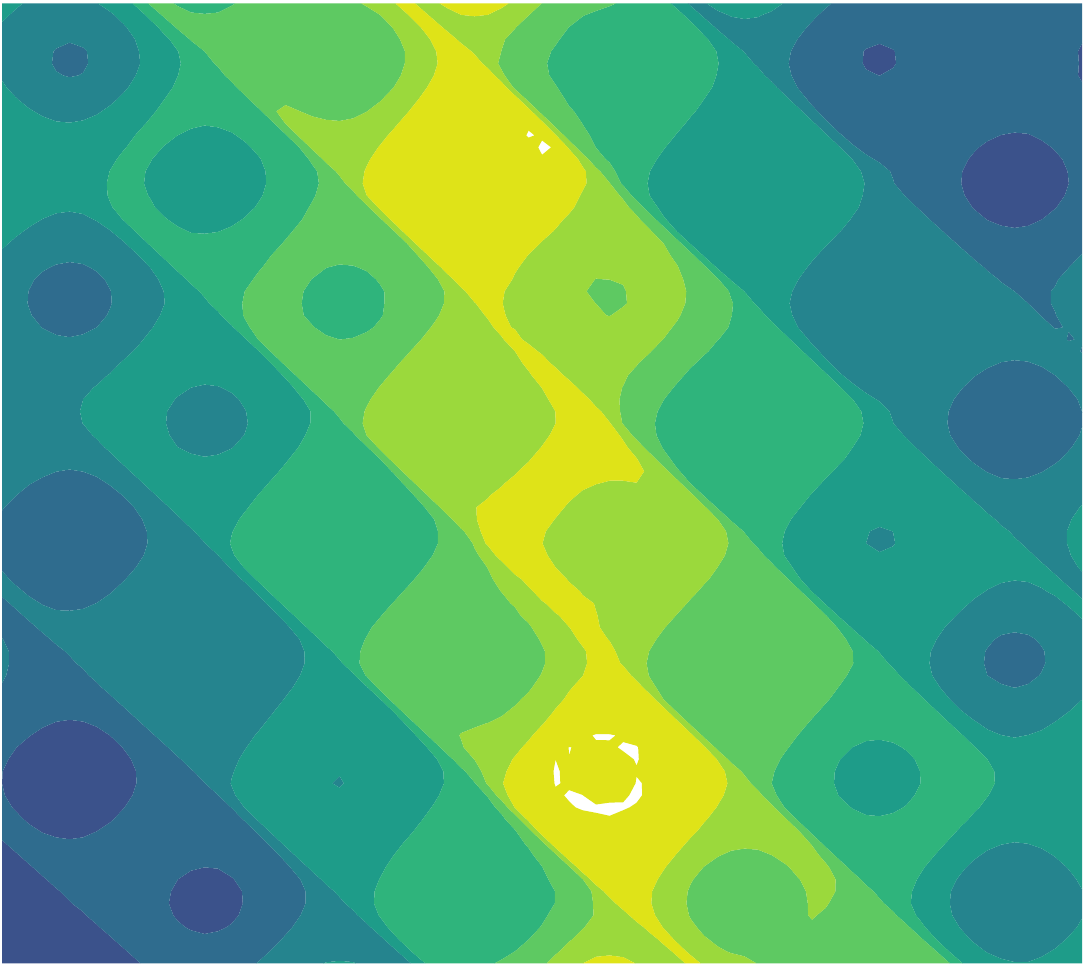}}
  \hspace{0.1in}
  \subfigure[{\tiny Prediction:~88\% data}]
    {\includegraphics[width = 0.2\textwidth]
    {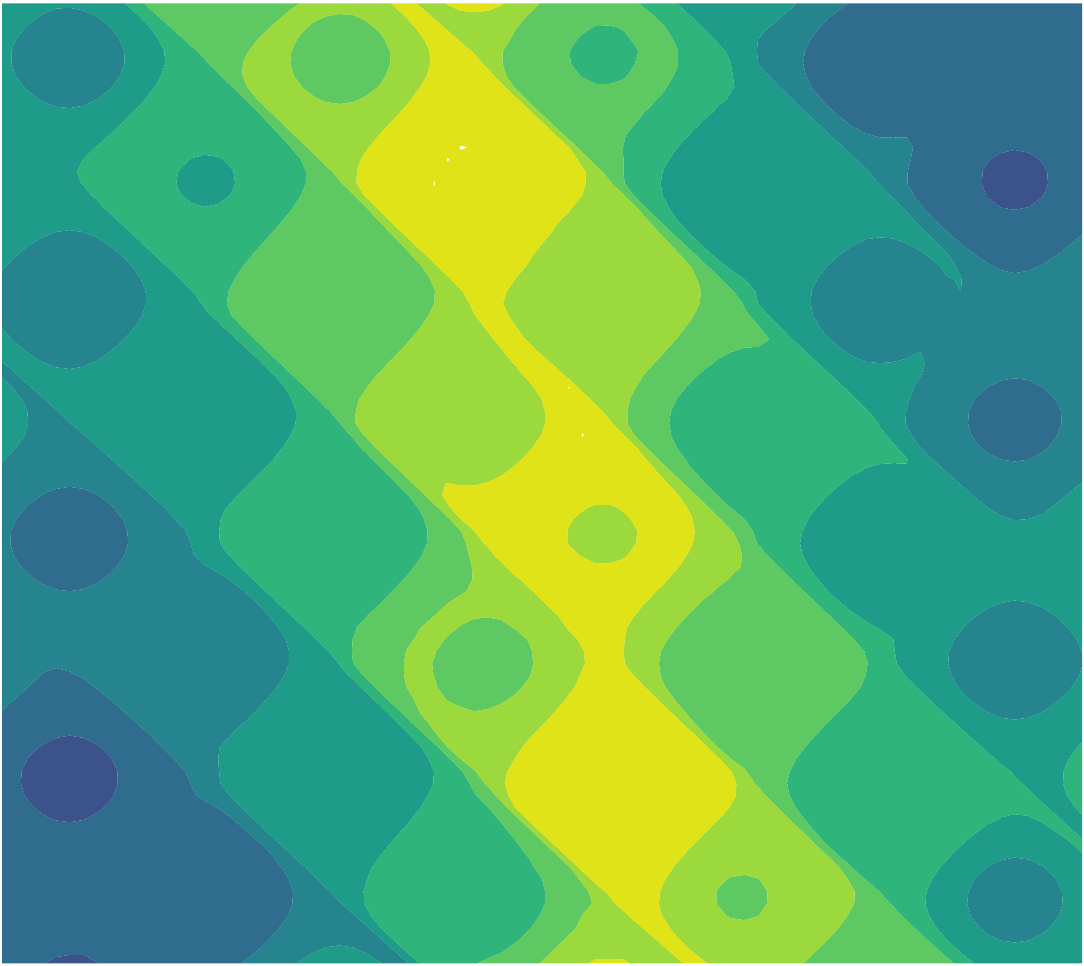}}
  \hspace{0.1in}
  \subfigure[{\tiny Prediction:~96\% data}]
    {\includegraphics[width = 0.2\textwidth]
    {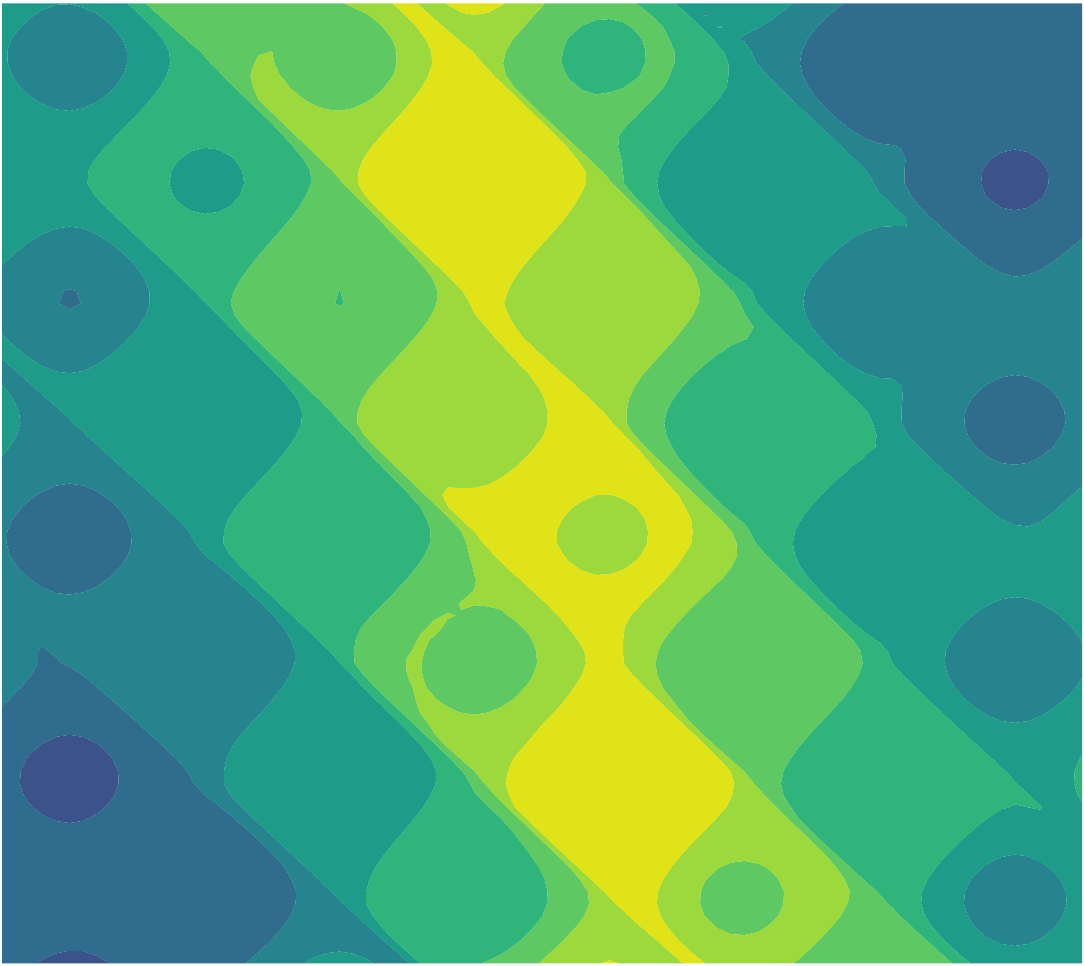}}
  \caption{\textsf{Predictions for $\kappa_fL = 4$:}~This figure compares the ground truth and predictions from the trained non-negative CNN-LSTM models at $t = 1.0$.
  From this figure, it is evident that 72\% ground truth data is reasonably sufficient to qualitatively capture different mixing patterns (e.g., preferential mixing due to small-scale features in the velocity field).
  \label{Fig:DL_RT_Pred_kfL4}}
\end{figure}

\begin{figure}
  \centering
  \subfigure[{\tiny Prediction error:~8\% data}]
    {\includegraphics[width = 0.285\textwidth]
    {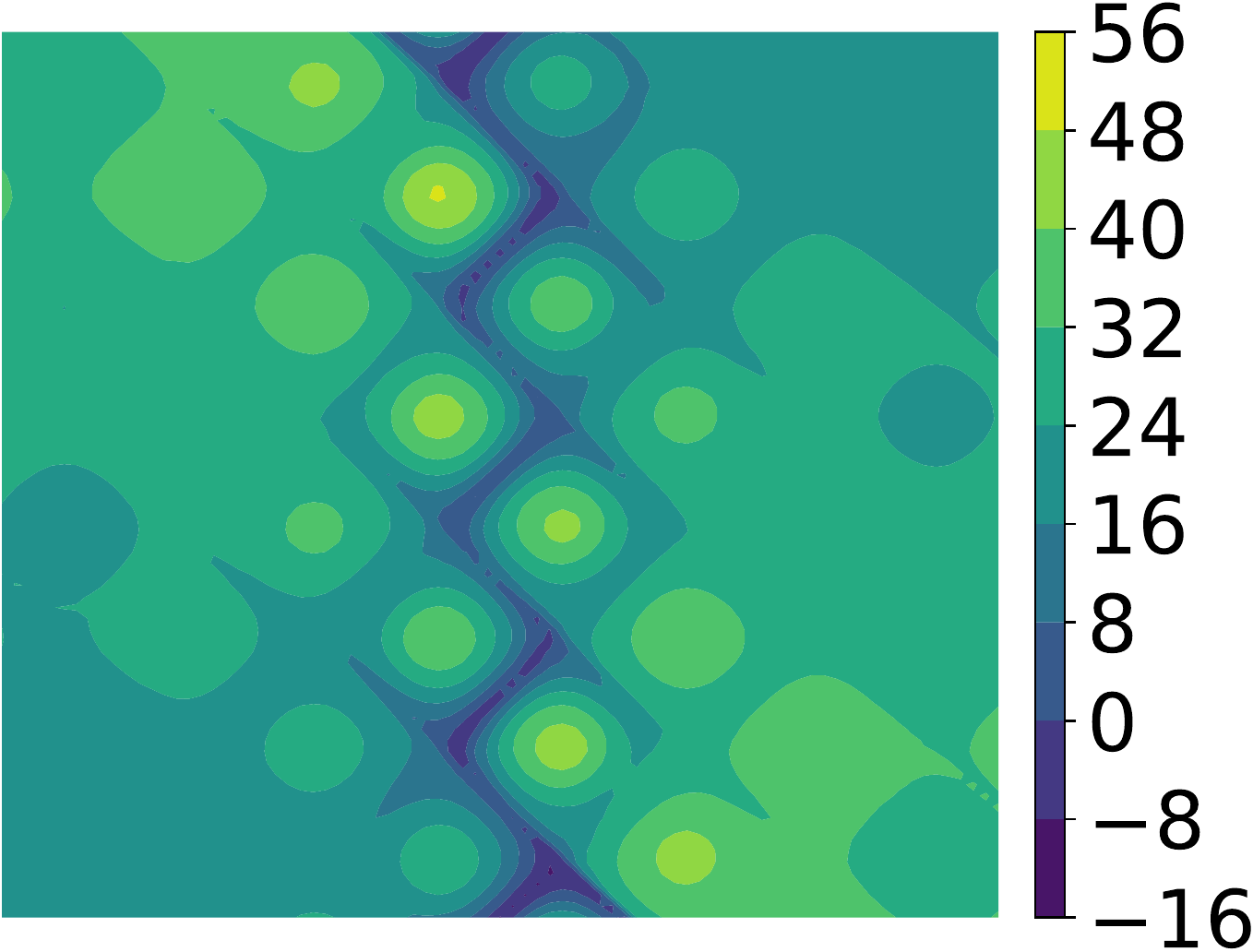}}
  \hspace{0.25in}
  \subfigure[{\tiny Prediction error:~16\% data}]
    {\includegraphics[width = 0.285\textwidth]
    {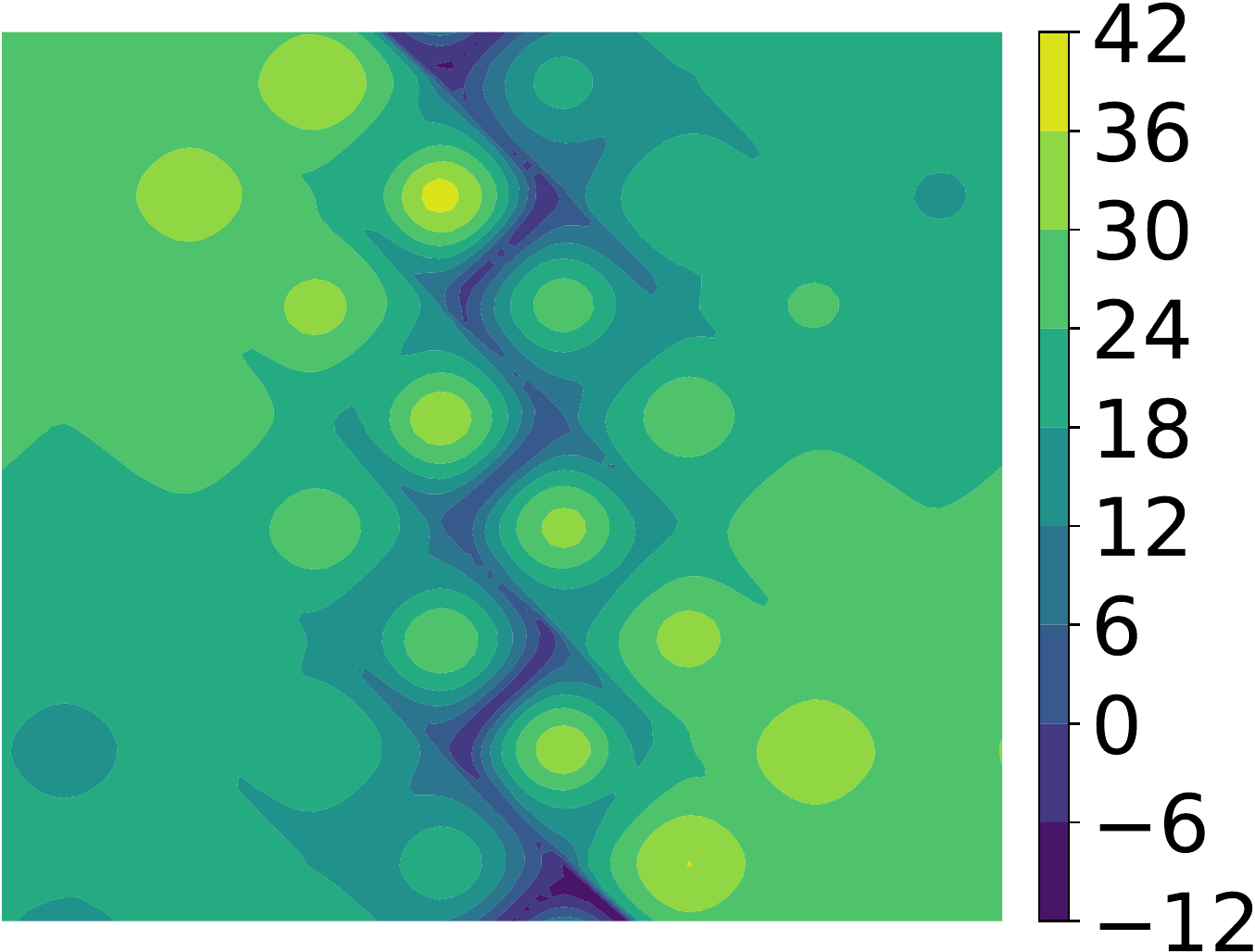}}
  \hspace{0.25in}
  \subfigure[{\tiny Prediction error:~24\% data}]
    {\includegraphics[width = 0.285\textwidth]
    {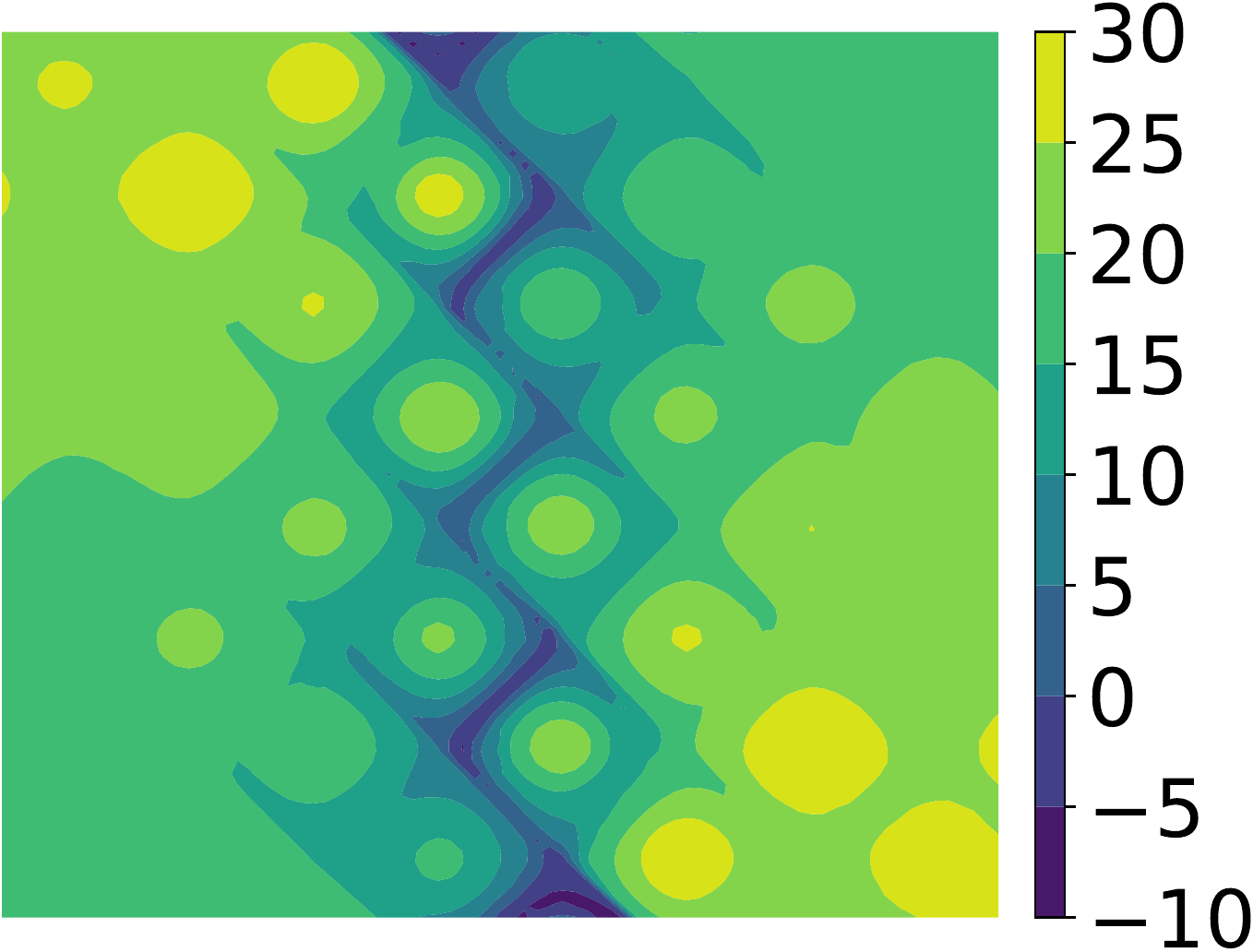}}
  \subfigure[{\tiny Prediction error:~32\% data}]
    {\includegraphics[width = 0.285\textwidth]
    {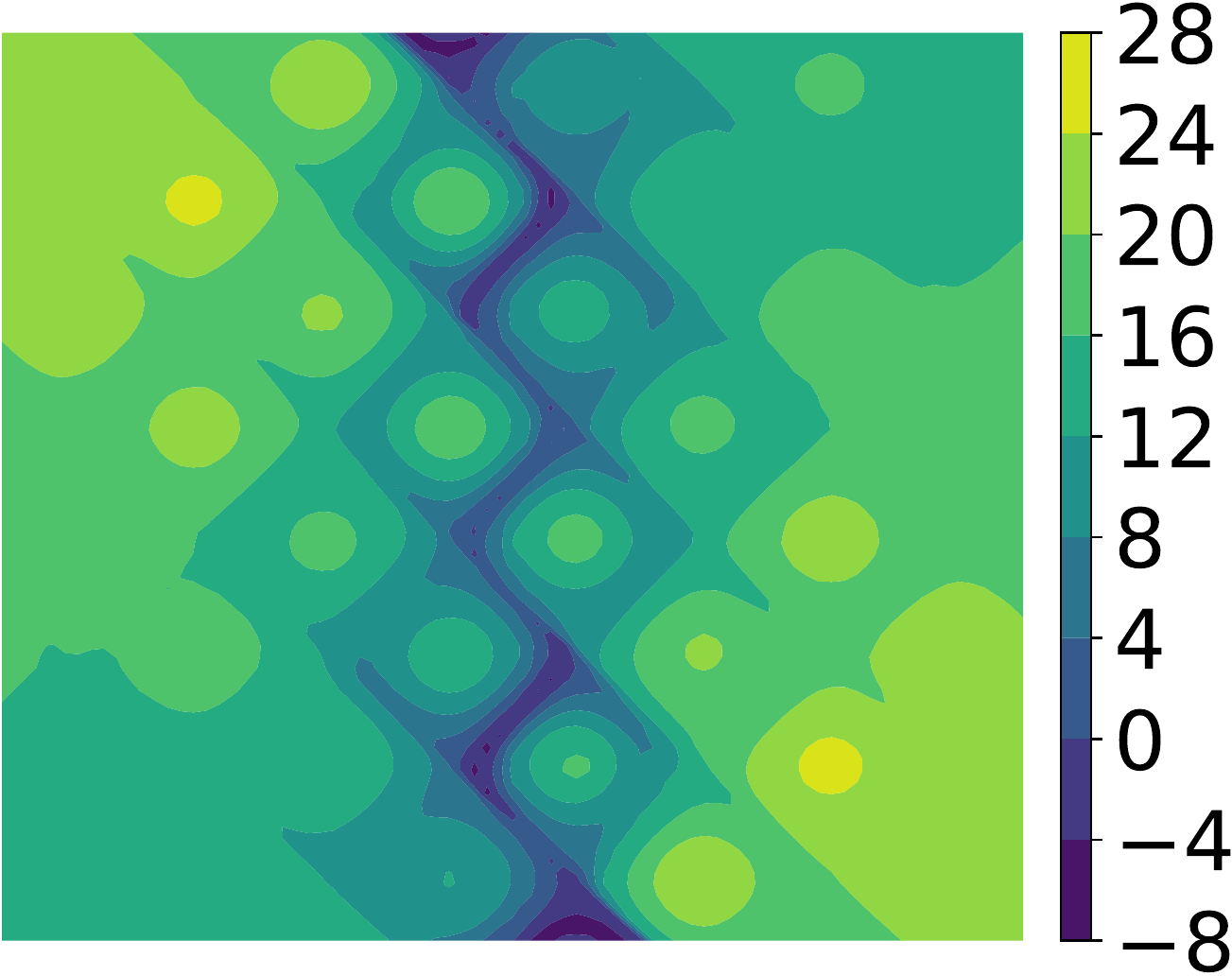}}
  \hspace{0.25in}
  \subfigure[{\tiny Prediction error:~40\% data}]
    {\includegraphics[width = 0.285\textwidth]
    {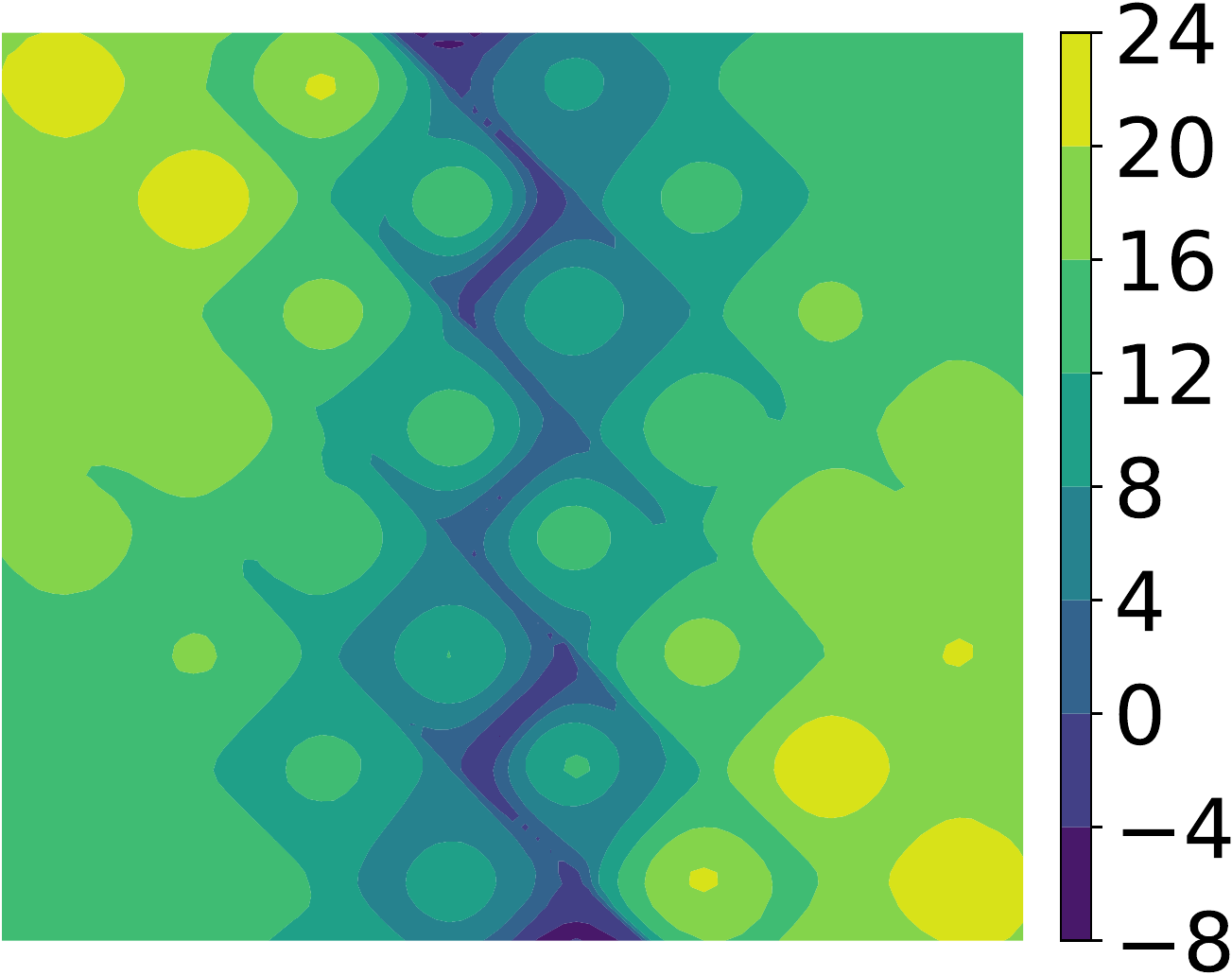}}
  \hspace{0.25in}
  \subfigure[{\tiny Prediction error:~48\% data}]
    {\includegraphics[width = 0.295\textwidth]
    {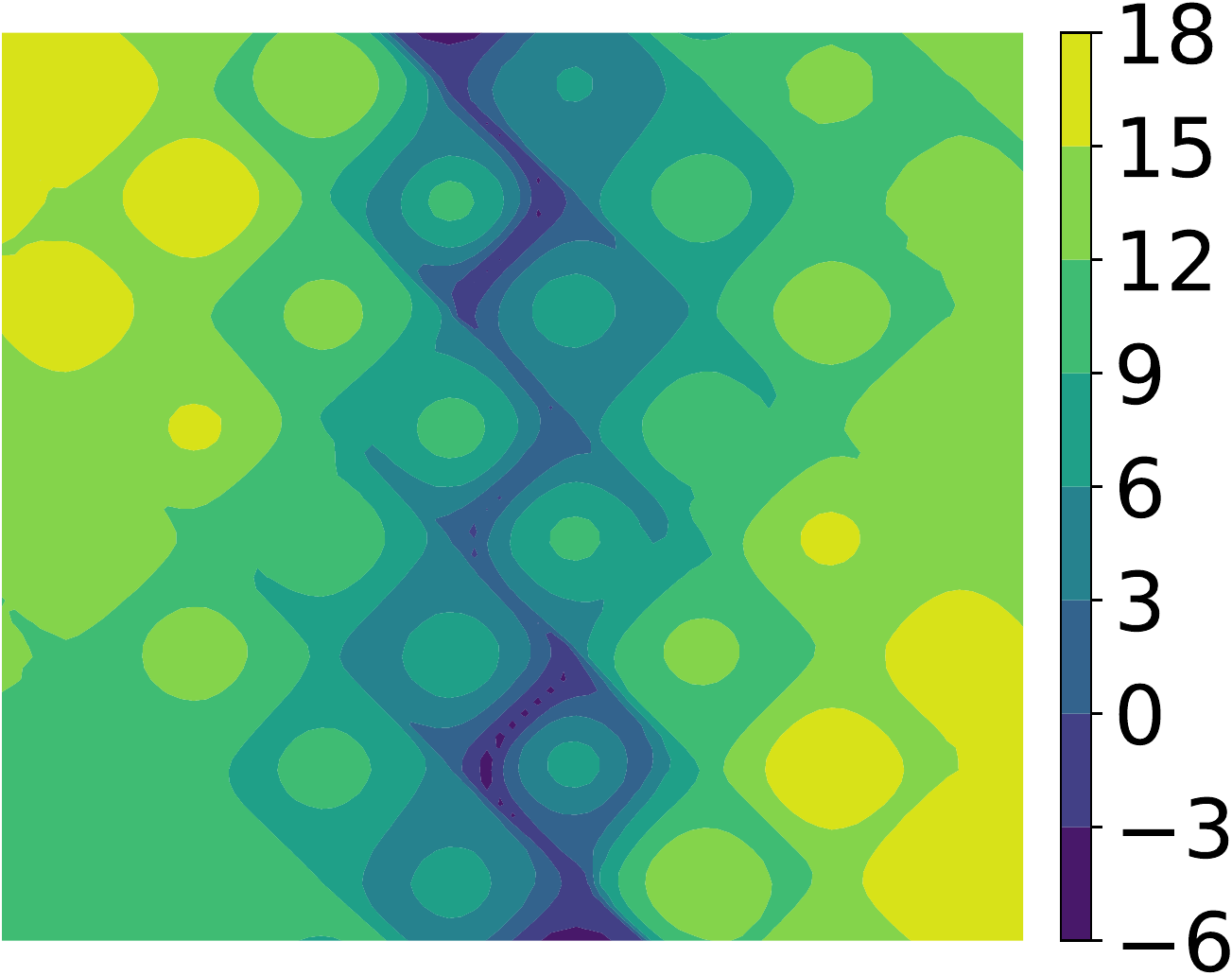}}
  \subfigure[{\tiny Prediction error:~56\% data}]
    {\includegraphics[width = 0.295\textwidth]
    {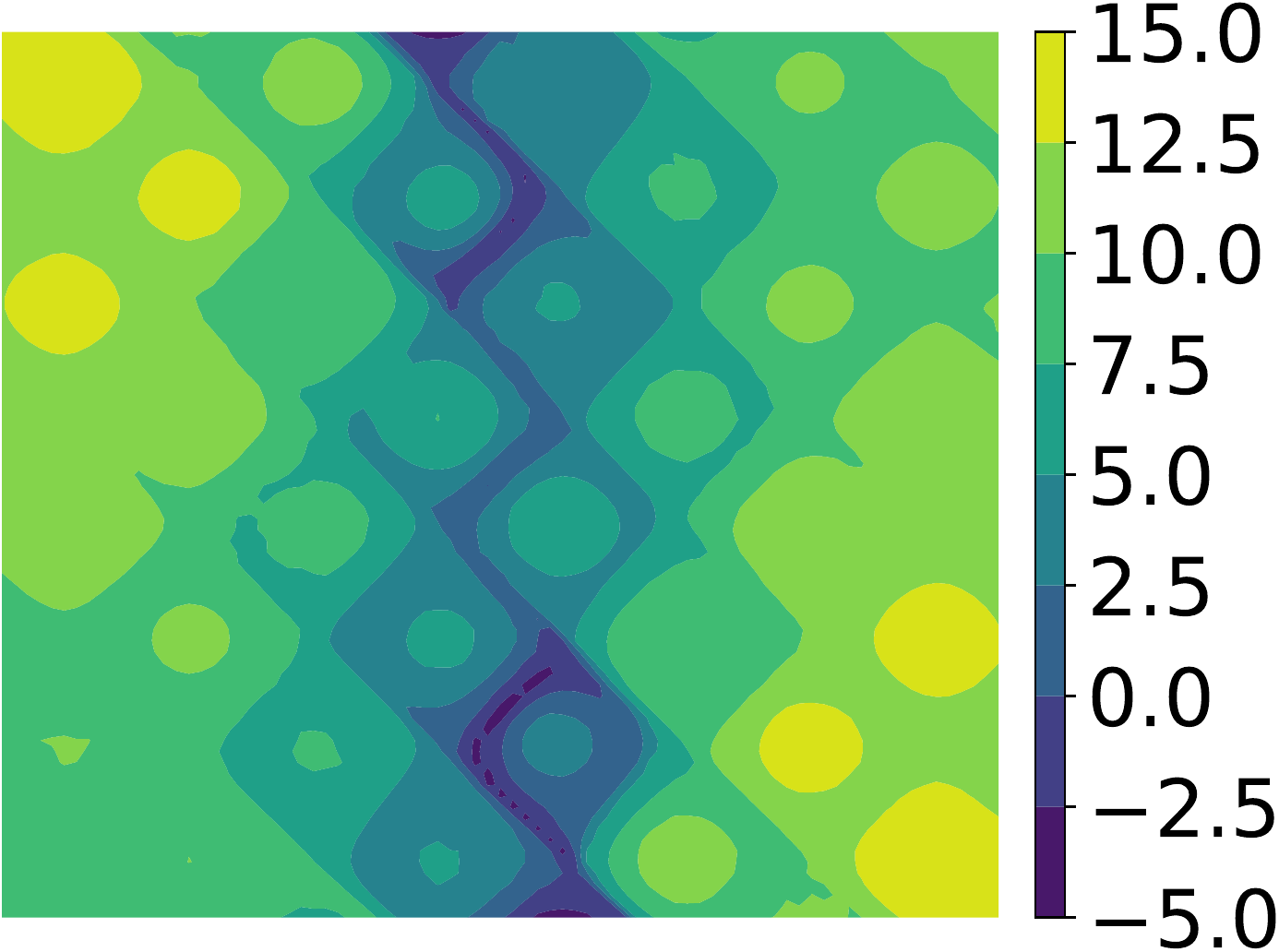}}
  \hspace{0.25in}
  \subfigure[{\tiny Prediction error:~64\% data}]
    {\includegraphics[width = 0.295\textwidth]
    {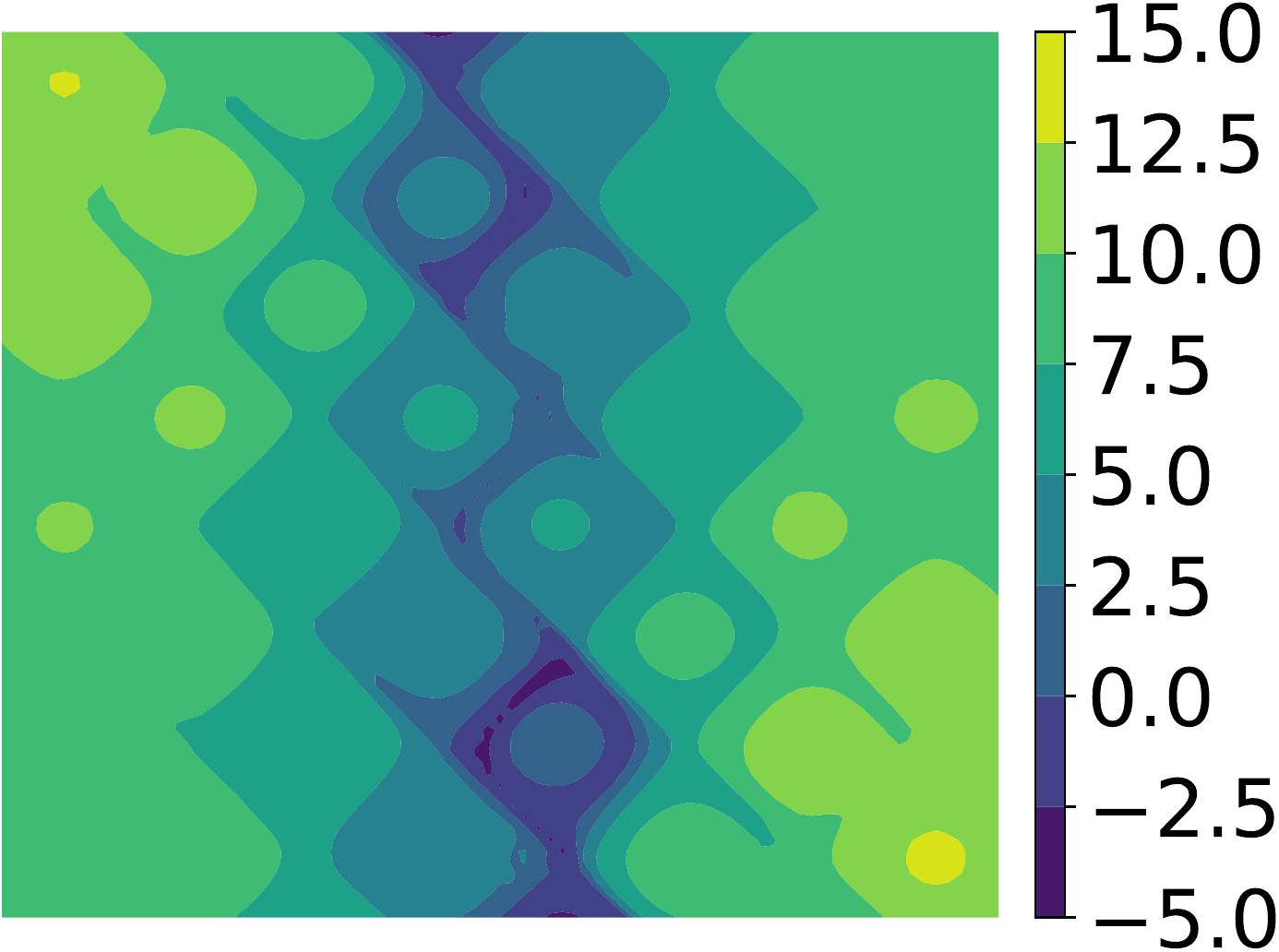}}
  \hspace{0.25in}
  \subfigure[{\tiny Prediction error:~72\% data}]
    {\includegraphics[width = 0.285\textwidth]
    {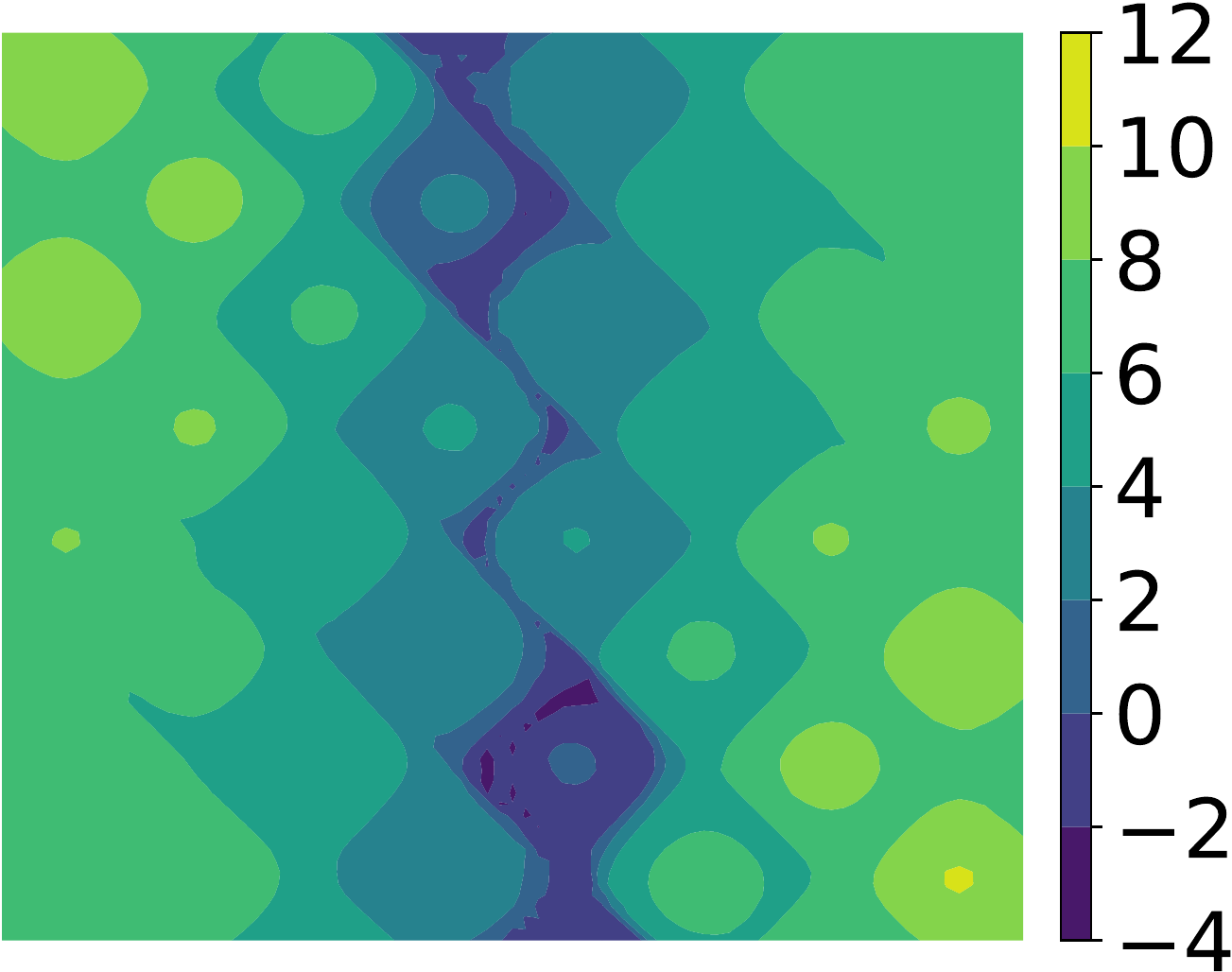}}
  \subfigure[{\tiny Prediction error:~80\% data}]
    {\includegraphics[width = 0.295\textwidth]
    {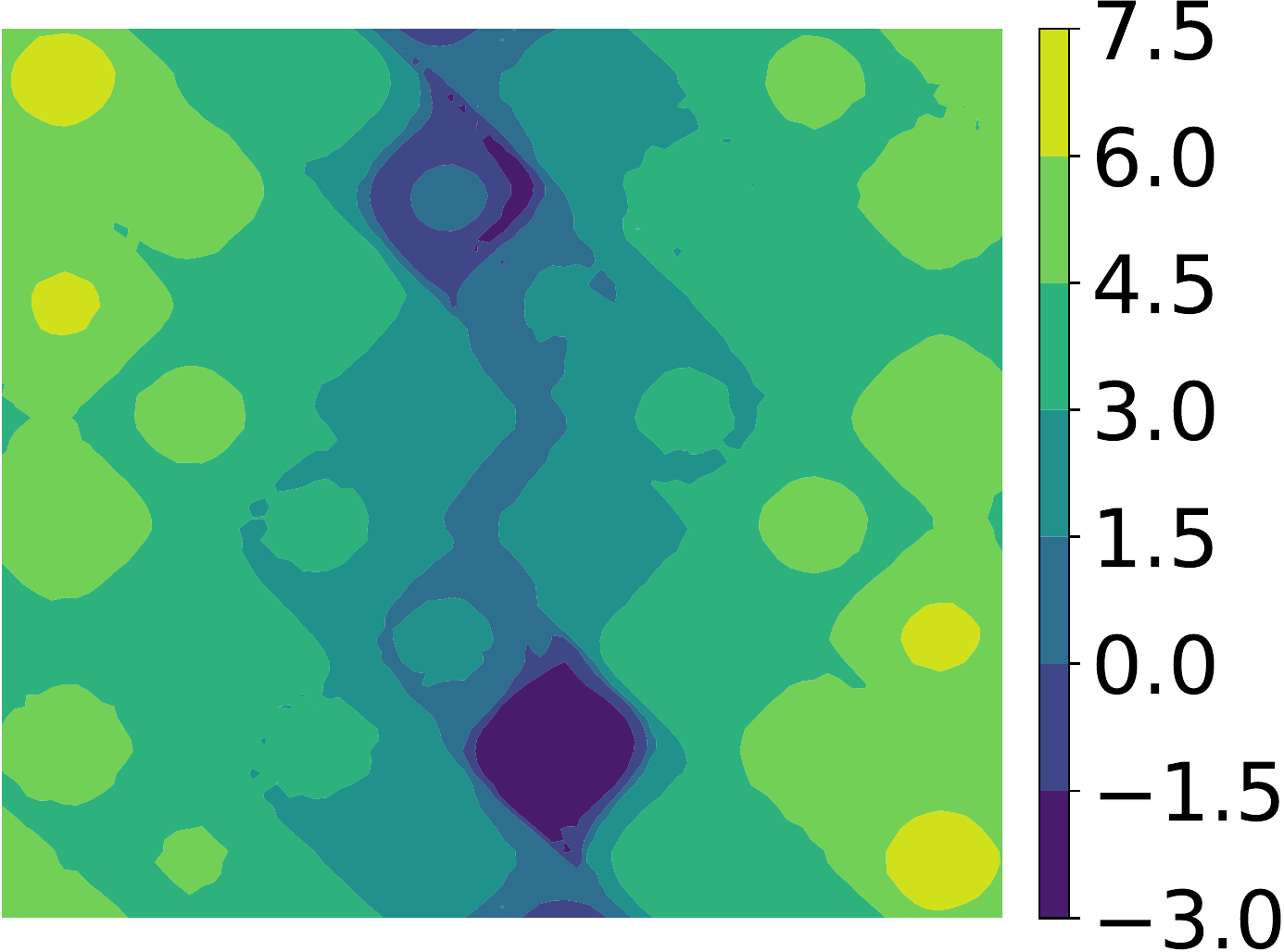}}
  \hspace{0.25in}
  \subfigure[{\tiny Prediction error:~88\% data}]
    {\includegraphics[width = 0.285\textwidth]
    {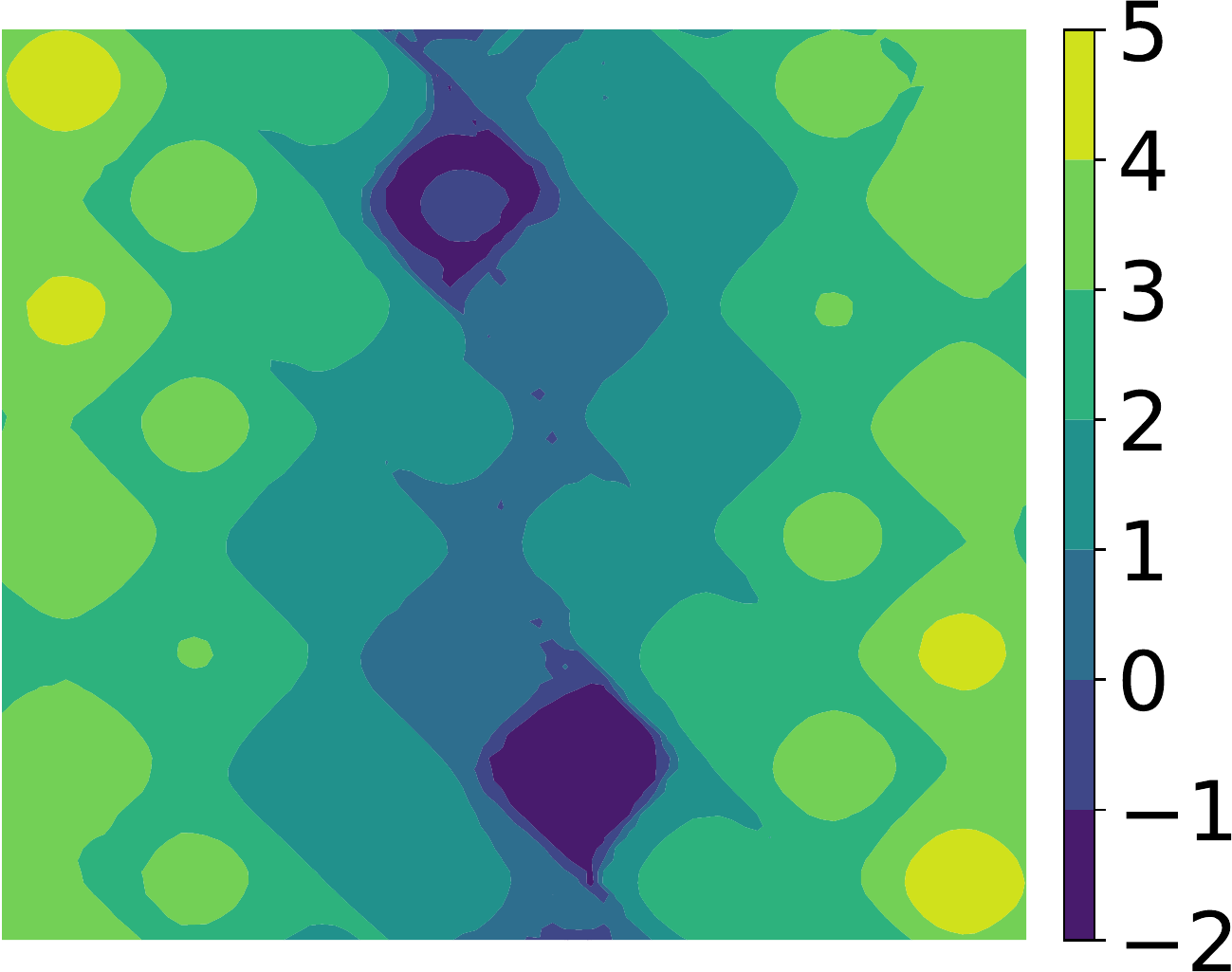}}
  \hspace{0.25in}
  \subfigure[{\tiny Prediction error:~96\% data}]
    {\includegraphics[width = 0.285\textwidth]
    {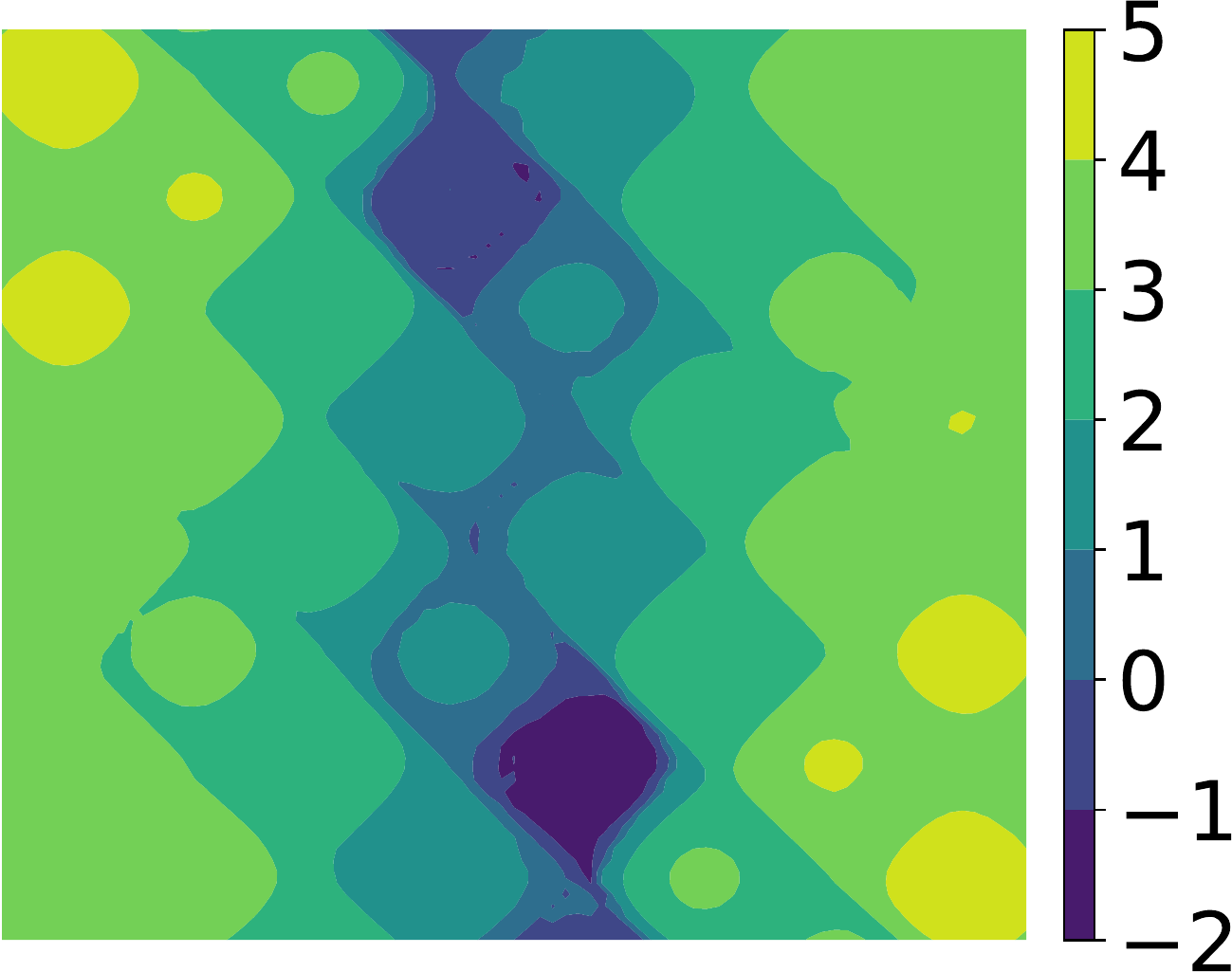}}
  \caption{\textsf{Prediction error in percentage for $\kappa_fL = 4$:}~This figure compares the prediction errors in the entire domain at $t = 1.0$ for different amount of training data.
  Based on the error values (e.g., $\leq 10\%$), it is evident that with 80\% training data, we can accurately capture small-scale mixing patterns with an accuracy less than 10\%.
  \label{Fig:DL_RT_Pred_kfL4_Errors}}
\end{figure}

\begin{figure}
  \centering
  \subfigure[Ground truth at $t = 1.0$]
    {\includegraphics[width = 0.285\textwidth]
    {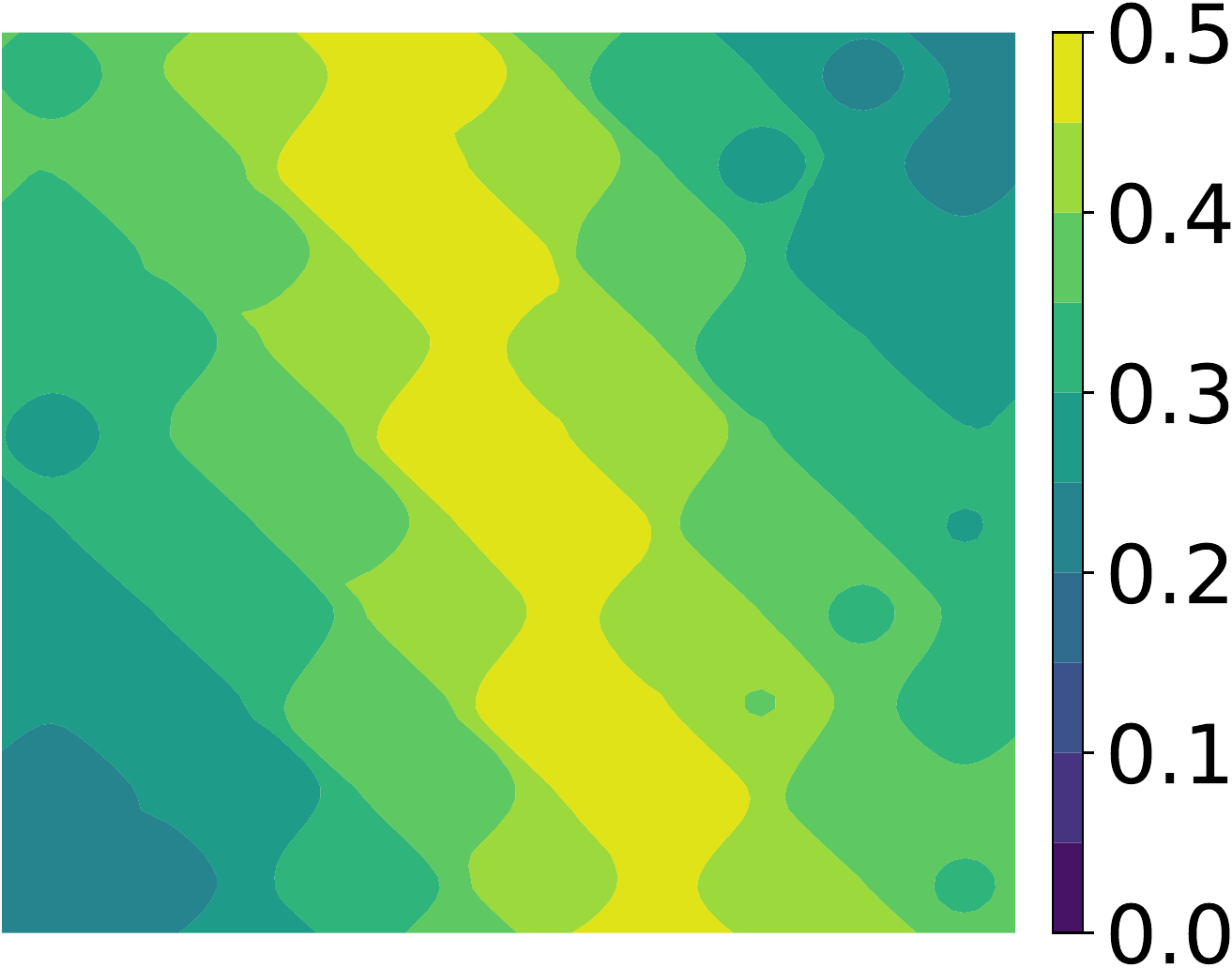}}
  \hspace{3.5in}
  \subfigure[{\tiny Prediction:~8\% data}]
    {\includegraphics[width = 0.2\textwidth]
    {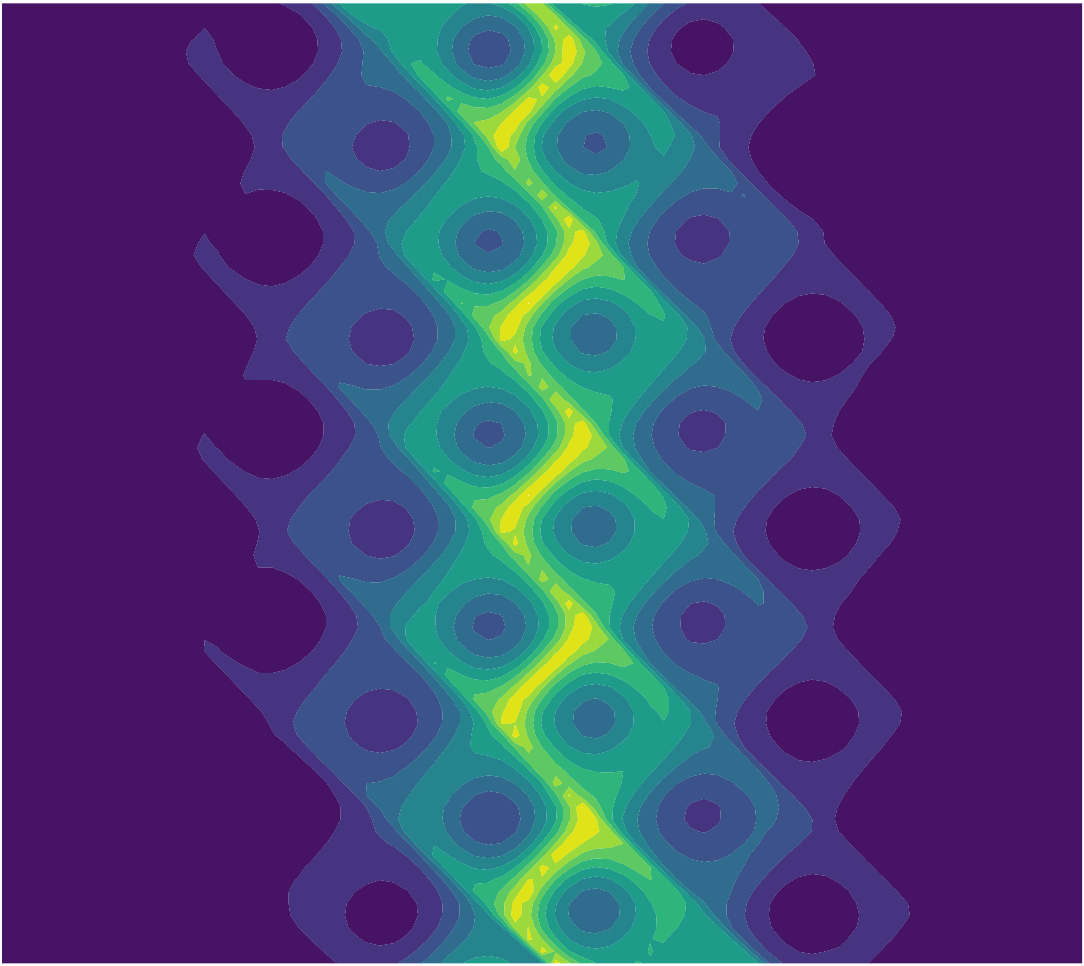}}
  \hspace{0.1in}
  \subfigure[{\tiny Prediction:~16\% data}]
    {\includegraphics[width = 0.2\textwidth]
    {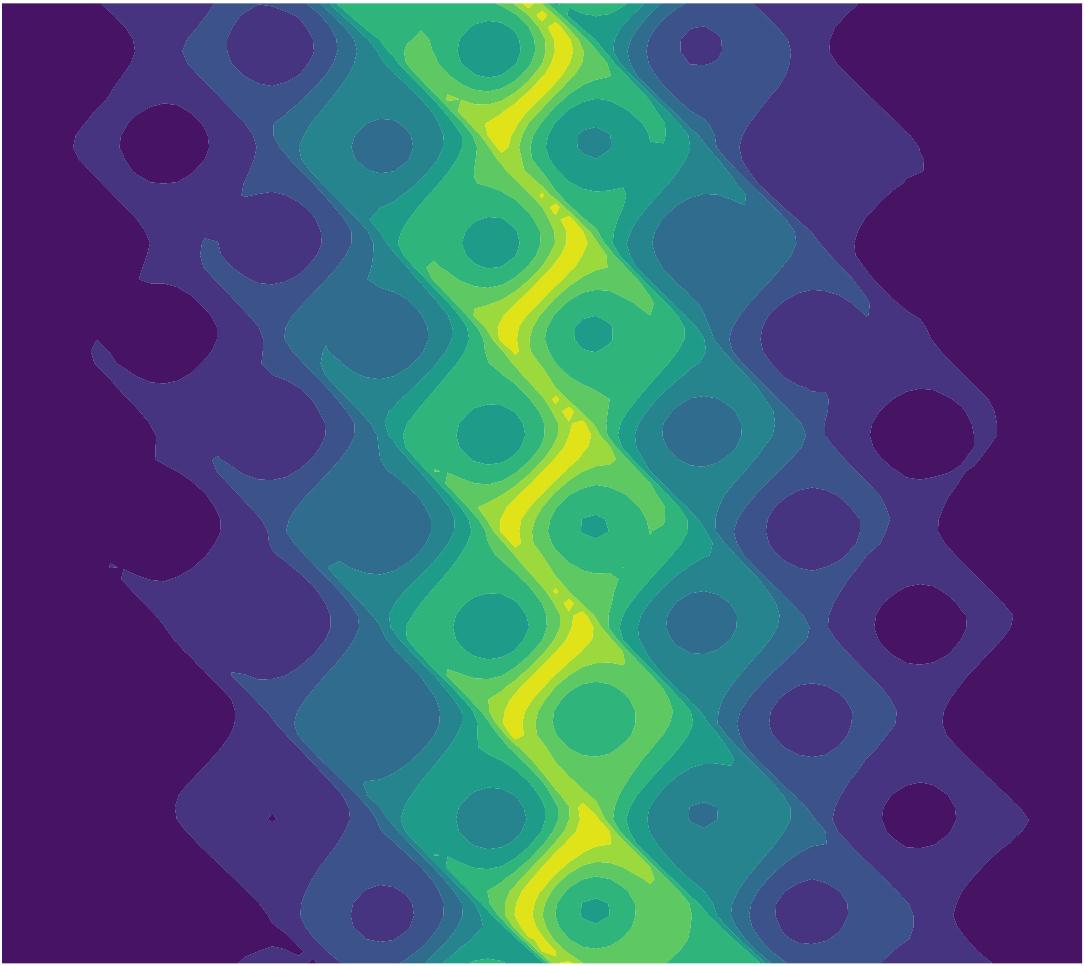}}
  \hspace{0.1in}
  \subfigure[{\tiny Prediction:~24\% data}]
    {\includegraphics[width = 0.2\textwidth]
    {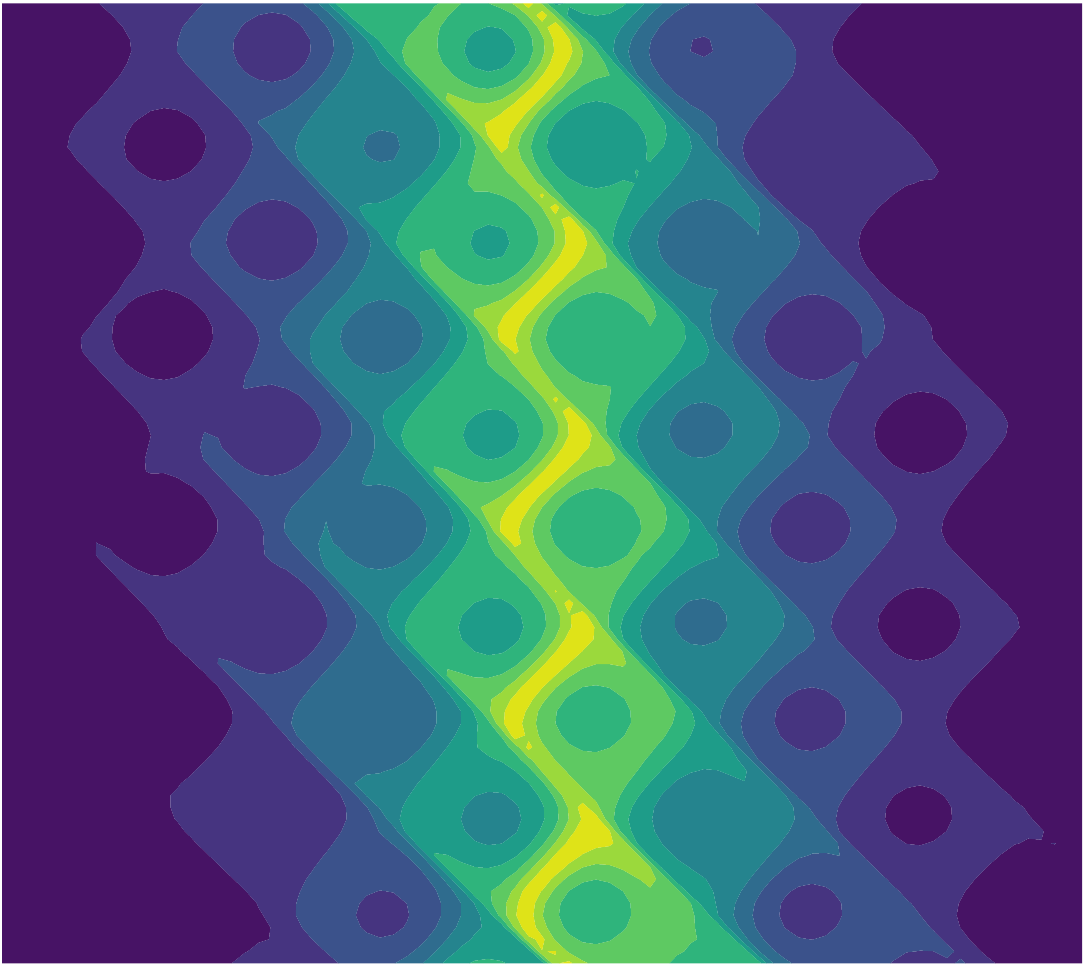}}
  \hspace{0.1in}
  \subfigure[{\tiny Prediction:~32\% data}]
    {\includegraphics[width = 0.2\textwidth]
    {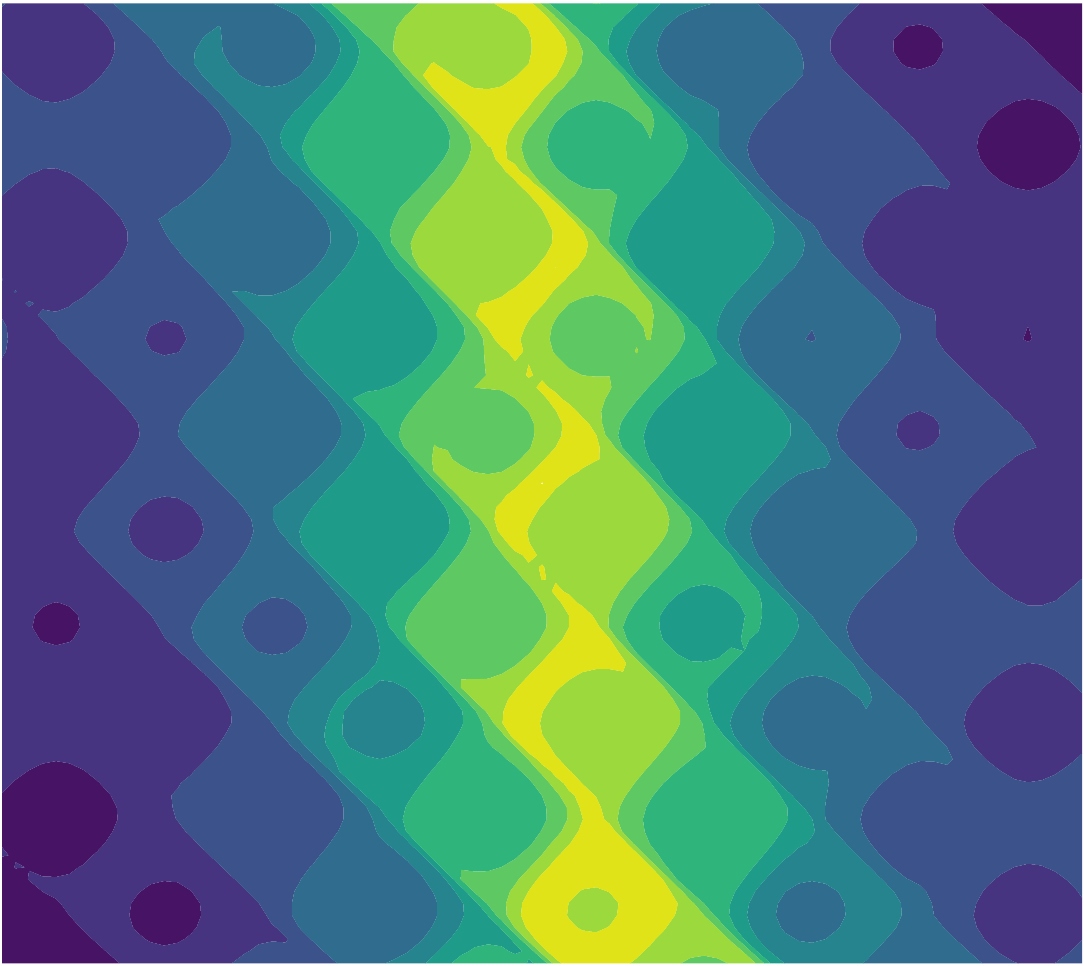}}
  \subfigure[{\tiny Prediction:~40\% data}]
    {\includegraphics[width = 0.2\textwidth]
    {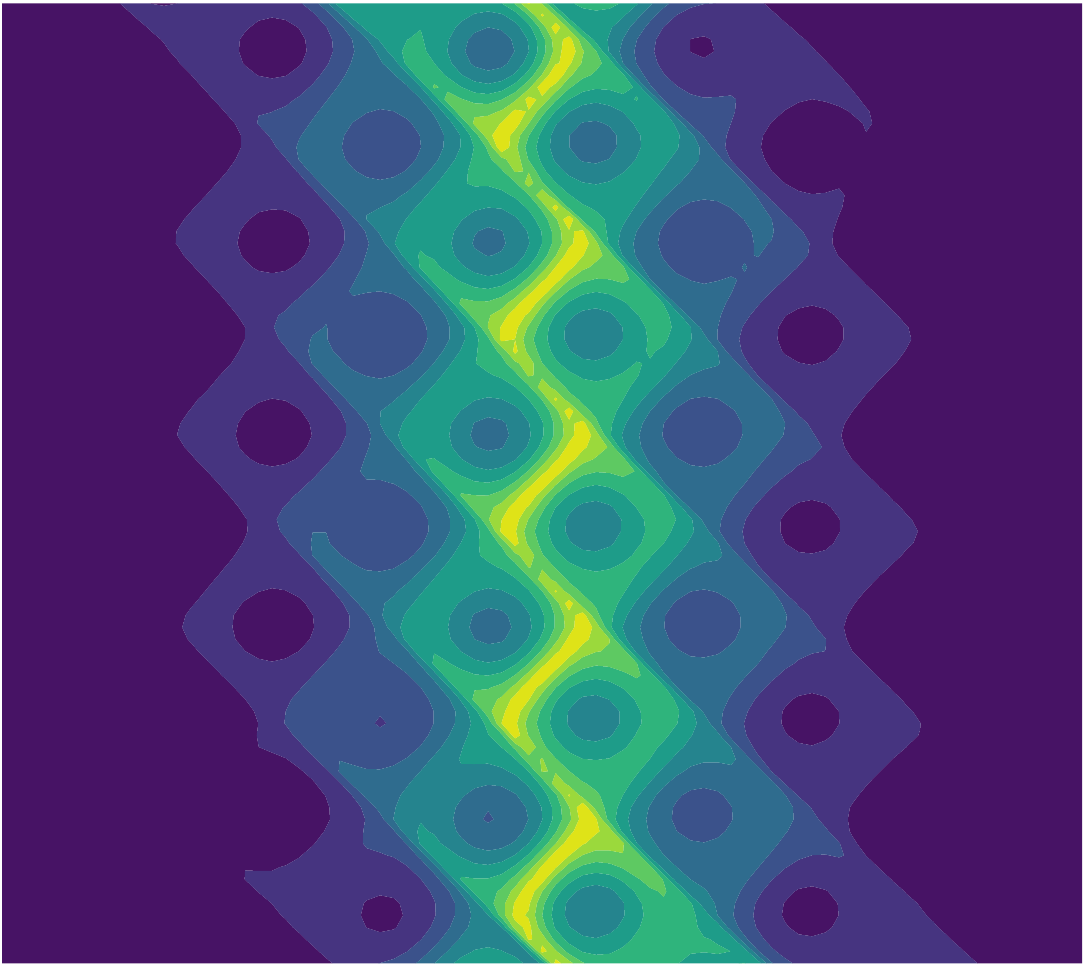}}
  \hspace{0.1in}
  \subfigure[{\tiny Prediction:~48\% data}]
    {\includegraphics[width = 0.2\textwidth]
    {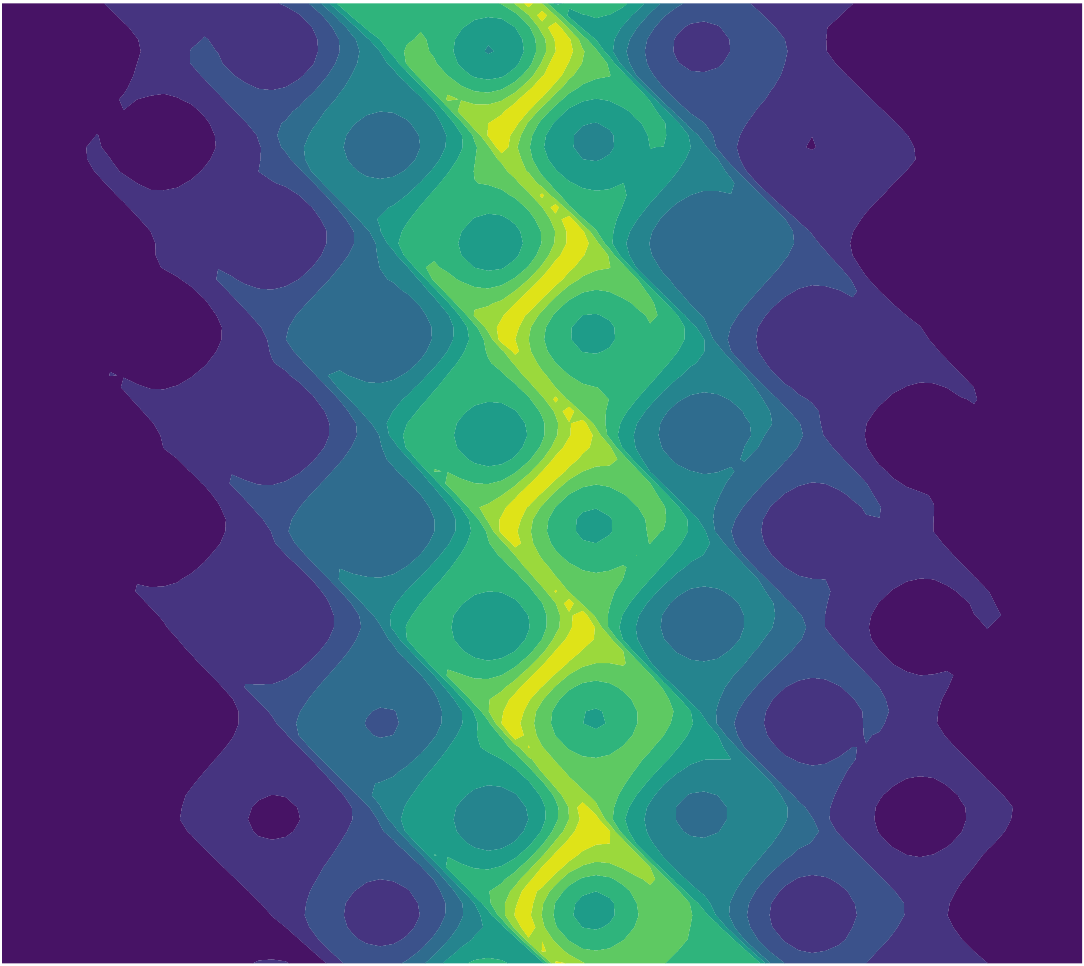}}
  \hspace{0.1in}
  \subfigure[{\tiny Prediction:~56\% data}]
    {\includegraphics[width = 0.2\textwidth]
    {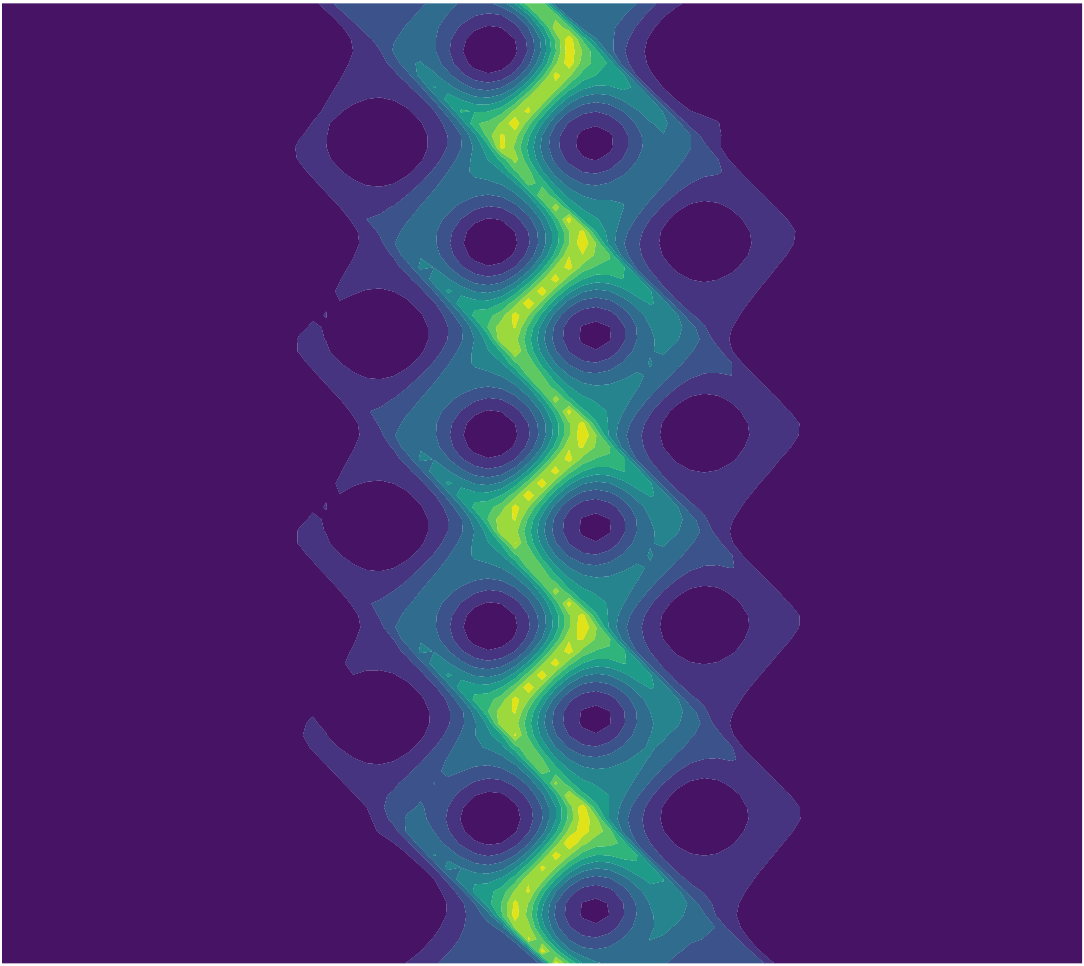}}
  \hspace{0.1in}
  \subfigure[{\tiny Prediction:~64\% data}]
    {\includegraphics[width = 0.2\textwidth]
    {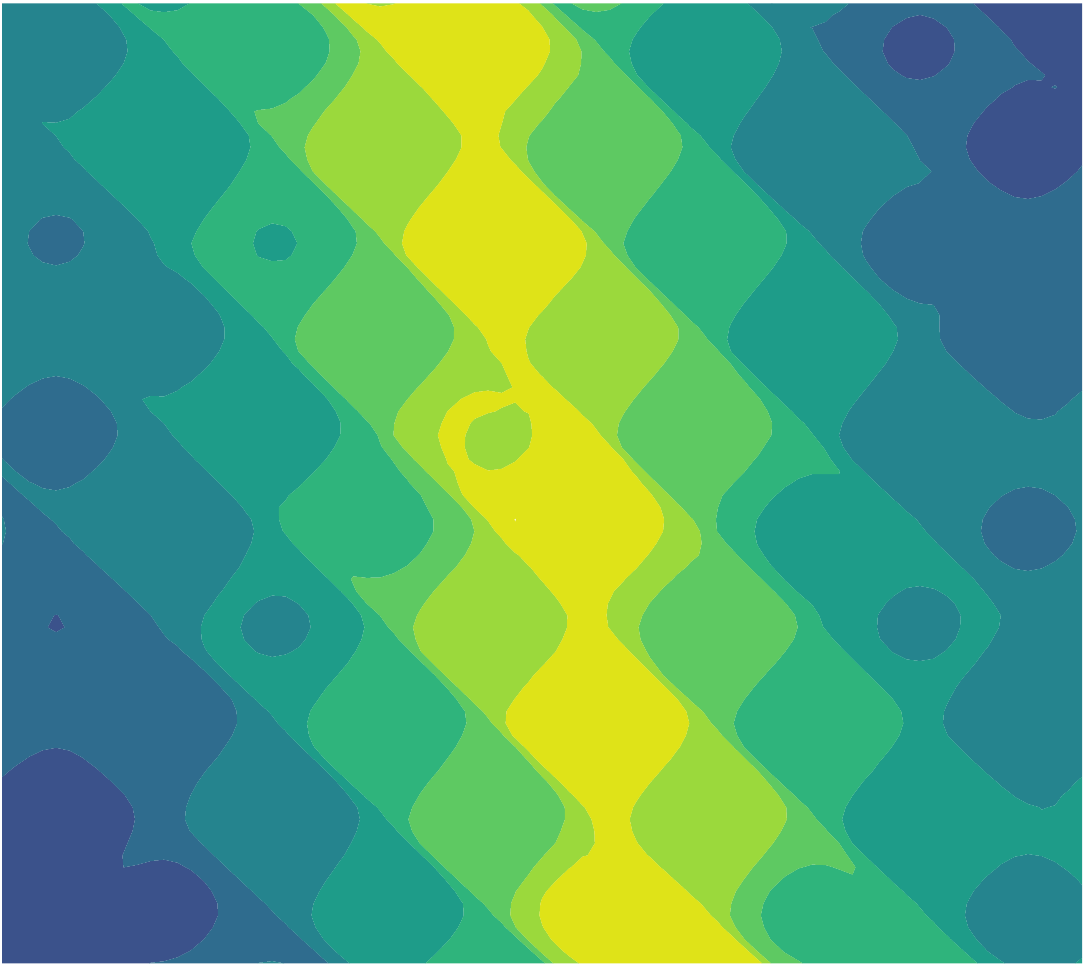}}
  \subfigure[{\tiny Prediction:~72\% data}]
    {\includegraphics[width = 0.2\textwidth]
    {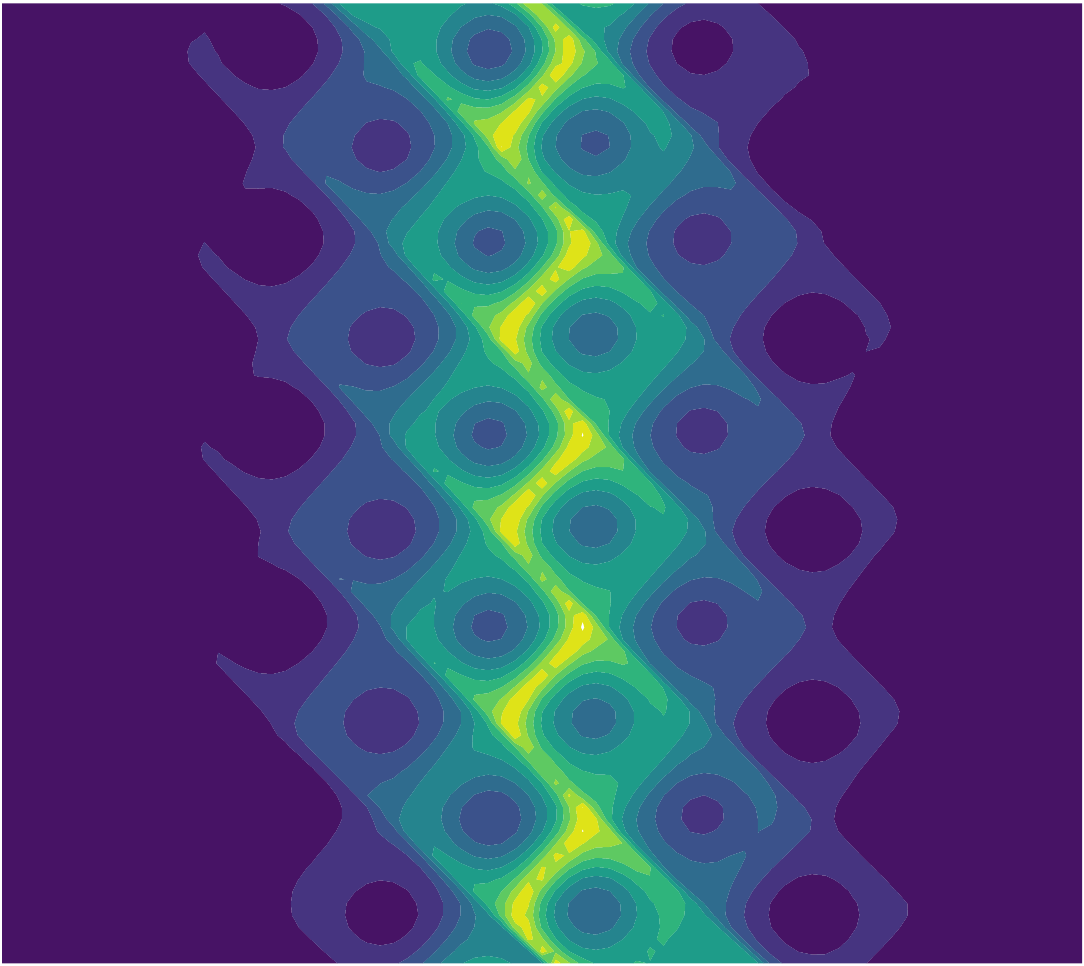}}
  \hspace{0.1in}
  \subfigure[{\tiny Prediction:~80\% data}]
    {\includegraphics[width = 0.2\textwidth]
    {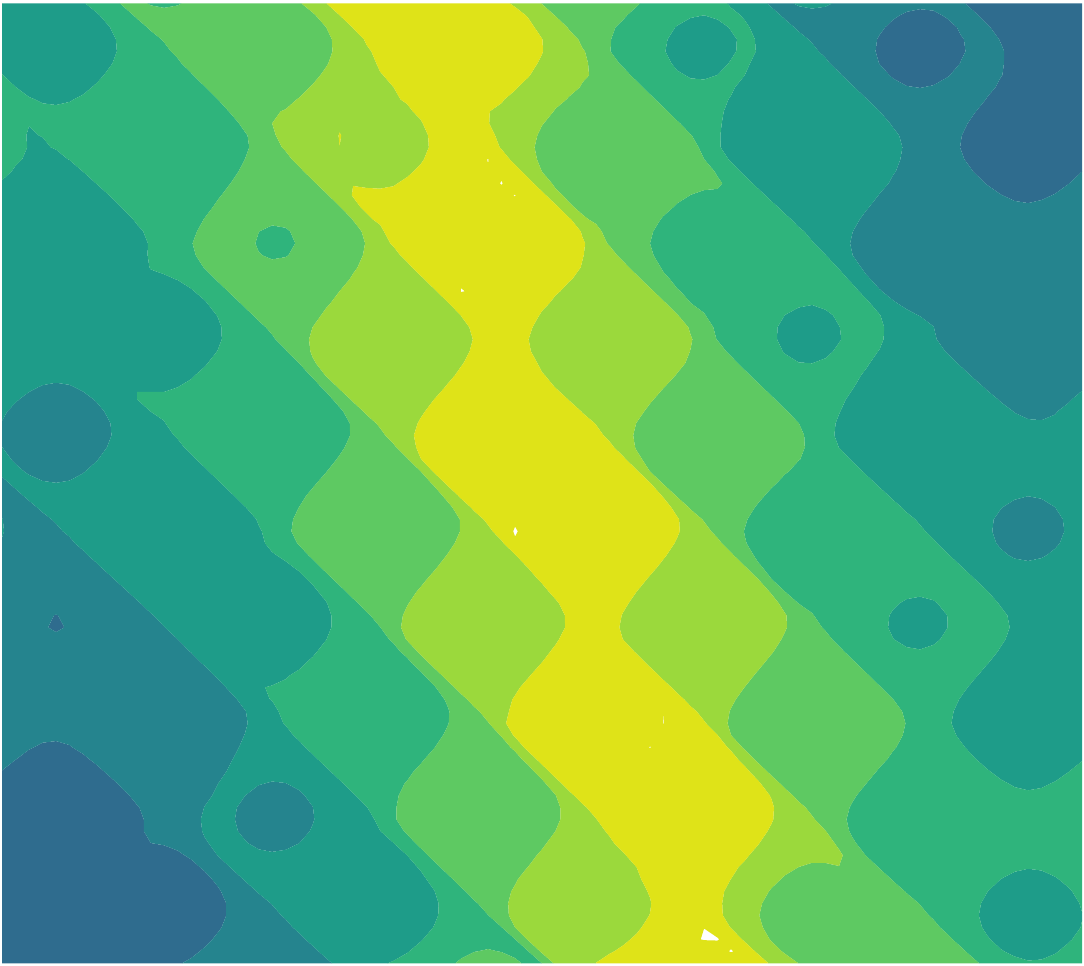}}
  \hspace{0.1in}
  \subfigure[{\tiny Prediction:~88\% data}]
    {\includegraphics[width = 0.2\textwidth]
    {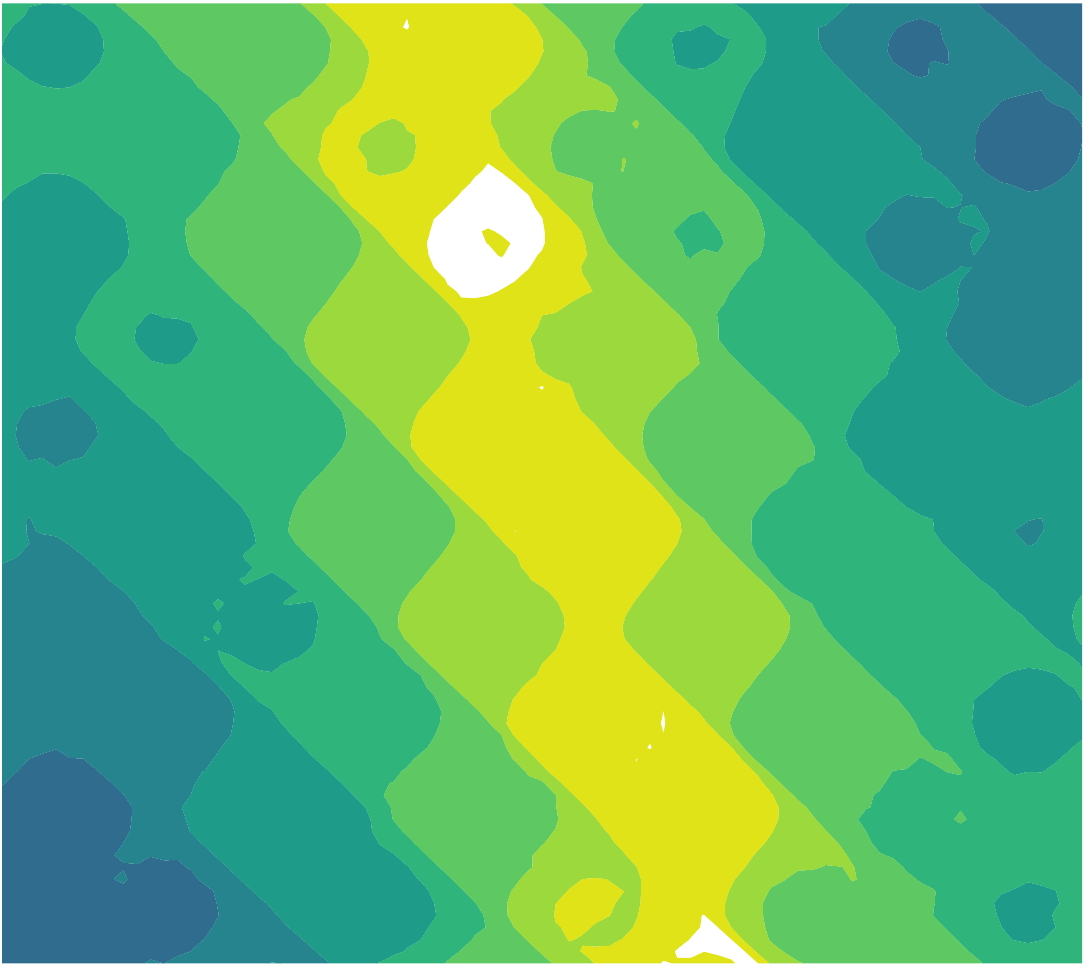}}
  \hspace{0.1in}
  \subfigure[{\tiny Prediction:~96\% data}]
    {\includegraphics[width = 0.2\textwidth]
    {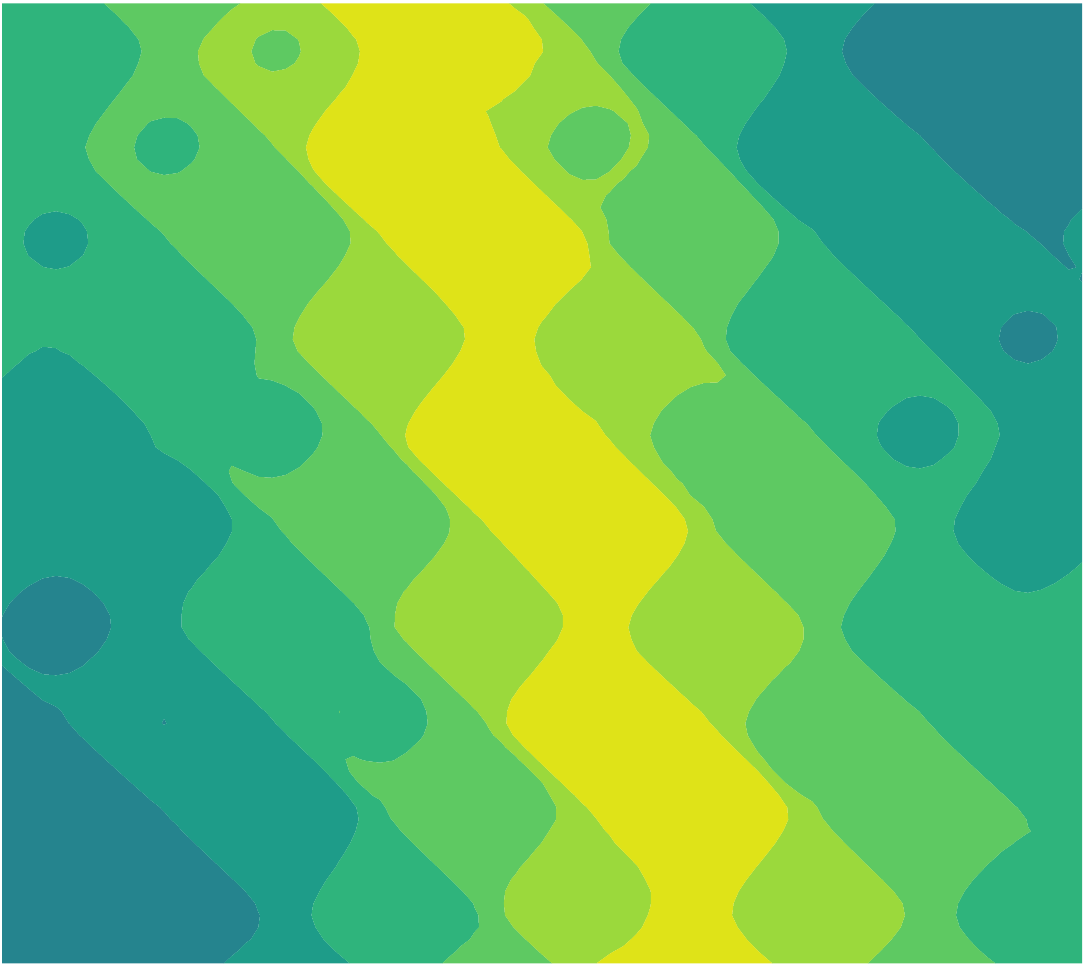}}
  \caption{\textsf{Predictions for $\kappa_fL = 5$:}~~This figure compares the ground truth and predictions from the trained non-negative CNN-LSTM models at $t = 1.0$.
  From this figure, it is evident that 64\% ground truth data is needed to reasonably capture product formation in the entire domain (e.g., preferential mixing patterns, mixing away from interface).
  \label{Fig:DL_RT_Pred_kfL5}}
\end{figure}

\begin{figure}
  \centering
  \subfigure[{\tiny Prediction error:~8\% data}]
    {\includegraphics[width = 0.285\textwidth]
    {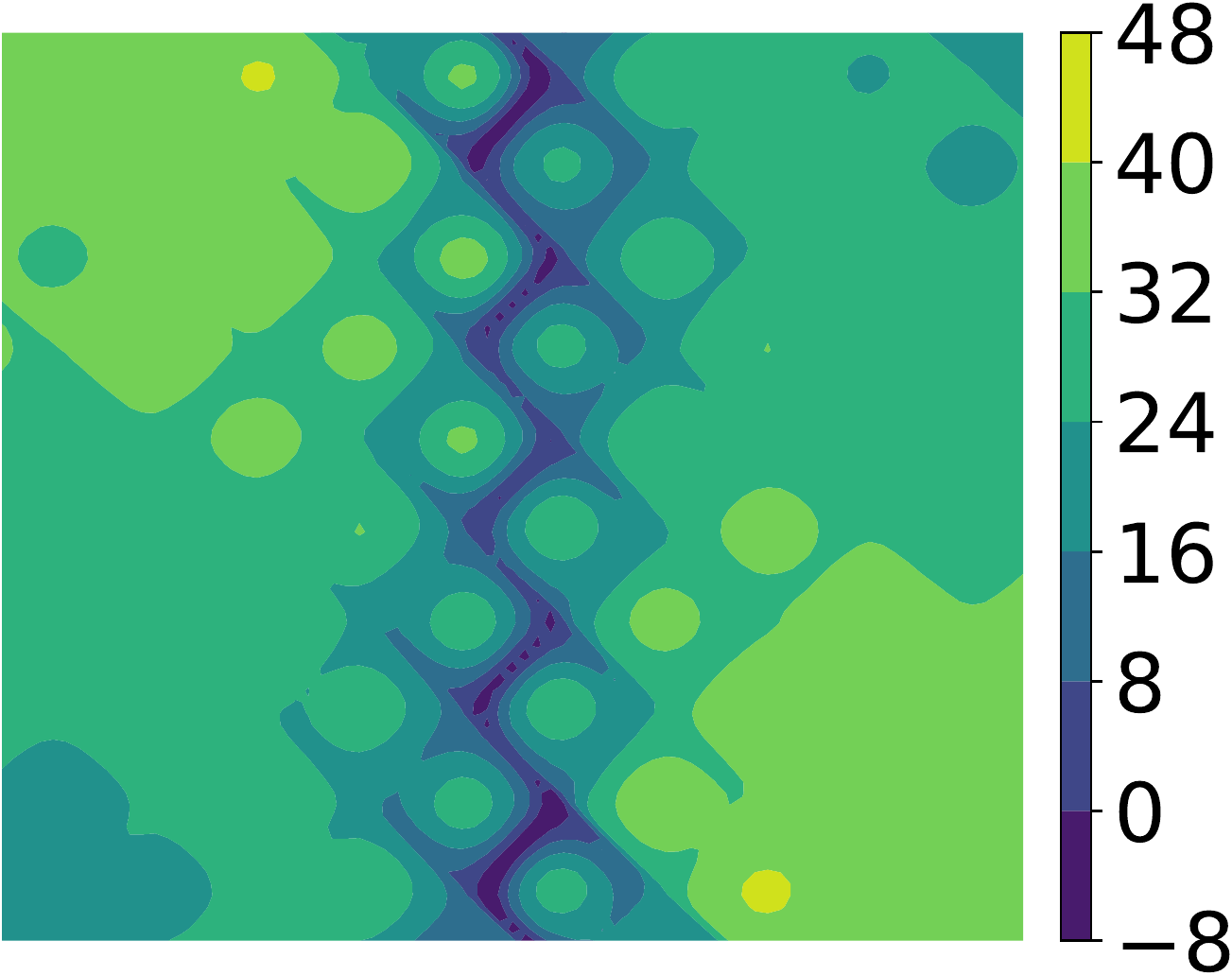}}
  \hspace{0.25in}
  \subfigure[{\tiny Prediction error:~16\% data}]
    {\includegraphics[width = 0.295\textwidth]
    {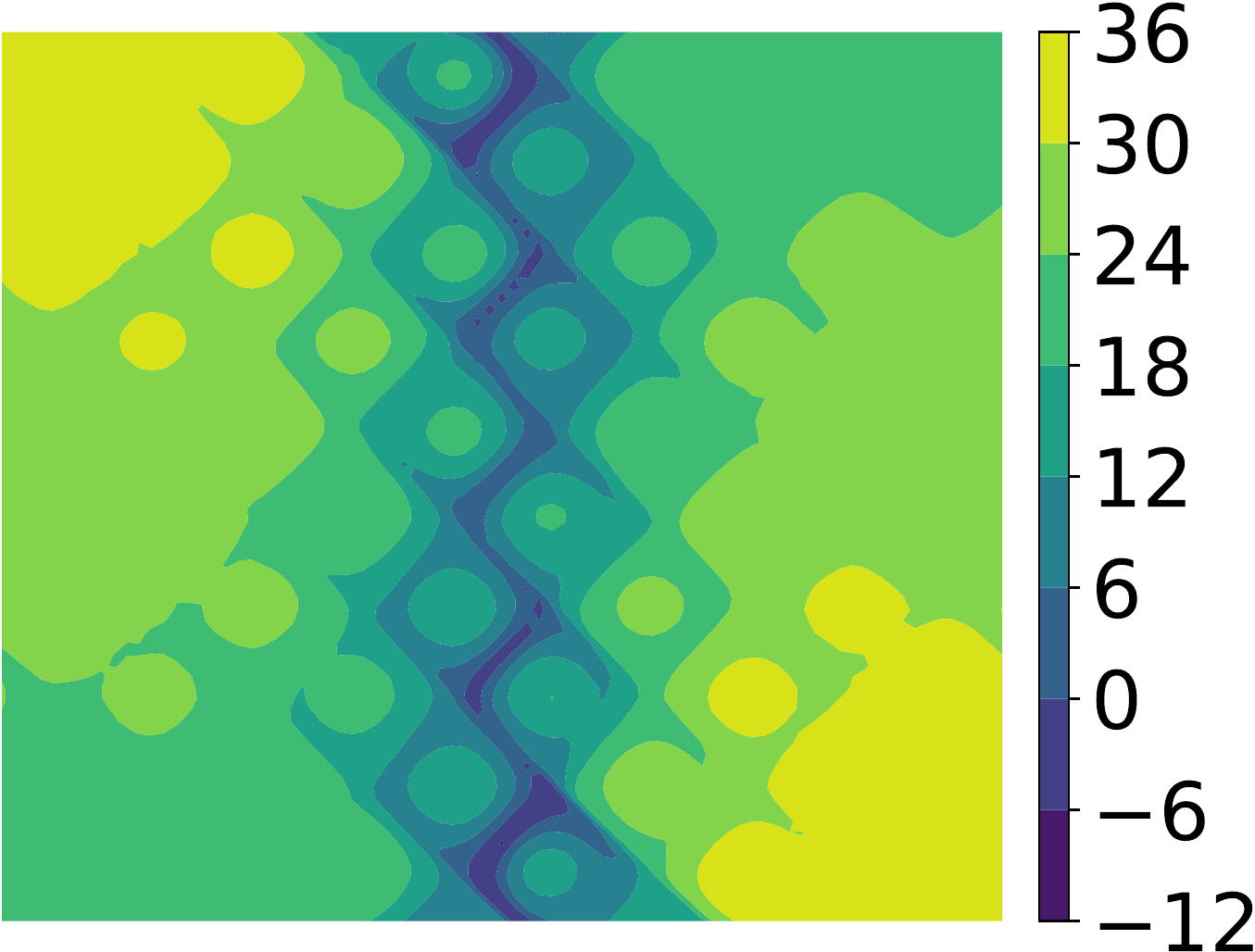}}
  \hspace{0.25in}
  \subfigure[{\tiny Prediction error:~24\% data}]
    {\includegraphics[width = 0.295\textwidth]
    {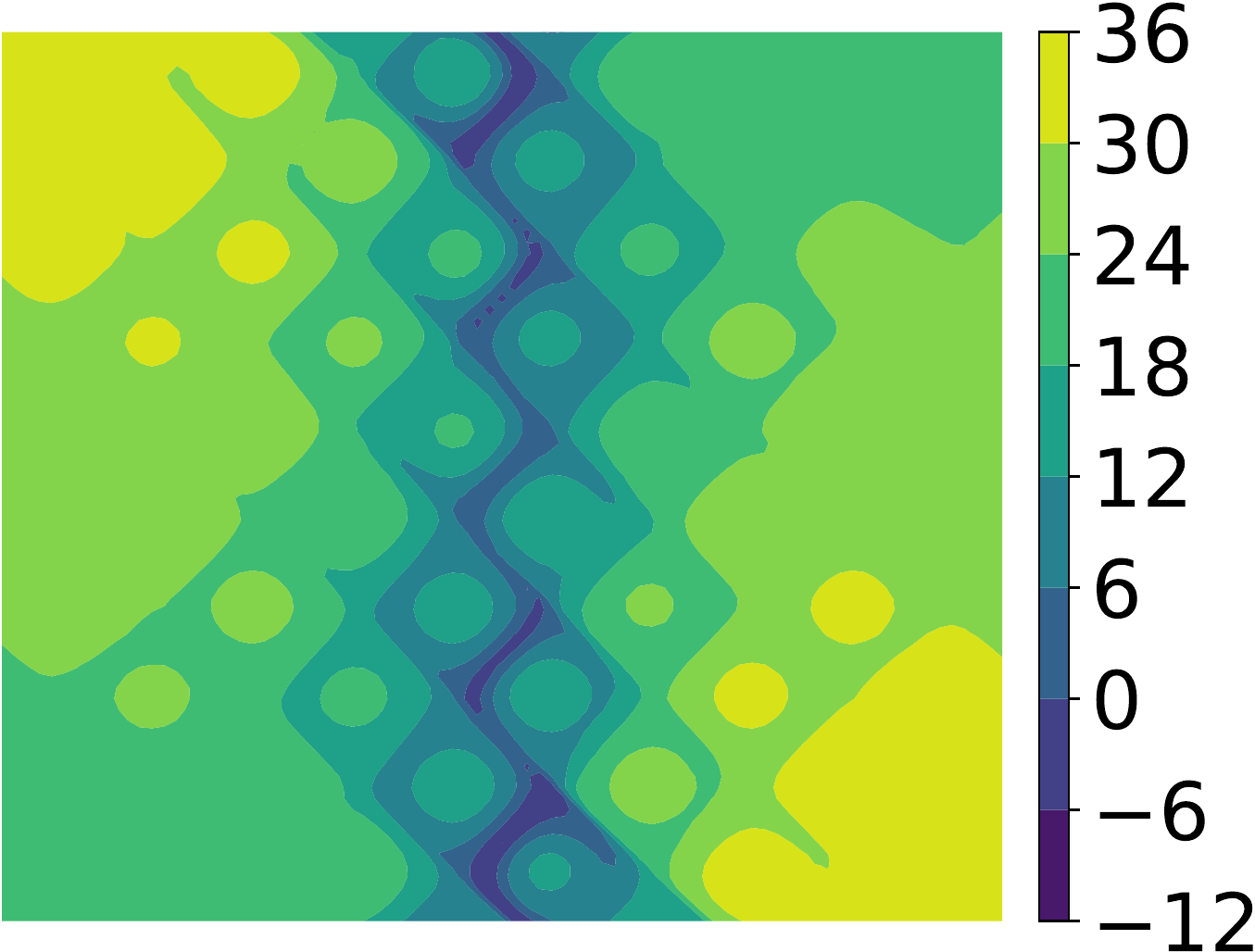}}
  \subfigure[{\tiny Prediction error:~32\% data}]
    {\includegraphics[width = 0.295\textwidth]
    {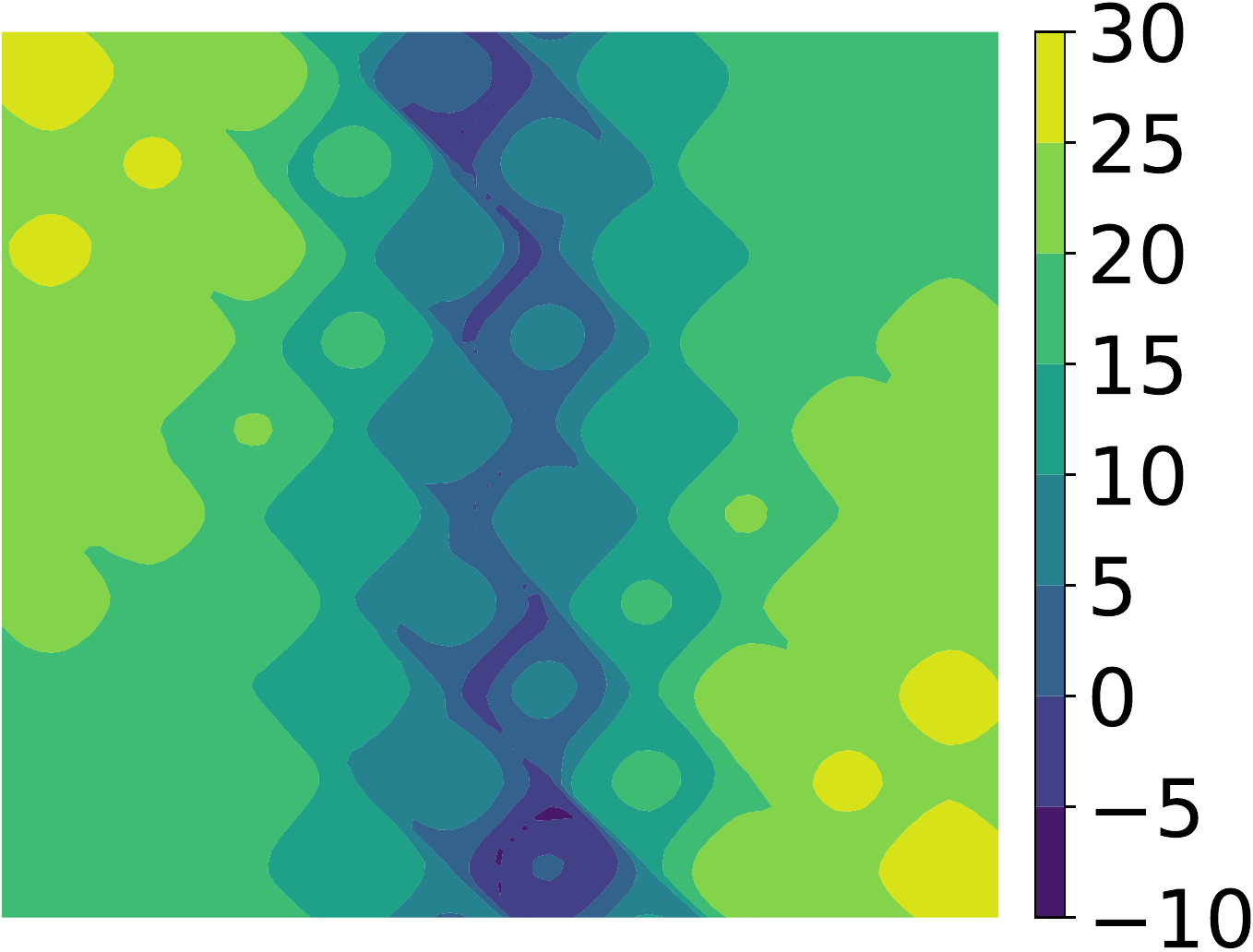}}
  \hspace{0.25in}
  \subfigure[{\tiny Prediction error:~40\% data}]
    {\includegraphics[width = 0.295\textwidth]
    {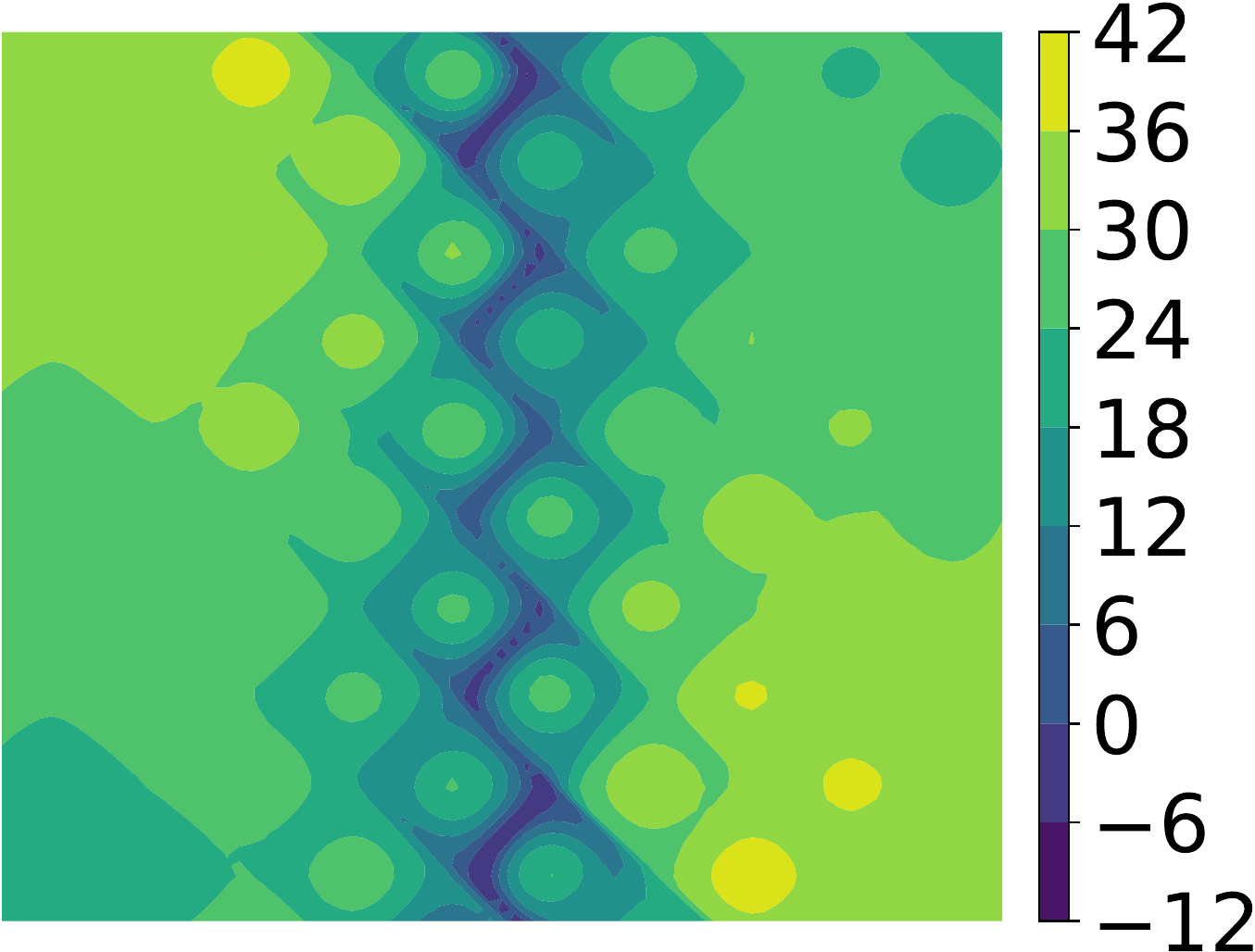}}
  \hspace{0.25in}
  \subfigure[{\tiny Prediction error:~48\% data}]
    {\includegraphics[width = 0.295\textwidth]
    {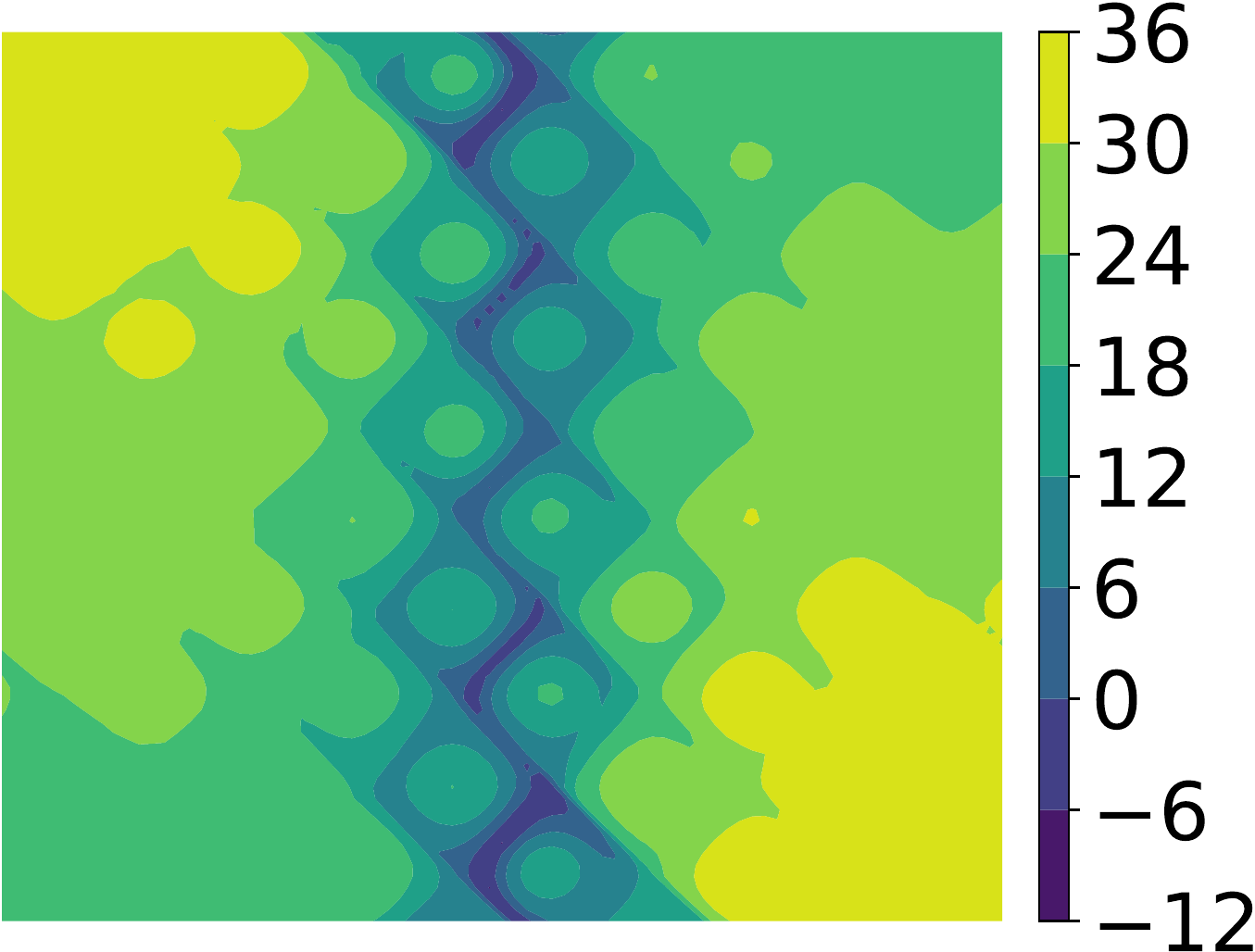}}
  \subfigure[{\tiny Prediction error:~56\% data}]
    {\includegraphics[width = 0.295\textwidth]
    {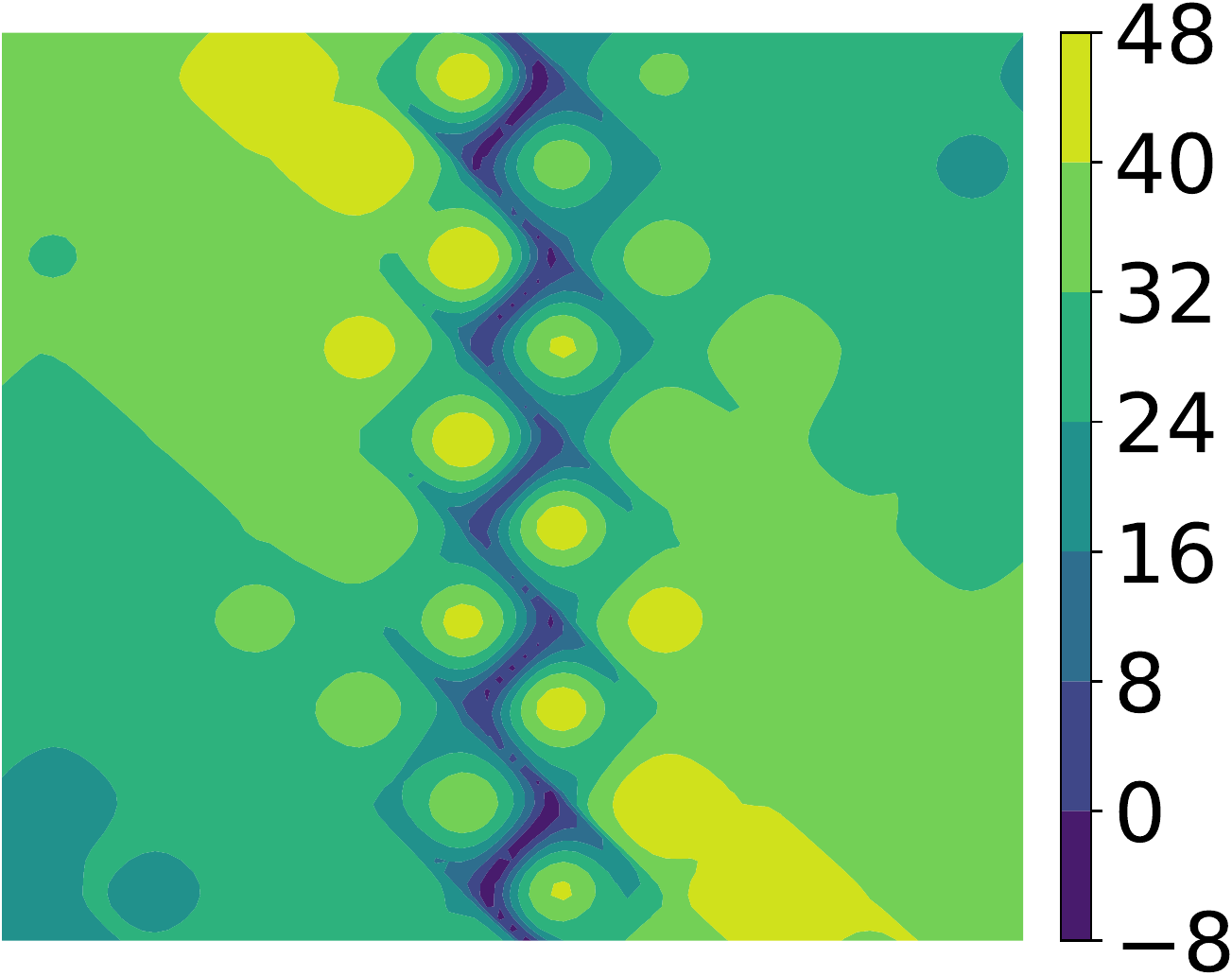}}
  \hspace{0.25in}
  \subfigure[{\tiny Prediction error:~64\% data}]
    {\includegraphics[width = 0.295\textwidth]
    {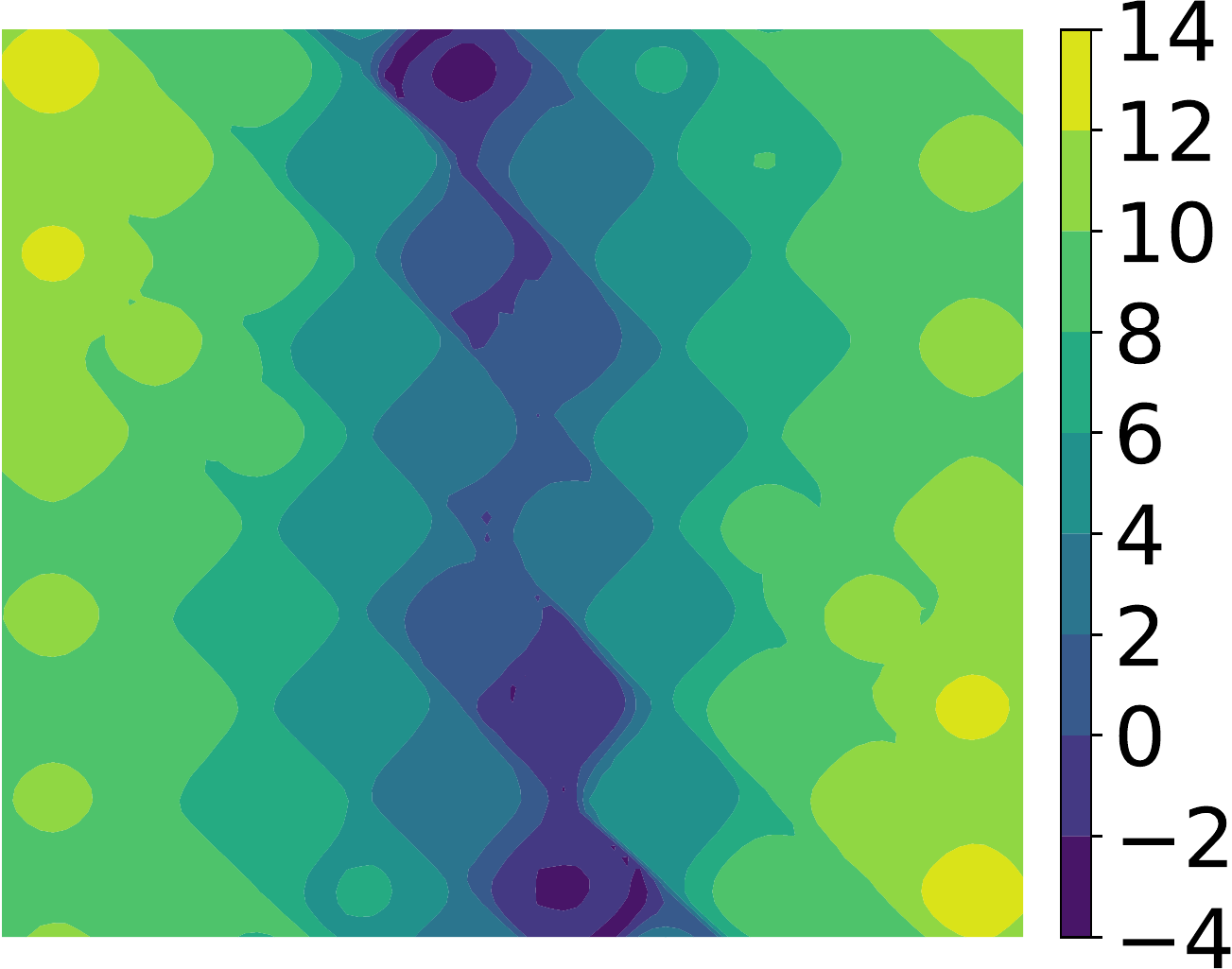}}
  \hspace{0.25in}
  \subfigure[{\tiny Prediction error:~72\% data}]
    {\includegraphics[width = 0.305\textwidth]
    {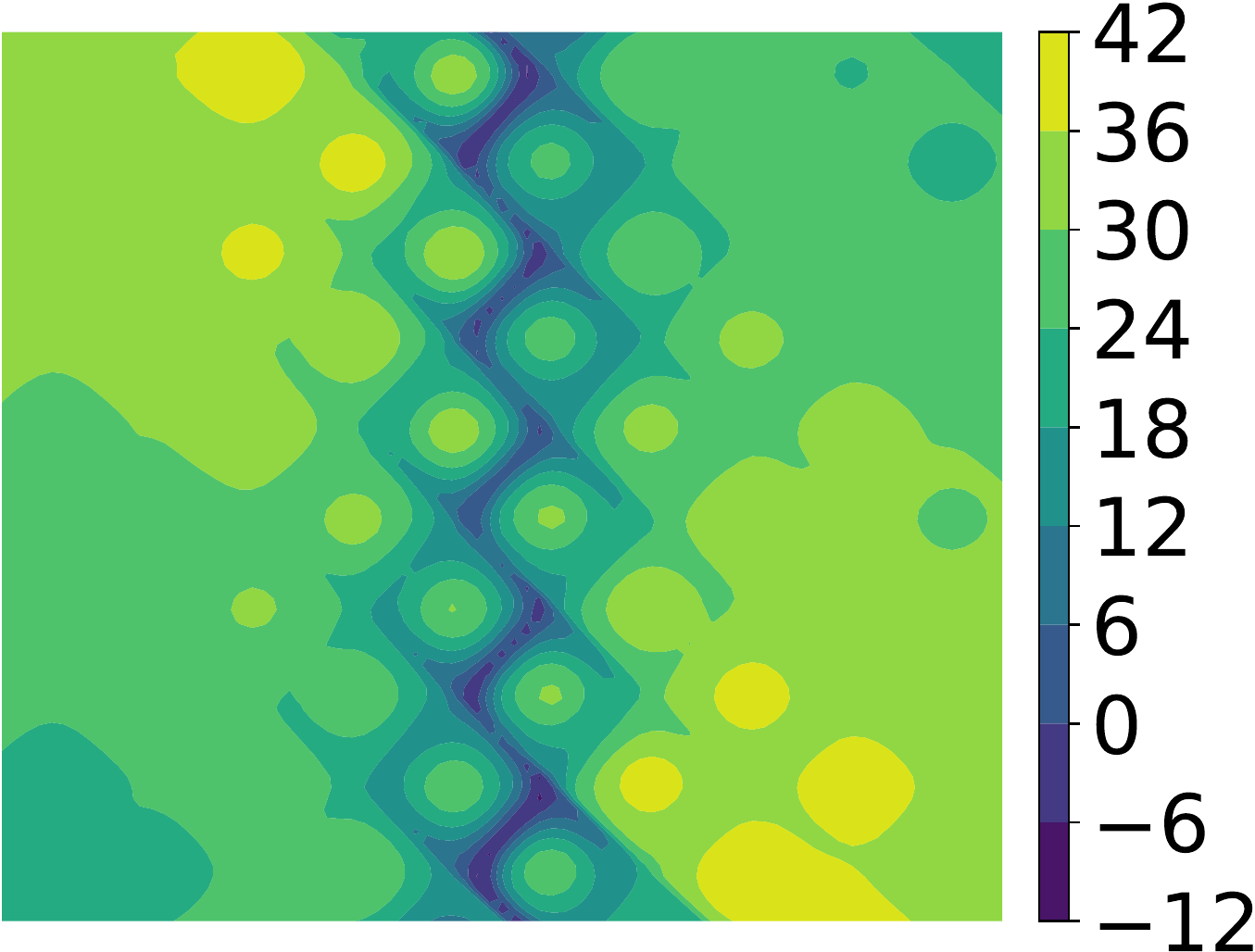}}
  \subfigure[{\tiny Prediction error:~80\% data}]
    {\includegraphics[width = 0.295\textwidth]
    {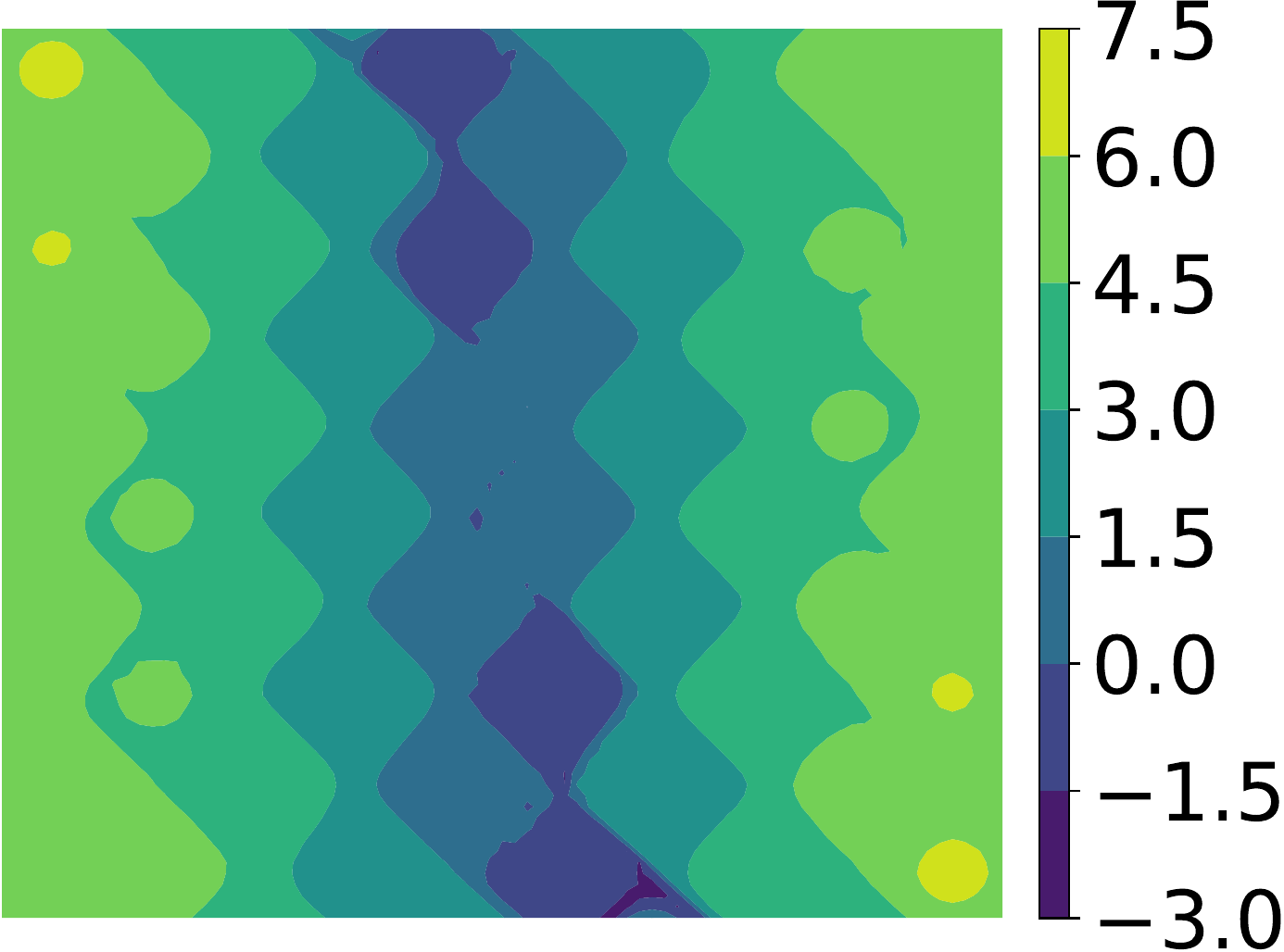}}
  \hspace{0.25in}
  \subfigure[{\tiny Prediction error:~88\% data}]
    {\includegraphics[width = 0.285\textwidth]
    {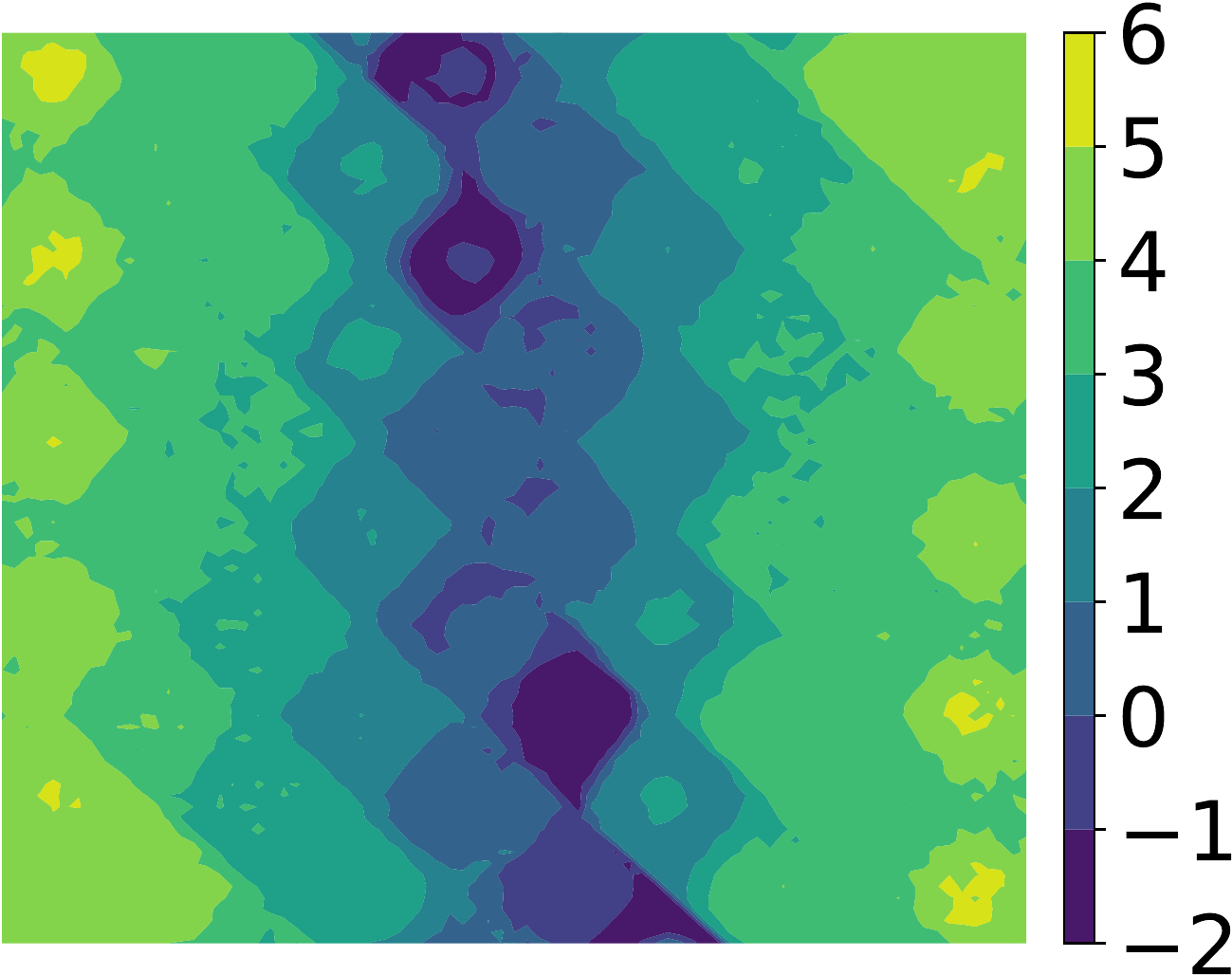}}
  \hspace{0.25in}
  \subfigure[{\tiny Prediction error:~96\% data}]
    {\includegraphics[width = 0.295\textwidth]
    {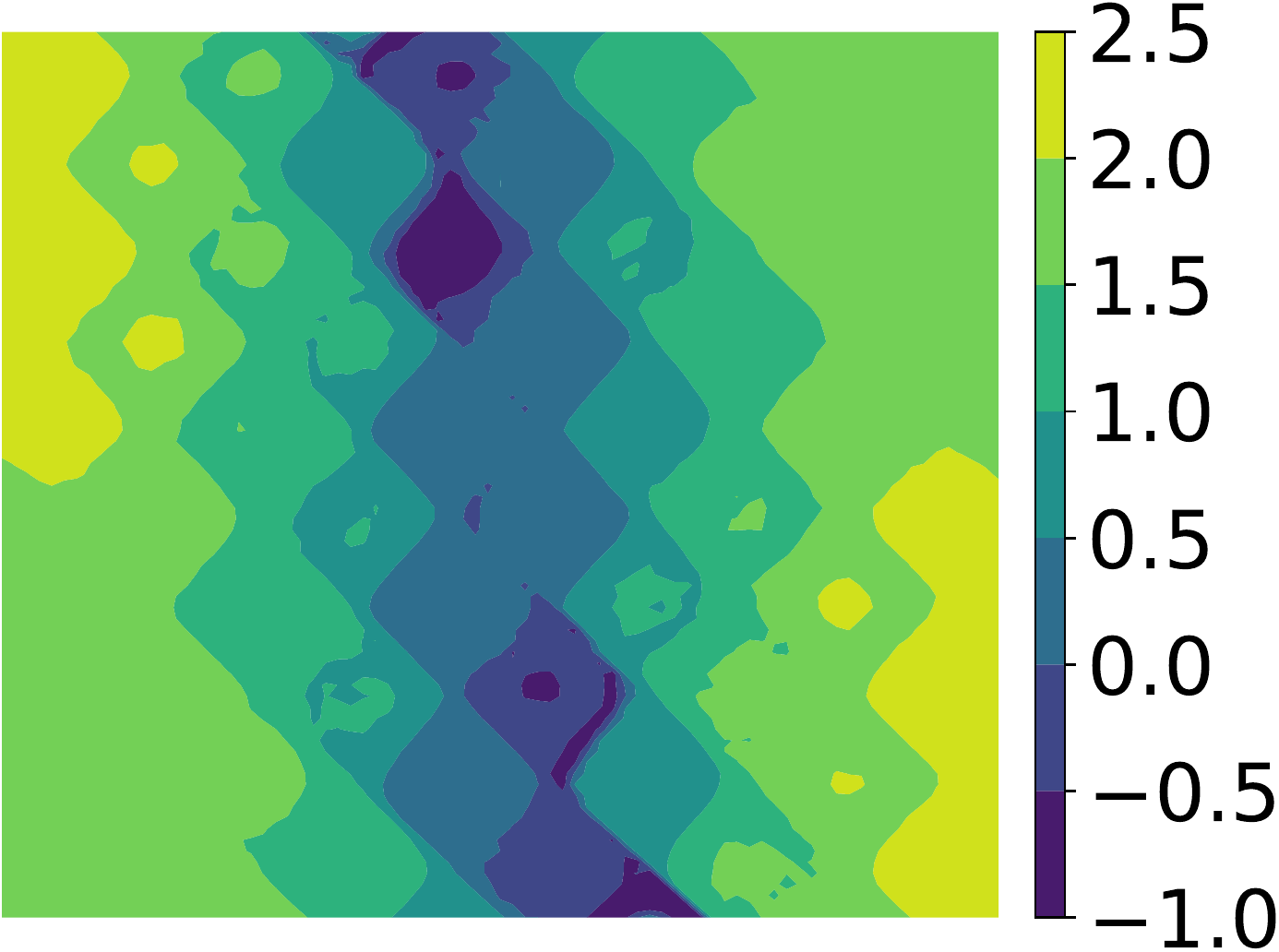}}
  \caption{\textsf{Prediction error in percentage for $\kappa_fL = 5$:}~This figure compares the prediction errors in the entire domain at $t = 1.0$ for different amount of training data.
  Based on the error values (e.g., $\leq 10\%$), it is evident that with 80\% training data, we can accurately capture product formation in the entire domain.
  To accurately predict preferential mixing patterns (e.g., with an accuracy less than 5\%), 64\% of the ground truth is sufficient.
  \label{Fig:DL_RT_Pred_kfL5_Errors}}
\end{figure}
\end{document}